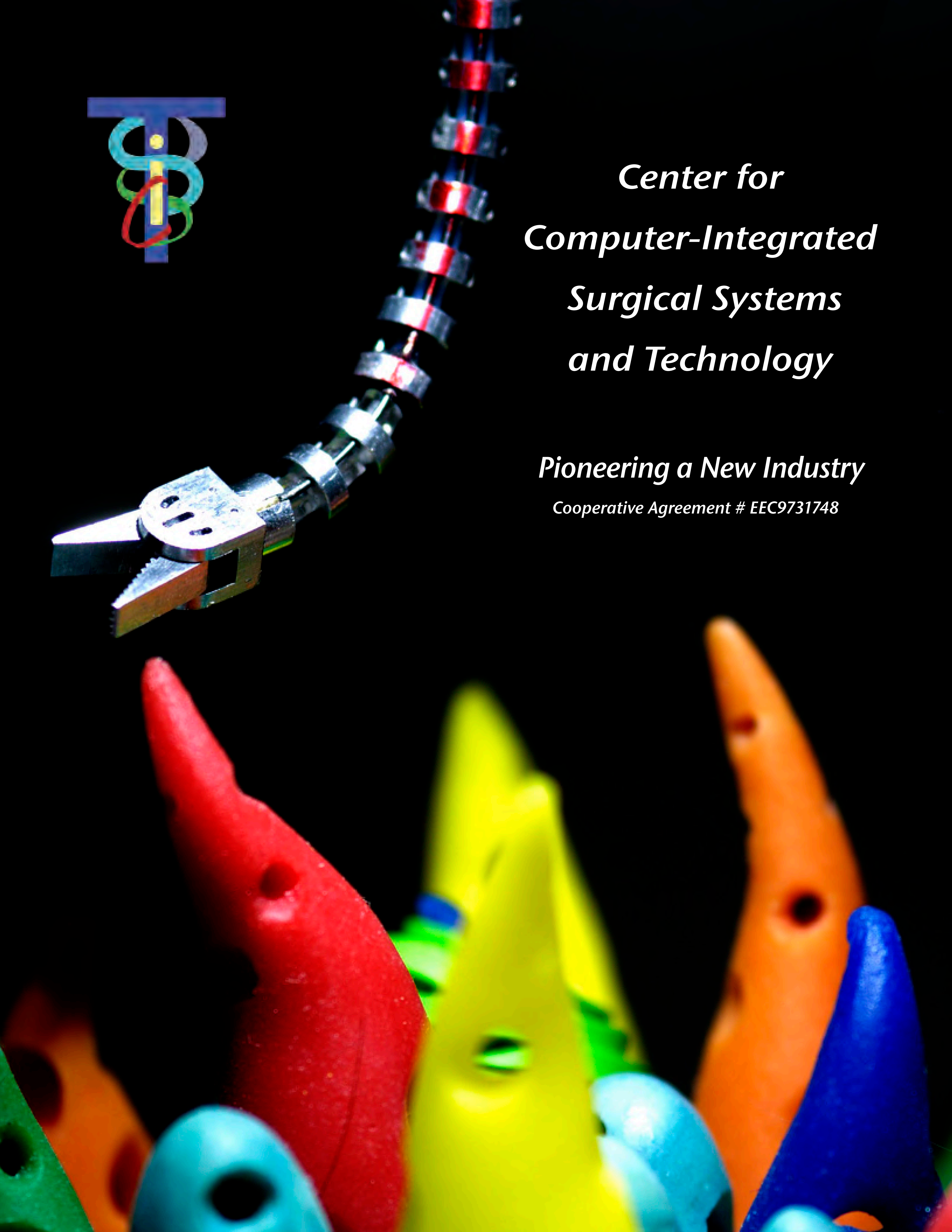

# Center for Computer-Integrated Surgical Systems and Technology

## *Pioneering a New Industry*

*Cooperative Agreement # EEC9731748*

# Center for Computer-Integrated Surgical Systems and Technology

*A National Science Foundation Engineering Research Center*

*Cooperative Agreement # EEC9731748*

*Final Report*

*CISST ERC — Pioneering a new industry*

Center for Computer-Integrated Surgical Systems and Technology

A National Science Foundation Engineering Research Center

Johns Hopkins University

Carnegie Mellon University

Massachusetts Institute of Technology

Brigham and Women's Hospital

Final Report

Director: Russell H. Taylor, Ph.D., JHU

Deputy Director and Associate Director for Research: Gregory D. Hager, Ph.D., JHU

Associate Director for Education: Ralph Etienne-Cummings, Ph.D., JHU

W. Eric Grimson, Ph.D., MIT

Ron Kikinis, M.D., Brigham and Women's Hospital

Cameron Riviere, Ph.D., Carnegie Mellon University

# Table of Contents

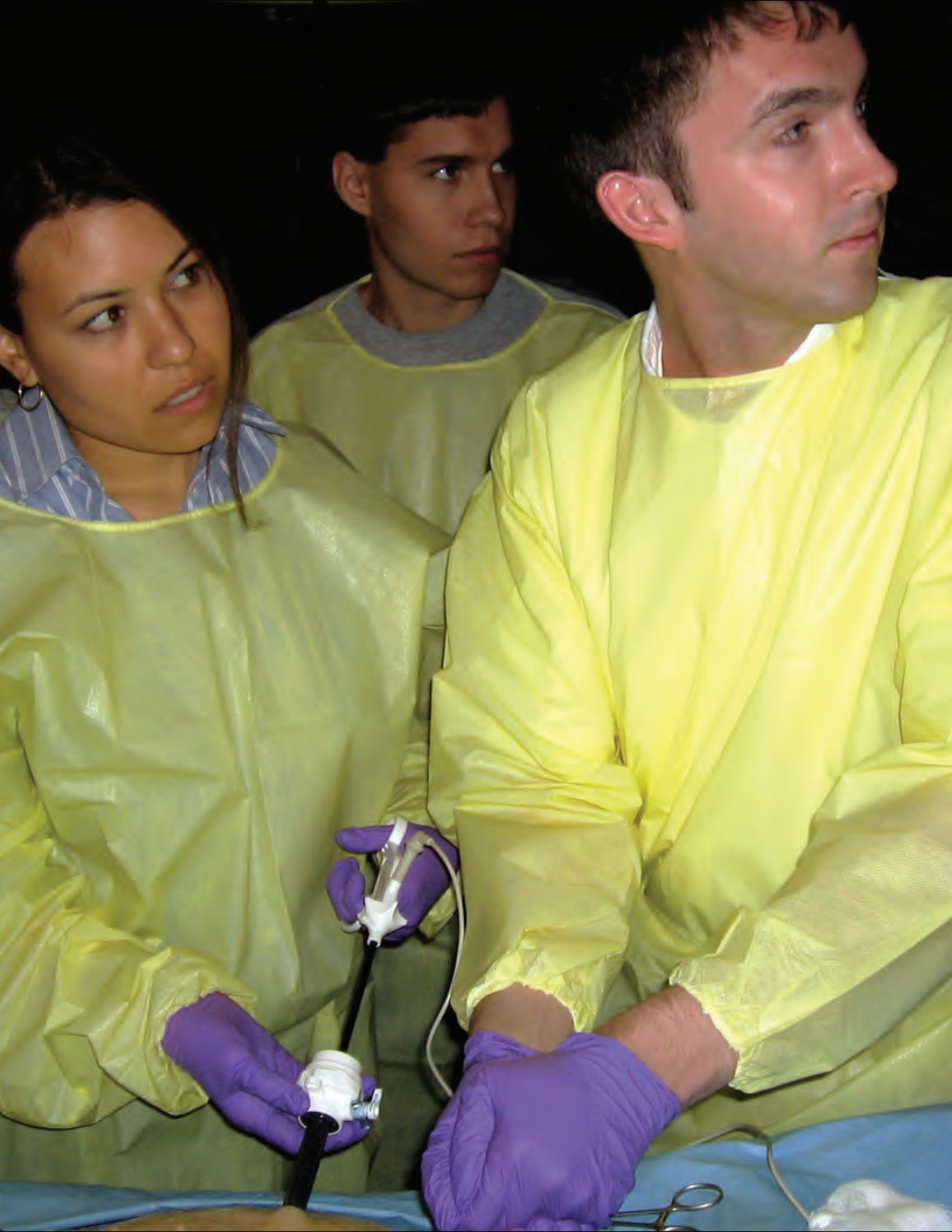

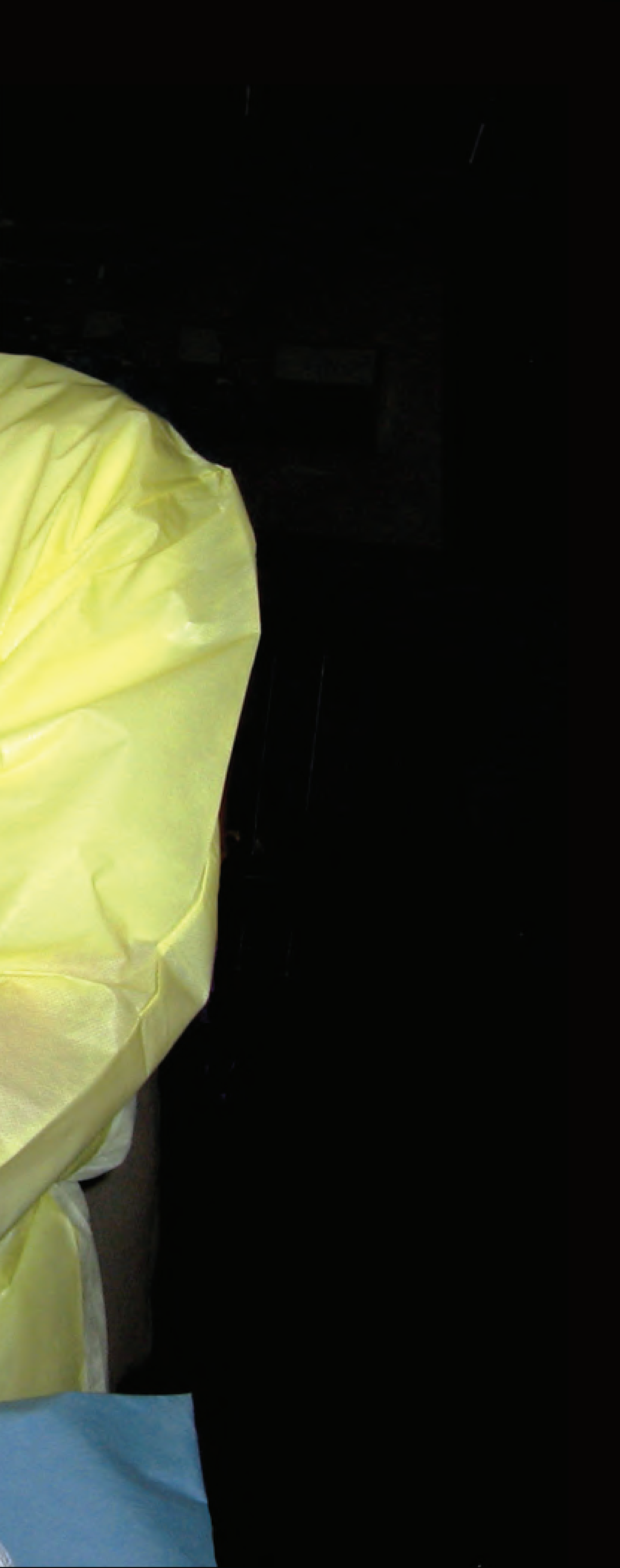

In the last ten years, medical robotics has moved from the margins to the mainstream. Since the Engineering Research Center for Computer-Integrated Surgical Systems and Technology was launched in 1998 with National Science Foundation funding, medical robots have been promoted from handling routine tasks to performing highly sophisticated interventions and related assignments.

The CISST ERC has played a significant role in this transformation. And thanks to NSF support, the ERC has built the professional infrastructure that will continue our mission: bringing data and technology together in clinical systems that will dramatically change how surgery and other procedures are done.

The enhancements we envision touch virtually every aspect of the delivery of care –

- More accurate procedures
- More consistent, predictable results from one patient to the next
- Improved clinical outcomes
- Greater patient safety
- Reduced liability for healthcare providers
- Lower costs for everyone – patients, facilities, insurers, government
    - Easier, faster recovery for patients
    - Effective new ways to treat health problems
    - Healthier patients, and a healthier system

The basic science and engineering the ERC is developing now will yield profound benefits for all concerned about health care – from government agencies to insurers, from clinicians to patients to the general public. All will experience the healing touch of medical robotics, thanks in no small part to the work of the CISST ERC and its successors.

*Computer-integrated surgical [CIS] systems redefine the context and connections between the surgeon and the patient.*

Look at a surgical procedure analytically and it quickly becomes clear that if computers are integrated into the process – including both planning and execution – the potential benefits make it well worth the effort that will be required to overcome daunting challenges. This is what drives the research strategy of the ERC and has led us to create multi-disciplinary teams of scientists, engineers and clinicians to carry it out – and to develop an educational program to prepare engineers and clinicians to be future members of such teams.

Their over-arching challenge: create not merely a system but a family of systems – each exquisitely complex – that will work harmoniously to give the surgeon greater powers than could be imagined until recently. These include algorithms to guide computer processing, imaging devices and sensors to inform planning and execution, robotic devices to perform the desired actions, and human/machine interfaces that give the clinicians precise control of the action.

### *The Anatomy of CIS Systems*

In a CIS system, computer-based information technology provides a functional link between the preoperative plan and the tools used by the surgeon. The computer has access to anatomical atlases and statistical databases, as well as to specific images and other data obtained from the patient. Drawing on all this data, it performs modeling and analysis tasks that are vital to effective planning and monitoring of procedures. The computer also relies on the collective data to assist physically in the intervention,

# Systems Vision

communicate with the surgeon and others in the operating room, and help with follow-up.

In our approach to developing CIS systems, the ERC refers to systems involved in modeling and planning that are coupled with computer-assisted execution and follow-up as Surgical CAD/CAM. For those CIS systems that interact directly with surgeons to extend human capabilities, the ERC uses the term Surgical Assistants. The two classes of systems are, of course, complimentary and often draw upon the same technologies.

*Surgical CAD/CAM Systems: Overcoming Limitations*

The primary focus of the ERC's work with Surgical CAD/CAM is in minimally-invasive, localized biopsies and therapies delivered percutaneously through needle-like instruments. Such therapeutic interventions – including radioactive seed implantation, RF ablation, cryotherapy, localized drug and nanoparticle injections – are performed freehand by the physician. So even with imaging used in planning and during the procedure, errors, variability and inconsistency are common. What's more, if X-ray imaging is used, there is the added problem of dose accumulation. With 3D imaging, the need for two trips to the imaging suite (one for planning, one for the procedure) increases costs and makes it harder to register plan to procedure.

*Data flow is the life blood of the system…creating a patient-specific model used in surgical planning… registering the model to the current reality, and updating it as the procedure continues… enabling a robot to perform as directed by the surgeon, or assisting in manual execution…and providing post-op analysis to guide follow-up and evaluation of the treatment plan.*

*Building the Infrastructure for the Future*

ERC director Russell Taylor has always regarded the NSF grant, large as it is, as seed money for building the infrastructure, culture and collaborative network that would put computer-integrated surgery research and development on the fast track and extend well beyond graduation.

"The funding has allowed us to build 'people' infrastructure," he says. "Engineering is a systems discipline. Our systems are complex, combining computation with action and interfacing with people. To build them," he continues, "we've brought together an infrastructure of professional engineers, the best students, and faculty who have been 'hires of the year' at Hopkins."

Taylor feels that ERC students, as part of the diverse collaborative network, have had a great advantage. He notes that Hopkins has long been known for involving undergraduates in research. "When I was an undergraduate here, I was writing 'traveling salesman codes,' which was an amazing opportunity. But," he notes, "it was with one faculty member. Undergrads in the ERC lab are exposed to many different faculty."

With the ERC now evolving into the Integrating Imaging, Intervention and Informatics in Medicine, or I4M initiative, Taylor feels he has now achieved a major milestone in his career.

The CISST ERC has developed two Surgical CAD/CAM themes to overcome such limitations:

*One-Stop Shopping* – the full integration of planning, execution, assessment and follow-up of minimally-invasive interventions in one setting. Suitable for a range of clinical conditions and organ systems, and as convenient as outpatient diagnostic procedures are today.

*Plug-and-Play* – a modular family of subsystems, from imagers to end effectors, that can be combined quickly to create complete interventional systems for the effective, predictable and certifiable performance of a variety of therapies for multiple organ systems.

### *Surgical Assistant Systems: Enhancing Physician Performance*

While our Surgical CAD/CAM systems are designed for localized procedures, we envision Surgical Assistants playing a major role in procedures that involve more complex manipulation of tools and tissues and the need to make more intraoperative decisions. Given human limitations, Surgical Assistants will be invaluable in minimally-invasive surgery and microsurgery, and likely applications include retinal surgery, ENT surgery, neurosurgery and various endoscopic procedures.

Today, such interventions rely on the surgeon's hand-eye coordination, in spite of natural problems such as hand tremor, sensory limitations and the difficulty of mentally merging multiple sources of information. Even successful telerobotic systems like the daVinci® systems made by our industry affiliate Intuitive Surgical Systems are large and costly, have limited haptic capabilities and dexterity, and still depend on manual control of instruments with video feedback, much like conventional laparoscopic surgery.

The key to reaching the next level in medical robotics is to exploit the ability of computers to fully integrate information. The ERC is pursuing this goal for Surgical Assistants with two kinds of systems:

*Human Augmentation Systems* – using robotic devices to enable the average surgeon to perform as skillfully as a gifted one, by extending and refining ordinary sensory motor skills. These systems will make it possible to perform interventions that cannot be done at all with existing technology.

*Information-intensive Surgery Systems* – Surgical Assistants that interface with an information-rich surgery environment capturing real-time navigational data and fusing multiple information streams to constantly update the modeled anatomy as it is deformed by the procedure and to present critical real-time information to the surgeon.

***Meeting the Research Challenge***

Each module of an envisioned CIS system poses its own formidable research challenge, and there are several modules that must work in concert if the system is to perform effectively and realize its potential. These basic modules are:

- Imaging methods and devices such as novel MRI sensing devices, ultrasound monitoring of ablative therapy, and compact optical and acoustic sensors
- Modeling and analysis algorithms, to plan interventions, process images and give real-time feedback about patient and environment during procedures
- Robotic devices, compatible with a range of imaging methods and operating environments, to execute commands
- Human-machine interfaces to link reality and virtual reality effectively, augmenting and improving operator performance
- Architecture and specifications for joining system components together

***Growing Webs of Collaboration***

Harvard radiology professor Clare M. Tempany, MD, has been involved in a long-term collaboration between the Surgical Planning Lab (SPL) at Brigham and Women's Hospital and the ERC. It began when the SPL needed to move their MR image-guided therapy program for prostate cancer – placing radiation seeds into the prostate gland – into a closed-bore environment requiring robotic technologies.

They turned to Gabor Fichtinger at the ERC, who was working on robotic devices for various applications. Before long the SPL/ERC team, along with industrial collaborator Acoustic Medical Systems, got a Bioengineering Research Grant from the National Cancer Institute. This initial collaborative effort has spawned many others. Tempany, now co-leader of the NIH-funded National Center for Image-Guided Therapy, continues working with Fichtinger and many others originally drawn together by the ERC.

As she says, "The graduates of the ERC have moved on and are establishing similar centers at new sites across the country. The networks of collaboration become more dense," she continues, "the spider webs are being woven across the country, arising from what was a wonderful arrangement in the beginning – the ERC."

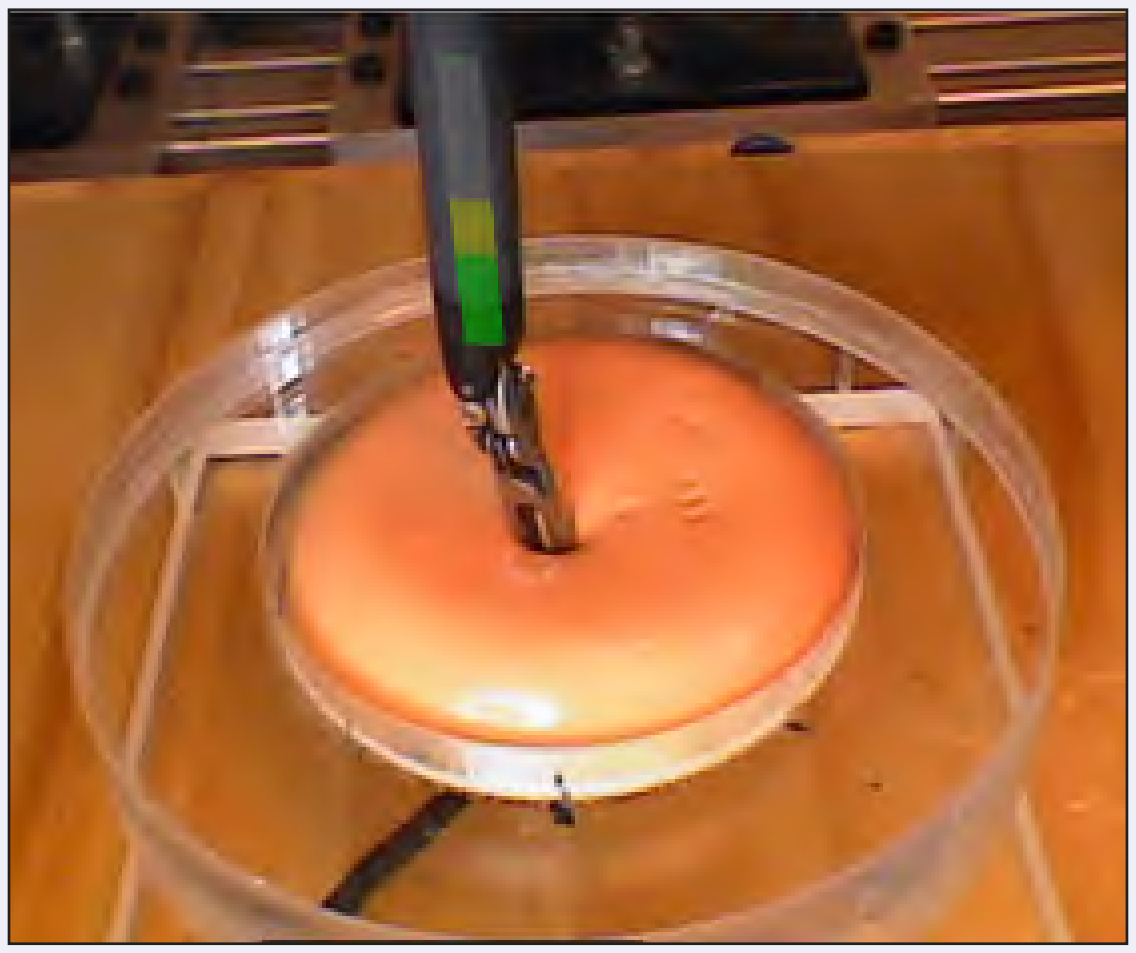

Fig. 1

***Visual Force Feedback***

Sensory substitution is one approach to giving surgeons feedback about the force they apply to tissues using a teleoperated surgical robot. While some research at the CISST ERC involves applying forces directly to the surgeon's hands, sensory substitution replaces force with vision. With it, surgeons can see how much force they are exerting on tissue -- they can tell by watching the color and height of a bar displayed on a monitor. Green means the device is applying little or no force (fig. 1); yellow, moderate force; and red, a surgeon is in danger of damaging tissue or breaking a suture. This approach has been shown to improve the accuracy of surgical tasks performed on artificial tissues, and is less expensive and easier to implement on an existing commercial robot than direct force feedback.

Though instances of these engineered systems can now be created, actualizing our CIS system vision will require a multidisciplinary effort that yields significant advances in fundamental knowledge and enabling technology. We have been pursuing advances in three areas.

*1. Modeling and Analysis*

For planning and executing surgical procedures, we must have new algorithms and representational methods for modeling the patient and the surgical environment. The categories of research in this area are: extracting and correlating data from multiple sources for modeling patient anatomy; combining functional and geometric information; representing and reasoning about uncertainty; and managing complexity. Of immediate interest is the development of methods for intraoperative patient images with preoperative images and atlases, because this is essential to our One-Stop Shopping Surgical CAD/CAM theme.

*2. Interface Technology*

A number of devices and techniques will be needed to ensure that the virtual reality of computers and plans correspond accurately with the reality of operating room, patient and surgeon. The ERC is exploring three types of interface technology:

- Imaging methods and sensors to gain better information about the patient
- Robotic technology that can extend human precision, geometric accuracy and performance in confined spaces
- Human-machine interfaces that improve communications between surgeon and system, including haptic interfaces and superimposed visual displays

# Systems Vision

*3. Systems*

Engineered systems are needed for validating research and providing context for new developments, ensuring system safety and reliability, characterizing expected performance in the presence of uncertainty, analyzing interaction of components, and validating system performance. Additional system needs for Surgical Assistants include cooperative control, information display and teleoperation.

### *Our Evolving Vision: Driven by Information*

From the beginning, the CISST ERC concentrated on the development of the closed, patient-centered information loop that is central in our approach to interventional medicine. Encompassing patient modeling, planning, execution and follow-up, this loop in itself required the capturing and processing of massive amounts of data.

Early on, we realized that there was potential for another closed loop – a process loop that would assimilate the information from numerous individual patient loops (Fig. 2). This would provide the inputs for statistical analysis that could show ways to improve treatment processes, safety and efficiency for future cases. We believe developing this process loop can have a profound impact on national health.

In order to focus resources more effectively, and because the basic tools for modeling and data handling had to come first, the ERC has concentrated on the patient loop. Now, with enabling tools in hand thanks to NSF core funding, we have turned to the process loop, using radiation oncology as the initial application. Information-driven interventional medicine, progressing along both the patient loop and the process loop, will transform health care over the next decade.

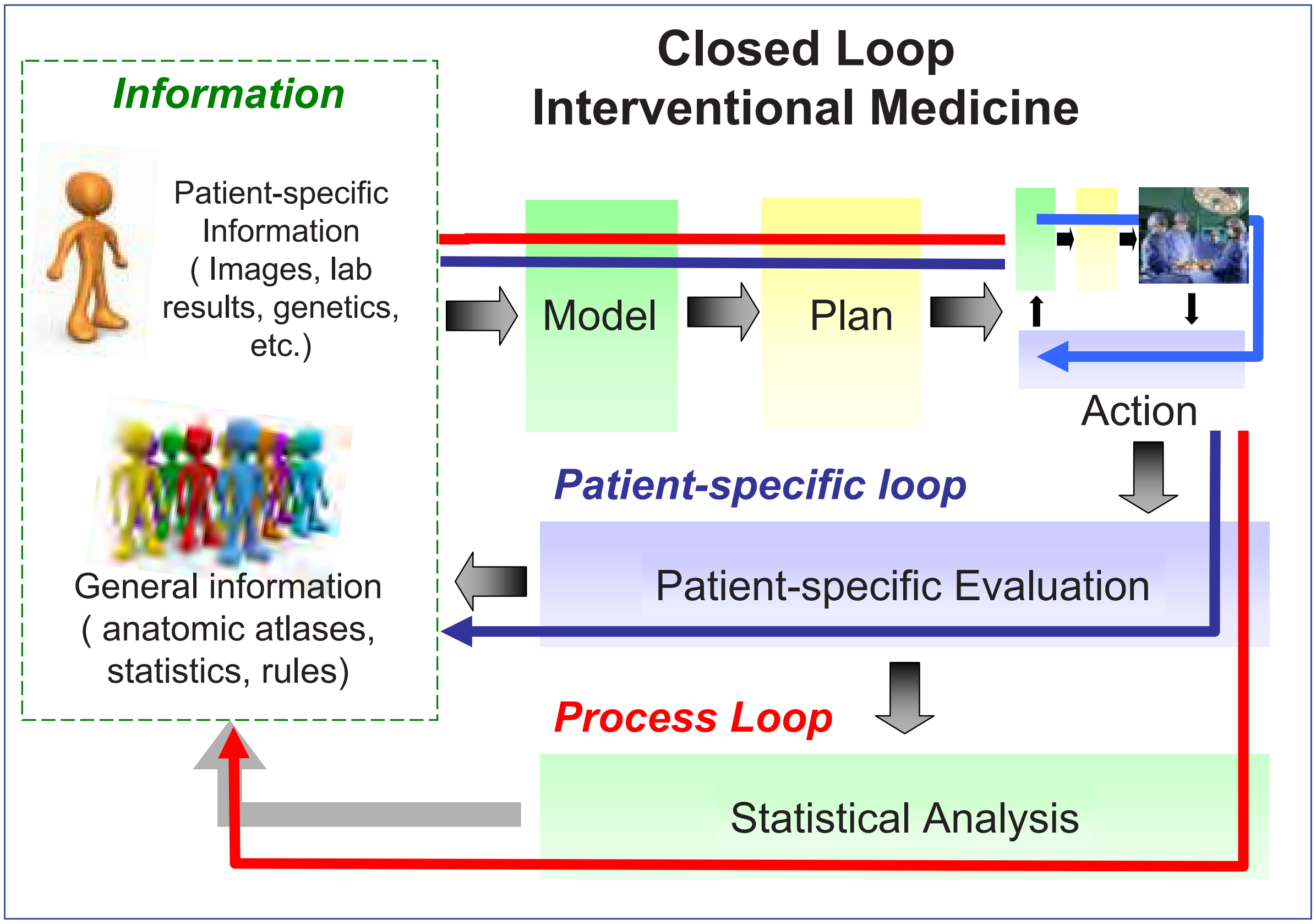

Fig. 2: View of closed loop processes in interventional medicine: Note the multiple time scale nature of information-model-plan-action-evaluation loops in treating an individual patient. Computer-based systems have the inherent ability to retain information created in the course of treating patients and enable subsequent analysis to improve treatment plans and processes.

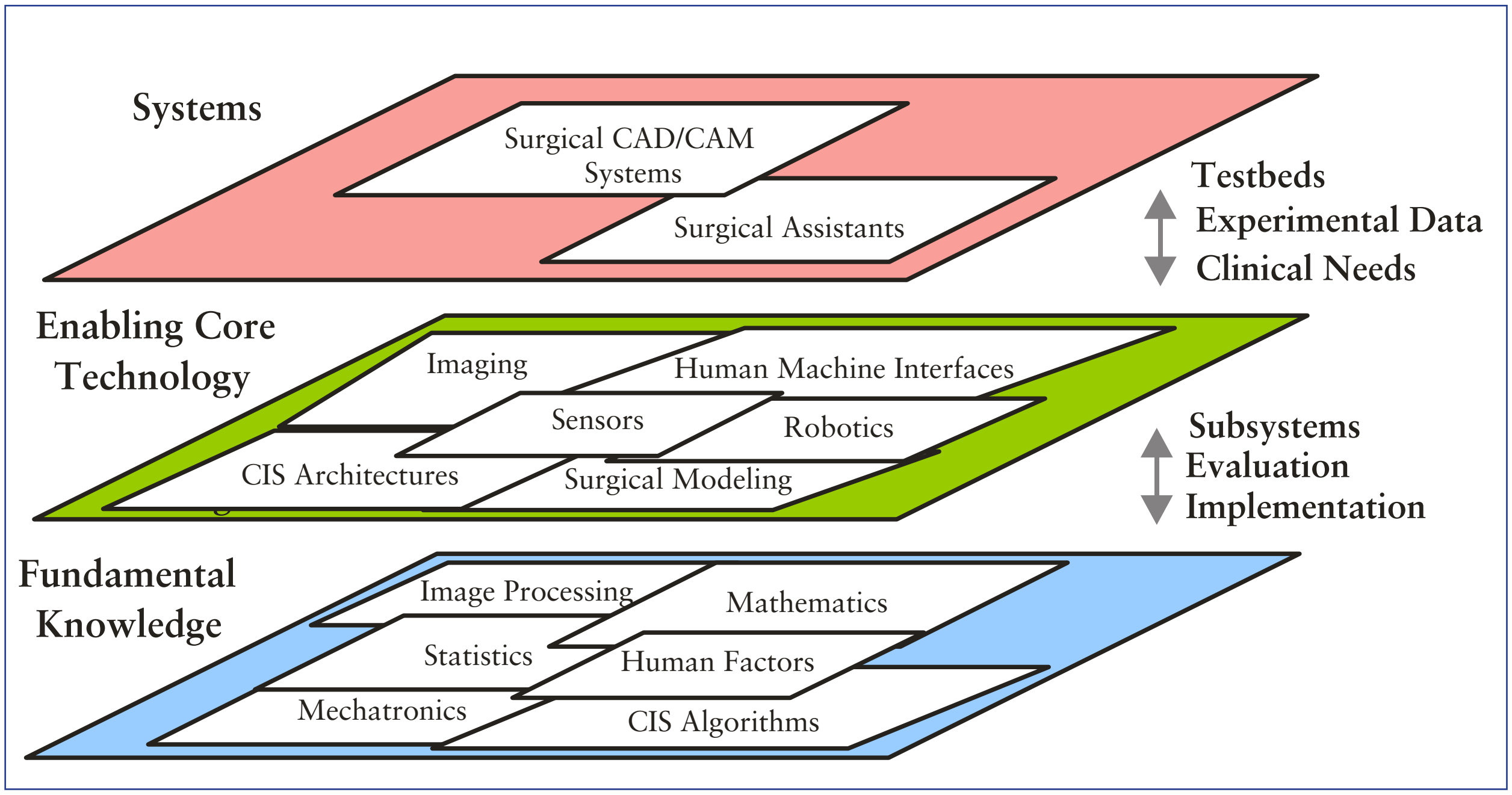

Fig. 3: Systems Driven Research Strategy

*Create a family of highly capable systems that share the same standards, software and devices, for maximum clinical applicability and adaptability.*

When the ERC began its ten-year mission, we identified specific applications, both medical and scientific, where we believed we could make major contributions. Concentrating on these opportunities drove our work from the start, and even as we adjusted our research targeting along the way – based on what we were learning – this focus has sharpened our efforts in evaluating systems, opening new paths for research, and pursuing commercialization.

Along with this decision to create systems for sharply-defined applications, we also made a strategic commitment to devising common infrastructure that could be readily adapted for diverse systems. Our approach has been to develop a set of standards, a body of software and a family of devices that can be used by teams of researchers pursuing different goals. They can simply draw from our core infrastructure as needed, rather than reinventing it, thereby saving an enormous amount of time and using human and financial resources more efficiently.

### *Mapping the Search, in Three Planes*

How ERC investigators approach technological challenges is best visualized as a process that plays out across and between three planes (fig. 3):

At the bottom is the broad intellectual foundation: the fundamental knowledge we are able to gather, create and exploit.

In the middle is the realm of the core technologies that ERC teams develop and draw upon as they conceptualize computer-integrated solutions for the delivery of clinical care.

On the top plane are the engineered systems themselves – the ERC deliverables designed to perform surgical CAD/CAM tasks or serve as surgical assistants.

By positioning key research leaders in this three-planed context, the ERC sets the stage for creative interaction. While focusing intensely in their own areas of expertise on one of the planes, investigators become much more aware of the full array of contributions needed to bring concepts to fruition. There is a constant, productive interplay of people and ideas, and it has grown to include clinicians and industrial partners as well. This has yielded benefits in many areas.

Individual investigators have broadened their horizons. Many now conduct research on more than one strategic plane. This has helped shape the overall ERC program, achieve milestones, develop testbeds and attract additional funding.

We have learned how to reach out to more institutions. As a result, ERC teams are not only multi-disciplinary but also multi-institutional. With this added dimension, their resources and impacts have been broadened.

Graduate and undergraduate students have learned to be better problem solvers, seeing the larger context as it extends from fundamental theories to clinical realities.

The inclusion of clinical and industrial collaborators provides ongoing reality checks about applicability, identifying true opportunities that drive research activities.

Several testbeds are operational, providing the means of testing and evaluating ERC work and pinpointing needs for additional research.

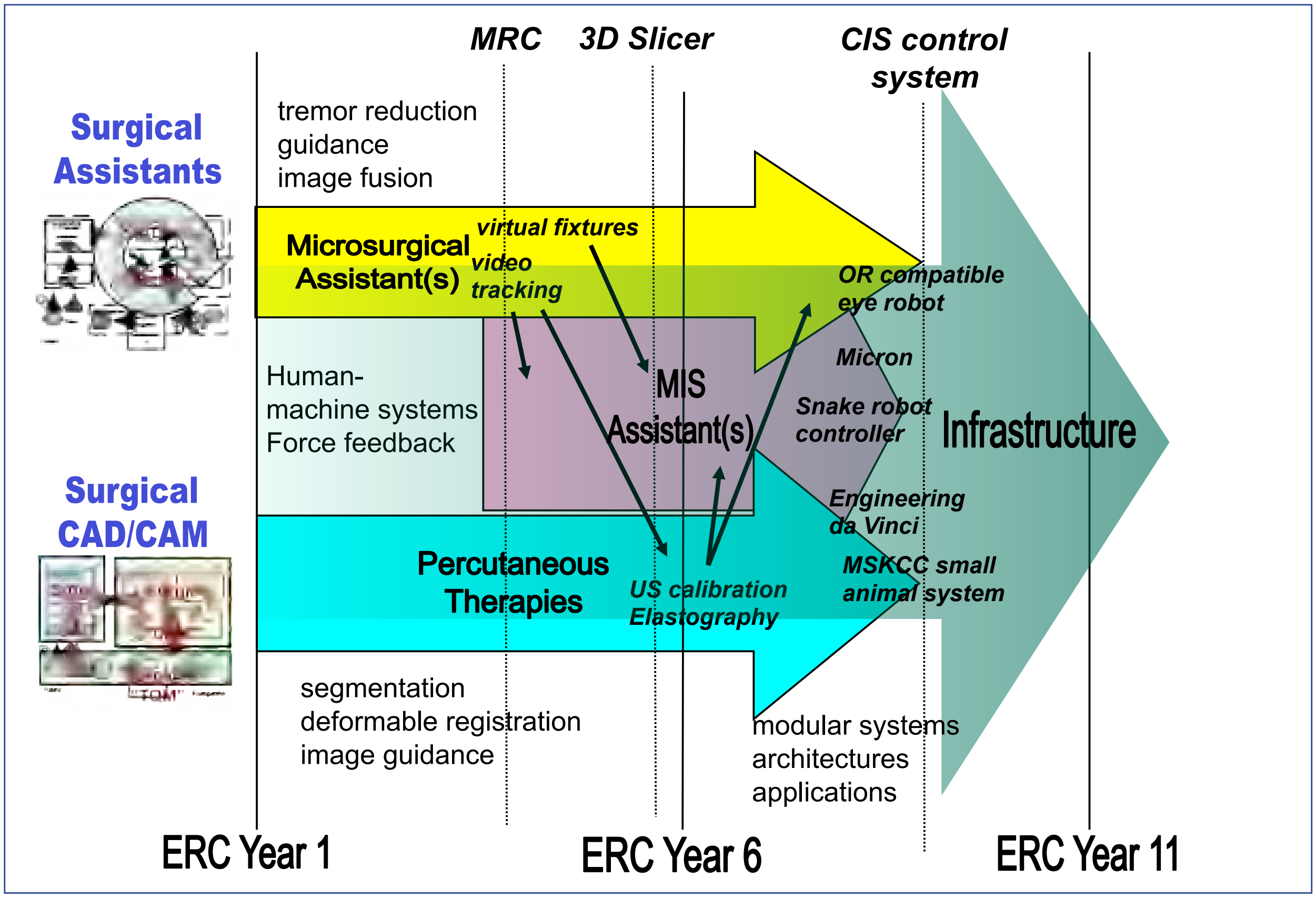

Fig. 4 :Thrust Evolution

*Using the excitement of medical robotics to interest young people – especially females and minorities – to follow in our footsteps.*

If the United States is to continue playing a leading role in the development of new technologies, bright young students must be persuaded to consider careers in science and engineering. Educational outreach has therefore been an important part of the ERC's mission. Fortunately, ours is a field that has a built-in advantage when it comes to generating interest among young people. "Robots" are fascinating. Though popular conception of robotics may not bear much resemblance to reality, it nevertheless provides an educational starting point that can set the path towards understanding and leadership.

Another ERC advantage is that, in keeping with the long-standing Hopkins tradition, we have established strong connections between education and research, creating an environment in which students learn through research.

The ERC conducts programs to reach six communities:

***K through 12 Schools***

The CISST ERC has been very active in the NSF's Research Experiences for Teachers (RET) program, which brings K-12 teachers right into research labs. Our RET program has become a national model, giving over 300 teachers first-hand lab experiences they can take back to their classrooms. Our TeamRET program has further enhanced the RET experience, and a number of RET spin-offs are now also encouraging students to use science, math and technology to solve problems.

# Education & Outreach

***Pre-College Outreach***

The ERC has partnered in presenting a number of events and programs designed especially for students preparing for college. These have included events like Robotics Summer Camp and Robotics System Challenge, Computer Mania Day, and tours of ERC labs, as well as FIRST: Future Inventors, Researches, Scientists and Teachers, with the National Inventors Hall of Fame.

***Undergraduate Students***

The ERC has initiated a number of programs to heighten interest in CIS systems among undergraduates. Most involve coursework, including an introduction to CIS and the opportunity to Minor in CIS at JHU, a Surgery for Engineers summer course, and an agreement that allows students to enroll for a semester in an ERC-partner school other than their home institution. Our 10-week Research Experience for Undergraduates (REU) summer program gives students a first-hand encounter with CIS systems research at Hopkins. The REU program now also includes Louis Stokes Alliance Minority Partnership participants, assuring greater ethnic diversity. Finally, the Computer-Integrated Surgery Student Research Society, which has existed as long as the ERC itself, plays an active role –promoting student research, helping with recruitment, and sponsoring a growing number of activities for members. It is worth noting that even students who are not otherwise associated with the ERC, but who are looking for research opportunities, have been welcomed in our labs.

***Graduate Students***

The ERC values its graduate students for the many tasks they handle, from recruitment to research, and from supervision to mentoring. Every effort is made to maintain a collegial atmosphere in which interaction with faculty and other students is routine, whether in the lab, in seminars or at social events. Within that framework, graduate students can also take advantage of enrichment programs. Weekly seminars feature research presentation by faculty, students and guest speakers. The Friday Student Seminar Series takes a less formal approach, with ERC graduate students taking on a range of topics including ERC research, broader engineering topics, and career development. Also, by participating in the NSF program, GK-12 BIGSTEP (Broader Impact for Graduate Students to Teach Engineering), graduate students not only help the K-12 students and schools in science, technology, engineering and mathematics (STEM) education, but also sharpen their own teaching and interpersonal skills.

*Practitioners*

A major reason for the ERC's success is the collaboration between those who create CIS systems with those who use them. This has inspired us to develop two related courses at JHU: Engineering for Surgeons, and Surgery for Engineers. A more compressed version of the latter is being planned to accommodate the schedules of professionals who work for our industrial affiliates.

In all of these areas, particular attention is paid to involving those who are typically under-represented in science and engineering disciplines: women, and members of minority groups. This is important for two practical reasons. One is that mixing people with diverse backgrounds and perspectives is a formula for greater creativity. The other reason is that a more inclusive talent pool will be a larger one, and as the international market for good scientists and engineers becomes more competitive, the U.S. will need to draw on a larger pool of local talent.

*"What are we aiming at?*

*The encouragement of research... and the advancement of individual scholars, who by their excellence will advance the sciences they pursue, and the society where they dwell."*

*~ Daniel Coit Gilman*
*JHU Inaugural Address*
*February 22, 1876*

### *Intuitive Surgical's Perspective on ERC Value*

The CISST ERC relationship with Intuitive has produced benefits that cut across many research programs and interaction mechanisms, creating several collaborative research projects. These projects have included work led by Intuitive with subcontracts to CISST ERC, and vice versa. Intuitive is licensing technology from CISST, and has hired several CISST ERC graduates. One CISST ERC graduate even returned to JHU as research faculty after a period of employment on research projects at Intuitive. Technical work has spanned the fields of novel surgical robot mechanisms, image-guidance, haptics, and computer vision.

As an example of the collaboration, in 2003, Intuitive's da Vinci robot had no image guidance capability. Prof. Russ Taylor worked with the newly established Applied Research Group at Intuitive, championing the cause of ultrasound integrated with da Vinci. This resulted in a joint research proposal to NIH, involving two engineering faculty, one medical school faculty, and several students and Intuitive researchers, and became a project that has spanned three years, with three generations of prototypes. It resulted in a transfer of technology not just from the CISST ERC to Intuitive's Applied Research Group, but also to Intuitive's product development group, which plans to release the first da Vinci ultrasound instrument to help enable robotic partial nephrectomy.

Currently, the vast majority of the 30,000 nephrectomies performed for kidney cancer each year are full nephrectomies, which remove the entire kidney. Robotic assistance is helping to drive more and more of these procedures to be partial nephrectomies, which spare part of the kidney to preserve function, reducing morbidity and extending life. The procedure depends heavily on ultrasound for localization of the tumor. Through several series of laboratory trials, we have learned that the articulated ultrasound probe driven by the robot console surgeon is far superior to conventional ultrasound maneuvered by an assistant. We expect robotic ultrasound to reduce the amount of time that the kidney must survive during surgery without blood flow, improving the preservation of kidney function for the patient. This kind of outcome takes years of joint effort between the university and a company.

Intuitive has found that only a few universities, among dozens the company has interacted with, possess the broad engineering capability and clinical collaborations required to make an effort like this succeed. Productive academic-industrial relationships take a great deal of effort. In addition to creating cross-cutting capability, pulling resources into an ERC arrangement makes it much more efficient for a company to leverage its investment in a university relationship to cover many different projects, essentially creating a "repeat customer" for the university's translational research capabilities. These traits have made CISST ERC the most productive of Intuitive's worldwide collaborations.

*~ Chris Hasser, Intuitive Surgical*
*ERC Industrial Member*

# *Clinical & Industrial Collaboration*

If new technologies are to revolutionize the way surgery and other interventions are done, there must ultimately be a strong partnership that includes not just researchers but also clinicians and industrial affiliates who can put ERC solutions to the test and take those solutions to the market.

Our clinical partners realistically define problems and potentials, trouble-shoot our development processes, conduct trials and ultimately validate CIS systems.

*Finding the partners who can move innovative systems from the research lab to the operating room through commercialization and marketing.*

Industrial partners bring more to the process than production and marketing capabilities. Their knowledge of end users' needs and how to meet them is indispensable, making them important team members long before the commercialization stage. And they have the financial resources to help speed up the R&D process and give researchers greater responsiveness and flexibility as new opportunities are revealed.

For all these reasons, the ERC is continuously refining its Industrial Affiliates Program to attain maximum, mutual benefit from these partnerships.

***Acoustic Medsystems – CISST***

The CISST ERC and Acoustic Medsystems enjoy a long and fruitful collaboration. Both parties are responsible for boosting the capabilities of the other throughout the relationship. JHU gives Acoustic Medsystems the advanced technology and experienced faculty it needs to help create new products, while Acoustic Medsystems offers funding for projects, training for students, and a gateway for the commercialization of mutually produced technologies.

The collaboration has been responsible for the framework and "underpinnings" for several research projects in the areas of prostate brachytherapy and acoustic ablation. Projects include "Transperineal Brachytherapy", "Transperineal Interventions in Closed MRI", and "Ultrasound Ablation of Bone Cancer under CT Fluoroscopy". The research is geared toward clinical applications and provides a hands-on learning environment for students, faculty, and industrial researchers.

In one project, Acoustic Medsystems and the CISST ERC worked together to increase the potential for improved patient outcomes in ultrasound therapy. The collaborating team also found ways to reduce pain, morbidity, treatment and recovery times, and treatment costs. These improvements are all possible within an outpatient procedure with only a local anesthetic.

During this collaboration, which has been successful both clinically and commercially, both the CISST ERC and Acoustic Medsystems have developed and improved technologies. Working together has advanced a diverse group of fields and has created the potential to revolutionize the treatment approach for localized diseases, cancer, and other conditions.

# *Infrastructure*

EVEN BEFORE THE CISST ERC WAS ESTABLISHED, Hopkins was widely recognized for successful engineering-medicine collaborations. This made it the right setting for the ERC, which would need to build on that quality. Because our work is mission-focused and concentrated here, the ERC today stands as the prime example of the collaboration paradigm, extending not only through different Hopkins divisions but also among the ERC partner institutions and industry affiliates.

Thanks in great measure to the generous, long-term support of our primary sponsor, NSF, the ERC has been able to people this collaborative environment with a diverse and highly-talented array of scientists, engineers, clinicians and students. Each has come with distinctive technical skills and personal attributes, yet all have shared a focus on the mission and a personal commitment to collaboration.

To maintain focus and continuity, the ERC has established a core of professional engineers to work with faculty and students. It has also recruited a number of excellent faculty members, who have enabled the ERC to reach "critical mass" in medical robotics. And, with the assurance of NSF funding over ten years, we were able to entice the best students into our enterprise – not only graduate students but also undergraduates, who benefit from the Hopkins practice of giving them hands-on research experience.

The recently completed Computational Science and Engineering Building at Hopkins will serve as the venue for our continued development of CIS systems. ERC robotics and vision faculty, engineers, postdocs, administrative staff and about 50 gaduate students will now be drawn together in a 15,000 square foot section of this research facility, which features nine individual labs adjoining a 2,500 square foot shared lab space. a machine shop, a large optical bench, and a mock operating room for testing integrated systems. This excellent facility will intensify the intellectual exchange and support that characterize our work.

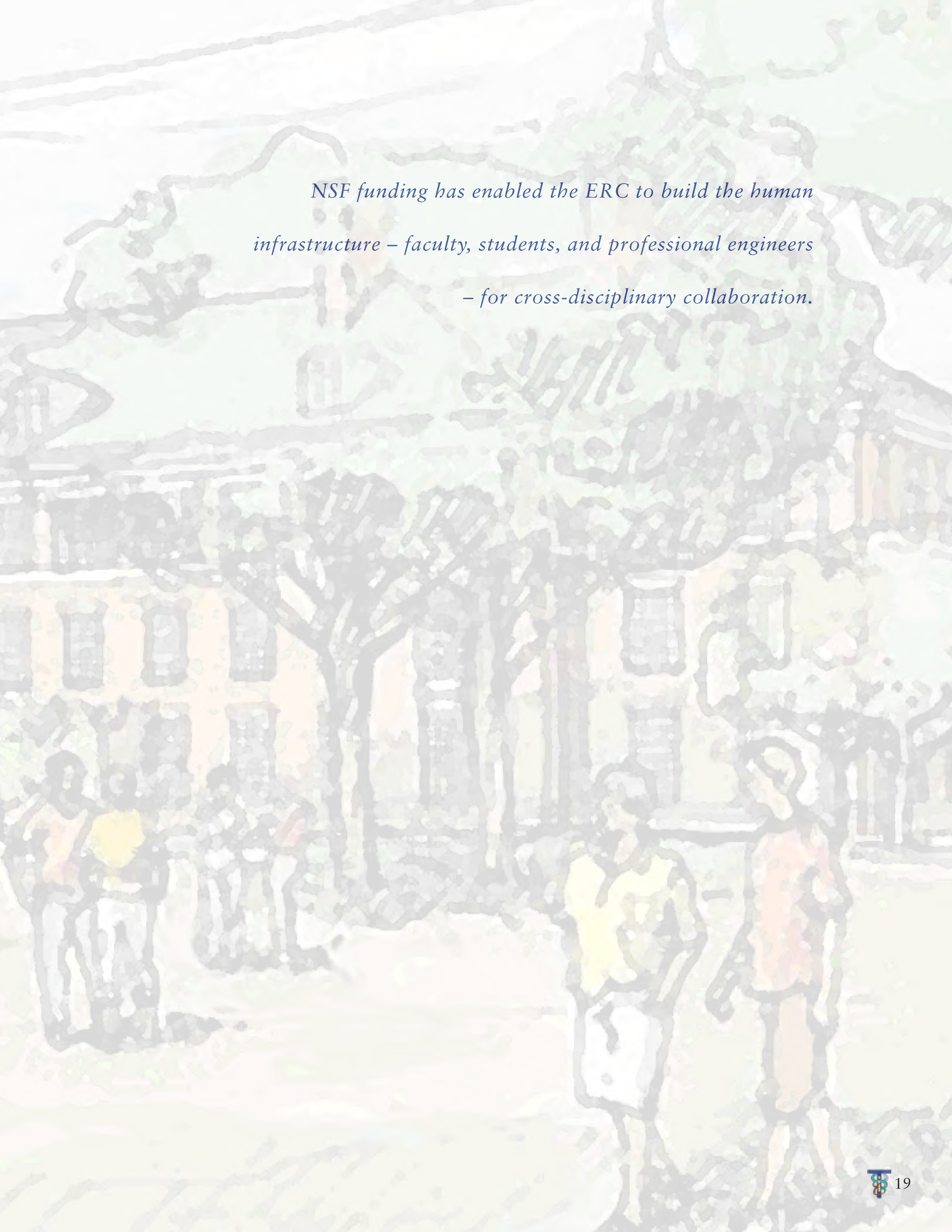

*NSF funding has enabled the ERC to build the human infrastructure – faculty, students, and professional engineers – for cross-disciplinary collaboration.*

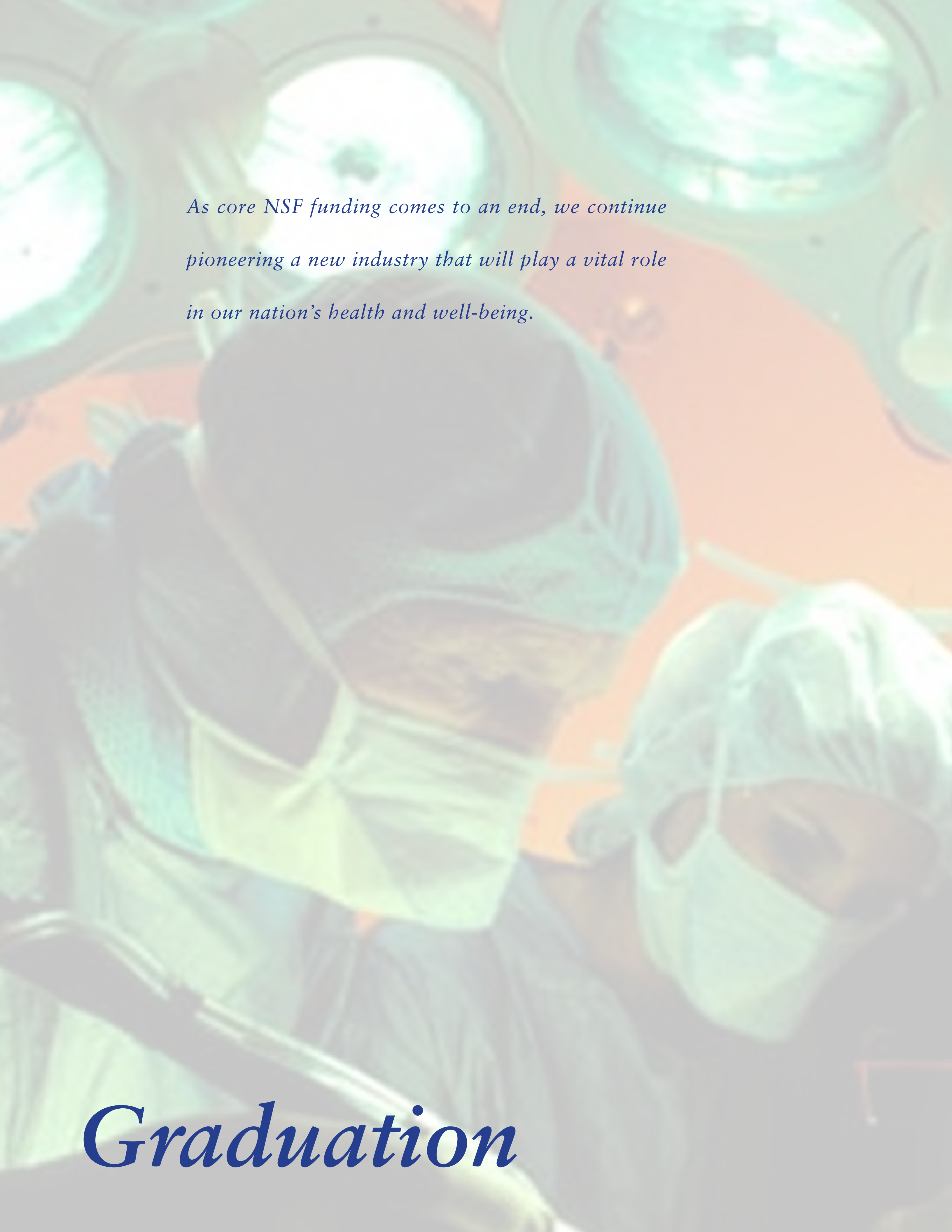

*As core NSF funding comes to an end, we continue pioneering a new industry that will play a vital role in our nation's health and well-being.*

# *Graduation*

From the beginning, we regarded the NSF support as seed money. Now, the seeds have been planted, results are validating our research strategy, and a whole network of ERC alumni is in place at universities, medical institutions, corporations and government agencies. We are confident that we will be able to sustain and amplify our focus on clinical applications of novel, high-performance CIS systems.

We have learned how to win multi-institution grants and have received a major grant from the National Institutes of Health (NIH), as well as grants from other government and private entities. Our aim is to maintain our core engineering and management infrastructure, clinical testbeds, education and diversity outreach initiatives, and connections with industry affiliates. As we move closer to clinical trials and commercial deployment, we will be putting greater emphasis than ever on relationships with medical and corporate affiliates.

A major component of our continuation strategy is the new Hopkins multi-divisional program known as Integrating Imaging and Information in Interventional Medicine, or I4M. As the name suggests, this program fully incorporates the informatics aspect of our CIS developments, which we believe will have a profound impact on the effectiveness of surgical procedures and other interventions. I4M will also be broader than the ERC organizationally. In addition to continuing collaboration with other institutions, it will now bring in the capabilities of the Johns Hopkins Applied Physics Lab and other groups at the university.

Also bolstering the research process that began with the ERC is the new Laboratory for Computational Sensing and Robotics (LCSR), which will house the robotics faculty of JHU's Whiting School of Engineering (WSE) and extensive lab facilities for their research activities and those of collaborators in medicine, biology and computation. The ERC will become a "Center" within LCSR.

By bringing these multidisciplinary programs, people, and resources together under one roof in our new Computational Science and Engineering Building (CSEB), we are multiplying the synergistic possibilities dramatically.

This final report of the Engineering Research Center for Computer-Integrated Surgical Systems and Technology marks a transition rather than an ending. Thanks to the generous financial support and guidance provided by the National Science Foundation, the United States is now a leader in the development of medical robotics. Perhaps more important, we can now foresee a time when patients with health problems will be better served by healthcare providers armed with more effective tools and techniques founded on CIS technology.

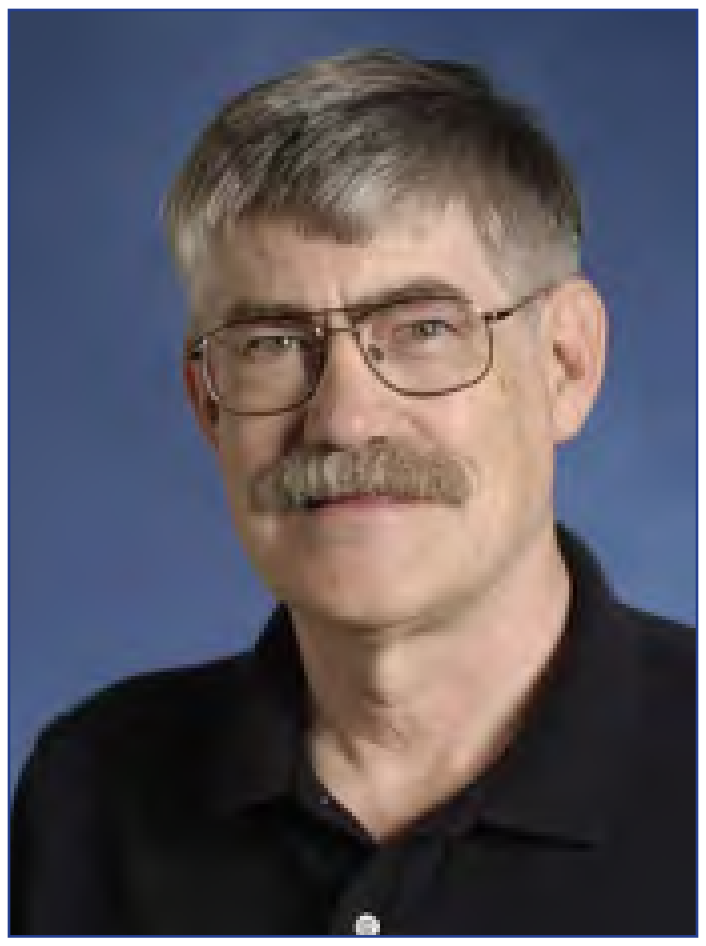

***Russell H. Taylor***

Director

***Russell H. Taylor*** received his Ph.D. in Computer Science from Stanford in 1976. He joined IBM Research in 1976, where he developed the AML robot language and managed the Automation Technology Department and (later) the Computer-Assisted Surgery Group before moving in 1995 to Johns Hopkins, where he is as a Professor of Computer Science with joint appointments in Mechanical Engineering, Radiology, and Surgery, and is Director of the NSF Engineering Research Center for Computer-Integrated Surgical Systems and Technology. He is the author of more than 230 refereed publications, a Fellow of the IEEE (1994), of the AIMBE (1999), and of the Engineering School of the University of Tokyo (2009). He is also a recipient of the Maurice Mueller Award for excellence in computer-assisted orthopaedic surgery (2000) and of the IEEE Robotics and Automation Pioneer Award (2008), as well as several earlier awards for his work at IBM.

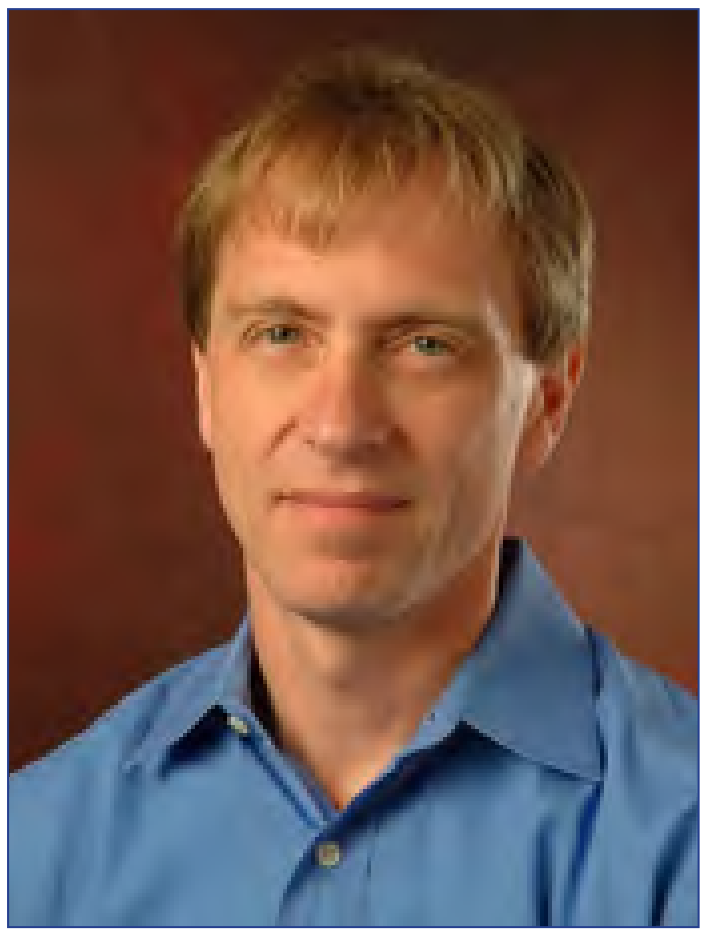

***Gregory D. Hager***

Deputy Director

Associate Director for Research

***Gregory D. Hager*** received a PhD in Computer Science from the University of Pennsylvania 1988. From 1988 to 1990, he was a Fulbright junior research fellow at the University of Karlsruhe and the Fraunhofer Institute IITB in Karlsruhe, Germany. From 1991 until 1999, he was with the Computer Science Department at Yale University. In 1999, he joined the Computer Science Department at Johns Hopkins University, where he is Deputy Director of the Center for Computer Integrated Surgical Systems and Technology. Professor Hager's current research interests include visual scene analysis, visual tracking, and vision-based control with applications to medical robotics, robot manipulation, and to human-computer interaction. Professor Hager is a fellow of the IEEE for his contributions in Vision-Based Robotics.

# *Biographies*

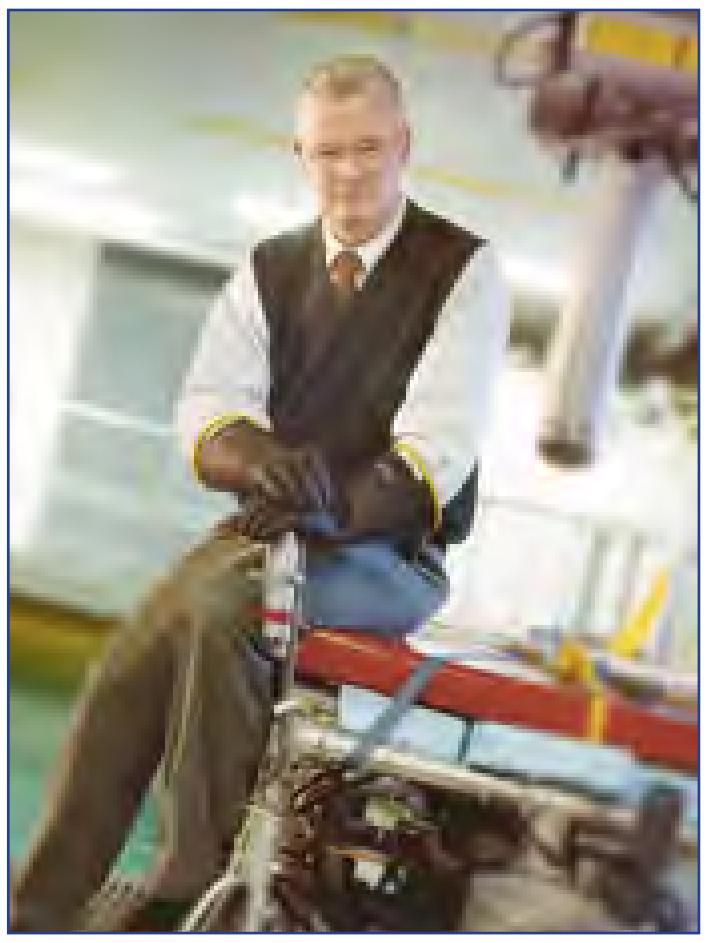

*Louis L. Whitcomb*
Acting Deputy Director for Year 10
and
Director of the LCSR

*Louis L. Whitcomb* received his B.S. and Ph.D. degrees at Yale University in 1984 and 1992, respectively. His research focuses on the dynamics and control of robot systems – including industrial, medical, and underwater robots. He is a principal investigator of the Nereus Project which recently developed a robotic vehicle capable of performing unmanned oceanographic research missions to the deepest depths of the world's oceans – to 10,900 meters (6.77 miles) depth. Whitcomb has received numerous teaching awards at JHU, and was awarded a NSF Career Award and an Office of Naval Research Young Investigator Award. He has numerous patents in the field of robotics, and is a Senior Member of the IEEE. He is the founding Director of the JHU Laboratory for Computational Sensing and Robotics. Whitcomb is a Professor in the Department of Mechanical Engineering, with joint appointment in the Department of Computer Science, at the JHU's Whiting School of Engineering.

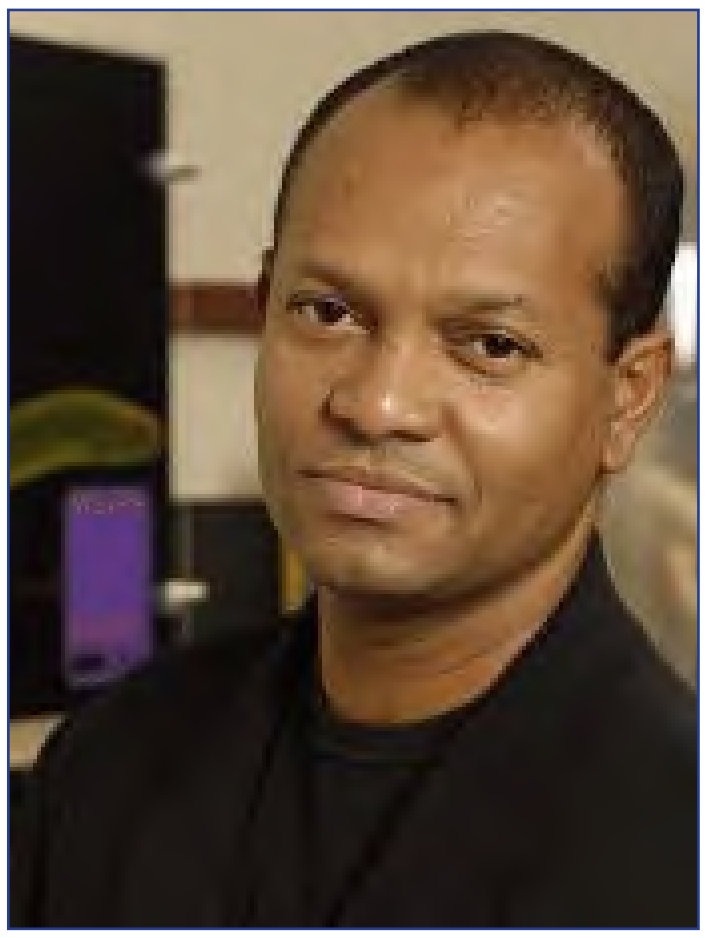

*Ralph Etienne-Cummings*
Associate Director for
Education and Outreach

*Ralph Etienne-Cummings* received his M.S.E.E. and Ph.D. in electrical engineering at the University of Pennsylvania in 1991 and 1994, respectively. He served as Chairman of the IEEE Circuits and Systems (CAS) Technical Committee on Sensory Systems and on Neural Systems and Application, and was re-elected as a member of CAS Board of Governors in 2006 - 2009. A member of Imagers, MEMS, Medical and Displays Technical Committee of the ISSCC Conference from 1999 – 2006 and a recipient of the NSF's Career and Office of Naval Research Young Investigator Program Awards, he was also named a Visiting African Fellow and a Fulbright Fellowship Grantee at the University of Cape Town, South Africa (2006). He has also been recognized for his activities in promoting the participation of women and minorities in science, technology, engineering and mathematics.

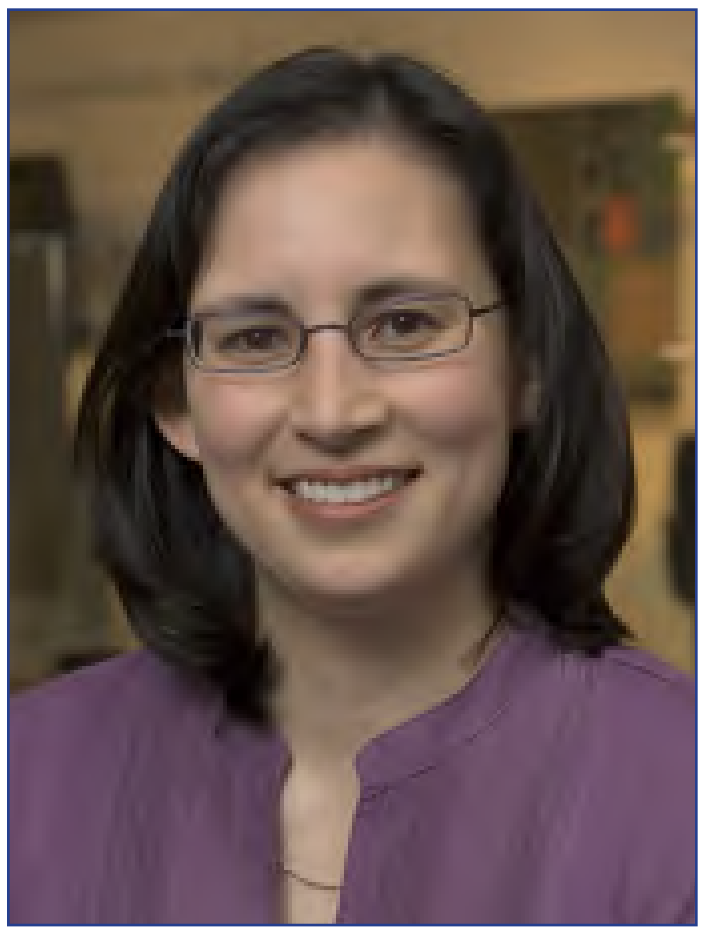

***Allison Okamura***

Surgical Assistants Thrust

(Thrust 1)

***Allison M. Okamura*** received the BS degree from the University of California at Berkeley in 1994, and the MS and PhD degrees from Stanford University in 1996 and 2000, respectively, all in mechanical engineering. She is currently a professor of mechanical engineering and the Decker Faculty Scholar at Johns Hopkins University. She is associate director of the Laboratory for Computational Sensing and Robotics and a thrust leader of the NSF Engineering Research Center for Computer-Integrated Surgical Systems and Technology. She is an associate editor of the IEEE Transactions on Haptics, and a senior member of the IEEE. Her awards include the 2009 IEEE Technical Committee on Haptics Early Career Award, the 2005 IEEE Robotics and Automation Society Early Academic Career Award, the 2004 US NSF CAREER Award, the 2004 JHU George E. Owen Teaching Award, and the 2003 JHU Diversity Recognition Award. Her research interests are haptics, teleoperation, medical robotics, virtual environments and simulators, prosthetics, rehabilitation engineering, and engineering education.

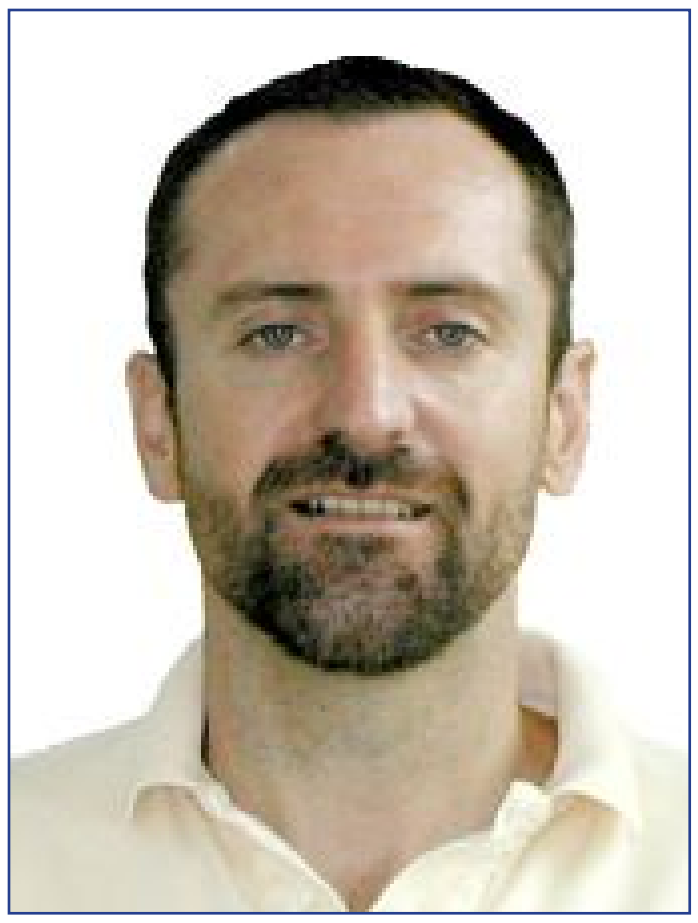

***Gabor Fichtinger***
Surgical CAD/CAM
(Thrust 2)

***Gabor Fichtinger*** received Ph.D. in Computer Science from the Technical University of Budapest, Hungary, in 1990, and received postdoctoral training at the University of Texas at Austin (1990-1992) in high performance computing and biomedical visualization. Before joining JHU in 1999, he architected radiotherapy and neurosurgery navigation systems at George Washington University and he worked in the medical device industry and developed image-guided radiation therapy systems. Dr. Fichtinger focuses his research on image-guided therapy and surgery. His specialty is needle-based percutaneous (through the skin) surgery, with a strong focus on interventional oncology. Now an Associate Professor of Computer Science at Queen's University, he holds multiple cross and adjunct appointments at Queen's and JHU and serves on the Board of Directors of the International Society of Medical Image Computing and Computer Assisted Interventions (MICCAI) and the editorial board of the Journal of Medical Image Analysis.

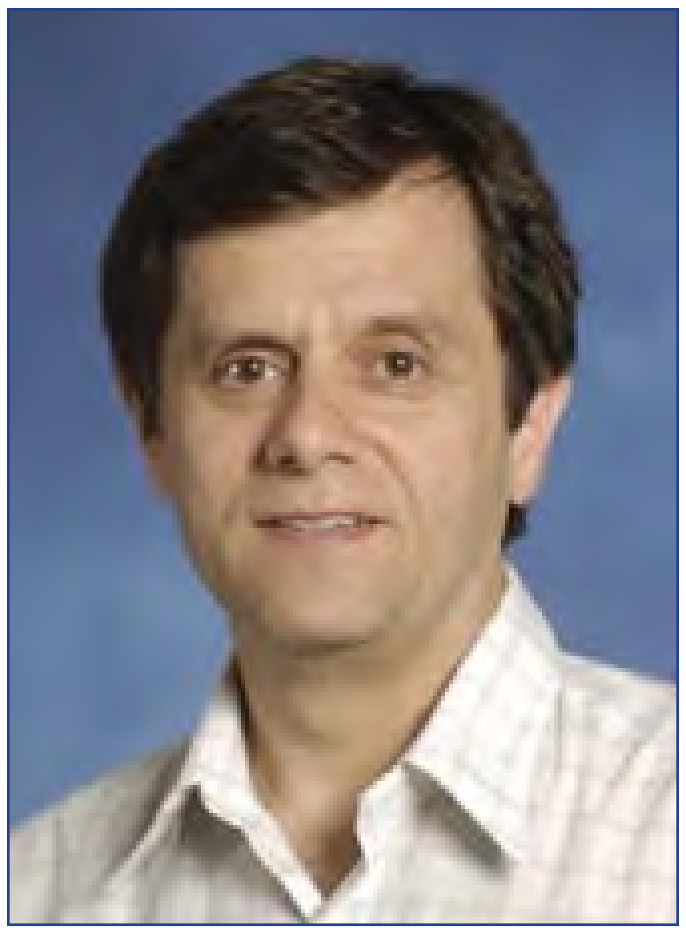

***Peter Kazanzides***
Infrastructure Thrust
(Thrust 0)

***Peter Kazanzides*** received his Ph.D. in electrical engineering from Brown University in 1988 and subsequently began working in the area of surgical robotics, with Dr. Russell Taylor, as a postdoctoral researcher at the IBM T.J. Watson Research Center. Dr. Kazanzides then co-founded Integrated Surgical Systems (ISS) in November 1990 to commercialize the robotic hip replacement research performed at IBM and the University of California, Davis. As Director of Robotics and Software, he was responsible for the design, implementation, validation and support of the ROBODOC® System, which has been used for more than 20,000 hip and knee replacement surgeries. Dr. Kazanzides joined the CISST ERC in December 2002, where he leads the engineering infrastructure (Thrust 0). He currently holds an appointment as an Associate Research Professor of Computer Science at JHU.

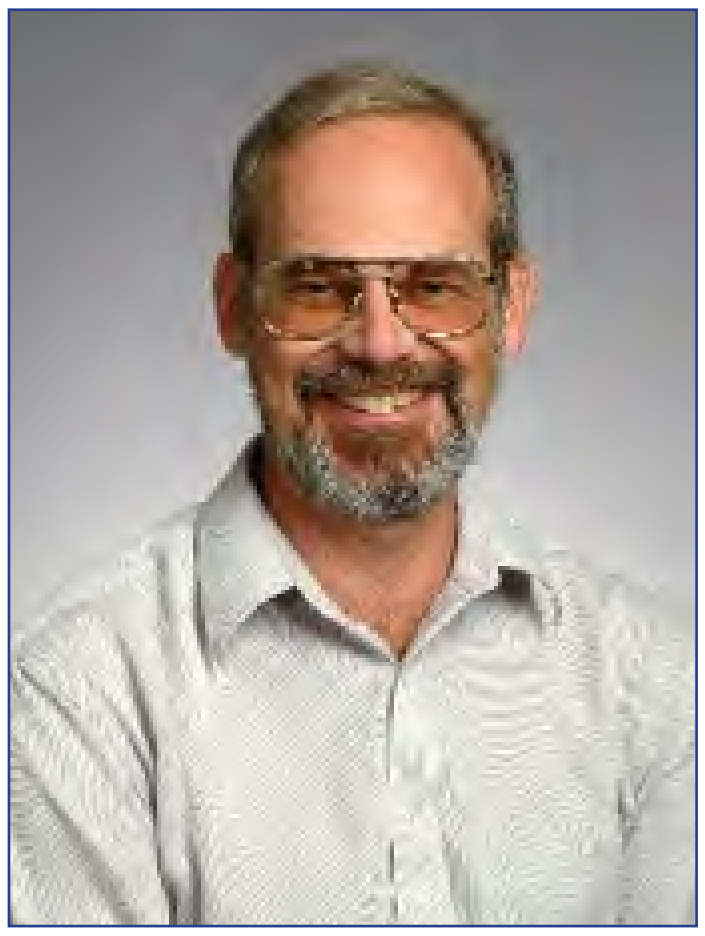

***Eric Grimson***

Massachusetts Institute of Technology

***Eric Grimson*** is the Bernard Gordon Professor of Medical Engineering in the Department of Electrical Engineering and Computer Science, at the Massachusetts Institute of Technology. He is also the Head of the EECS Department, and is a Lecturer on Radiology at Harvard Medical School. Professor Grimson received his B.Sc. from University of Regina and his Ph.D. from MIT. He has over thirty years experience as a researcher in computer vision and medical image analysis, having published two books, 200 refereed articles, and six patents in this area, and has supervised more than 45 doctoral theses. In recognition of these contributions, he was elected a Fellow of the American Association of Artificial Intelligence, and a Fellow of the IEEE. He has also won the Bose Award for Excellence in Teaching at MIT.

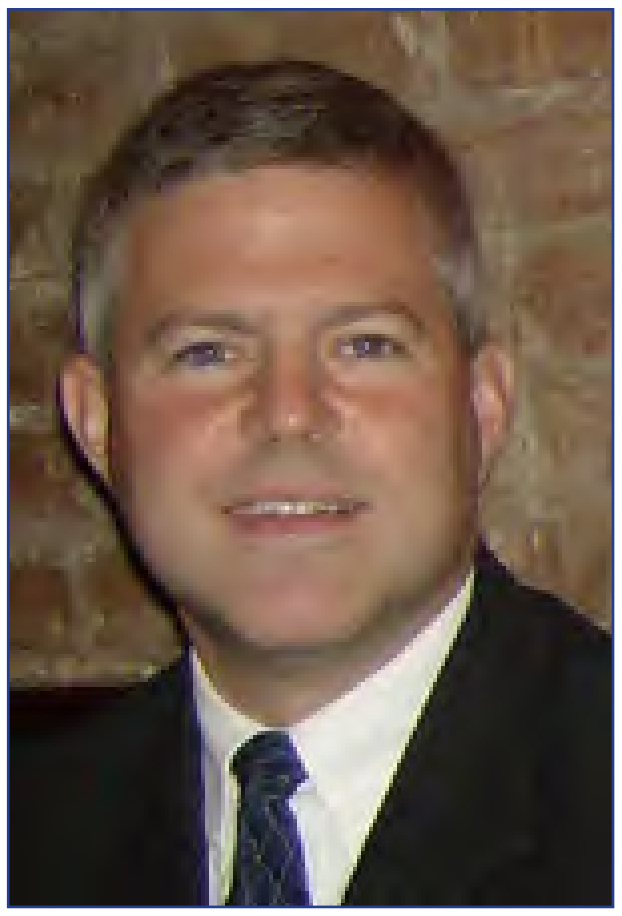

*Cameron Riviere*

Carnegie Mellon University

***Cameron Riviere*** received the Ph.D. in mechanical engineering from The Johns Hopkins University in 1995, and joined the Robotics Institute at Carnegie Mellon University the same year. He is presently Associate Research Professor of Robotics and Biomedical Engineering, and the Director of the Medical Instrumentation Laboratory. Since 1998 he has also been Adjunct Professor in the Department of Rehabilitation Science and Technology at the University of Pittsburgh. His research interests include medical robotics, control systems, signal processing, learning algorithms, and biomedical applications of human-machine interfaces.

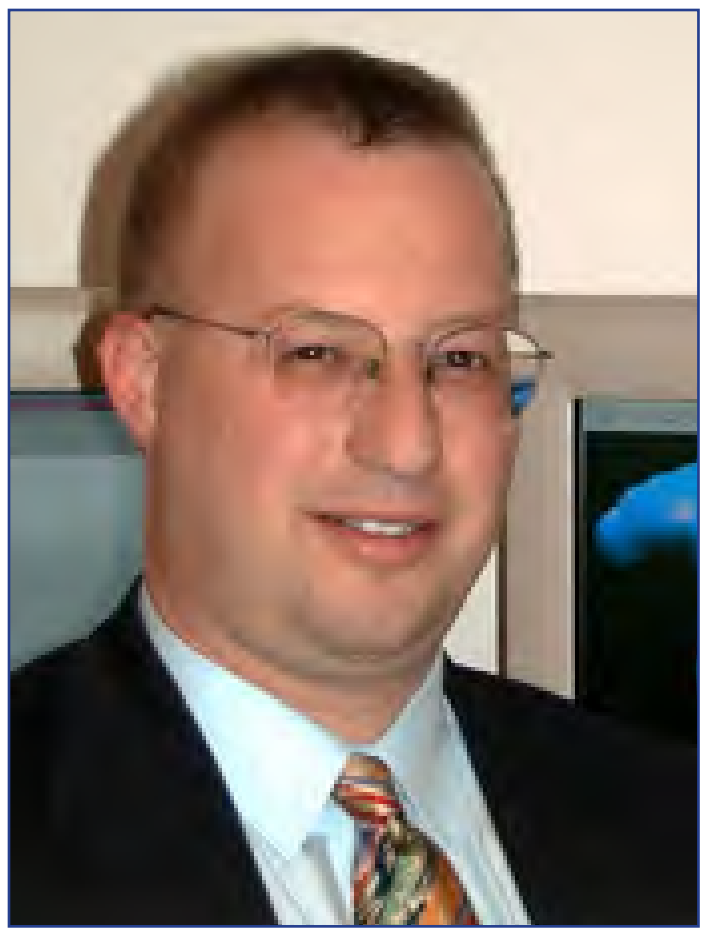

*Ron Kikinis*

Brigham and Women's Hospital

***Dr. Ron Kikinis'*** scientific interest is in image processing algorithms and their use for extracting relevant information from medical imaging data. His activities include technological research (segmentation, registration, visualization, high performance computing), software system development (most recently the 3D Slicer software package), and biomedical research in a variety of biomedical specialties. The majority of his research is interdisciplinary in nature and is conducted by multidisciplinary teams.

| Faculty Awards include: | |
|---|---|
| Allison Okamura | Whitaker Foundation Biomedical Engineering Grant; NSF CAREER Award; George E. Owen Teaching Award; Early Academic Career Award; Decker Faculty Scholar, JHU; Diversity Recognition Award |
| Eric Grimson | Fellow of the American Association of Artificial Intelligence; Bose Award for Excellence in Teaching, School of Engineering, MIT; IEEE Fellow |
| Gabor Fichtinger | Capers and Marion McDonald Award for Excellence in Mentoring and Advising; Teaching and Learning Enhancement Fund Award, Queen's University; William R. Kenan, JR Fund Award for Innovative Undergraduate Teaching |
| Greg Hager | IEEE Fellow; Yale Junior Faculty Fellowship; Fulbright Junior Research Award |
| Jerry Prince | Maryland's Outstanding Young Engineer; JHU Certificate of Recognition; IEEE Fellow; NSF Presidential Faculty Fellow |
| Louis Whitcomb | JHU Student Council Award for Excellence in Teaching; George E. Owen Teaching Award; Appointed Senior Member, IEEE; University Alumni Excellence in Teaching Award |
| Noah Cowan | William H. Huggins Excellence in Teaching Award, JHU |
| Polina Golland | Distinguished Alumnus (1964) Career Development Chair; NSF CAREER Award |
| Ralph Etienne-Cummings | Science Spectrum Trailblazer Award for Top Minorities in Science; Diversity Leadership Council Diversity Award, JHU; National Academics of Science Kavli Frontiers in Science Fellow; NSF CAREER Award; Young Investigators Program Award, Office of Naval Research; Visiting African Fellowship Award, University of Cape Town; Fulbright Fellowship Award to South Africa |
| Russ Taylor | Maurice E. Müller Award for Excellence in Computer-Assisted Surgery; IEEE Third Millenium Medal; SPIE, Medical Imaging; IEEE Robotics and Automation Society Pioneer Award; IEEE Fellow |

CISST-ERC faculty have traveled world-wide to take part in hundreds of workshops, invited talks, symposiums and conferences, many of which they helped organize.

The CISST ERC has 60 faculty and 49 clinicians.

# *Awards*

CISST-ERC faculty have served as editors and on numerous editorial boards, including:

- IEEE Trans. on Robotics and Automation
- IEEE Sensors Journal
- IEEE Trans. Biomedical Circuits and Systems
- IEEE Transactions on Image Processing
- IEEE Transactions on Medical Imaging
- IEEE Engineering in Medicine and Biology Society
- IEEE Transactions on Haptics
- Medical Image Analysis Journal
- Journal of Computer Aided Surgery
- IEEE Sensors Journal, Special Issue on Array Processing in VLSI
- IEEE Trans. Biomedical Circuits and Systems, Special Issue on BioCAS 2007
- Kluwer's AICSP Journal, Special Issue on Smart Sensors
- IEEE PAMI
- IJCV
- International Journal of Robotics Research
- NeuroImage
- INE The Neuromorphic Engineer

Student Awards include:

- SPIE Fellowship, SPIE
- Siemens Corporate Research Fellowship
- NDSEG Fellowship
- more than 2 Link Foundation Fellowship
- National Outstanding Student Award, Ministry of Science and Technology, Thailand
- more than 9 NSF Research Fellowships
- more than 3 U.S. Department of Defense Predoctoral Traineeship Award

Faculty and Sudent Best Poster Awards and Best Paper Awards include, among others:

- 2 Finalist, Best Paper Awards at the IEEE/RSJ International Conference on Intelligent Robots and Systems
- 2003 Best Paper Award, EURASIP Journal of Applied Signal Processing
- 3 Best Student Paper Awards, SPIE
- 4 Best Poster Awards and 1 honorable mention at SPIE
- Best Paper Award, CAOS
- Best paper award, IEEE Computer Vision for Biomedical Images
- Best Paper Award, IEEE Signal Processing Society
- Best Paper Award, MICCAI
- Best Paper Finalist, IEEE International Conference on Electronics, Circuits, and Systems
- Best Paper Honorable Mention, North East BioEngineering Conference
- Best Poster Award at Medicine Meets Virtual Reality (MMVR 16): Misra
- Best Poster Award, Honorable Mention, IMPI
- Best Student Paper Finalist, International Symposium of Circuits and Systems
- Best Young Investigator Paper Award, MICCAI
- Cum Laude Poster Award, Sixth Interventional MRI Symposium, Leipzig
- 2 Medicine Meets Virtual Reality Best Poster Award
- Young Author Best Paper Award in the Image and Multidimensional Signal Processing Area

# Outcomes & Impacts

CISST was founded on the premise that engineering and medicine must be equal partners in the development of computer-integrated surgical systems. Throughout the life of our ERC, we have learned how to create engineering teams of faculty, clinicians, and students and how to make those teams cohesive and self-sustaining research groups, such as the percutaneous prostate group, the eye microsurgery group, and the interventional ultrasound group.

The CISST ERC has developed a web of collaborations that extends throughout the engineering and medical campuses, unites several major universities and companies. Morgan State joined the center in 2001 and has been doing work on image registration and infrastructure. The University of Pennsylvania has played a large role in registration for percutaneous therapy and human-machine systems. Columbia is now a partner for the development of the snake robot. Harvard has been active in haptics for teleoperation. We have also begun to work with two international partners: Queens University of Canada, and the Technical University of Munich, in Germany.

Our industrial collaboration has led to a series of patents and clinical systems. We have developed strong working relationships with Burdette Medical Systems, Siemens Inc., and Intuitive Surgical which have led to several patents and a variety of collaborative research and development projects.

As our research has matured toward clinical use, it has become clear that our clinical collaborators want and need to take a larger leadership role in future research directions. As a result, in 2006, the Committee for Integrating Imaging, Intervention, and Informatics in Medicine (I4M) was formed. This committee provides a forum that brings together the ERC leadership with the chairs of several major departments in the medical school, including Radiology, Surgery, Interventional Radiology, and Biomedical Engineering. The WSE and the JHU Medical School have both made substantial time and financial commitments to this initiative.

The ERC's leadership in the field is exemplified by excellence in research, presentations at major conferences and industrial forums, and prizes and awards (see pages 22-23). Our director, Dr. Russell Taylor, has been honored with the IEEE Robotics and Automation Society Pioneer Award, the Maurice E Mueller Award for Excellence in Computer-Integrated Surgery, and the IEEE Third Millenium Medal. Dr. Allison Okamura, one of our ERC faculty hires, has been awarded the NSF CAREER award, the Robotics and Automation Society Early Career award, and numerous awards for teaching, service, and diversity. These are but two examples of an outstanding record of awards and honors by the faculty of the ERC.

ERC faculty have been extremely active in furthering the field of Computer-Integrated Surgery as a whole. We have organized tutorials on CIS at several major conferences, and many faculty hold one or more editorial duties at major journals. Over the life of the center, faculty and students have delivered hundreds of speeches and lectures to a diverse set of audiences around the world.

Most recently, the leadership of the ERC has worked to organize and host a tutorial on Medical robotics,

continuing a series of similar tutorials given worldwide. This week of lectures featured many ERC investigators and collaborators, and allowed our group to place our vision of Computer-Integrated Surgery in front of a large and influential set of international guests.

The ERC has been instrumental in the emergence of CIS specifically and robotics in general as major forces at JHU. The center started with three full-time faculty in 1996 and ERC funding resulted in two additional tenure-line faculty hires. JHU recognized that a critical mass was now established in these fields and supported two tenure line faculty hires in related areas. In addition, the ERC is home to three full-time research faculty as well as several staff engineers and postdoctoral scholars, making it one of the largest and most successful research groups in the WSE.

Our growth and success led us to form the Laboratory for Computational Sensing and Robotics (LCSR) in 2007. The university administration realized the advantages of uniting the broader robotics program under one roof and as a result designed a 15,000 sq. ft. laboratory facility in the new CSEB to specifically house LCSR. A Mock Operating Room, generously donated by the Swirnow family, allows faculty, students, and clinicians to come together for research. CSEB, occupied in Summer 2007, now houses nearly 100 personnel.JHU's contribution makes it clear that the inter-relationship between engineering and medicine will play a large role in the future of the university. The I4M committee and development of the collaboration with the Fraunhofer Society are signs of future growth and commitment.

The CISST ERC's Education, Outreach and Diversity Committee's (EODC) activities have had significant impact both within the JHU community and in the greater global community. Computer integrated surgery (CIS) content curricula have augmented degrees in various engineering departments. The cross disciplinary nature of the ERC has helped develop CIS and Robotics undergraduate minors in several departments, as well as unique courses such as Surgery for Engineering, which brings engineering students to the operating room, requiring them to perform basic surgical operations on animal cadavers. We plan to continue developing our CIS curriculum beyond graduation.

We have programs which impact all educational levels. Some examples include our summer robotics camp for middle school students, and the Systems Robotics Challenge, along with the Research Experience for Teachers (RET). Our major impact at the undergraduate level has been through our Research Experience for Undergraduate programs.

Our programs have diversified the composition of ERC Leadership, faculty and students. When we initiated our EODC in 2003, there were very few women and minorities in the ERC. Since then we have seen a dramatic increase in the number of women in our graduate programs, and we have included women and minorities in our leadership. They have received university- and nation-wide honors for their research, teaching and diversity activities. Our Strategic Plan to Promote Recruitment/Retention, Community and Achievement has received commendation both within JHU and outside. The EODC is now a permanent component of the LCSR and will continue after the ERC's graduation.

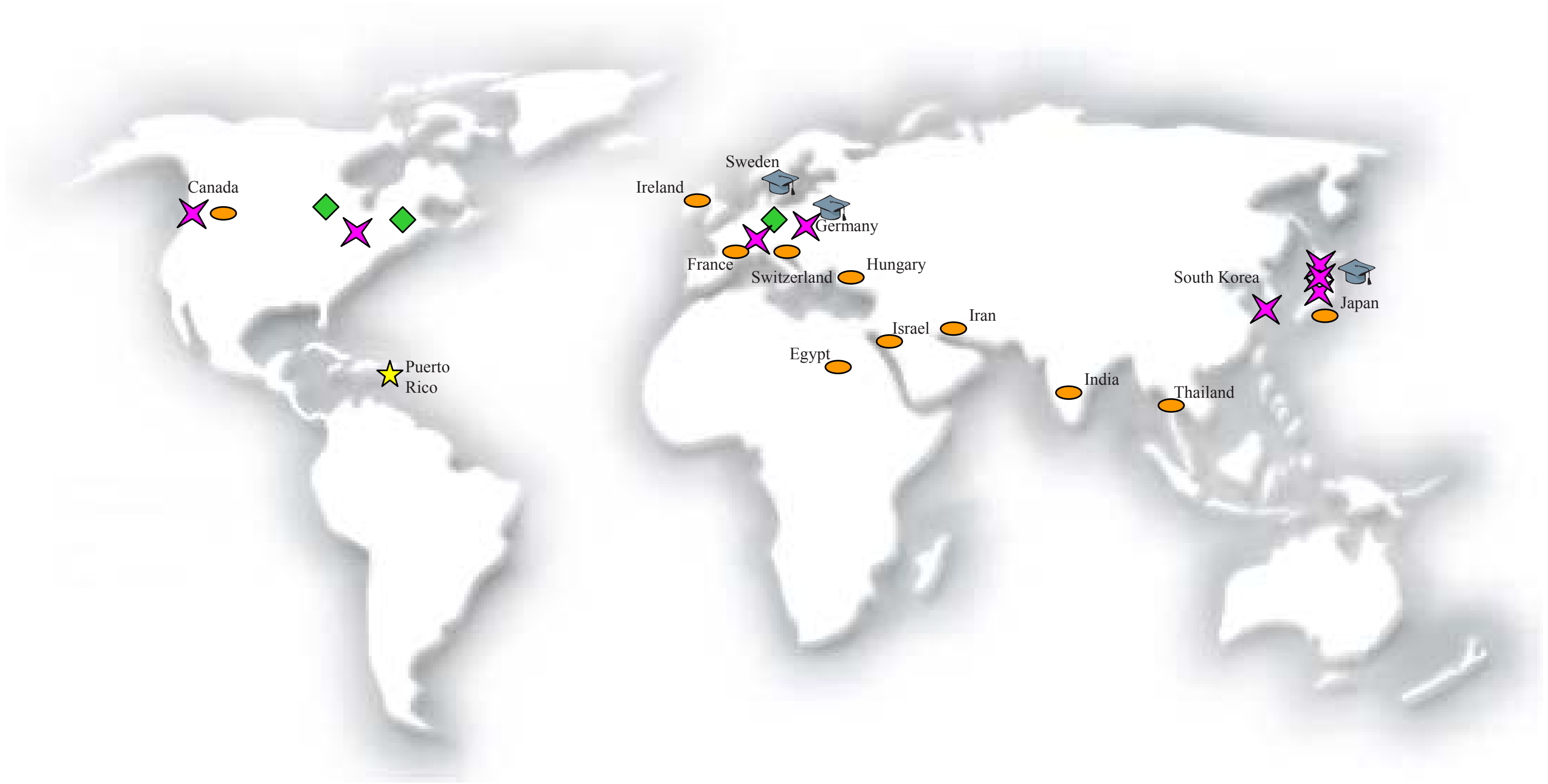

Lead Institution

Core Partners

Federal Government

K-12

LSAMP & Non ERC REU Students

Industry

Outreach

Private Foundations

Other NSF Diversity Program Awardees

Former ERC Students

# *Impact World Wide*

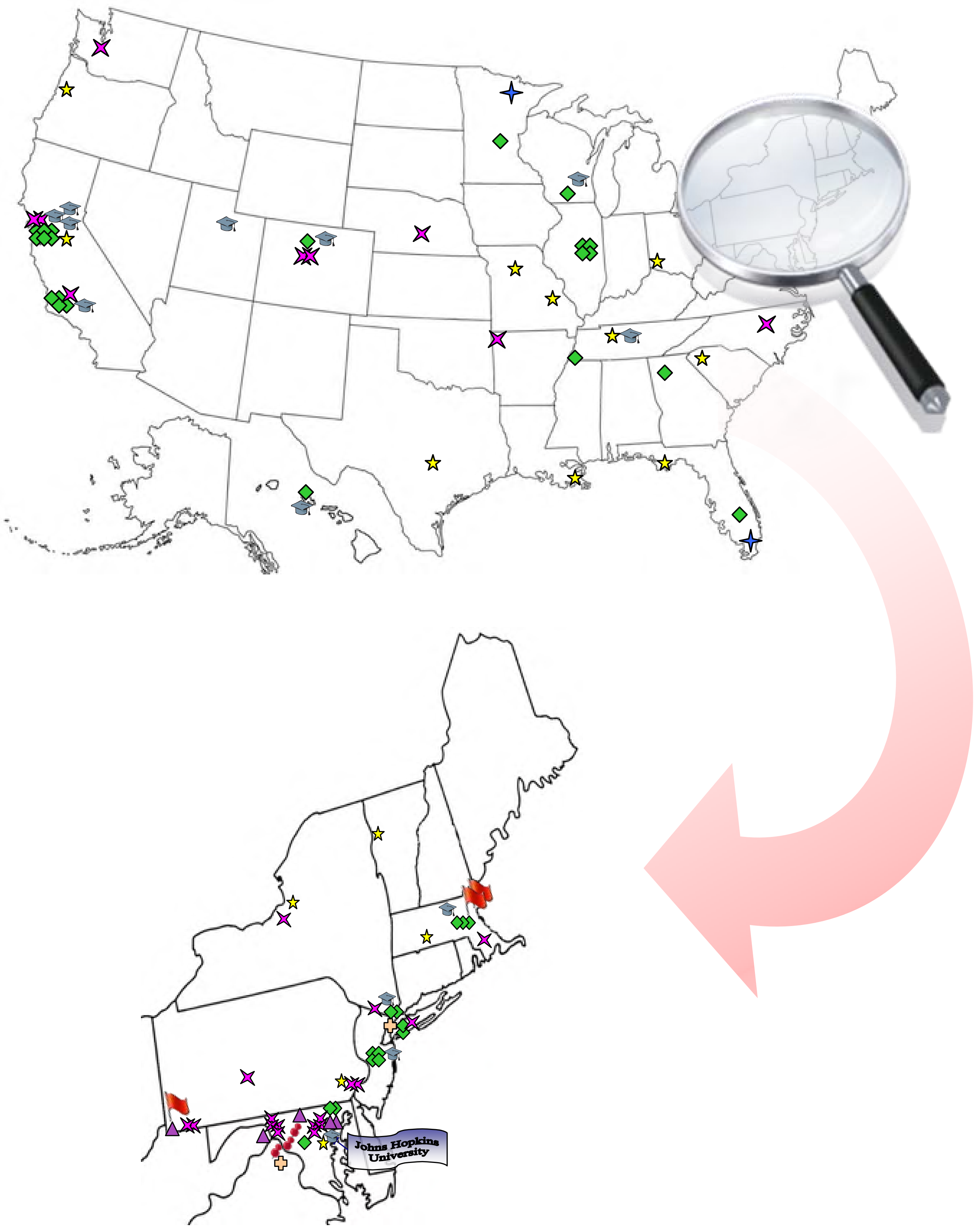
Johns Hopkins University

# Research Impact

Our research strategy has focused on two testbed-oriented research thrusts: Surgical Assistance, and Surgical CAD-CAM. In addition, we later added a third thrust, Infrastructure, that is designed to further develop our self sustaining core CISST systems infrastructure.

In developing the strategy for each thrust, we first identified clinical requirements and created a system evolution plan. We then started building technology and basic science blocks, often seed-funded from the ERC core grant. Some core resources were also directed toward prototype system development. As projects matured, grants (typically from NIH and DoD) were acquired to fund clinical trials. In addition to their basic science aims, these grants usually provided some resources to develop our technology and basic science blocks, thereby furthering our infrastructure development and helping to connect research to practice.

The impact of the CISST research strategy can be easily measured by the data contained in our indicator tables: we have produced roughly 700 publications, graduated over 230 students, and generated more than 10 million dollars of external research funding. What is less obvious is the evolution of our thinking about CIS systems and the implications that had for our research program and for the field as a whole.

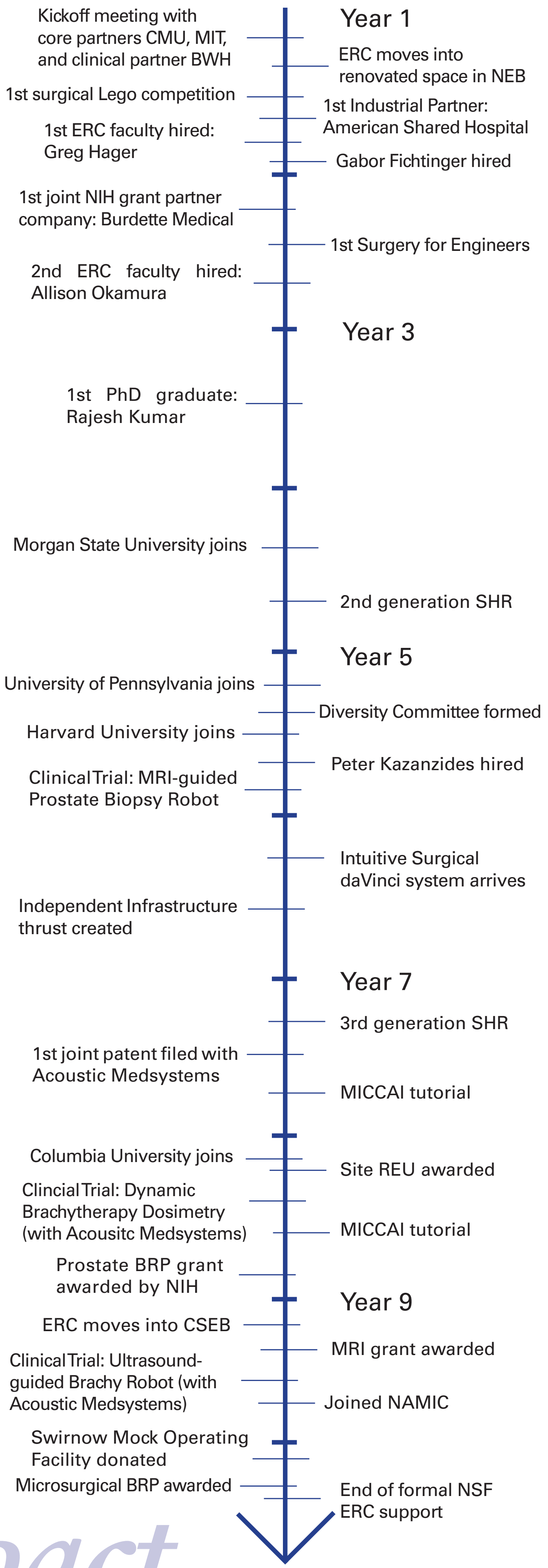

*Surgical Assistance Thrust (Thrust 1)*

The most important accomplishment of the Surgical Assistance thrust has been to demonstrate that surgical workstations, which integrate robotic devices with imaging, information processing, and advanced visualization and human-machine interfaces, comprise powerful tools that are more than the sum of their individual components. These workstations allow surgeons to perform today's procedures with super-human capabilities, but also are new tools for advancing the state of the art in surgery and, more broadly, in biomedical research as a whole.

One of our principle application areas has been retinal microsurgery, in which surgeons must perform tasks at the fringe of human sensing and manipulation capabilities. With ERC support we have developed two microsurgical manipulation devices that enhance physical precision: the JHU "steady-hand" robot and the Micron system. These systems have evolved into complete micro-surgical assistant workstations. They embody a family of guidance principles, called virtual fixtures that enhance precision and safety. The basic science of virtual fixtures lies in the development of passive control methods that limit the robot's movement to restricted regions and/or influence its movement along desired paths. Our workstations also include new visualization tools that combine preoperative images with the image from the surgical microscope (see fig. 5).

Throughout the life of this project we made several fundamental contributions in mechatronic design, control, and visualization. Recent

## Thrust 1: Surgical Assistants

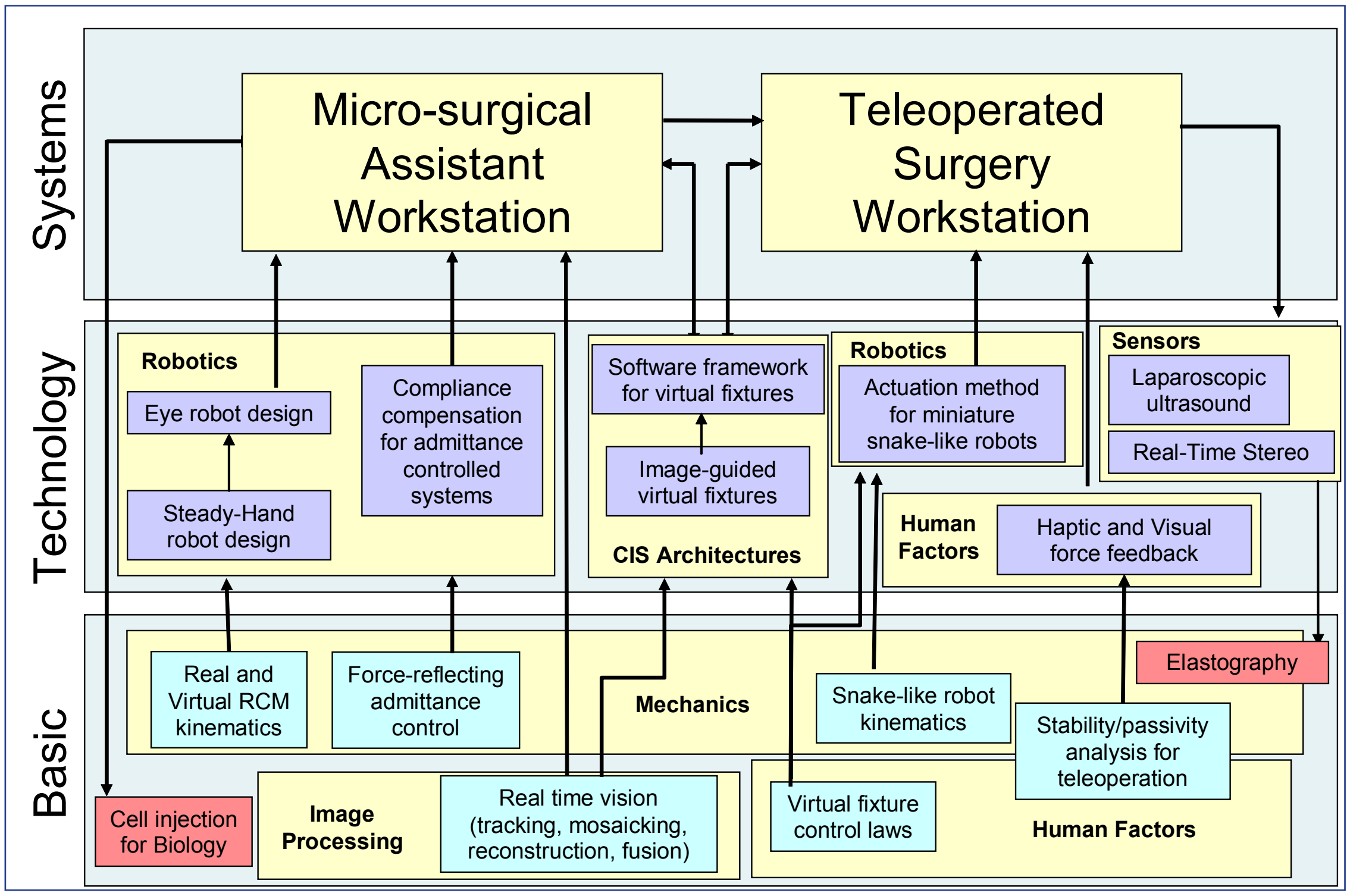

Fig. 5: 3 plane chart for Thrust 1

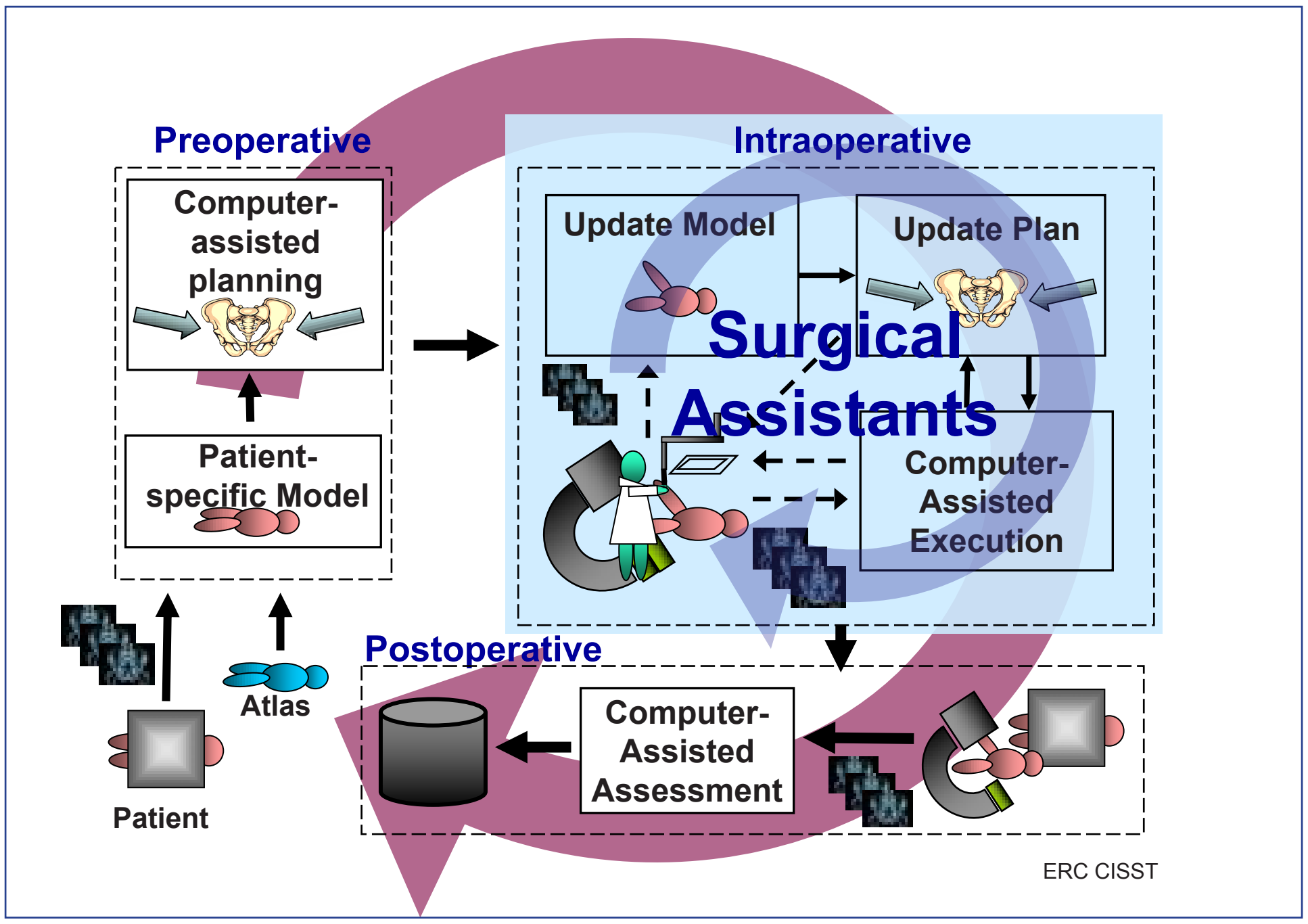

Fig. 6

generations of these systems have been optimized for surgical ergonomics and workflow though a series of human use studies. They now form the core of a newly awarded NIH Biomedical Research Partnership grant in the LCSR at JHU.

Our second major testbed, the teleoperated surgical workstation allows us to create and test systems for minimally invasive surgeries which require high dexterity and visualization at locations deep within the body. An integral part of our collaboration with Intuitive Surgical Inc., this testbed led to advances in basic robotic technology, such as the JHU snake robot, and advances in information registration and visualization and human-machine systems.

Like the microsurgery testbed, further development of virtual fixtures was a key element in creating these systems. The development of multiple workstations using virtual fixtures facilitated, in turn, a set of basic science investigations on human-machine interaction with specific devices and controller. The results of these and other studies drive the design of new devices and algorithms, with the later generations also considering the ergonomics of human-machine interaction and the constraints of surgical workflow.

In order to provide meaningful virtual fixtures relative to patient anatomy, we developed new algorithms for processing video, ultrasound, CT, and MRI. Many of these results are now integrated into our SAW software framework. This architecture was the key driver in our further collaboration with Intuitive Surgical, and the development of an international collaboration with the Technical University of Munich.

Some of our workstation technologies like the snake robot are being commercialized through licensing arrangements. This testbed has major funding from NIH to study haptic feedback and to develop an ultrasound assistant for the da Vinci system, NSF to study the "language of surgery," and DoD to develop information overlay technologies.

### *Surgical CAD/CAM Thrust (Thrust 2)*

Our work in percutaneous therapy experienced a similar evolution of its research program. This thrust particularly excelled at developing several basic principles which have been engineered into systems that are now under clinical evaluation or are being commercialized with our industrial partners.

Percutaneous therapy's overarching mandate is to develop and clinically deploy Surgical CAD/CAM systems that share common architecture, technology and basic science foundation. The most important accomplishments of Surgical CAD/CAM are proving it is a practical and useful paradigm and that a family of clinical systems specialized for specific clinical use can be derived from generic architecture and technological building blocks.

The MR-guided Prostate Intervention Systems family has grown into a large international clinical-engineering-industry consortium during the ERC funded years. Clinical trials are in progress at BWH, JHU, NIH, and Princess Margaret Hospital (Toronto). Industry partners include Acoustic Medsystems, Siemens, and Sentinel Medical, while core engineering faculty spans over JHU, UPenn, BWH, and Queen's. The consortium is funded by an extensive portfolio of technology development grants, several clinical grants and most importantly by a Biomedical Research Partnership grant. The ERC's contribution has been seminal and widely acknowledged in MR-compatible mechatronics, MR image segmentation and registration and statistical biopsy atlases.

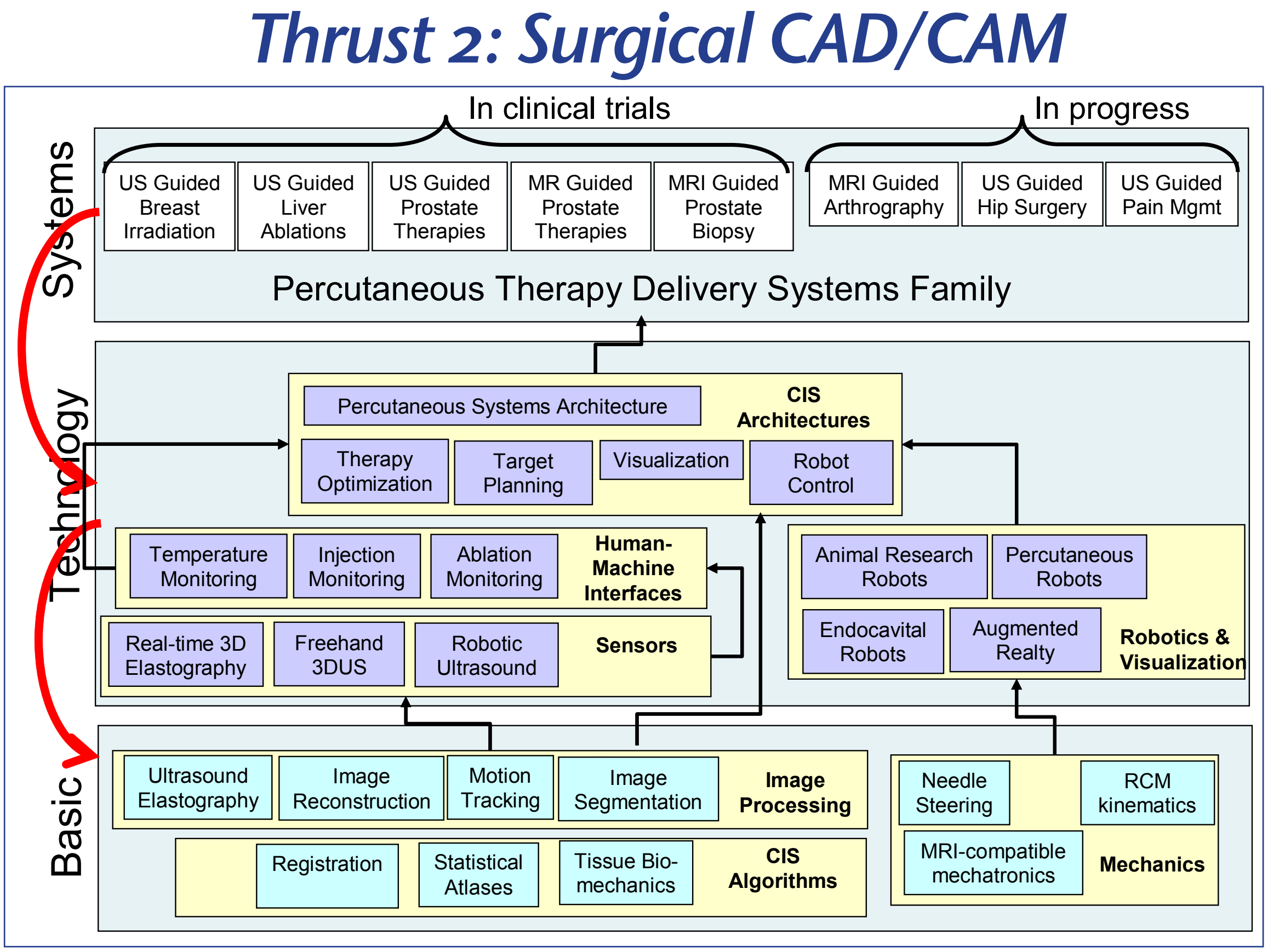

Fig. 7: 3 Plane chart examples for Thrust 2

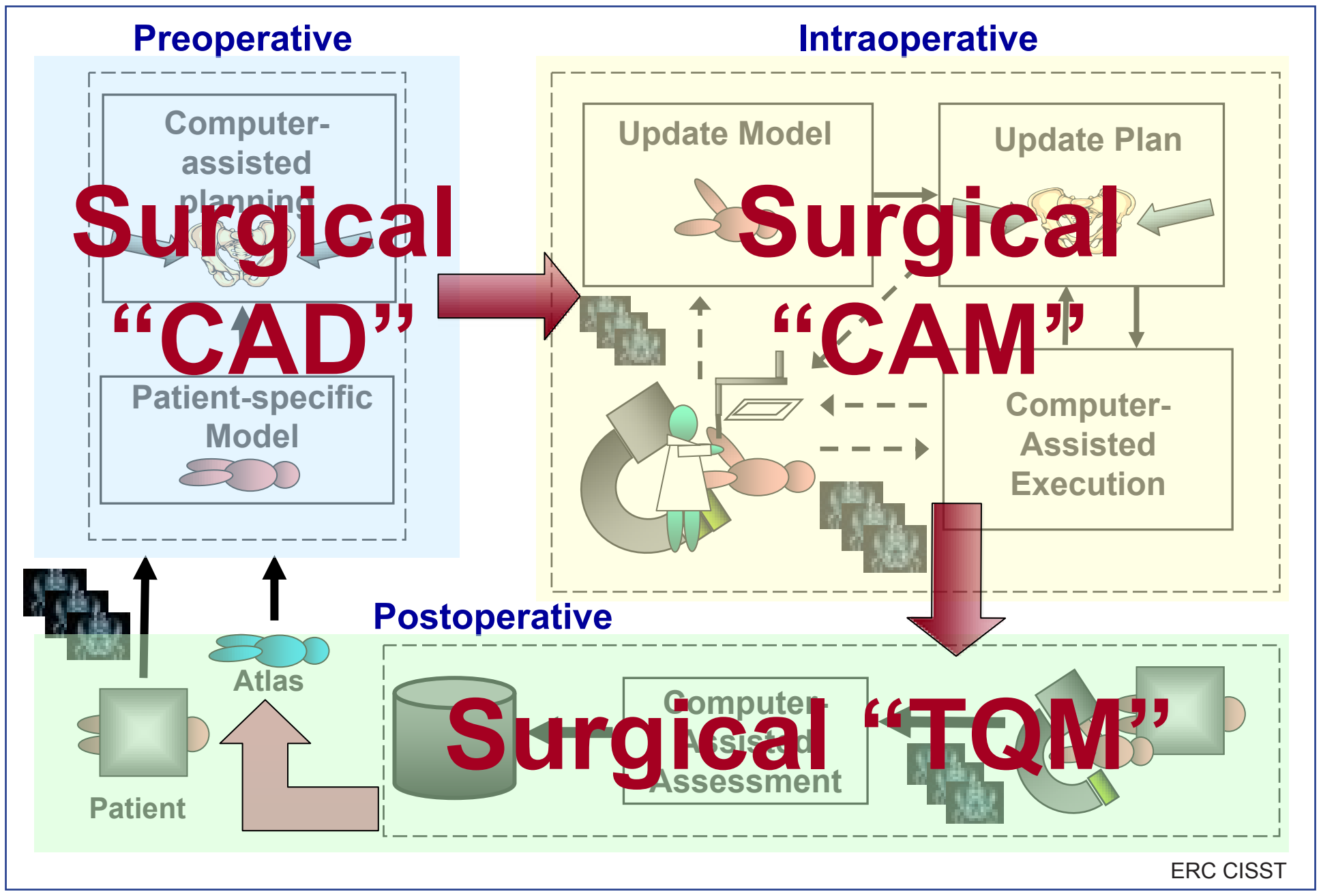

Fig. 8

Ultrasound-guided Prostate Interventions Systems mark our ERC's first industrial partnership with Burdette Medical, predecessor of Acoustic Medsystems. The major aim was to achieve optimal prostate cancer brachytherapy under transrectal ultrasound guidance. We brought two clinical systems to trials, featuring robotic augmentation and quantitative fluoroscopic dosimetry that opened new lines of clinical research in radiation oncology. Our efforts were funded by six industry-academia partnership grants and a DoD clinical trial grant. Most notable basic science contributions included ultrasound segmentation tomosynthesis, fluoroscopic implant reconstruction, ultrasound-fluoroscopy registration, and RCM robots. This clinical testbed also absorbed output from the NIH-funded needle steering and tissue biomechanics program.

In response to an early NSF site visit suggestion, we created our ultrasound-guided breast irradiation and liver ablation clinical application testbeds. Today these systems are funded steadily from external resources combining clinical trial grants, industry contributions, NIH research grants, and fellowship grants. The evolution of these testbeds marks the career of our junior radiology faculty member, Emad Boctor, who joined the ERC as a graduate student in 2001. Emad was the main driver behind our interventional ultrasound program and today is the leader of this portfolio. Ultrasound elastography, ultrasound calibration and speckle characterization are the program's fundamental science contributions to ultrasound imaging.

An extensive international program is emerging in musculoskeletal interventions, championed at Queen's and financed chiefly from Canadian grants. These efforts are based on the Image Overlay technology which besides entering NIH-funded clinical trials at JHU, have been adopted in the Ontario Consortium for Adaptive Cancer Interventions. The ERC's contribution to the ultrasound-guided hip surgery and pain management clinical systems also include statistical anatomical atlases, ultrasound motion tracking, and speckle recognition.

*Infrastructure Thrust (Thrust 0)*

A broad family of related surgical CAD/CAM and surgical assistant systems requires a strong and open systems infrastructure of versatile components and subsystems. Engineering research results must be embodied in robust, modular and reusable software and hardware components and subsystems. Standard interfaces are essential for collaboration within the ERC and between the ERC and other researchers and industrial companies. The Research Infrastructure thrust provides the resources and a common control point for architecture definition, software library maintenance, shared laboratory infrastructure, and other elements. It also provides training and integration of new students into the technical activities of the ERC, since these students are mentored by experienced staff engineers

The key technical accomplishments of the research infrastructure were to define system

*Exciting Research Opportunities*

Carol Reily has been a member of the ERC team since 2004, when she began her master's studies in Associate Professor Allison Okamura's Haptics Exploration Lab. She worked on a sensory substitution project in which surgeons relied on visual feedback – on-screen color variations – to tell them how much force they were applying as they tied knots robotically.

Reily's Computer Science PhD project with Professor Greg Hager also involves surgical performance skills, in this case mathematically modeling surgical motions as performed with a da Vinci® robot-assisted minimally-invasive surgical system. The aim is to provide quantitative measures for evaluating and training surgeons.

Her involvement with the ERC has enabled Reily to do research at Stanford University, co-teach an undergraduate course in haptics, win an NSF fellowship, mentor other students, become a leader in ERC student groups –and decide on a career in research and teaching. "The ERC has helped me grow as a student, a researcher and a contributor to society in a way that I could not have imagined," she says. "I feel fortunate to be a part of a research center where I've encountered such diverse people, cutting-edge equipment, and community outreach opportunities."

# Thrust 0: Infrastructure

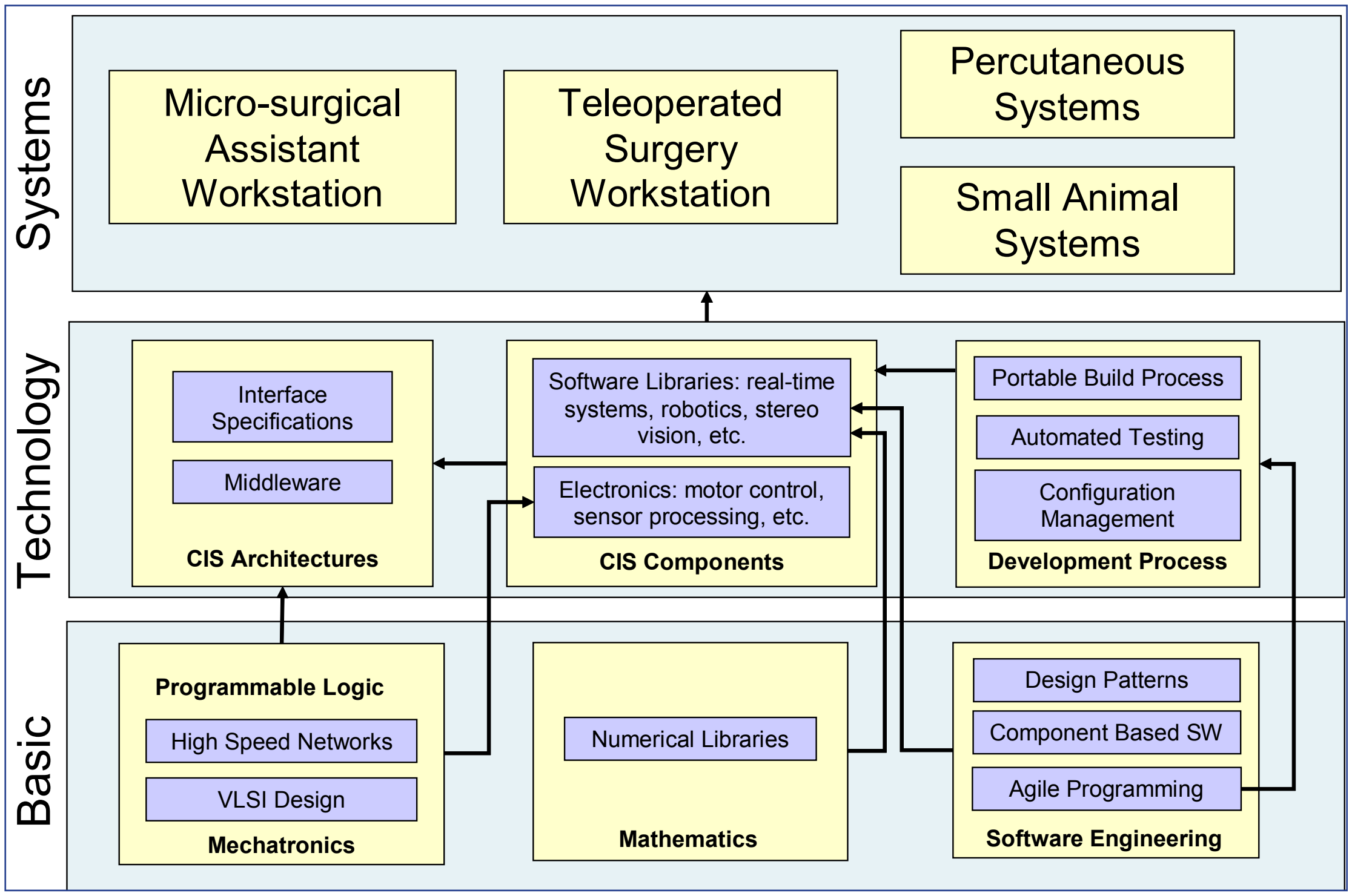

Fig. 9: 3 plane chart examples for Thrust 0

architectures, provide documented and tested components, and establish the development process and tools. The architectural design included the specification of standard interfaces to all major components. For example, there are many technologies (e.g., optical, magnetic, mechanical) that can be used to track the position of a surgical instrument, but they all provide the same fundamental capabilities. Defining a generic interface to these devices leads to a modular architecture that enables "plug and play" development of medical devices. The SAW described earlier is a concrete example that defines standard interfaces to robots and other devices and provides a pipeline for real-time image acquisition, processing and display. This architecture is central to much of our research and has garnered interest from other groups, producing a collaboration with the Technical University of Munich.

Thrust 0 has produced software libraries and custom hardware. The CISST software package provides components and tools that focus on the integration of physical devices, such as robots and sensors, with computation and human/machine interaction. This package is available under an open source license, and portions of it can be downloaded directly from our web site (www.cisst.org/cisst). The hardware components include a custom motor controller that provides accurate sensing and control of small, low-current motors, such as those employed by the daVinci and the JHU Snake. The most recent version uses Firewire to provide a high-speed interface between the control computer (PC) and the hardware. This enables researchers to use a familiar development environment, rather than having to learn the idiosyncrasies of embedded system design.

The third key accomplishment was the creation of a development process and the adoption of tools to help manage the process. This process includes documentation, programming standards, change control, test suites, bug tracking and an audit trail. Although a formal software development process is generally not necessary for a specific research project, it is appropriate when the goal is to create a core development platform that can support many research projects and can ultimately be used in clinical trials.

The Research Infrastructure Thrust has been heavily involved in and benefited from diversity initiatives. Four years ago, an ERC Fellowship was created to support an incoming diversity student for their first two years of graduate school. The student works within the Research Infrastructure, thereby obtaining practical skills as well as a broad knowledge of ERC activities. This Thrust also had extensive collaboration with Morgan State University (MSU). The Morgan State team contributed to the distributed systems infrastructure and conceived and led the development of a system for intelligent management and dissemination of research information. These two projects supported seven Morgan State students, producing one M.S. degree and one Ph.D. degree so far, and have involved many others via related class projects. We will continue the collaboration with MSU and already have a pending joint proposal.

*"Having a specific application in mind helps me focus on research that will have an impact on patients in my lifetime – maybe even within the next five years."*

*~Allison Okamura*

Over its ten year lifetime, the CISST ERC has made substantial research accomplishments in numerous areas of computer-integrated surgery. The driving systems vision of the center is to integrate cutting edge technologies into systems that are able to perform existing interventions with unprecedented levels of effectiveness, and to enable new procedures that would not otherwise be possible.

## *Medical Robotics*

The CISST ERC has established world leadership in the development of medical robots for a wide spectrum of applications including microsurgery, minimally invasive surgery, percutaneous therapy, and a variety of non-medical biological applications. In contrast to manufacturing, where a single design paradigm has dominated the field, medical robotics has inspired a wide variety of designs. In every case the system has been developed with the unique aspects of its particular application in mind. As a result, the systems described in this section provide a unique core of design concepts and technologies for the medical field.

### *The JHU Steady-Hand Robot*

In mid-1990s, Dr. Louis Whitcomb, Dr. Russell Taylor, and Dr. Daniel Stoianovici developed the first generation of the JHU Steady-Hand robot, extending earlier work at IBM and JHU by Dr. Taylor. This system embodied two novel design components including a very high-precision, compact and high-mobility mechanism with a mechanically defined remote center of motion mechanism, and new methods for shared "hands on" control of surgical instruments.

The combination of these two components led to a system that is highly effective at reducing hand tremor and increasing motion precision for manipulation tasks at the microscopic level. This has particular relevance to retinal microsurgery, where manipulations at the boundaries of human capabilities are common.

The JHU steady-hand robot has gone through several generations, and each has improved its performance, precision, and ergonomics. During the life of the center, the steady-hand robot has played a role in numerous projects leading to many innovations and activities, including:

1. The development of admittance-control-based force scaling methods.
2. The development of a family of virtual fixture guidance algorithms.
3. The first demonstrations of human-machine shared control for retinal vein cannulation.
4. Human subjects tests of unaided vs. human-robot shared control for medical applications such as retinal cannulation, and non-medical applications such as micro-assembly (NSF SBIR with PicoSys, Inc.).

The most recent generation of this system is now part of a recent NIH Bioengineering Research Partnership (BRP) project to develop a complete eye surgery workstation.

# Research Accomplishments

A key related technology was developed by Dr. Cameron Riviere at CMU. The Micron system also seeks to eliminate hand tremor for microscopic manipulation tasks, and thereby significantly improve precision in microsurgery, particularly in vitreoretinal procedures. The instrument handle incorporates an inertial motion-sensing module. Using these sensors, the velocity of the instrument tip is computed and integrated to obtain tip displacement. The tremulous component of this motion is then estimated, and a micromanipulator built into the instrument tip deflects the tool tip with an equal but opposite motion, compensating the tremor. As with the steady-hand robot, several generations of the instrument have been designed. The most recent provides a range of motion of more than 1 mm in each of the three coordinate directions, more than enough for canceling of physiological tremor during microsurgery.

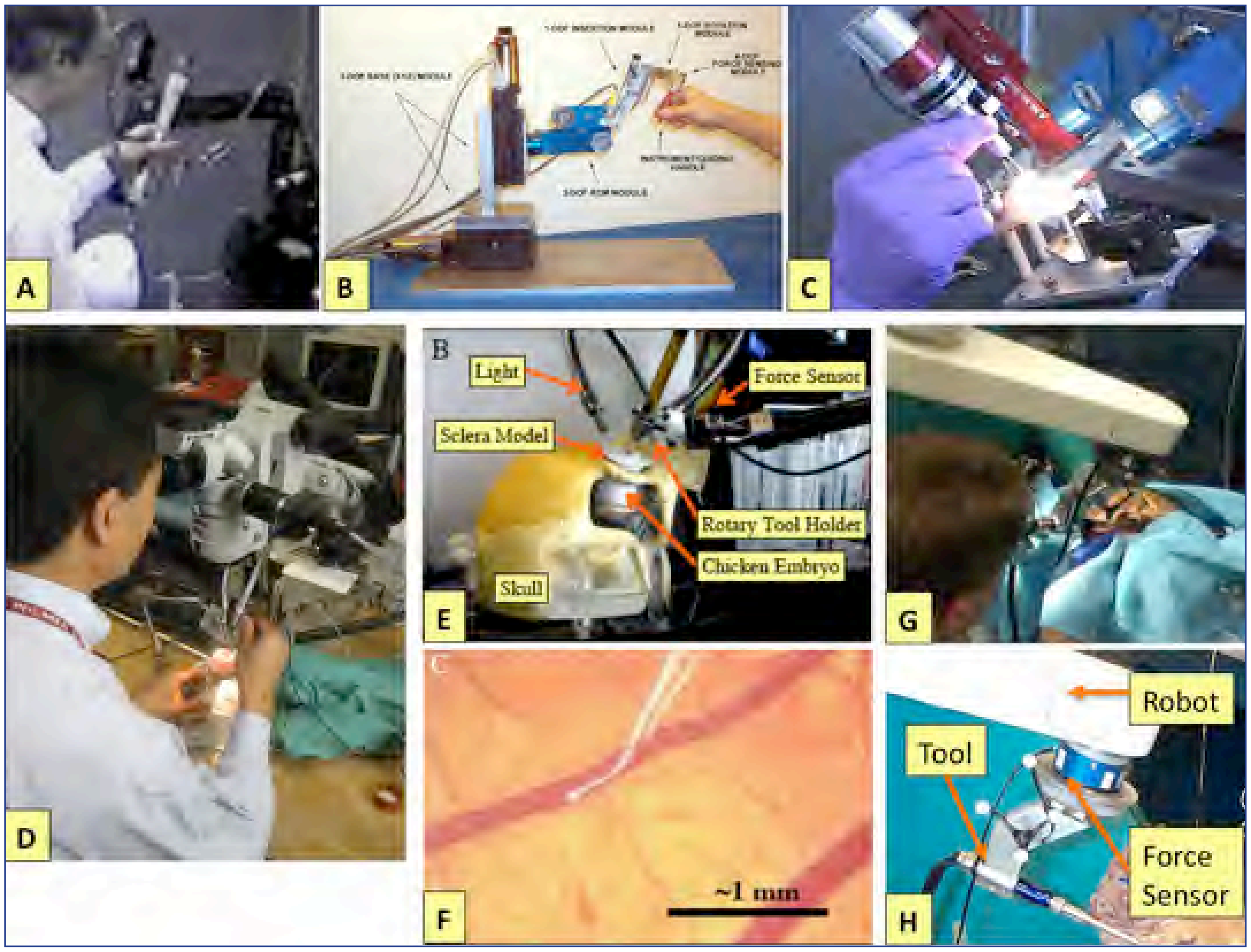

Fig. 10: JHU Steady Hand Robots: A) Early experiments with steady-hand evacuation of hematoma model with the IBM/JHU LARS robot, using a 5-bar linkage remote-center-of-motion (RCM) mechanism; B) Next generation Steady Hand Robot incorporating high dexterity chain drive RCM; C) Steady hand otology microsurgery experiments using the robot in B; D) Retinal surgeon James Handa using our current generation steady hand "Eye Robot" to cannulate 100 µm vessels in chick embryo; E) Detail of the robot and experimental setup for the Eye Robot; F) close up of the micro-pipette and blood vessels from the cannulation experiments; G, H) Steady-hand skull base surgery using a modified Neuromate™ neurosurgical robot. The systems shown in A-C have mechanically constrained remote-centers-of-motion, while the systems in D-G use software to enforce RCM constraints.

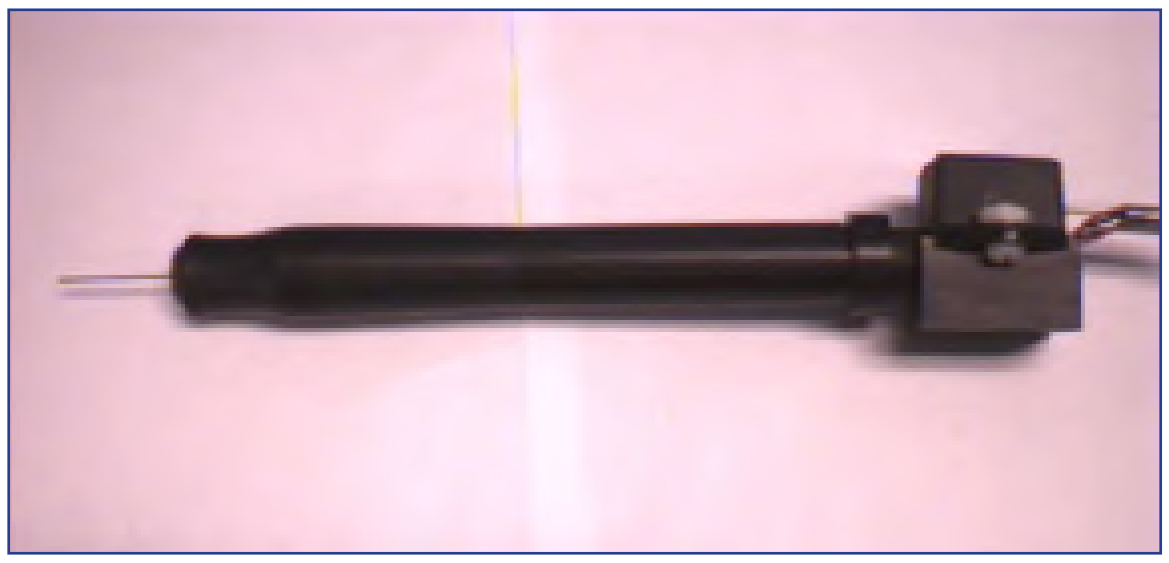

Fig. 11: First prototype Micron, an instrument for canceling hand tremor during microsurgery.

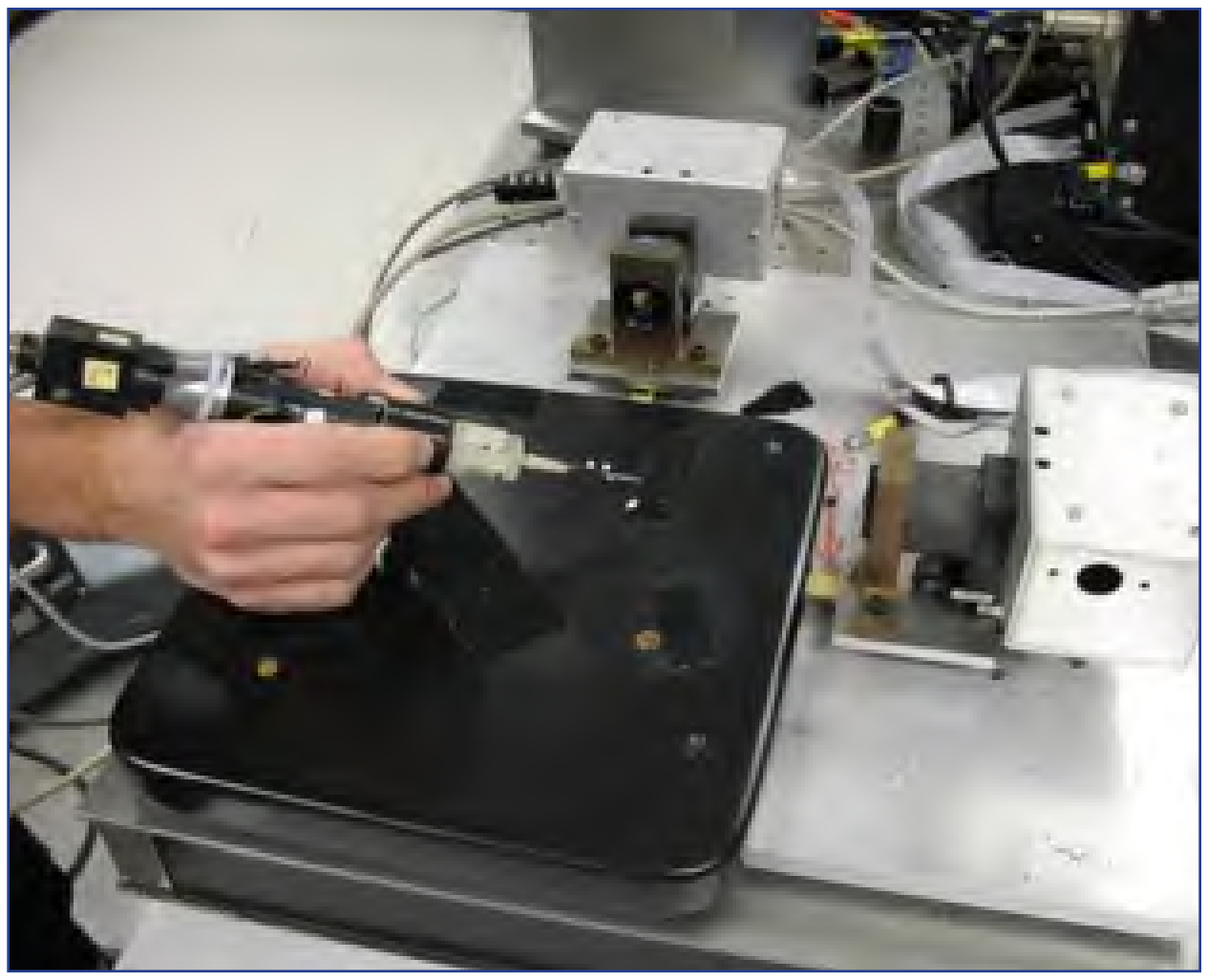

Fig. 12: Second Micron prototype with the ASAP measurement system.

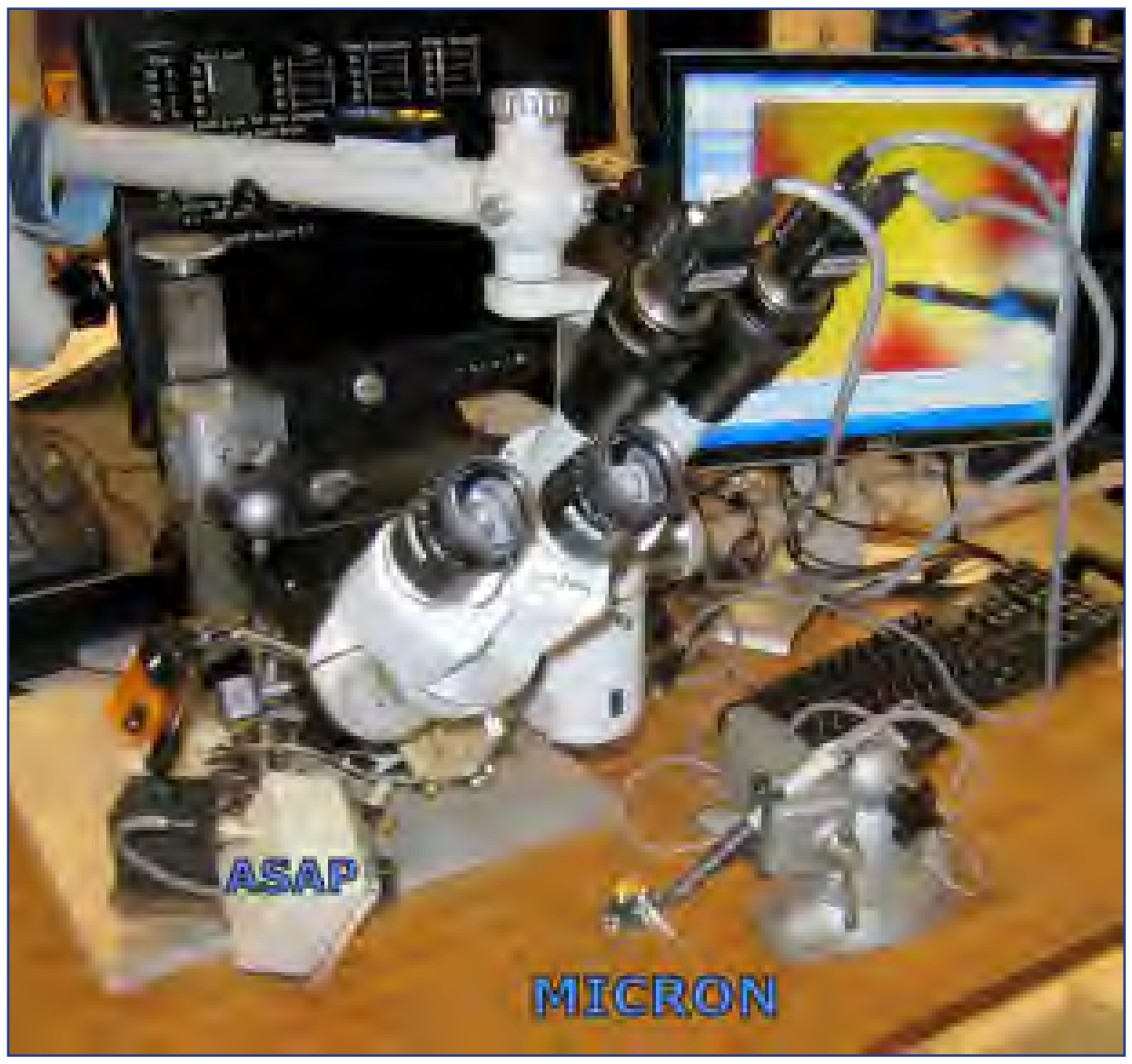

Fig. 13: Latest Micron and ASAP systems with visual tracking provided by JHU (shown on screen).

Canceling of tremulous disturbances has been demonstrated in the laboratory. The most recent version of the system is now being designed for its first in-vivo testing. Also, in collaboration with Dr. Gregory Hager at JHU, a vision-based tip tracking system is being developed. Using this system Micron will be able to implement both motion scaling and virtual fixture style behaviors. Micron is an integral part of the new NIH BRP award on retinal microsurgery.

### *Heartlander*

Most robot systems are "grounded," meaning they are physically attached to a stable base that is independent of the patient. However, systems attached to the patient can have great benefit when the organ of interest is highly mobile. As an example, Dr. Cam Riviere and his colleagues at our partner institution, Carnegie Mellon University (CMU), developed HeartLander, a minimally invasive robotic device for heart surgery that can adhere to the heart surface and navigate to any desired work site under the control of a surgeon. It uses suction to fasten itself to the heart and crawls like an inchworm across the surface. The device incorporates a videoscope to provide visual feedback to the surgeon, who controls it through a joystick interface. The device has a working channel through which various tools can be introduced for surgical procedures such as electrode placement, tissue ablation, drug or tissue injection, and anastomosis. The HeartLander attaches directly to the surface of the heart. It can be used to perform high-precision procedures without requiring compensation for heartbeat motion. It can crawl or walk to reach any point on the heart surface from any incision in the pericardial sac, so it can be inserted through an incision below the ribcage and does not require general anesthesia. Since it is compatible with local or regional anesthesia, it could enable ambulatory outpatient heart surgery for the first time.

The prototype has been tested on the beating hearts of four live pigs, and has demonstrated successful prehension, turning, and walking. The HeartLander project has been possible because of the tight coupling between engineers and surgeons that has been facilitated by the ERC. A patent has been filed for the device and CMU is considering the establishment of a start-up company to commercialize it.

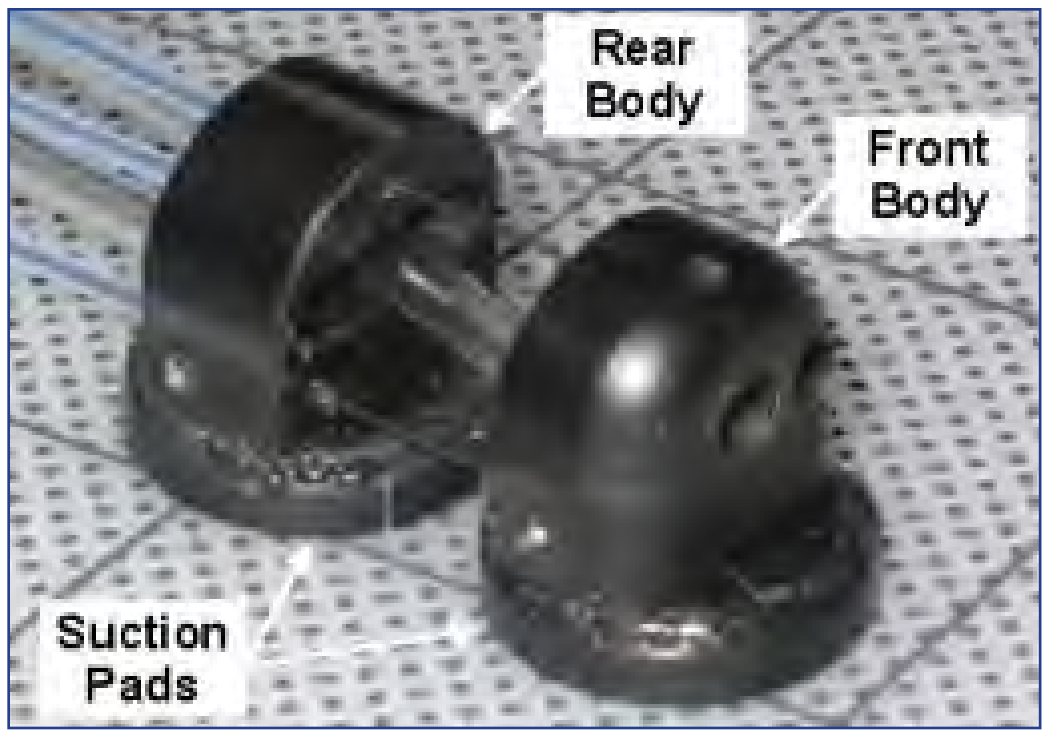

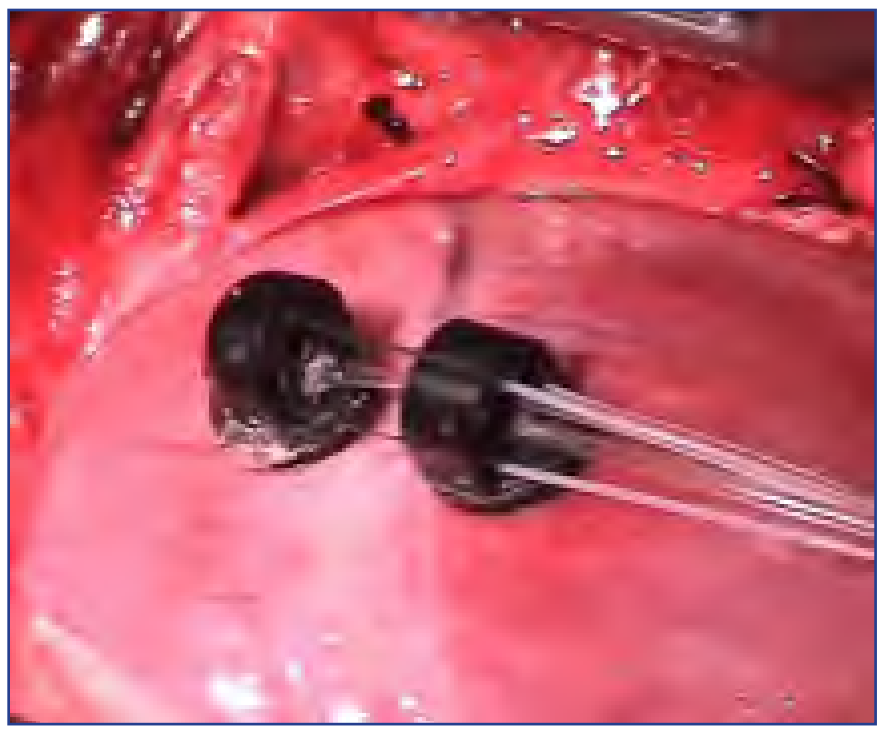

Fig. 14 and 15: HeartLander Robot: (top) Prototype. (bottom) Testing in vivo on beating heart of pig.

*A Snake-Like Robot for Minimally Invasive Surgery*

Robotics for minimally invasive surgery have made huge strides in the past ten years. This is particularly exemplified by the recent success of the da Vinci system from Intuitive Surgical Inc. However, there are many minimally invasive procedures for which the current da Vinci design approach is unsuited.

Within the ERC, Dr. Nabil Simaan (Columbia University), Dr. Russell Taylor and students Ankur Kapoor , Kai Xu, and Wei Wei, in collaboration with Dr. Paul Flint (JHU Otolaryngology) have developed a snake-like robotic system for minimally invasive surgical (MIS) procedures. Snake-like systems are particularly advantageous in procedures such as laryngeal surgery that require high dexterity in a constrained workspace. Current manual instrumentation is awkward,

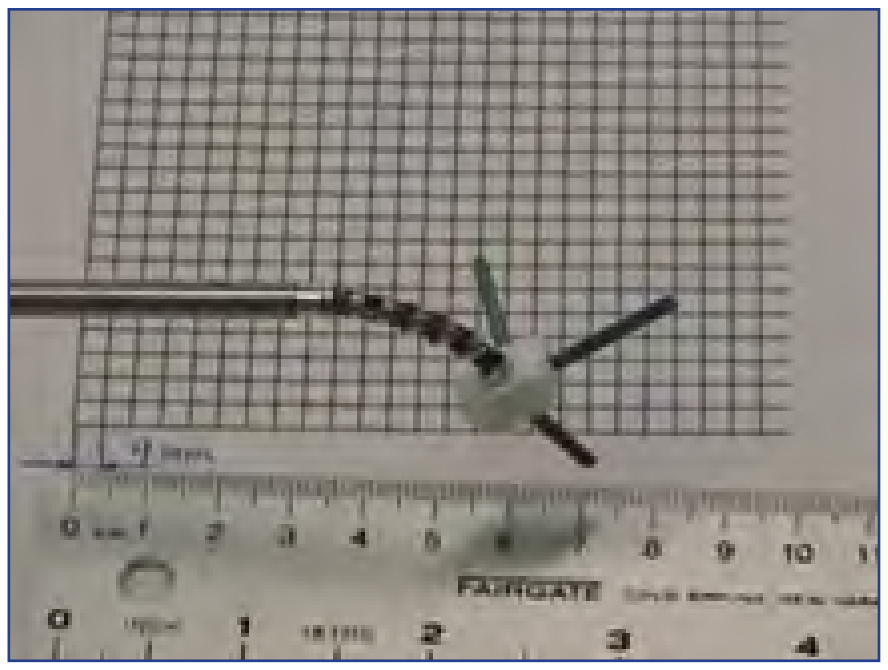

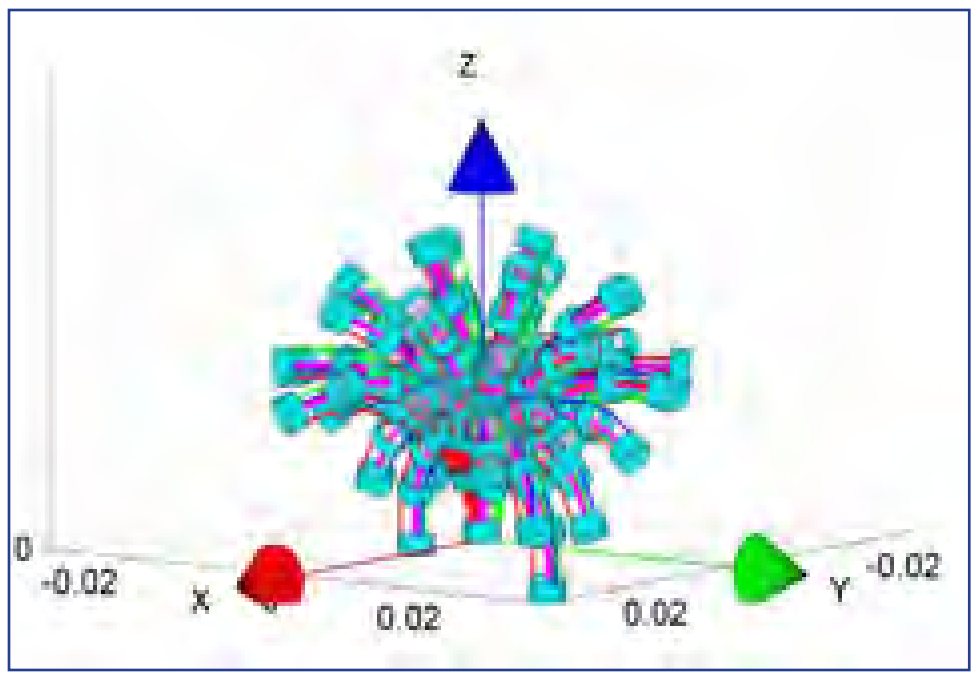

Fig. 16 and 17: Prototype (left) and workspace analysis (right) of a snake-like robot for laryngeal surgery hard to manipulate precisely, and lacks sufficient dexterity to permit common surgical subtasks such as suturing vocal fold tissue. Size limitations require a small diameter robot, less than 5 mm, for suturing purposes inside the throat.

This robot is based on a new design concept for snake-like robots using flexible members and redundant actuation. This innovative design makes it possible to build extremely small robots that can apply the forces necessary for surgical instrument manipulation. The final system is a hybrid system combining a low-degree-of-freedom tool manipulation unit (TMU) and several distal dexterity units (DDUs) consisting of a snake-like robot and a detachable “milli-parallel” wrist. The TMU provides a mobile base from which multiple DDUs can be controlled through a single entry point, thereby minimizing the complexity of coordinating several DDUs manually. Figure 16 shows a prototype of a 4 mm snake-like robot actuated by a continuous multi-backbone structure, as well as a workspace analysis revealing the high dexterity of a single end-effector.

The most recent version of the system has been integrated with a master console that is normally used to run the commercially available da Vinci surgical robot. The overall system block diagram is shown in Figure 18. The method works as follows: (1) Individual joints of the manipulator are servoed with low-level controllers (PD or PID) to position set points. (2) The force applied by the user to the master results in a position error, which is scaled to give Cartesian velocity. (3) A constrained least square problem is solved for the joint velocities by the high-level controller. The least square problem has an objective function describing desired outcome. It may also include constraints that consider any motion constraints due to VF, joint limits, and velocity limits. (4) Numerically integrate the joint velocities to arrive at a new set of joint positions.

The implementation on the patient side is similar except that the desired Cartesian velocity is computed using the current master robot position and the patient-side robot position. The use of the constraint-controller-based teleoperator also allows us to easily incorporate collision avoidance between the two patient-side robots. In the current setup, the snake-like end effectors are mounted on five-bar mechanisms. We would like to avoid the collision of these five-bar mechanisms without limiting the workspace of the two arms, and also allow interaction between the grippers. Thus, each patient-side robot has a constraint that depends on the current position and velocity of the other robot and maintains the distance to be larger than a safe distance between the two five-bar mechanisms. The technology is licensed to Intuitive Surgical, Inc.

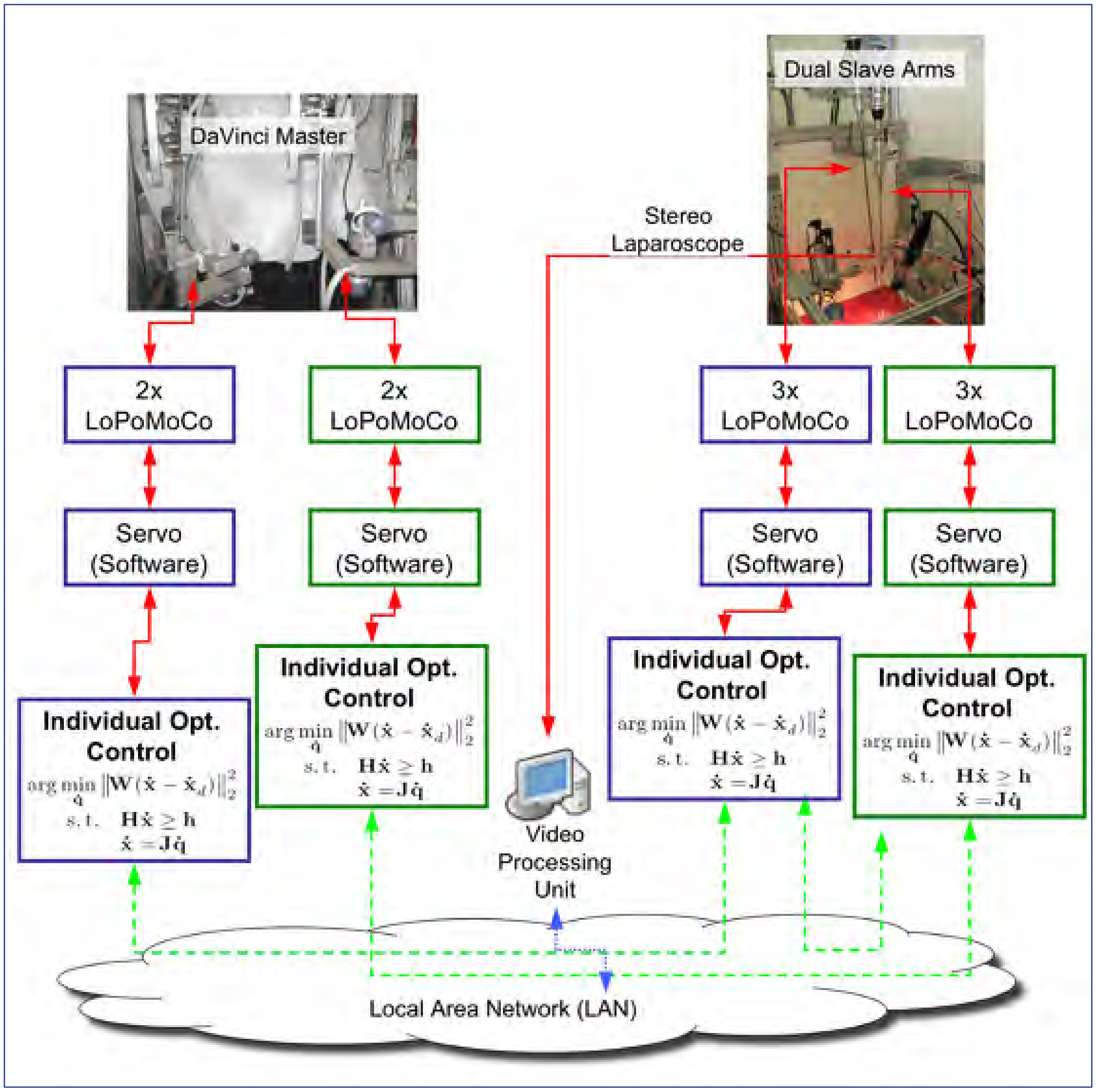

Fig. 18: The architecture of dual arm teleoperated snake-like robot.

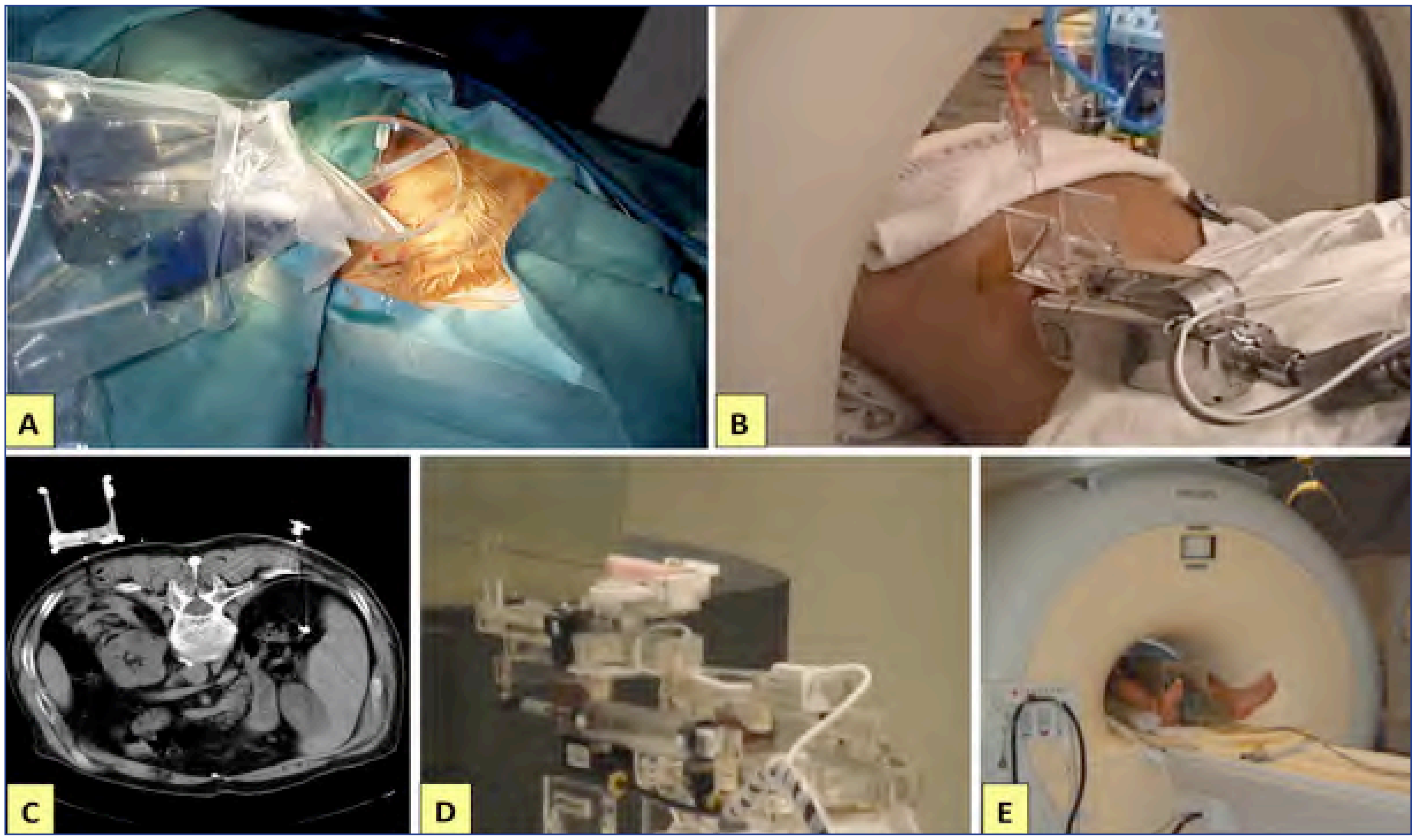

Fig. 19: Robots for in-imager percutaneous needle placement. A) JHU RCM robot with radiolucent PAKY needle driver for placing needles into the kidney under x-ray guidance; B) JHU RCM robot with modified PAKY needle driver for CT-guided biopsy; C) CT image of system in B showing the use of the fiducial structure to provide positional feedback of the needle driver from a targeting image; D, E) MRI-compatible robot for MRI-guided percutaneous prostate brachytherapy.

*Robots for percutaneous needle placement*

Most of the robots described above fall within the Surgical Assistance paradigm. However, accurate placement of needles and other therapy devices onto anatomic targets defined in medical images has been a major theme for our research over the life of the ERC. There are two general problems in developing such systems. First, the robots in question must often work in or near medical imaging devices, yet must not interfere with the operation of those devices. Second, it is common that the most effective therapeutic path makes use of natural orifices within the body. Designing robots to operate in these spaces poses its own unique challenges.

Within the ERC, we have developed many robotic systems for use with medical imaging devices. Figure 19 illustrates some of the systems that we have developed for in-imager percutaneous applications. Many of these systems have incorporated (e.g., Figures 19A,B) the same mechanically constrained RCM mechanisms used in some of our Steady-Hand systems. A related theme, illustrated in Figures 19B,C, has been the development of unique needle drivers incorporating fiducials to facilitate accurate closed-loop alignment of needles to targeted anatomy.

Under the leadership of Dr. Gabor Fichtinger and Dr. Louis Whitcomb, our ERC has developed Magnetic Resonance Imaging (MRI) and Ultrasound (US)-guided needle placement medical robot systems. These robots take the safest and shortest access route to the site of disease through the rectal cavity. There are still formidable technological challenges to overcome. There is no room to maneuver in the cavity, where the surgical needle must "turn a corner" to enter the prostate gland across the wall of the cavity. There are also compatibility issues with the given imaging device.

MRI provides the best anatomical picture available today, but the high magnetic field (200,000 times stronger than the Earth's) excludes the use of metals and electronics inside the scanner, where the workspace is also very limited. Due to these limitations, such high-quality scanners were previously unavailable for surgical interventions. Our MRI-guided robot (see figure above) overcomes these problems and it enables precise anatomical targeting inside a closed high-field MRI scanner, with previously unprecedented intra-procedural image quality. The system is in multiple clinical trials for prostate biopsy and seed placement, led by Dr. Cynthia Menard at NIH. Clinical accuracy has been promising, based on early results in both phase-1 trials. We have encountered no complications and safety problems, both during and after the procedures.

Transrectal ultrasound (TRUS) is the most popular prostate imaging tool, yet the imager entirely occupies the rectal cavity, thus leaving no room for any kind of mechanical actuation. We

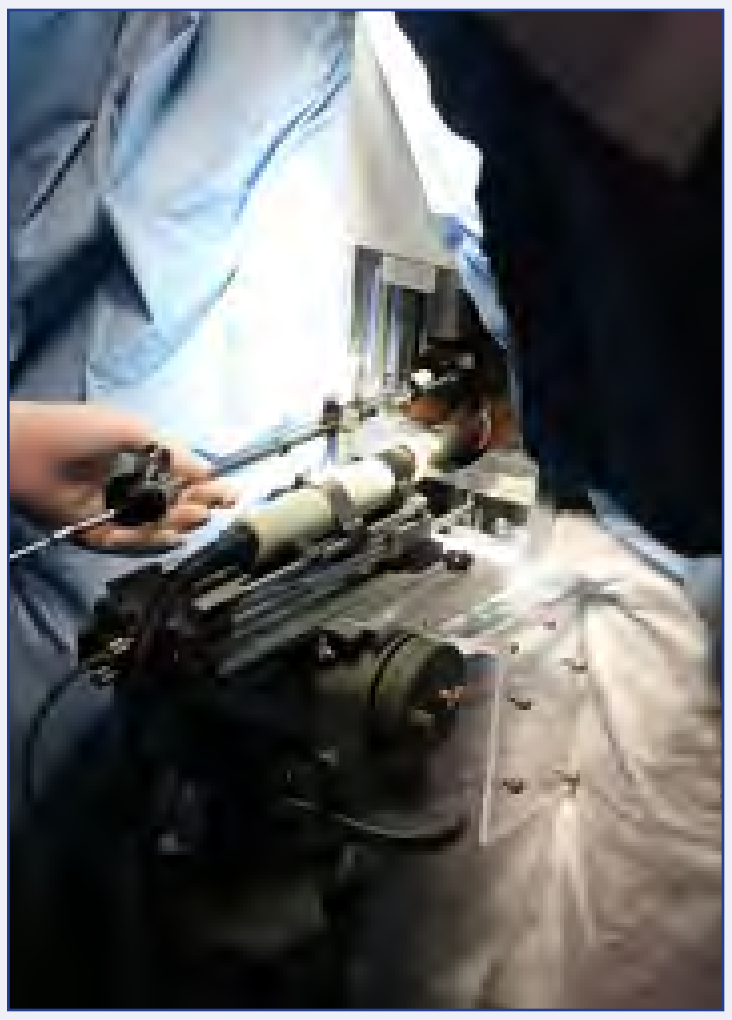

Fig. 20

In an NIH-funded collaborative project between ERC engineers, clinical partners, and industry partners, a robotic system for ultrasound-guided prostate brachytherapy has been developed. The system is integrated with a commercial product developed by Acoustic MedSystems and has been tested in a Phase-1 clinical trial. A recent paper received the Second Best Paper Award of the Journal of Medical Image Analysis, in the special issue on the 10th International Conference on Medical Image Computing and Computer Assisted Intervention (MICCAI).

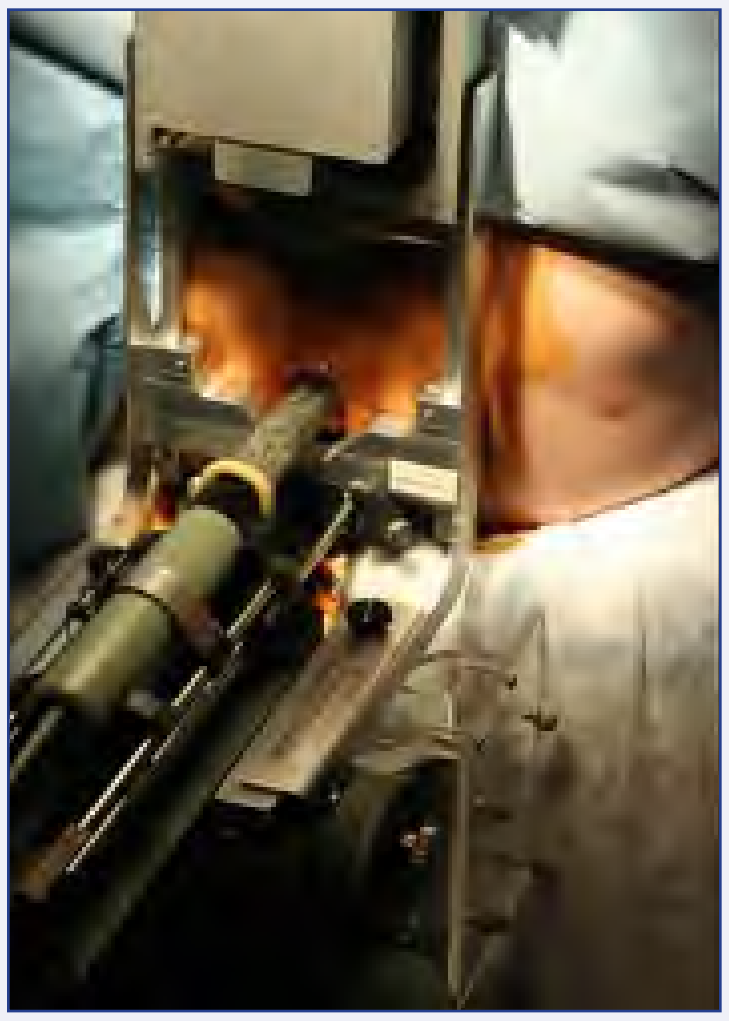

Fig. 21

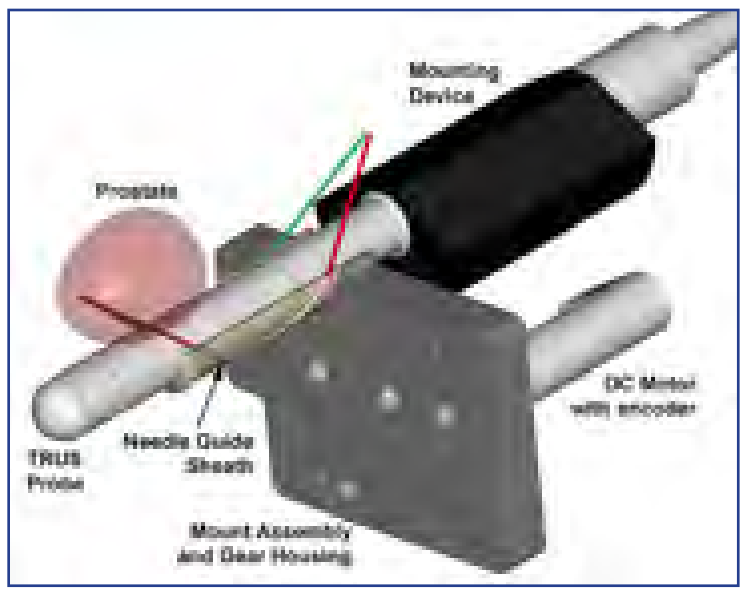

Fig. 22A

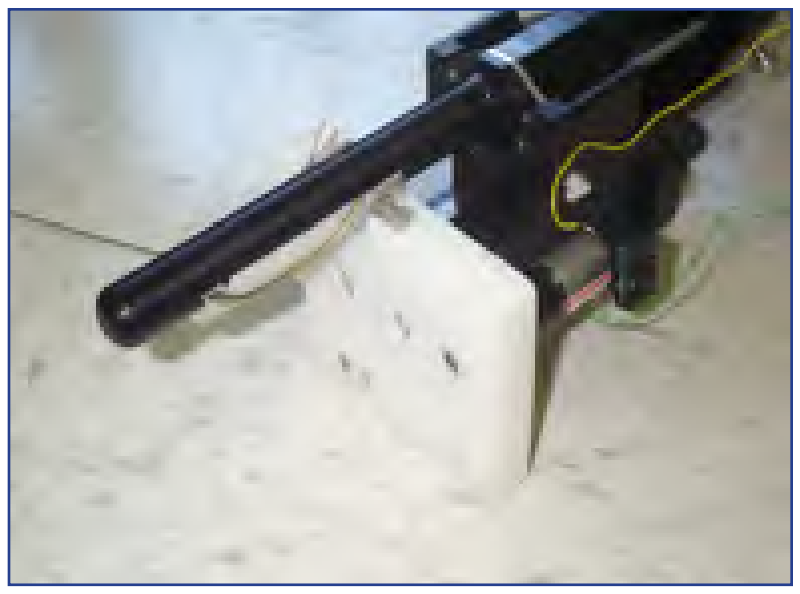

Fig. 22B

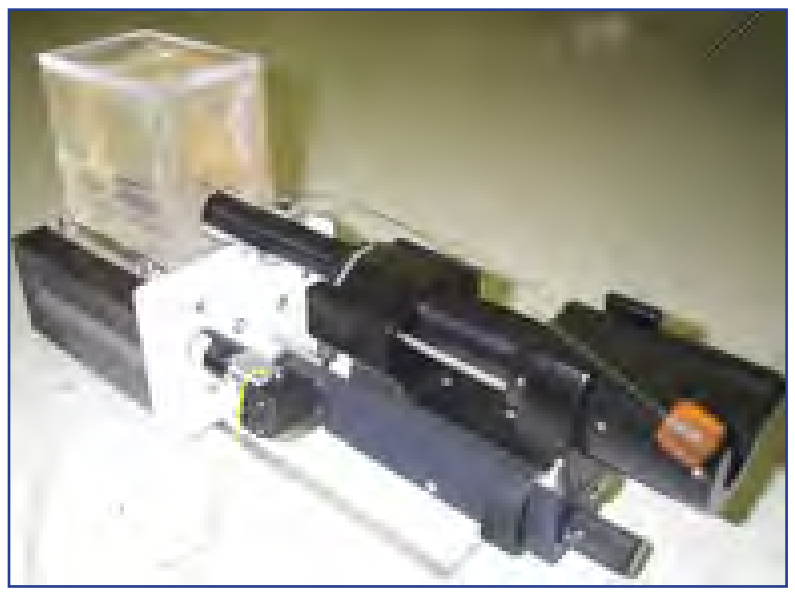

Fig. 22C

Fig. 22:TRUS-guided prostate robot as designed (A), assembled (B), and deployed in phantom experiment (C)

built a robot (Fig. 22) to overcome this problem by navigating a pair of flexible surgical needles in a ~2 mm space between a commercial, off-the-shelf, TRUS probe and the cavity wall, while keeping all actuation gear and mechatronic elements outside the patient's body. Initial phantom experiments suggest that targeting and needle placement accuracies are better than achieved manually. The prototype device is being adopted for ultrasonic ablation of recurrent prostate cancer, which is an inventive therapeutic procedure that is impossible to perform manually. This work is in partnership with Acoustic MedSystems, Inc., one of our industrial partners, financed from a Phase-1/ Phase-2 STTR grant from NIH.

## *Imaging and Registration*

The CISST ERC has developed new technologies in the area of imaging and registration for a variety of medical applications. One particular theme that the center has pioneered is the use of statistical atlases of imagery to support the development of patient specific models and treatment planning. We describe several of those innovations here.

### *Statistical Shape Models for Segmentation*

Robust and reliable automatic segmentation has been very challenging to date because of insufficient information and undesirable "noise" in the images being used. When the image contrast in the segmentation is weak, clinicians rely on their knowledge of anatomy to make up for these deficiencies. Better use of the computational modeling of anatomical shapes and their variability in the population can significantly improve the quality of automatic segmentation.

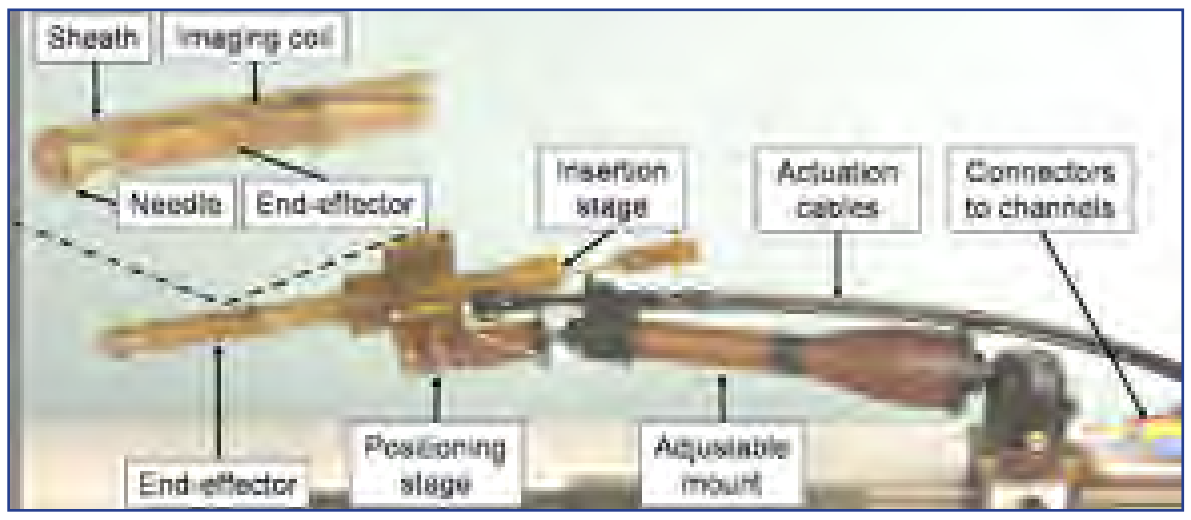

Fig. 23A

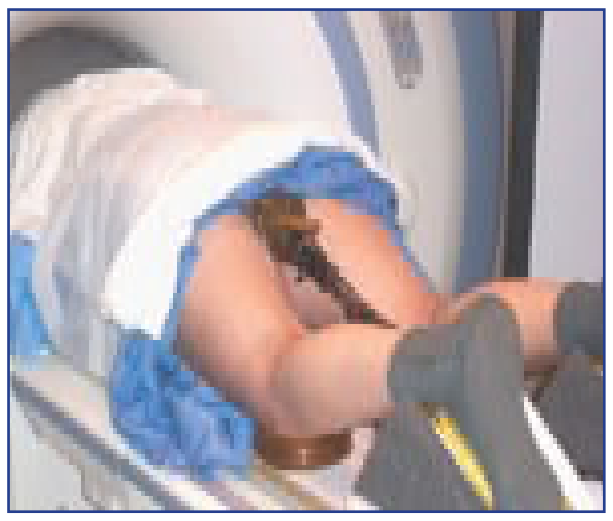

Fig. 23B

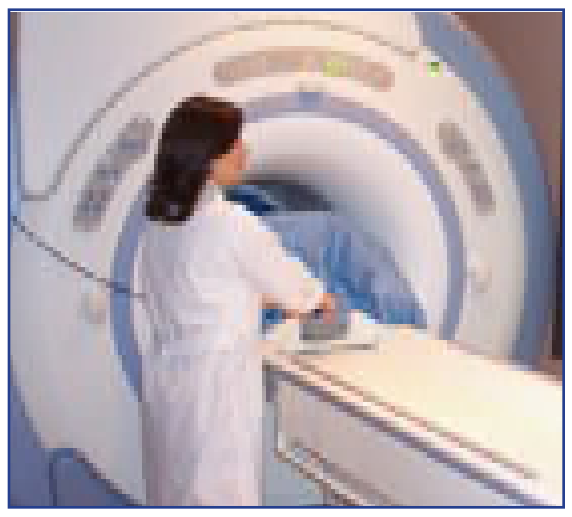

Fig. 23C

Fig. 23: MRI-guided prostate robot assembled for treatment (A), in treatment position (B), and operated by Dr. Menard (C)

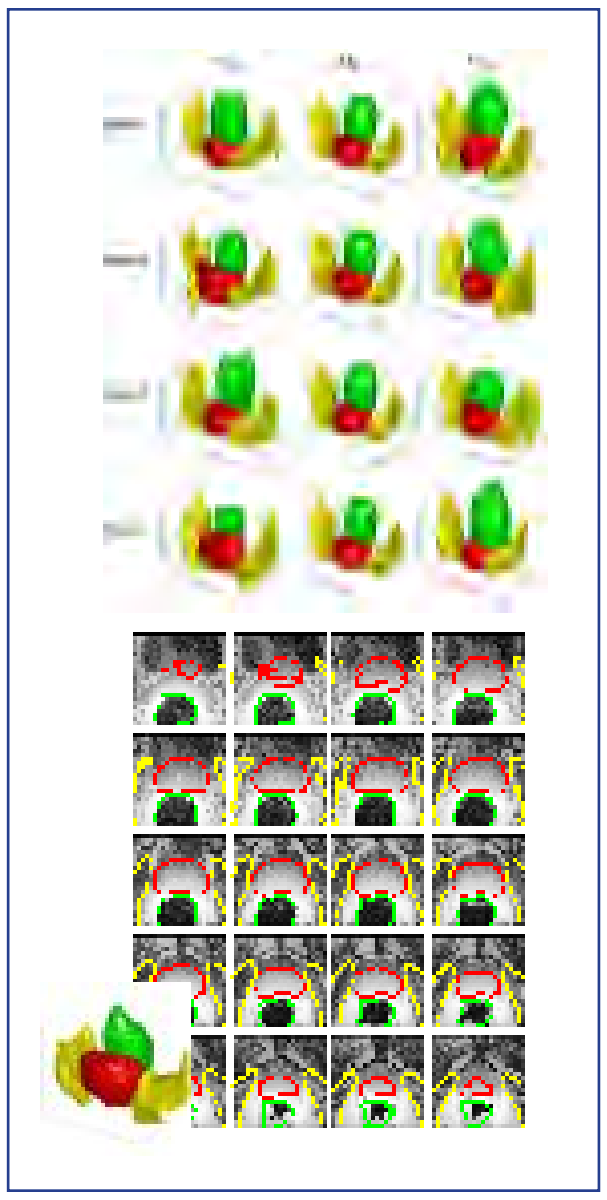

Fig. 24: Shape-based segmentation of the prostate.

A research team led by Dr. Eric Grimson at our partner institution, Massachusetts Institute of Technology (MIT), constructed a statistical model of prostate shape from previously segmented scans, by estimating the mean shape and the principal modes of its deformation within the population from which the image data was collected (Fig. 24). The top images in the figure illustrate such a model for the prostate (red), the colon (green) and the surrounding muscles (yellow). The array of 3D surfaces illustrates the mean shape (center column) and its variation (left to right) along the four most "important" directions estimated from the data. In this example, the model captures not only the organs' shape, but also their relative position and orientation. It is then used during the segmentation of new scans to constrain the shape of the hypothesized segmentation. The intensity information in the adjacent structures leads to a robust segmentation of the prostate, which would be very challenging if the prostate were considered in isolation.

The bottom images in the figure show a novel scan that was automatically segmented using this approach, as well as the 3D rendering of the segmented structures.

This automatic procedure greatly facilitates planning and navigation for minimally invasive treatment of prostate cancer. Moreover, the applications of the segmentation algorithms developed and demonstrated in this program range from surgical planning and navigation to population-based studies of neuroanatomy and the effects of diseases on the morphology of the brain.

***Targeted Prostate Biopsy Using Statistical Atlases and Mathematical Optimization***

Given a set of segmented images from a patient population, one of the remaining challenges is to make use of this information to guide interventions. Our partners at the University of Pennsylvania, led by Dr. Christos Davatzikos and Dr. Dinggang Shen, have constructed a statistical atlas of prostate cancer distribution and have shown that it can be used to significantly improve the determination of optimal biopsy sites.

Current imaging modalities have very poor performance in detecting prostate cancer. Therefore, prostate cancer diagnosis is currently performed rather blindly, under various protocols for needle placement. The most widely used protocol is the six-needle biopsy, performed under very approximate guidelines. Success rate of diagnosis and staging of prostate cancer can be greatly improved by developing statistical methods for sampling, provided that the statistics of the population can be determined.

Using histological sections from about 300 radical prostatectomy specimens, along with deformable registration methodologies, CISST and its partners – the University of Pennsylvania, Brigham and Women's Hospital, Georgetown University, and the Center for Prostate Disease Research – have developed a multivariate statistical model for determining optimized biopsy plans based on the spatial distribution of prostate cancer. This statistical atlas can be warped to the individual patient's space, under guidance from ultrasound images, for performing optimal biopsy for the particular patient (Figure 25). By applying this optimized biopsy protocol to a database of prostatectomy specimens, our detection rates, estimated from whole-mount histological stains, reached over 90% levels for 6-7 biopsy cores after applying cross-validation, which represents a significant improvement over standard random-systematic biopsy protocols that can only reach about 70% in the same patients.

***Deformable registration of statistical atlases to x-ray images***

The CISST ERC has been one of the leaders in developing ways to construct 3D anatomic models of bony anatomy from a limited number of x-ray images by "deformably" registering statistical atlases to x-ray images. X-ray images give two-dimensional projections of a three dimensional patient. The intensity at each point in the image is primarily a function of the total amount of x-ray absorbing material along a line from the x-ray source to that point on the x-ray film or detector. Human radiologists and surgeons use their knowledge of anatomy to infer 3D structural information about the patient from the 2D images. However, the process requires training and considerable experience.

The first step of the process is construction of the statistical atlas from CT images of multiple patients, as illustrated in Figure 26. In our case, we have chosen a data structure for the atlas to optimize performance when it is used for the purpose of creating patient-specific bone models from x-ray images. Bone shapes are represented as tetrahedral meshes, and bone density within each tetrahedron is represented by barycentric Bernstein polynomials. To date, we have produced atlases of the human male pelvis and male and female proximal femur, and work is underway for a female pelvis atlas.

The process of deformably registering the atlas to x-ray images is illustrated in Figure 27. We start with a set of x-ray images and an initial guess at the patient bone shape, density, and pose relative to the x-rays. These are used to produce a set of predicted x-ray images, which are compared to the observed images. An optimization process iteratively improves the shape and density estimate until a model is produced that closely predicts the observed x-ray images. The crucial step in performing this task efficiently is the prediction of x-ray projection images from deforming 3D models. Accordingly, we have developed novel, very fast methods for performing this step on inexpensive graphic processor cards found in modern PCs.

We have applied this method to several problems, including the hybrid tomographic reconstruction problem discussed in the next section. One promising use is ongoing work with our Industry affiliate, Hologic, Inc., to estimate 3D biomechanical properties of the proximal femur from a small number of 2D dual-energy x-ray (DXA) images. The most widely accepted diagnostic test for osteoporosis is an areal bone mineral density (BMD) measurement of the hip from DXA. While this is an excellent diagnostic tool in many ways (very low radiation dose, moderate cost, able to identify those at high risk for a future fracture), BMD is not definitive. More than 50% of fractures occur in people who are not classified by areal BMD as osteoporotic. Better diagnostic tests are needed. It has been widely remarked by researchers and physicians that a deficiency of areal BMD is a two-dimensional measurement, neglecting the three-dimensional structure of the bone.

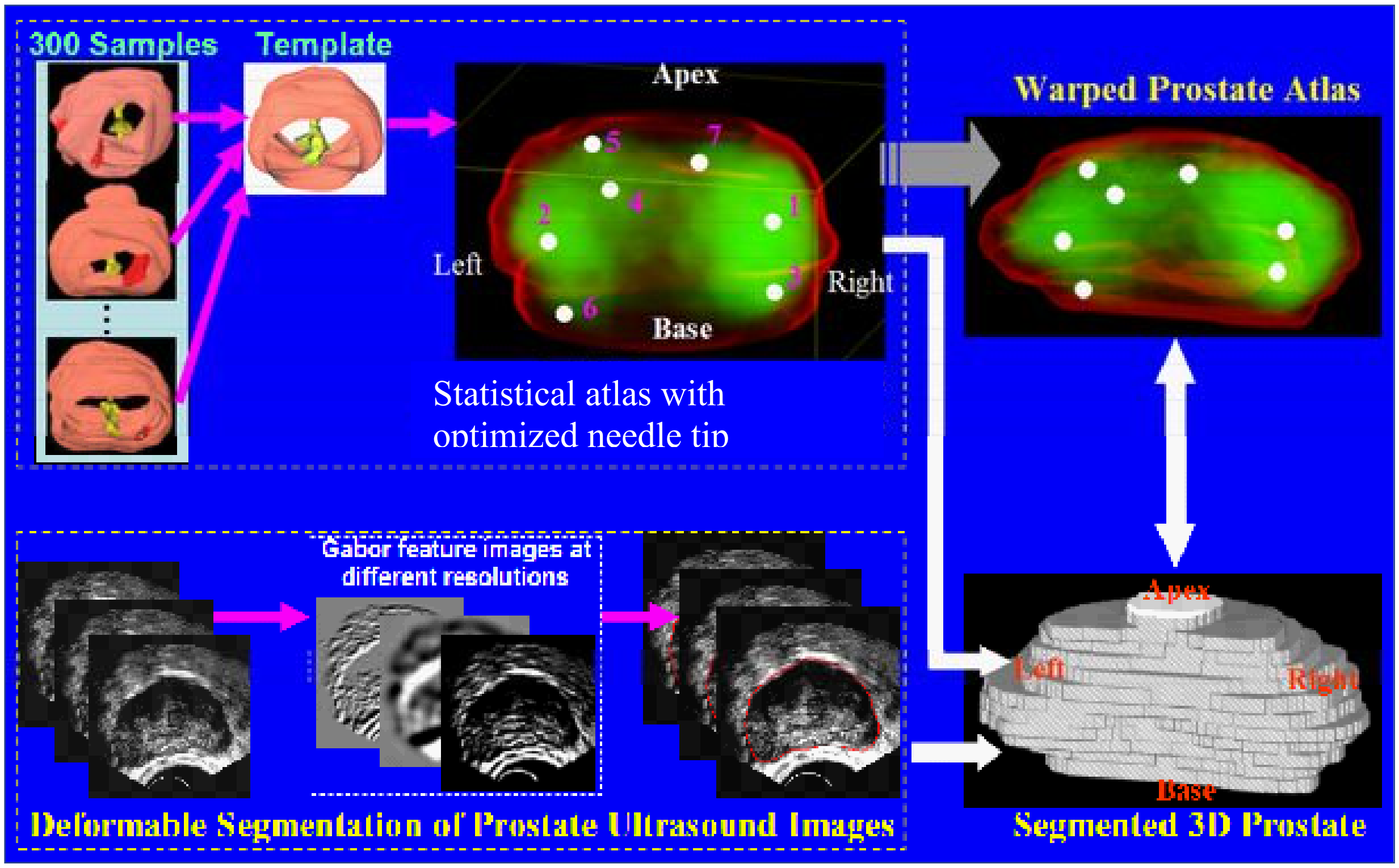

Fig. 25: Atlas-based prostate biopsy strategies.

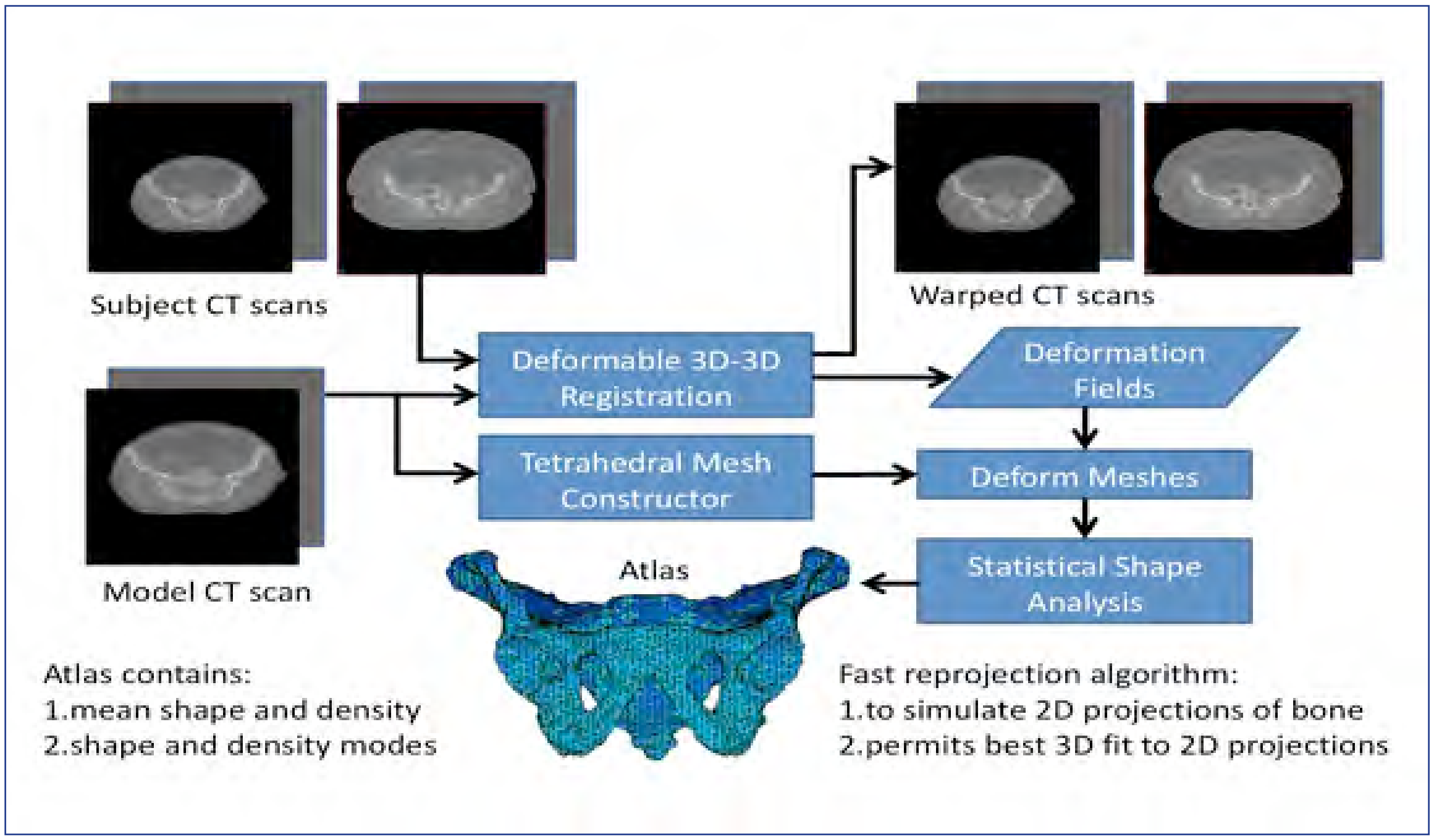

Fig. 26: Atlas Construction Process

While it is possible to measure volumetric BMD and three-dimensional structure of bone with CT, the cost and radiation exposure are both quite high, and it is not suitable as a screening test. CISST and Hologic have partnered together to combine a small number, currently four, of DXA images of the hip with a statistical atlas to produce a volumetric BMD patient specific model. This test uses a fraction of the radiation dose of a CT scan and can be done on equipment that is much less costly than a traditional CT. Additionally, it is additive to the physician's information as the physician still obtains the traditional areal BMD measurement from the DXA images.

Although this work is still in development stages, the preliminary results are encouraging, and we hope that it will eventually lead to an application to the FDA for permission to apply clinically as part of a commercially deployed system.

### *Hybrid Tomographic Reconstruction using Prior Models*

The CISST ERC has developed a method to combine prior information in the form of CT scans or statistical atlases with a limited collection of x-rays to produce CT-like reconstructions of the patient during surgery, using the mobile fluoroscopic x-ray "c-arms" commonly available in orthopaedic operating rooms (Fig. 27). Such reconstructions can have significant value to orthopaedic surgeons, since they can provide 3D feedback of changes to the patient during the procedure.

Three-dimensional reconstruction from x-ray projections is at the heart of x-ray computed tomography (CT) and is a well-studied problem. Standard reconstruction methods require the knowledge of x-ray projection intensity for every ray through every reconstructed volume element (voxel). "Cone beam" reconstruction methods of 3D volumes from multiple 2D x-ray projection images are known and are widely available on high-end x-ray systems (such as those found in angiography suites). Only

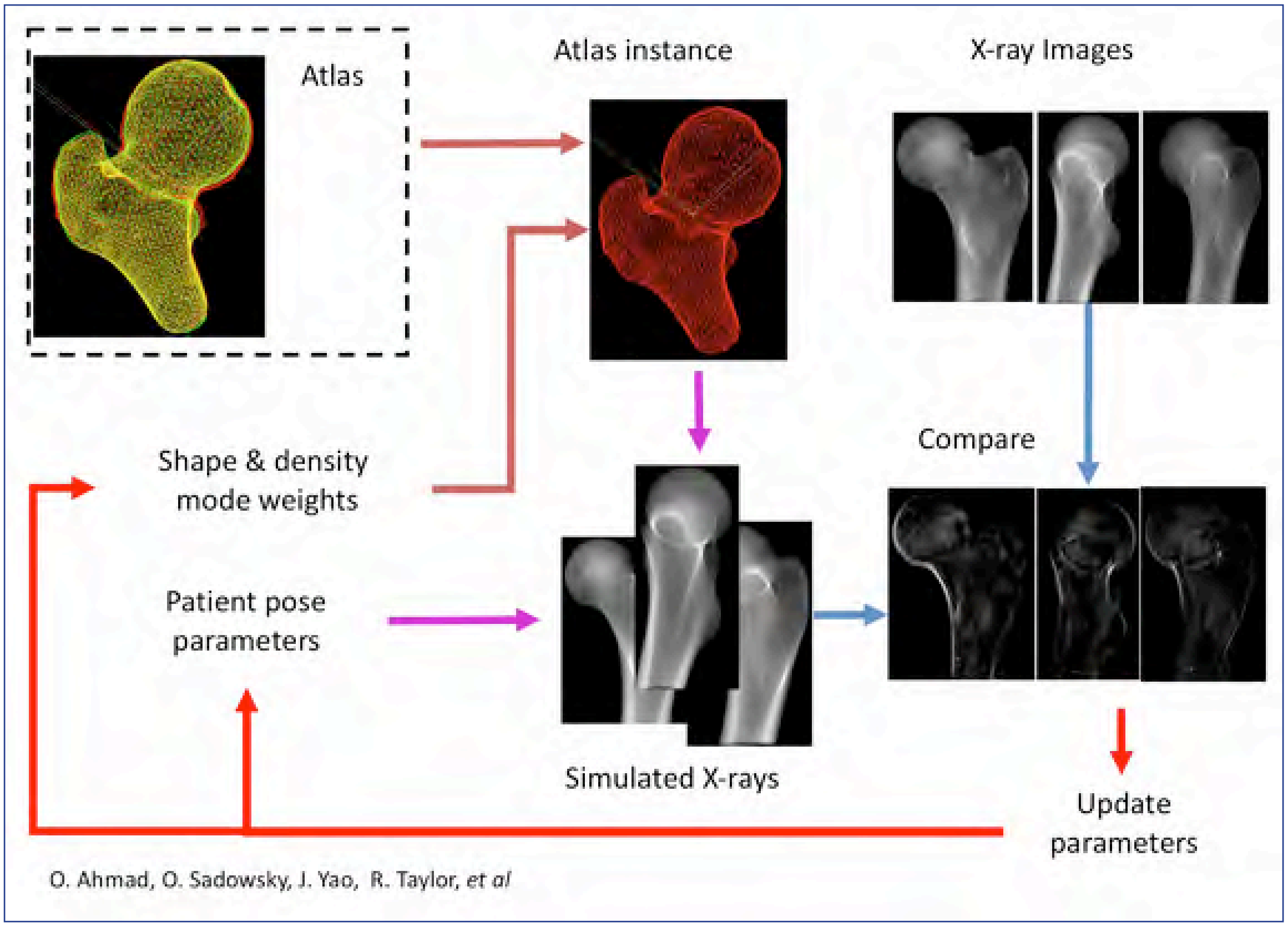

Fig. 27: Deformable 2D-3D registration method

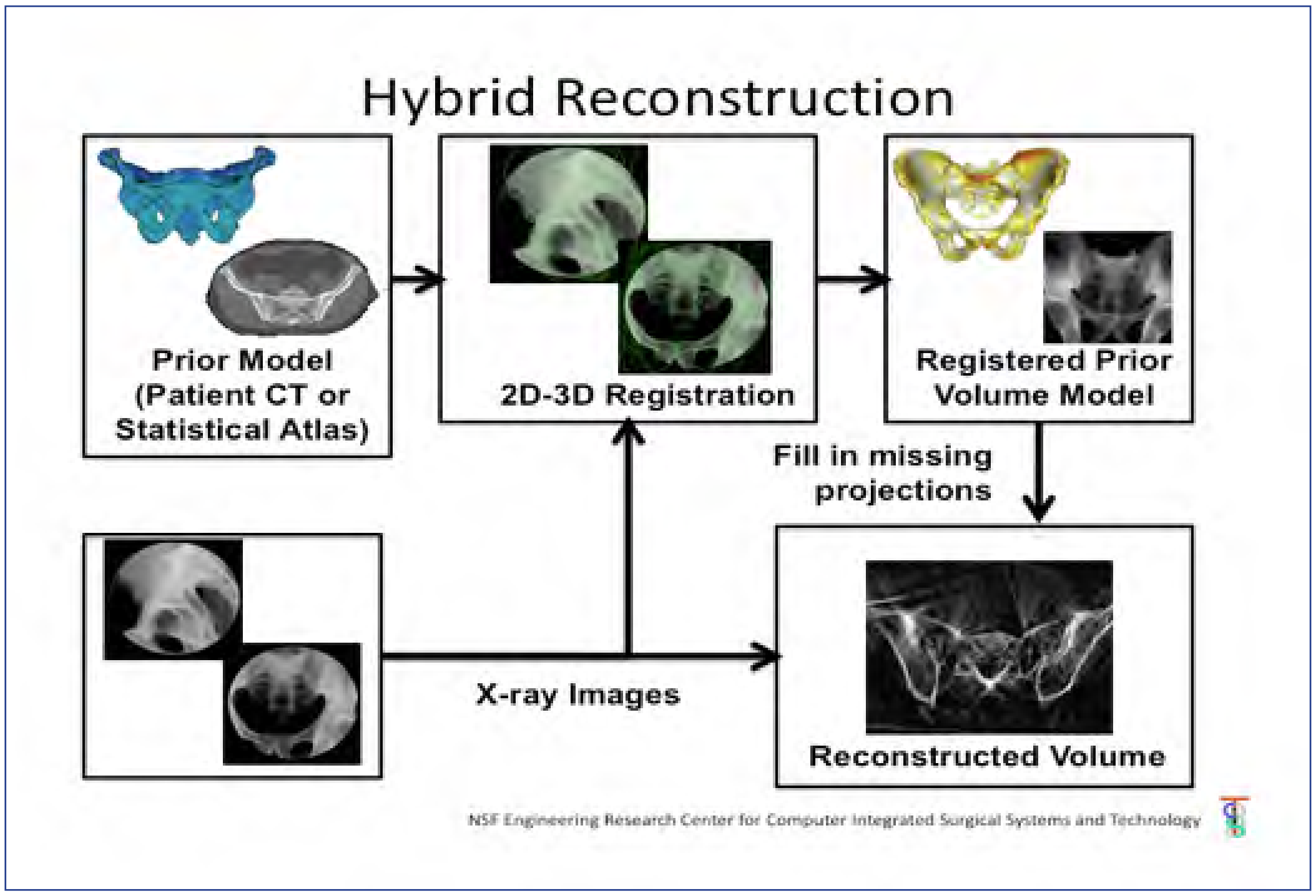

Fig. 28: Hybrid tomographic reconstruction from limited x-ray projection images and a prior CT scan of the patient, or statistical atlas of patient anatomy.

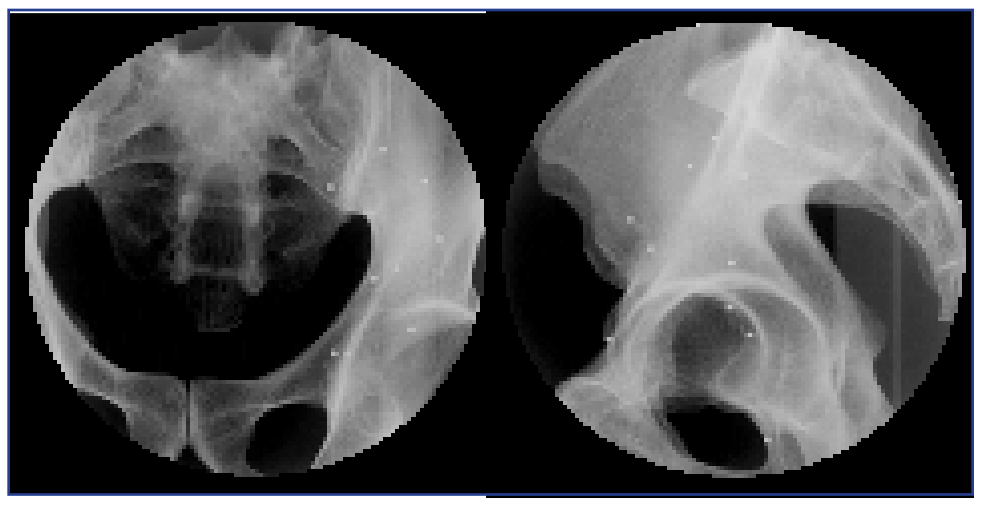

Fig. 29: X-Ray images of a dry pelvis bone taken with a mobile c-arm. Note image truncation.

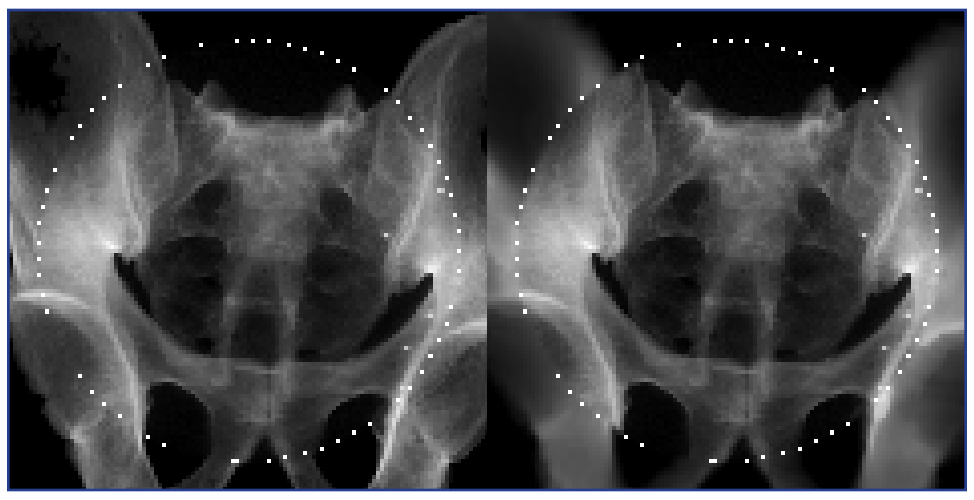

Fig. 30: Fusion of pelvis x-rays (in circle), registered to prior CT (left) and statistical atlas (right) outside the circle. The projections outside the circle compensate for image truncation.

*Research Accomplishments*

a relative handful of smaller mobile x-ray c-arms, typically found in orthopedic operating room systems, have such capabilities, and even in these cases there are many limitations.

To overcome these limitations, we have developed a hybrid tomographic reconstruction method that combines intraoperative x-ray images with prior information about the patient in the form of a CT scan or statistical atlas constructed from multiple patients (Fig. 28). The steps of this reconstruction method are as follows. First, a "scan" of x-ray images of the patient from multiple projection angles is obtained. Second, these images are corrected to account for image distortions and other geometric and intensity-response characteristics of the c-arm. Then the 2D projection images are registered to either the prior CT scan, rigid registration, or the atlas, non-rigid registration. In the current implementation, after the atlas is aligned to the available data the pixel intensities of the projected atlas images are matched to those of the x-ray images. This allows a smooth fusion of both modalities to be obtained (Fig. 30). To compensate for image truncation or missing x-ray views, the data from x-rays and atlas are blended together as inputs to a conventional cone-beam reconstruction algorithm.

Results of this method on a cadaveric human pelvis phantom using a typical mobile fluoroscopic c-arm (OEC 9600) are shown in Figure 30, using both a prior CT model and a shape/density atlas constructed from CT images of 110 normal adult male subjects. As this is written (Summer 2008) we are conducting preliminary cadaver trials to reconstruct intraoperative changes, such as cement injections in bone augmentation and are considering further extensions of this work.

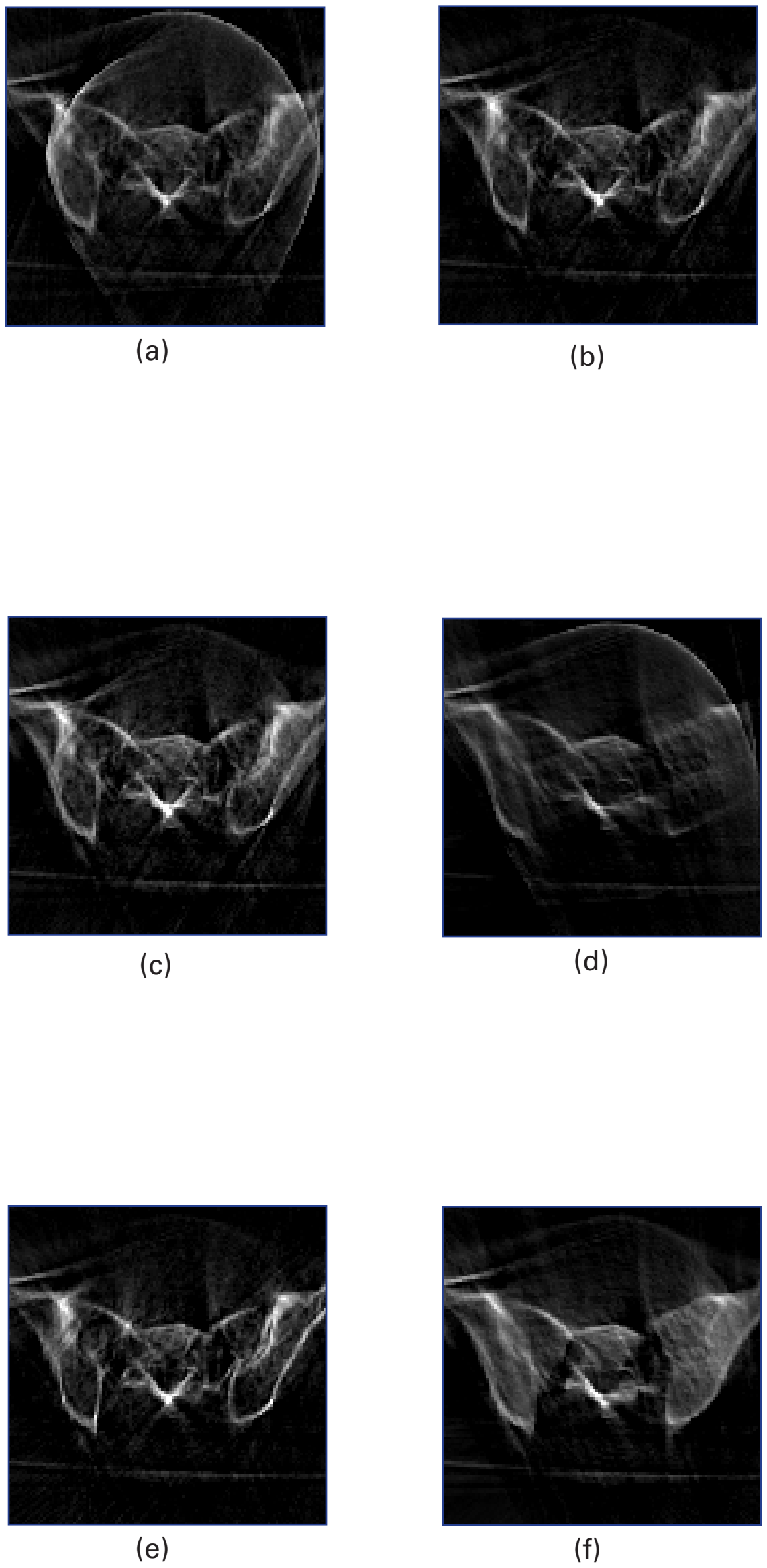

Fig. 31: Hybrid reconstruction of x-ray and prior models. Top row: using a 200° x-ray "short scan." Bottom: using 110° of x-ray scan and 90° of prior model.

(a) Cross section reconstructed from truncated x-rays only.

(b) Reconstruction from image fusion with prior CT.

(c) Reconstruction from fusion with statistical atlas.

(d) Reconstruction from 110° x-rays without prior.

(e) Reconstruction using 90° projections of prior CT.

(f) Reconstruction using 90° projections of deformable atlas.

## *Human-Machine Interfaces*

Nearly all clinical computer-integrated interventional systems couple a computationally driven device with a human partner. The goal of this partnership is to amplify or assist human capabilities during the performance of tasks that combine human judgment with robotic precision and information-enhancing visualization. We have thus coined the term "Human-Machine Collaborative Systems" (HMCS) to describe this area of work. Much of our work on HMCS is specifically aimed at surgical systems, but the basic principles apply to a wide variety of clinical applications within CIS.

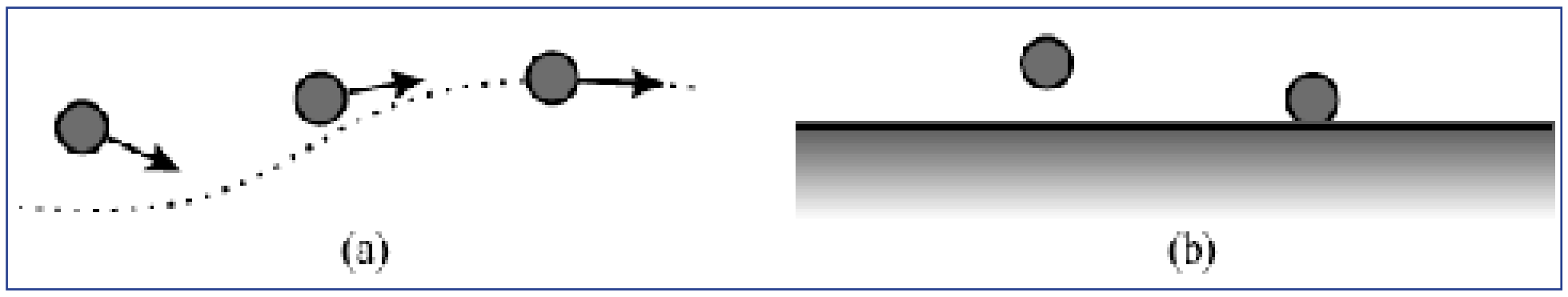

Fig. 32: (a) Guidance VFs and (b) Forbidden VFs

Our major accomplishments in this area include: (1) physical guidance using virtual fixtures; (2) development of a uniform mathematical framework for virtual fixtures incorporating both prior models and intraoperative information and permitting complex manipulation strategies to be built up from simple primitives; (3) improvements in haptic feedback and visualization during surgery; and (4) recognition of user actions and intent. In the remainder of this section, we briefly describe some of these achievements.

### *Virtual Fixtures*

Haptic virtual fixtures are software-generated force and position signals applied to human operators in order to improve the safety, accuracy, and speed of robot-assisted manipulation tasks. Dr. Russell Taylor, Dr. Gregory Hager, and Dr. Allison Okamura have led the establishment of the virtual fixture approach in computer-integrated surgery, with significant results in virtual fixture design, analysis, and implementation. We define two categories of virtual fixtures: guidance virtual fixtures, which assist the user in moving the manipulator along desired paths or surfaces in the workspace, and forbidden-region virtual fixtures, which prevent the manipulator from entering into forbidden regions of the workspace. Virtual fixtures have been used in both cooperative manipulation and telemanipulation systems, considering issues related to stability, human modeling, and applications. In cooperative manipulation, the human uses a robotic device to directly manipulate an environment. In telemanipulation, a human operator manipulates a master robotic device, and a patient-side robot manipulates an environment while following the commands of the master.

Virtual fixtures help humans perform robot-assisted manipulation tasks by limiting movement into restricted regions and/or influencing movement along desired paths. To visualize the benefits of virtual fixtures, consider a common physical fixture: a ruler. A straight line drawn by a human with the help of a ruler is drawn faster and straighter than a line drawn freehand. Similarly, a robot can apply forces or positions to a human operator to help him or her draw a straight line. However, a robot, or haptic device, has the additional flexibility to provide assistance of varying type, level, and geometry. Traditional cooperative manipulation or telemanipulation systems make up for many of the limitations of autonomous robots (e.g., limitations in artificial intelligence, sensor-data interpretation, and environment modeling), but the performance of such systems is still fundamentally constrained by human capabilities. Virtual fixtures, on the other hand, provide an excellent balance between autonomy and direct human control (Fig. 33). Virtual fixtures can act as safety constraints by keeping the manipulator from entering into potentially dangerous regions of the workspace, or as macros that assist a human user in carrying out a structured task. Applications for virtual fixtures include not only robot-assisted surgery, but also assembly tasks, inspection, and manipulation tasks in dangerous environments.

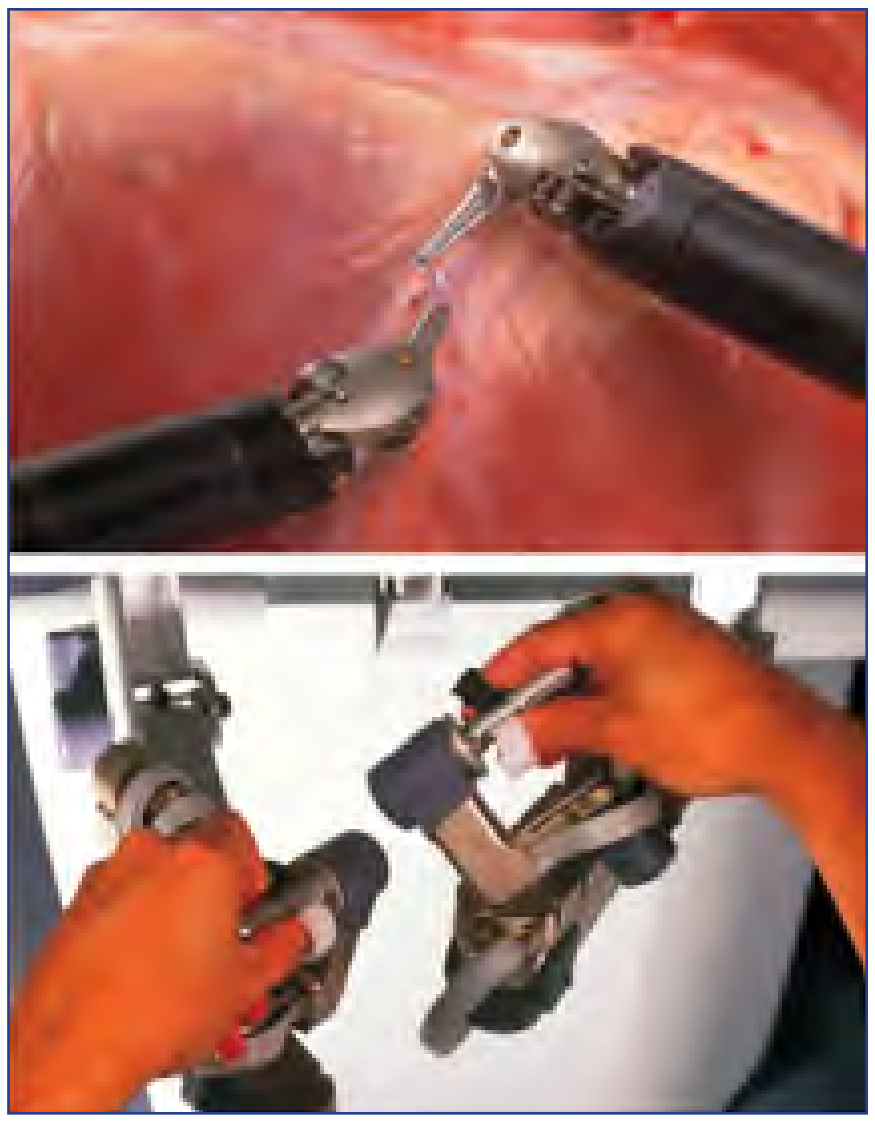

Fig. 33: JHU SHR and da Vinci

### *Ultrasound Calibration*

We have made three distinct contributions to the state of the art in ultrasound calibration. First, we have developed novel phantom and insonification sequence for rapid calibration with nonlinear optimization. Second, we have introduced a novel mathematical formulation for real-time calibration that obviates nonlinear optimization. Third, we have provided enhanced safety by developing an in-vivo quality control method to identify registration failures in a tracked US system during clinical procedures. In effect, we dynamically recalibrate the tracked US system. We detect any unexpected change by observing discrepancies in the resulting calibration matrix, thereby assuring quality and thus patient safety. We perform quality control in the background, transparent to the clinical user, while the patient is being scanned. Our system utilizes speckle tracking method based on the sum of squared differences (SSD). A unique and fundamentally important "side effect" of this work is that we now continuously calibrate the tracked US system during use, instead of using traditional off-line calibration on phantoms that may result in up to 5 mm depth error on patients of various constitutions.

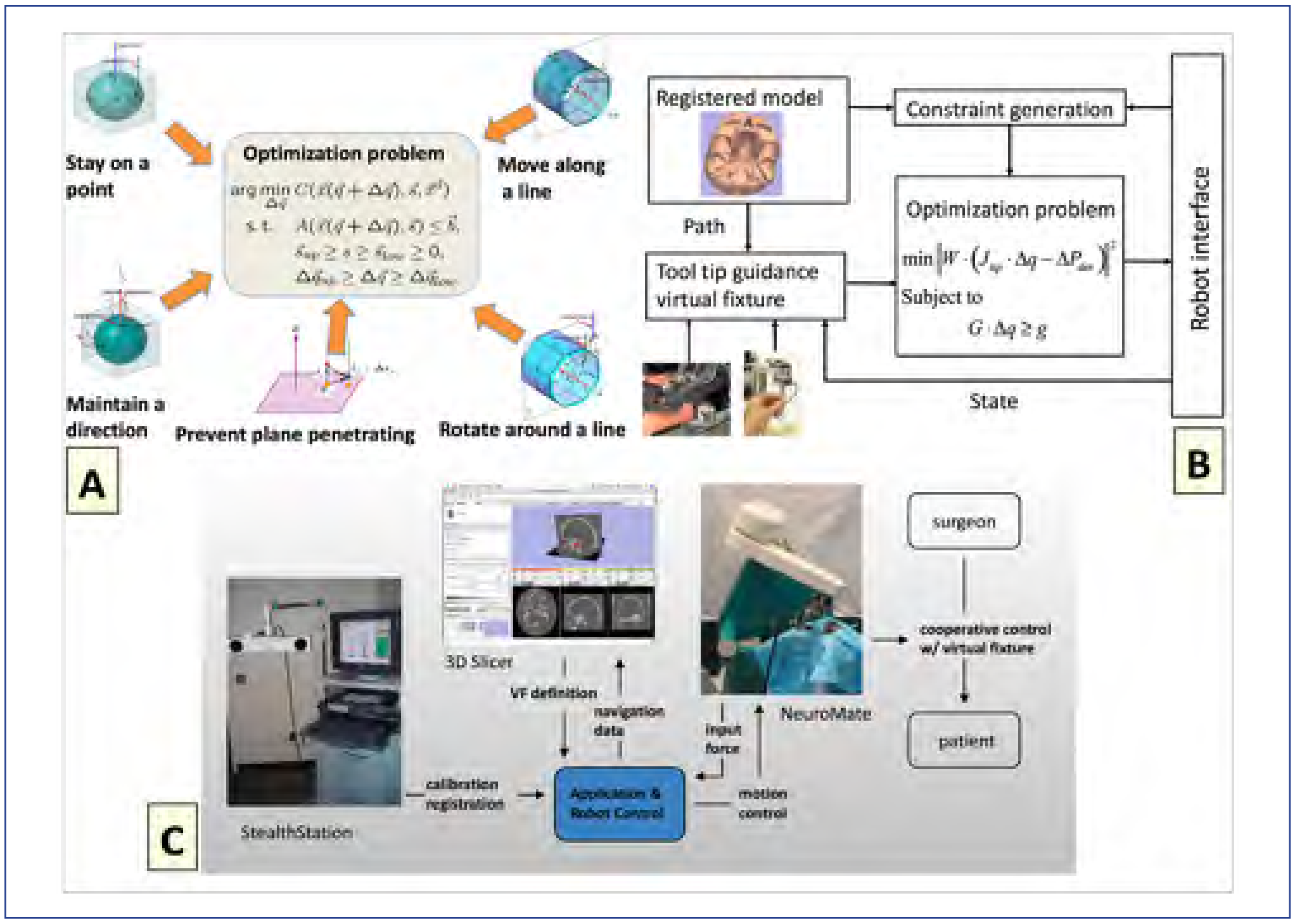

Fig. 34: Optimization-based virtual fixtures. A) Virtual fixtures for complex task steps can be synthesized by combining constraints and optimization criteria from multiple simple primitives. B) These virtual fixtures can be combined with additional constraints and optimization criteria derived from anatomic models, robot state, surgeon input, and real time sensor data. C) This example illustrates the use of virtual fixtures derived from preoperative medical images to enforce safety barriers in robotically assisted neurosurgery.

### *A mathematical toolkit for virtual fixture design*

We have developed a uniform mathematical framework for describing and implementing virtual fixtures. Briefly, we formulate the motion control of the robot as a quadratic optimization problem, in which constraints and optimization criteria are combined from multiple sources. These include: 1) joint limits and other kinematic constraints associated with the robot; 2) surgeon commands from a master hand controller, "hands-on" force sensor, or other input device; 3) real vision or other sensor data; 4) descriptions of desired behavior built up from simple primitives; and 5) registered anatomic models of the patient. We have demonstrated that these problems can be created and solved at interactive rates, typically from 20 Hz to 200 Hz depending on problem size, on modern surgical workstations. We have applied this formulation to a wide range of problems including teleoperation of complex robots such as our Snake robots, assistance in tasks such as suturing, implementation of safety regions in neurosurgery, and alignment aids in targeting tasks.

### *Haptic feedback for robot-assisted surgery*

Currently, a problem with commercially available robot-assisted surgical systems is the lack of significant haptic, or touch feedback for the surgeon, because they are not using their hands to operate on the patient directly. A technology allowing surgeons to receive more feedback from remotely controlled robotic tools would greatly enhance surgical procedures as well as the acceptance of teleoperated surgical-assistant robots by practitioners.

To help solve this problem, Dr. Allison Okamura has developed new feedback methods that allow a surgeon to physically feel surgical-environment forces relayed from his or her robotic tools, while eliminating the undesired resistant forces of the robotic manipulator itself, such as inertia and friction (Fig.35). With conventional robotic controllers, surgeons receive both kinds of information from the manipulator. Using the new haptic-feedback technology the surgeon will receive information directly from the surgical environment, without also receiving unwanted feedback from forces due to friction or inertia within the robotic tool itself. Through collaboration with Intuitive Surgical, Inc., the new haptic-feedback methods have been tested on a version of the da Vinci Surgical System (Figure 36), using customized software and hardware developed at the CISST ERC.

*Automated Recognition and Quality Assessment of Surgical Techniques*

One important aspect of effective assistance is not just knowing what to do, but also when to do it. The virtual fixtures described above provide physical assistance, but to be applied effectively they require contextual knowledge, an understanding of what the surgeon is trying to do. Dr. Gregory Hager has led a team to apply statistical modeling techniques, originally developed for speech recognition, to recognize common surgical maneuvers. The hypothesis is that a small basis set of surgical skills such as suturing, retraction, dissection and so forth, form the underlying vocabulary for the larger language of surgery. These skills not only provide the context for automated, appropriate assistance, but also provide a structured, objective basis for assessing the overall technical ability of a surgeon. Automated methods for recognizing basic manipulation skills can also provide a means for contextual information feedback and automated annotation of physiological or video data from a surgery. Furthermore, they can be used to provide diagnostic feedback to a surgeon-in-training, thus enabling automated, "intelligent" tutoring of surgical technique.

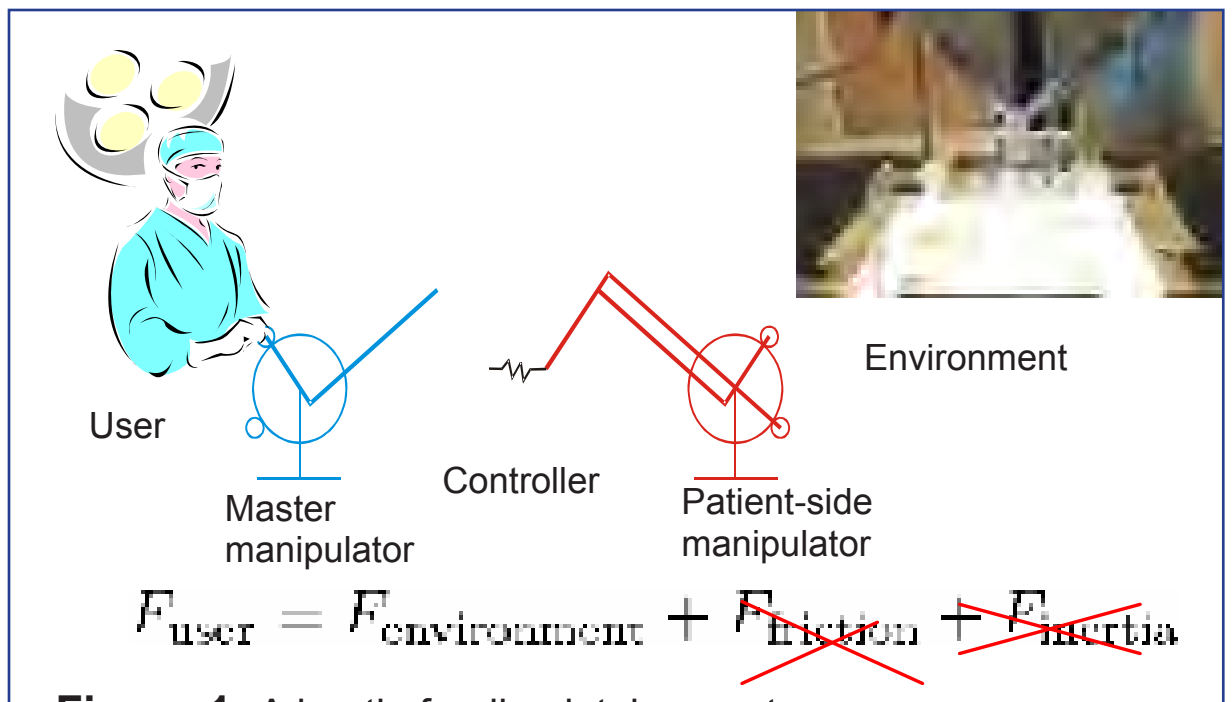

**Figure 1:** A haptic-feedback teleoperator

Fig. 35: A haptic-feedback teleoperator

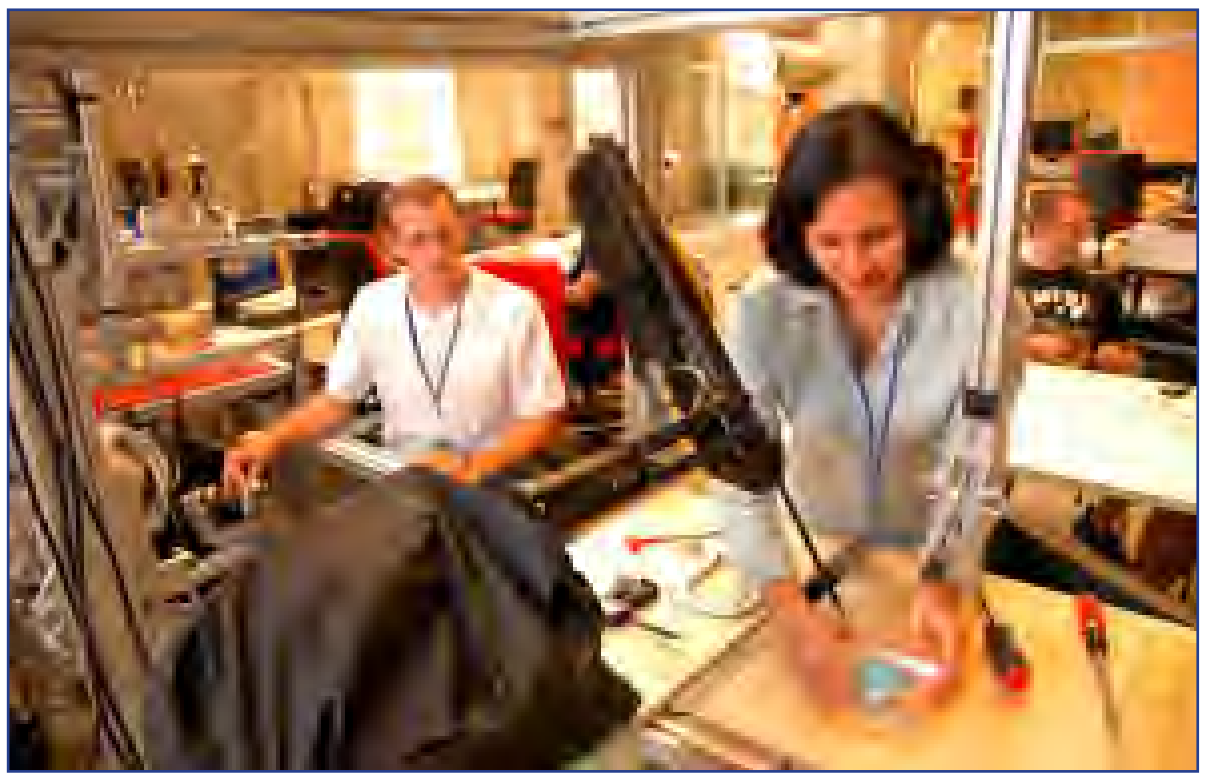

Fig. 36

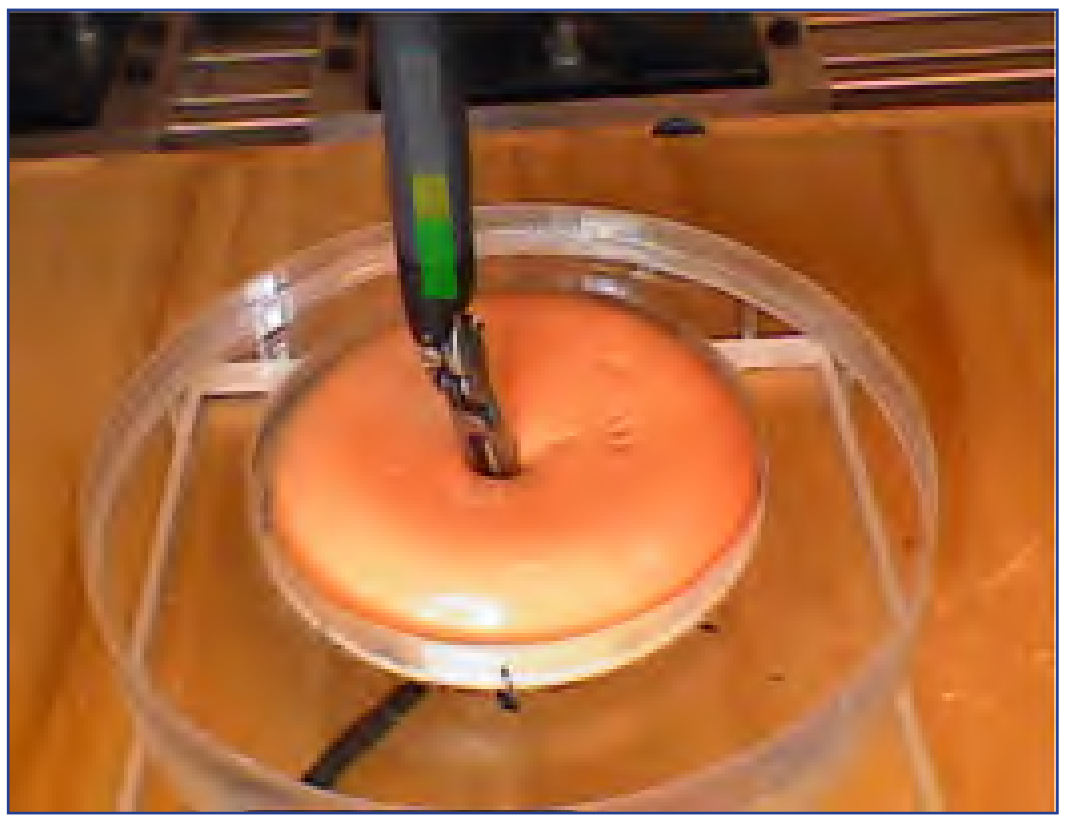

Fig. 37

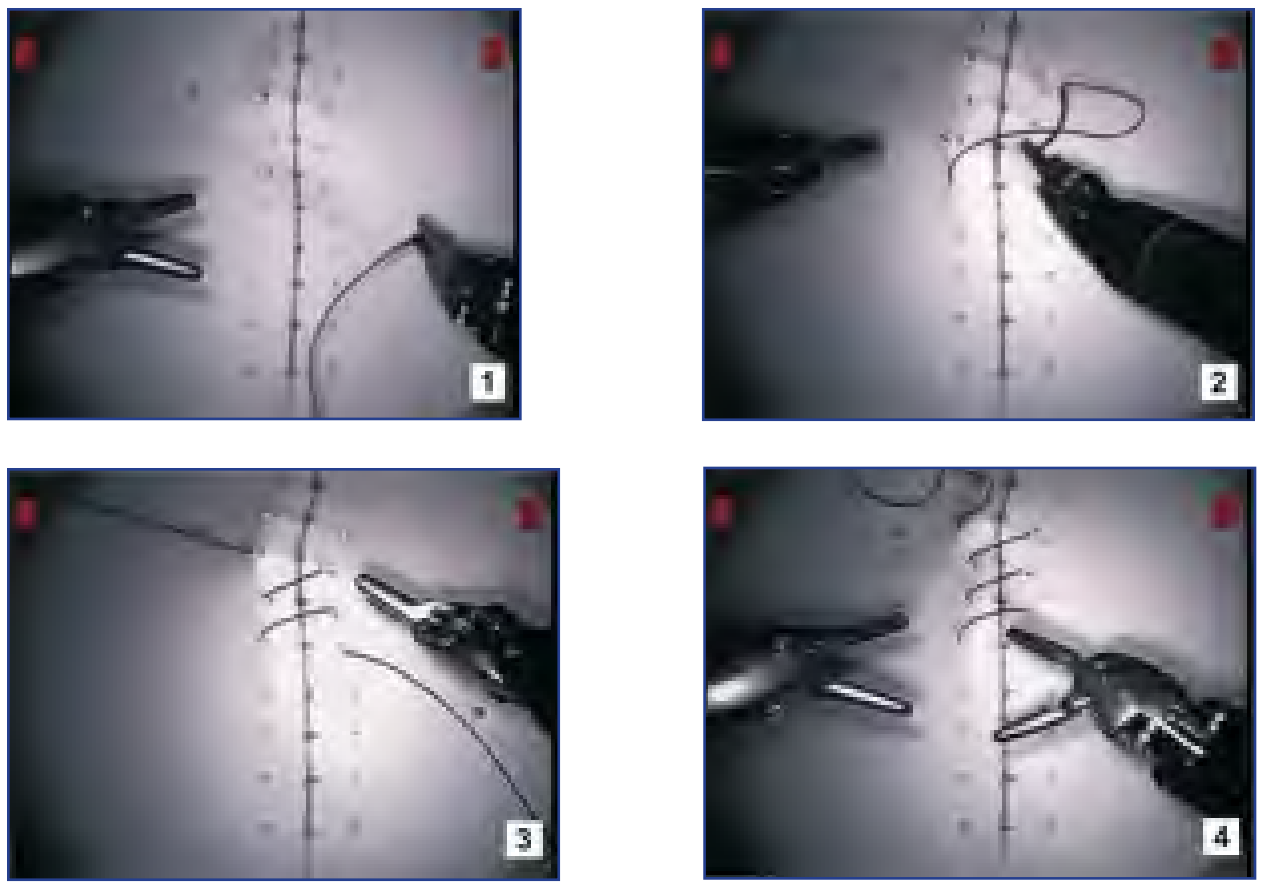

Fig. 38: Suturing task for recognition and evaluation: (1) Retrieving the needle from the starting position. 2) Inserting the needle with the right tool. (3) Pulling the suture through the sheet with the left tool. (4) The starting and ending position; the task is complete. Eight task "phases" were used in the segmentation.

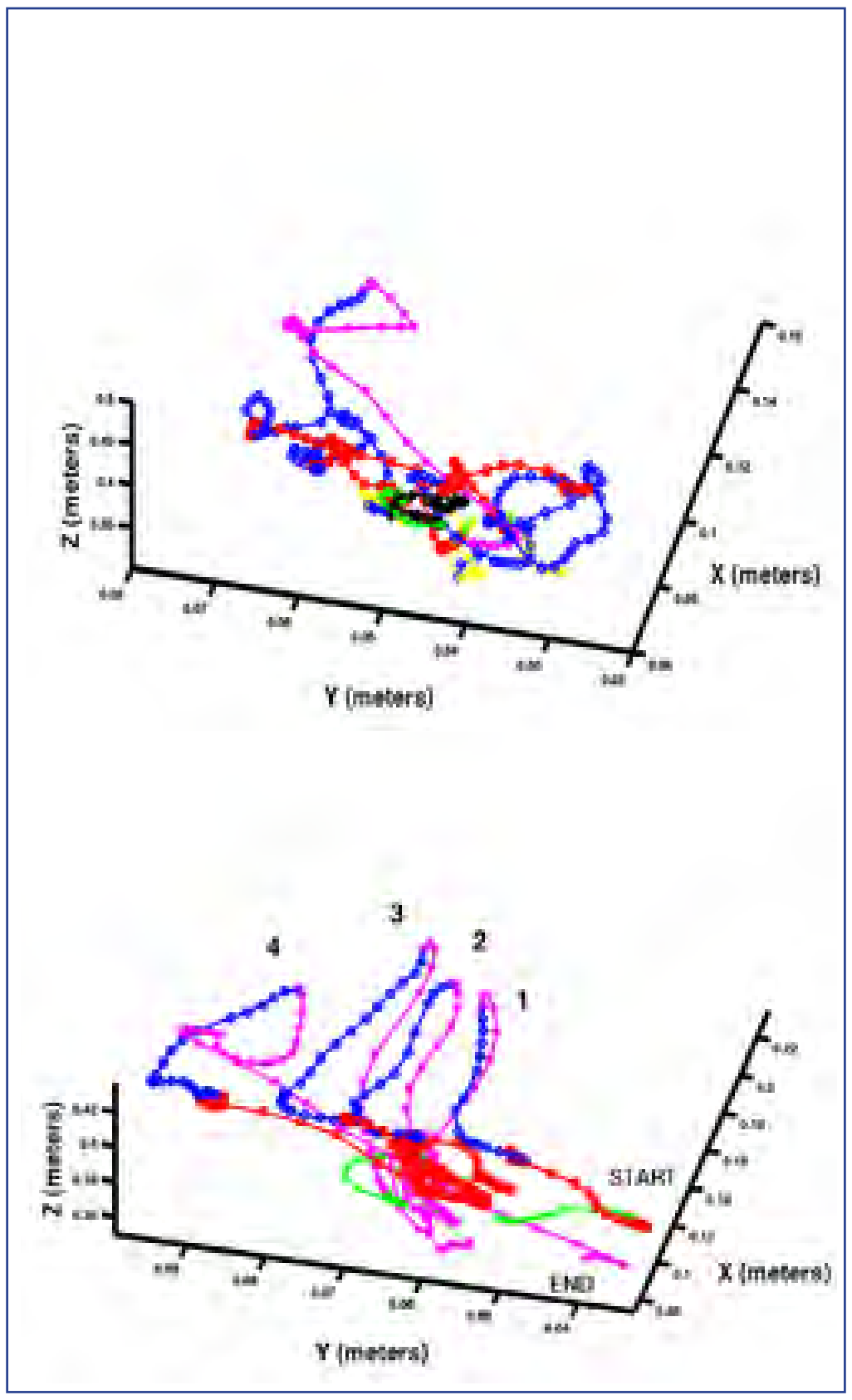

Fig. 39: Intermediate (top) and expert (bottom) surgeon performing suturing.

In the course of this project, it has emerged that Hidden Markov Models (HMMs), which have historically been used for modeling and recognizing human speech, are appropriate for automatic segmentation and recognition of user motions. Thus, it is reasonable to speak of a "Language of Surgery."

Dr. Hager and his team are in the process of further developing language-based methods for recognizing and assessing this language. Their most recent work has made use of data acquired using the da Vinci API (Application Programmer

Interface) furnished by ERC corporate partner Intuitive Surgical, Inc. Two representative data traces, one from an expert and one from an intermediate operator, are shown in Fig. 39.

With this data, they have shown that:

1. The most effective data for recognizing action is based on changes in contact state of the tools. This information appears to yield operator and task independent recognition of motion with very high (> 90%) recognition rates.

2. Some specific gestures are highly indicative of skill level, whereas others are relatively independent of skill.

3. The overall skill level of the expert can also be seen in the use of "surgical language," with expert users being far more efficient than novices.

An interesting example of the latter is shown in Fig. 40.

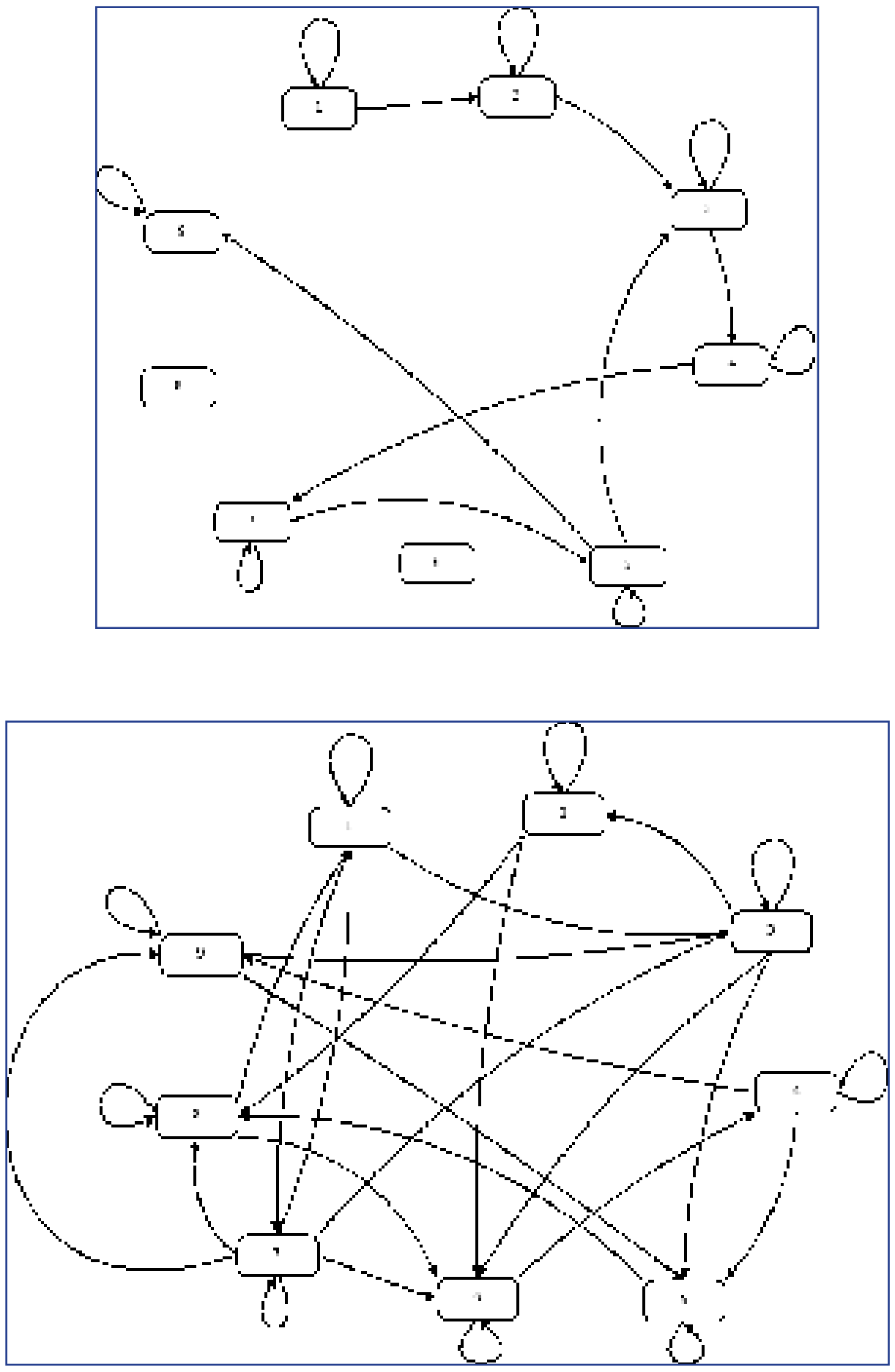

Fig. 40: The state transition diagram for an expert (top) and novice (bottom) user performing suturing

## *Image-Guided Systems*

In almost any intervention, clinicians need to see where they are going. One of the key advantages of a computer-integrated surgical system is enhancement of a clinician's ability to connect images from the patient to the real patient during the procedure, enabling him or her to perform an intervention with increased accuracy and confidence. In many cases, this task can be further improved with robotic guidance. In this section, we showcase some of our unique contributions in image guidance.

*Image Overlay for In-Scanner Interventions*

In collaboration with Dr. Ken Masamune of Tokyo and the Siemens Corporation, the CISST ERC has developed a practical, inexpensive 2D image overlay system to simplify, and increase the precision of, image-guided needle placements using conventional CT scanners.

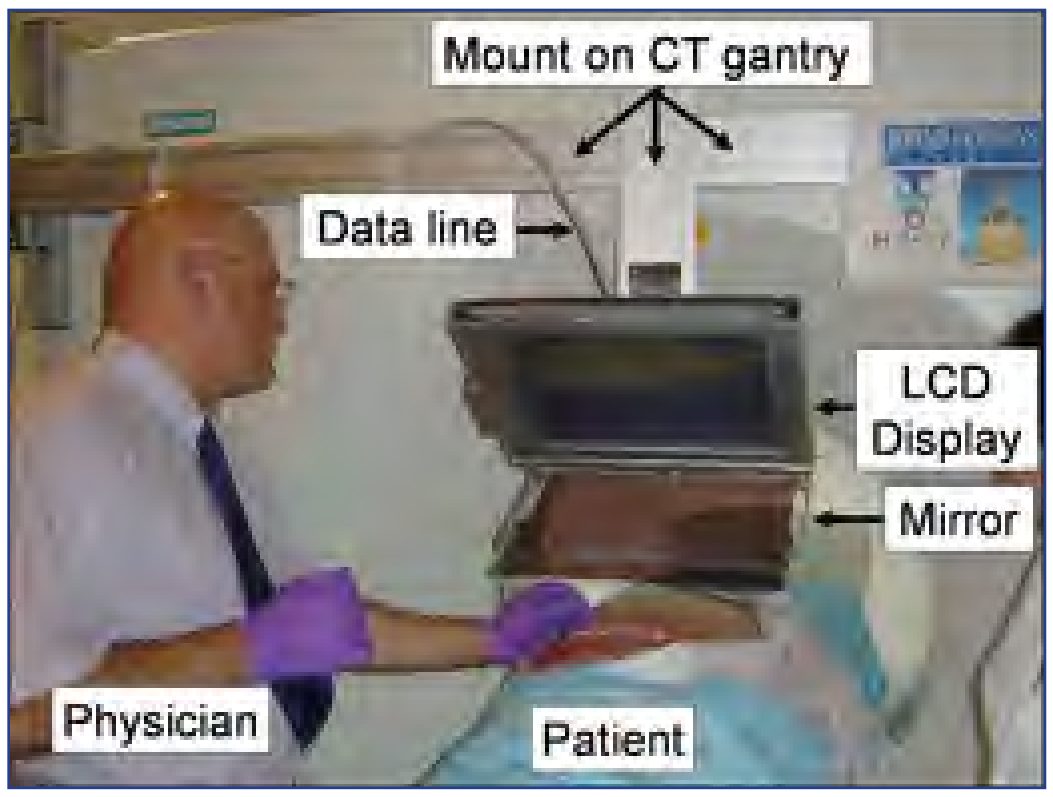

Fig. 41: James Zinreich, MD, performs a needle insertion experiment with the overlay device on a human cadaver

Numerous studies have demonstrated the potential efficacy of image-guided needle-based therapy and biopsy in a wide variety of medical procedures. However, most systems currently in use result in significant variability among practitioners with respect to accuracy and procedure time. Typically, the images are shown on the scanner's console, and the physician must then mentally register the images with the anatomy of the actual patient. A variety of virtual reality methods, such as head-mounted displays, video projections, and volumetric image overlay have been investigated, but all these require elaborate calibration, registration, and spatial tracking of all actors

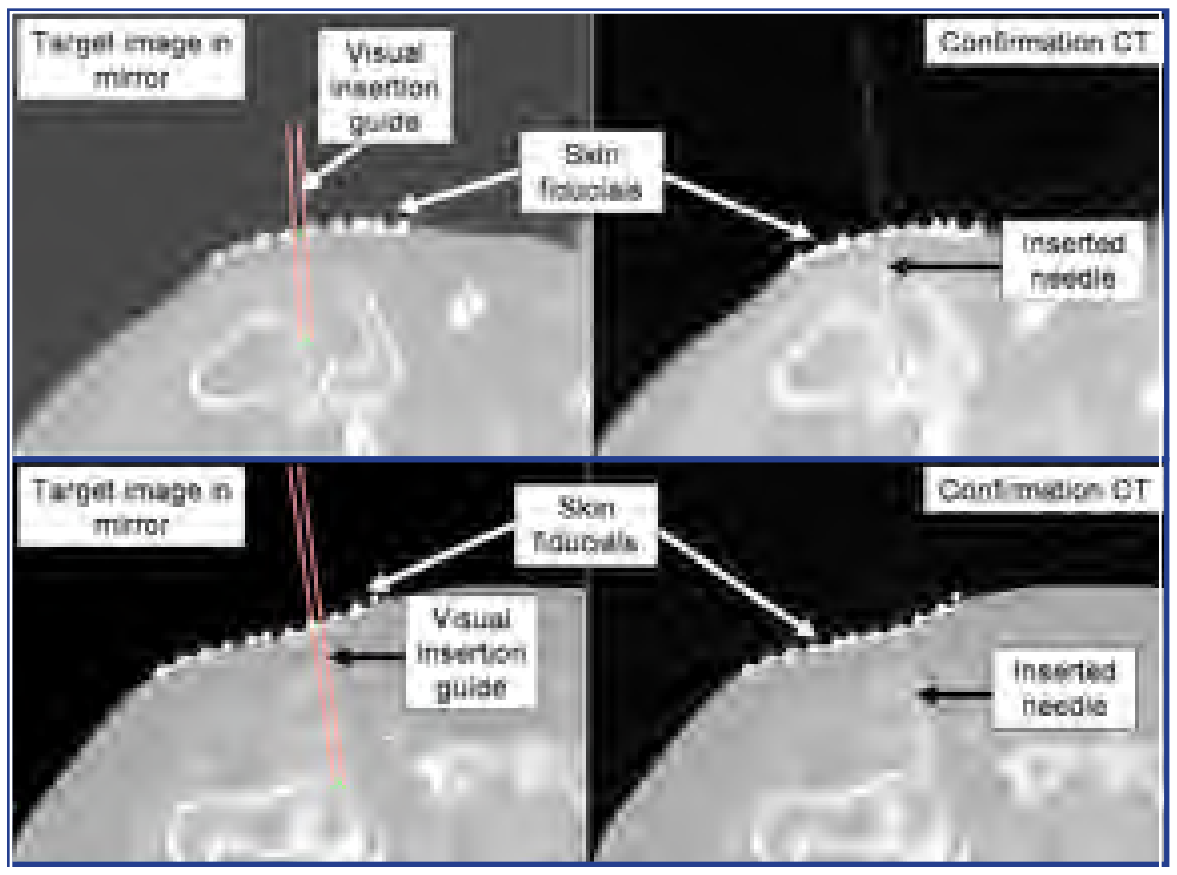

Fig. 42: Target and confirmation CTs in shoulder (top) and hip joint (bottom) arthography, proving millimeter-level accuracy

and components. This creates a rather complex and expensive engineered system.

The device developed at the CISST ERC consists of a flat LCD display and a half mirror, mounted on the gantry (see Fig. 41). When the practitioner looks at the patient through the mirror, the CT image appears to be floating inside the patient with correct size and position, thereby providing the physician with two-dimensional "x-ray vision" to guide needle placement procedures. The physician inserts the needle following the optimal path identified in the CT image, which is reflected in the mirror. The system increases needle placement accuracy and reduces the x-ray dose, patient discomfort, and procedure time by eliminating faulty insertion attempts. Cadaver studies have been conducted for several applications with a clinically applicable device. Dr. Laura Fayad at the Johns Hopkins Medical Institution has also performed joint arthography of the shoulder and hip joints, achieving millimeter-level accuracy in needle placement (see Fig. 42.) An IRB application for use of the CT-guided system and an MRI compatible prototype in humans is under development.

*Intra-Operative Dosimetry in Prostate Brachytherapy*

In many procedures, effective visualization of the treatment site allows a clinician to assess and adjust a pre-operative plan. For example, although numerous studies have demonstrated the efficacy and safety of prostate radioactive seed implants (a.k.a. brachytherapy), its success chiefly depends on the ability to intra-operatively tailor the radiation dose to the patient's individual anatomy, to adequately "cover" the prostate with radiation while avoiding too much radiation to surrounding organs. In contemporary practice, however, this level of precision is not always achievable. Many implants fail or cause severe side effects owing to faulty seed placement that cannot be identified or corrected in the operating room.

At our ERC, a multi-disciplinary team headed by Dr. Gabor Fichtinger has developed a method for intra-operative localization of the implanted seeds in relation to the prostate, to allow for in-situ dosimetric optimization and exit dosimetry. Their approach is to fuse ultrasound, which can view the prostate but not the seeds, with x-ray fluoroscopy, which is capable of viewing the seeds but not the prostate. Computer software identifies the seeds on the x-ray and projects their locations onto the ultrasound, allowing the physician to see exactly where the seeds are in relation to the prostate. This means that during the procedure the physician is able to optimize the planned seed locations and add additional seeds, thereby achieving an ideal seed distribution.

Thus, the team has developed novel methods for image-based fluoroscopy tracking with a specially designed radiographic fiducial and the reconstruction of implanted radioactive seeds. They have completed phase-1 clinical trials and their phase-2 protocol has been approved. In this regard, CISST is collaborating with one of its industrial partners, Acoustic MedSystems led by Dr. E. Clif Burdette, to integrate the CISST system with an FDA-approved prostate brachytherapy system in order to create a robust, commercially viable tool that is deployable in multi-center trials.

*Ultrasound Strain & Speckle Imaging for the Monitoring of Thermal Ablation*

Ablation of tumors is gaining momentum as a minimally invasive treatment option for some cancers. However, monitoring the ablation process in order to document the adequacy of treatment margins is a significant problem. Some ablative devices employ temperature monitoring using thermistors built within the ablation probes, but these only provide a crude estimate of the zone of ablation. MRI is too slow and expensive, and conventional ultrasound is not sufficiently sensitive. A promising approach our group is investigating capitalizes on the changes in tissue elastic properties that occur during tissue heating and protein denaturation.

The process of computing tissue stiffness with ultrasound, referred to as "elastography", involves the computation of the motion of tissue as an external force is applied. The stiffer the tissue, the less it is displaced for a given applied force. As a result, it is possible to resolve isoechoic anatomical structures that differ in stiffness from surrounding tissue (e.g., tumors) that would otherwise be invisible in a traditional ultrasound image (Figure 43). The major challenge in elastography stems from the fact that it is not the tissue displacements themselves that are used in the computation, but rather the change in displacement as a function of depth within the tissue. In order to reliably recover change in displacement, the tissue displacements themselves must be computed with extremely high accuracy.

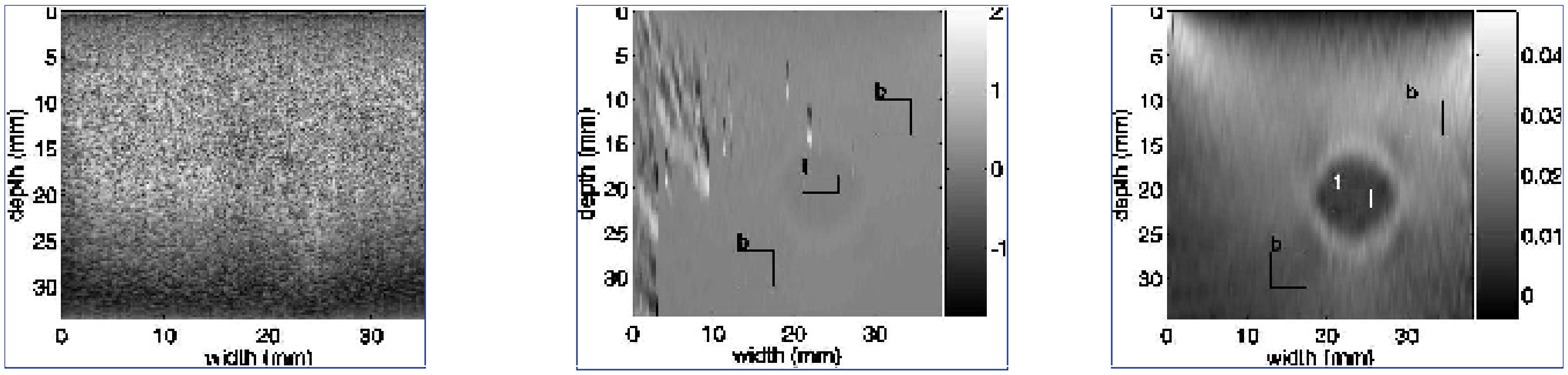

Fig. 43: Freehand strain images of elasticity breast phantom (using RF data). Our novel DP method, developed by investigators at the CISST ERC, shows robustness to noise and high-res definition (right.) Traditional NCC method shows substantial degradation of strain quality and low-res capability (middle). Note the isoechoic appearance of the lesion in the B-mode image (left.)

Dr. Emad Boctor, a former ERC Ph.D. student who is now appointed in Radiology, and Dr Gregory Hager have led a team that has developed a new method to achieve high quality elastography by efficiently computing very high resolution and high accuracy displacement maps. The approach makes use of an efficient global optimization technique known as dynamic programming. With this method, it is possible to compute displacements for each sample of the raw ultrasound data, RF data, without the artifacts introduced by the block structures of traditional methods. Further, it is possible to compute displacements to subsample precision and to include a constraint that enforces smoothness between samples. The method is quite efficient and can be computed to operate in near real time. Figure 43 shows the results comparing traditional matching methods with the new dynamic programming technique.

The new method shows great promise as both a diagnostic and an interventional imaging modality. Fig. 44 shows the results of computing elastography on an elasticity phantom made by CIRS Inc., locating lesions that would be invisible in normal CT images.

At the same time, we have begun to experiment with methods for directly monitoring the thermal heating of the tissue. As the tissue heats, the speed of sound through the tissue changes. These changes can be detected over time. Initial results using this method are shown below.

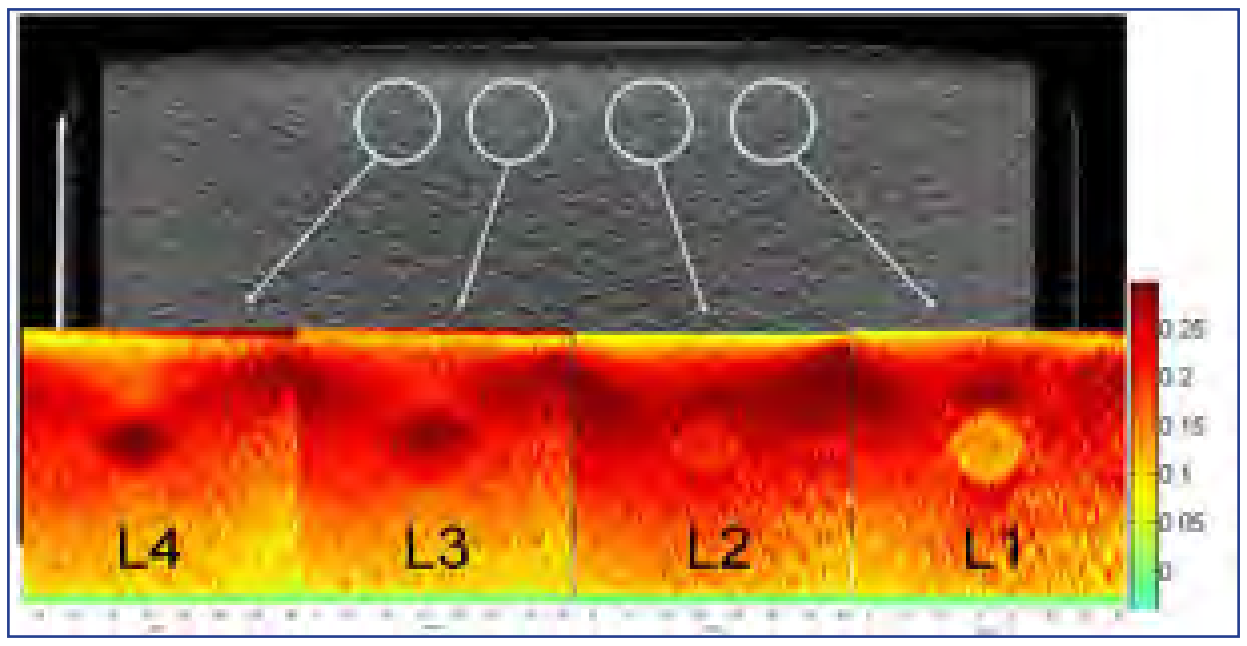

Fig. 44: CT (above) vs. US elastography (below) in elasticity phantom (CIRS Inc.). While CT shows the contours only, elasticity imaging is capable of better delineation and identification of relative stiffness of the four lesions. L1 is the hardest and L4 is the softest.

*Intraoperative Visualization of Anatomic Imagery*

Pre-operative volume imagery (CT or MR) not only are key diagnostic tools, but also provide guidance to a clinician during an intervention. Currently it is up to the clinician to register the pre-operative image to the surgical view. This can be particularly challenging during minimally invasive surgery due to the limited field of view, as well as the deformation of the organs in question.

We have been developing a series of tools for enhancing endoscopic views through automated registration of endoscopic video data to pre-operative MR or CT. Here, we describe a system we have implemented for the Intuitive Surgical da Vinci system.

The structure of endoscopic medical images is one of the major challenges in developing video-CT registration. Traditional real-time stereo systems produce very low data density and accuracy on stereo endoscopic images. In order to produce better quality results, we have developed a dynamic programming-based stereo system specially designed to operate effectively on video streams. This approach, which we term 4-D dynamic programming, makes strong use of both temporal and spatial continuity to improve data density, accuracy, and speed.

Given high-quality stereo data, it is possible to perform surface-to-surface registration of observed surfaces in video to surfaces segmented from MR or CT volumes. After an initialization phase, this registration is performed using an efficient locally adjusting ICP algorithm. Finally, a spring-model system is used to perform deformable registration. Several feature points are chosen and a tracking-based registration is performed. The latter has the advantage of a higher computational performance at the cost of slightly lower accuracy and robustness.

The complete system was integrated using a modular pipeline architecture that is part of the CISST Surgical Workstation libraries (further described below). The results of this system

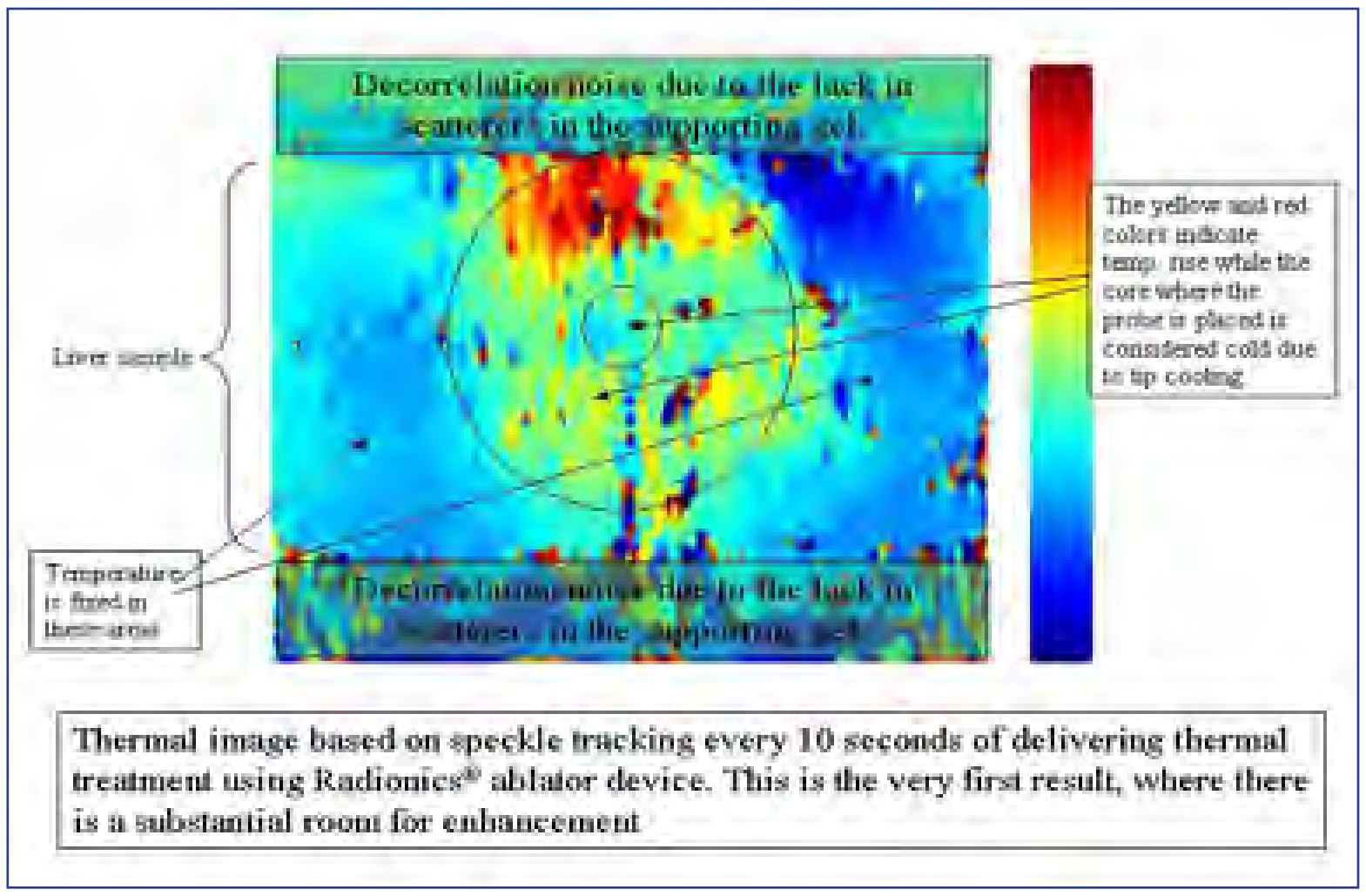

Fig. 45: Experimental setup and RF-image based speckle tracking of temperature shift

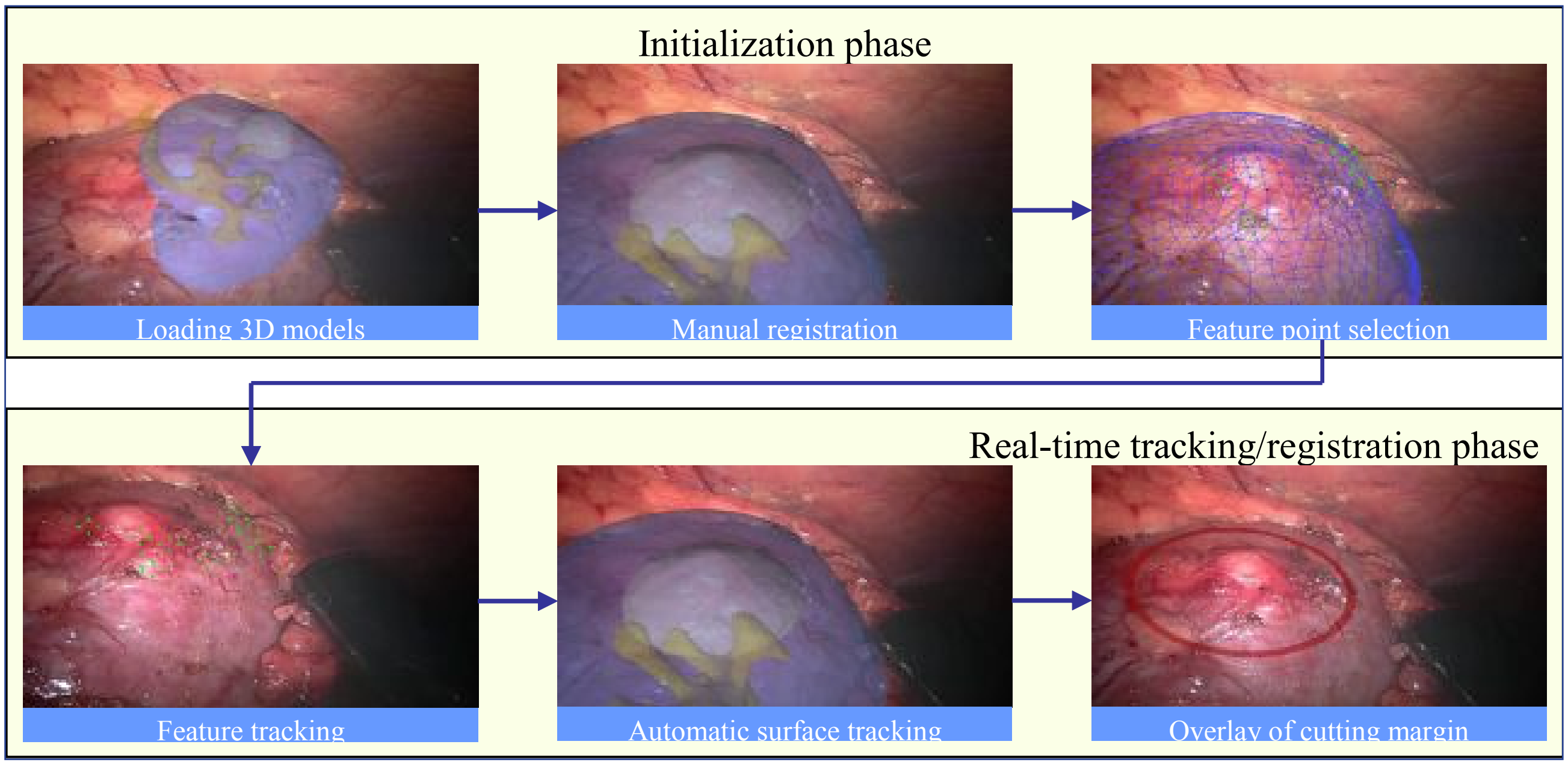

Fig. 46

for visualizing the dissection margins during laparoscopic nephrectomy is shown above:

Intraoperative ultrasound is another modality that can be exploited to aid in visualization during an intervention. In a NIH-funded Phase II STTR project, the CISST ERC, the JHU Department of Surgery, and Intuitive Surgical Systems (ISI) are working together to develop an integrated laparoscopic ultrasound capability for ISI's da Vinci surgical robot. Our immediate goals are to provide a laparoscopic ultrasound capability comparable to that currently available in open surgery and to enhance that capability through information fusion with the da Vinci visualization environment. The integrated laparoscopic ultrasound capability will provide the surgeon with a direct means for visualizing deeper structures within organs, overcoming the lack of tactile capability in minimally invasive surgery.

Liver surgery was selected as the initial clinical platform to develop and test this system, as this is where ultrasonography is most often clinically applied. The potential clinical applications are expansive including diagnostic and therapeutic procedures on the kidney, pancreas, bile ducts, intestinal tract, retroperitoneum, and mediastinum.

As of Summer 2008, an ISI-developed articulated laparoscopic probe has been integrated with a "S" model daVinci. A number of visualization paradigms have been prototyped using the "Surgical Assistant Workstation (SAW)" software being developed with NSF funding by the ERC and ISI. This laparoscopic ultrasound capability is a principal "use case" in the SAW development. Work is underway to develop advanced manipulation capabilities using the SAW infrastructure. This will allow the infrastructure to facilitate real time 2D and 3D ultrasound image acquisition, facilitate surgical tasks such as scanning solid organs for tumors, and aid in guiding the placement of biopsy needles and ablation probes. Preliminary surgeon evaluations for kidney and liver cancer procedures have been performed on the system at ISI, as well as procedures for gynecological applications. More formal evaluations comparing robotic system aided surgeon performance to freehand ultrasound are planned for fall and early winter.

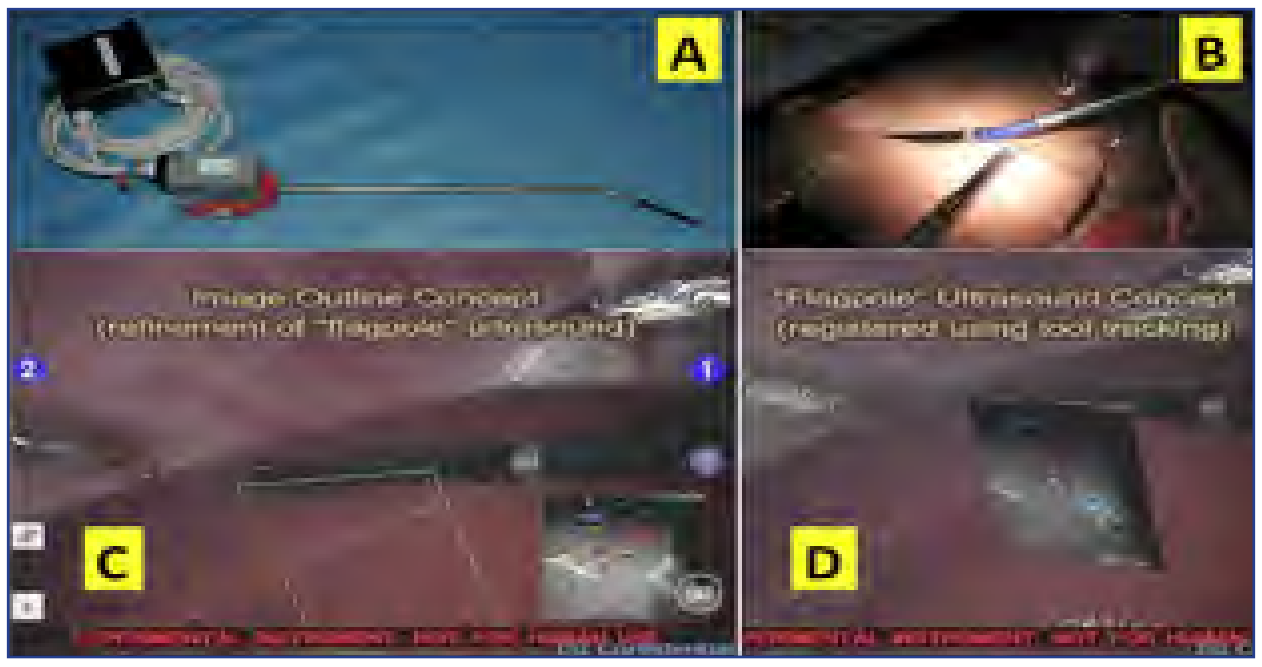

Fig. 47: daVinci Ultrasound system: A) dexterous laparoscopic ultrasound tool; B) intraoperative ultrasound scene; C, D) views through the daVinci stereo visualization system showing several video overlay concepts.

*Image-guided robotic systems for small animal research*

The potential impact of medical robotics and image-guided interventional systems on medicine extends well beyond their direct clinical application on human patients. The ability to accurately and repeatedly place needles or directed energy onto anatomic targets in laboratory animals can significantly enhance research into disease mechanisms and possible therapies. Preclinical research using well-characterized small animal models is providing tremendous benefits to medical research. These include enabling low cost, large scale trials with high statistical significance of observed effects. Although many mouse models of human cancer are currently available, existing clinical imaging and therapeutic systems are ill suited to such small subjects. The equipment is also in high demand and seldom available for lengthy laboratory trials.

Accordingly, the CISST ERC has collaborated in several research projects to develop image-guided robot systems for cancer research with small animals.

Image-guided needle placement: In a subcontract from Memorial Sloan Kettering Cancer Center (MSKCC) we have developed an image-guided needle-placement robot for small animals. The motivating problem was to automate and improve a tedious manual procedure used at MSKCC to identify hypoxic, oxygen-deficient tumor cells. Hypoxic cells are more resistant to radiation treatment, so a non-invasive identification method would allow clinicians to deliver higher radiation doses to those cells. MSKCC researchers have been validating the efficacy of PET image data, with an appropriate tracer, for identifying hypoxic cells by correlating multiple image data points with oxygen tension (pO2) measurements obtained invasively by a manually inserted probe.

The robot system consists of a mobile cart that houses the electronics, provides a tabletop for the four-axis robot and display monitor, and contains a pull-out drawer for the keyboard and mouse. The robot design consists of an X-Y platform that moves the rodent bed and a two-axis insertion stage (Z1 and Z2), as shown in Figure 48. The Z1 axis is used to position a cannula near the skin surface and the Z2 axis is used to drive the needle or measurement probe to the target. The measurement procedure is physically decoupled from the imaging procedure for maximum flexibility.

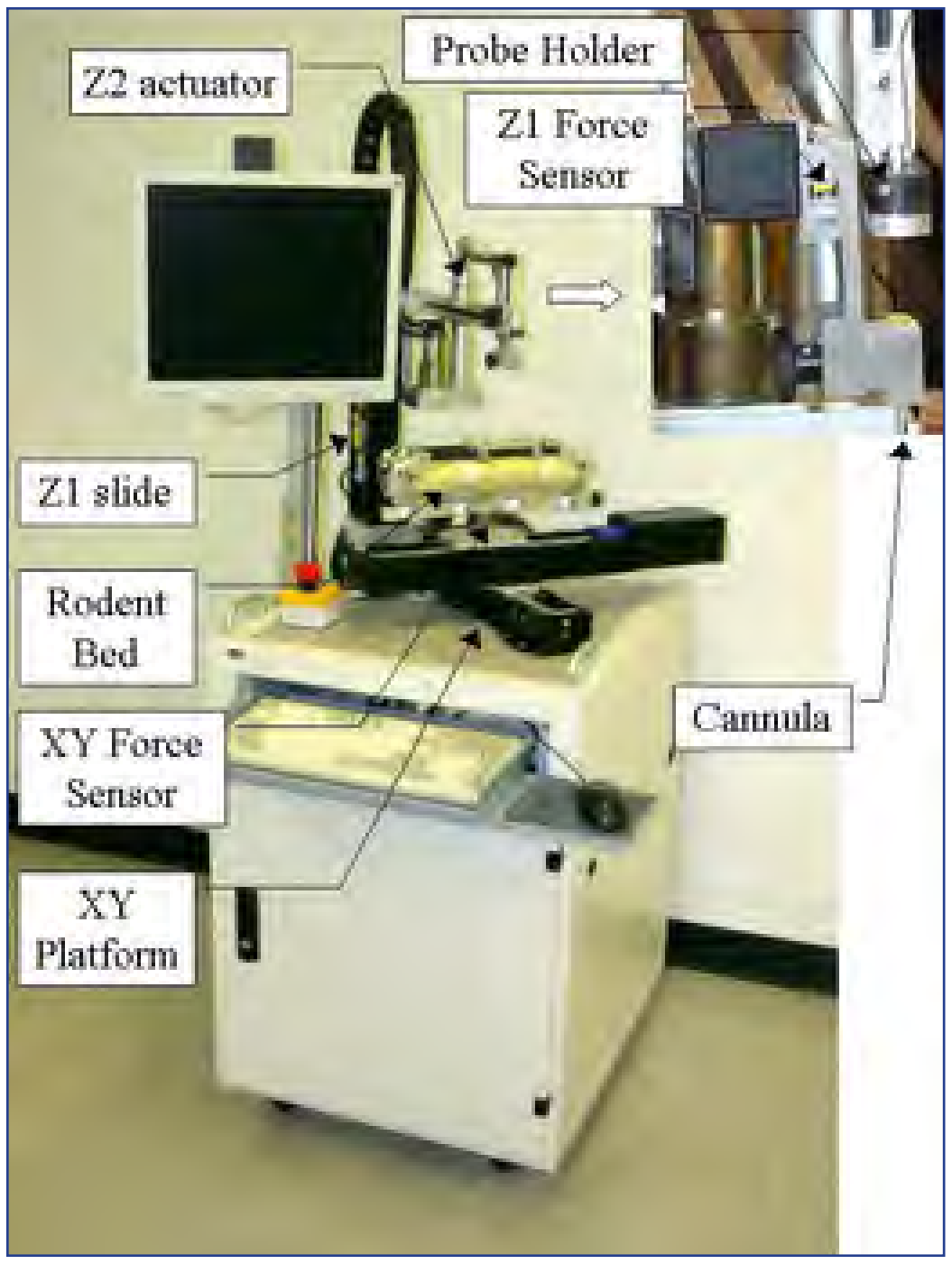

Fig. 48: Image-guided robot for small animal research

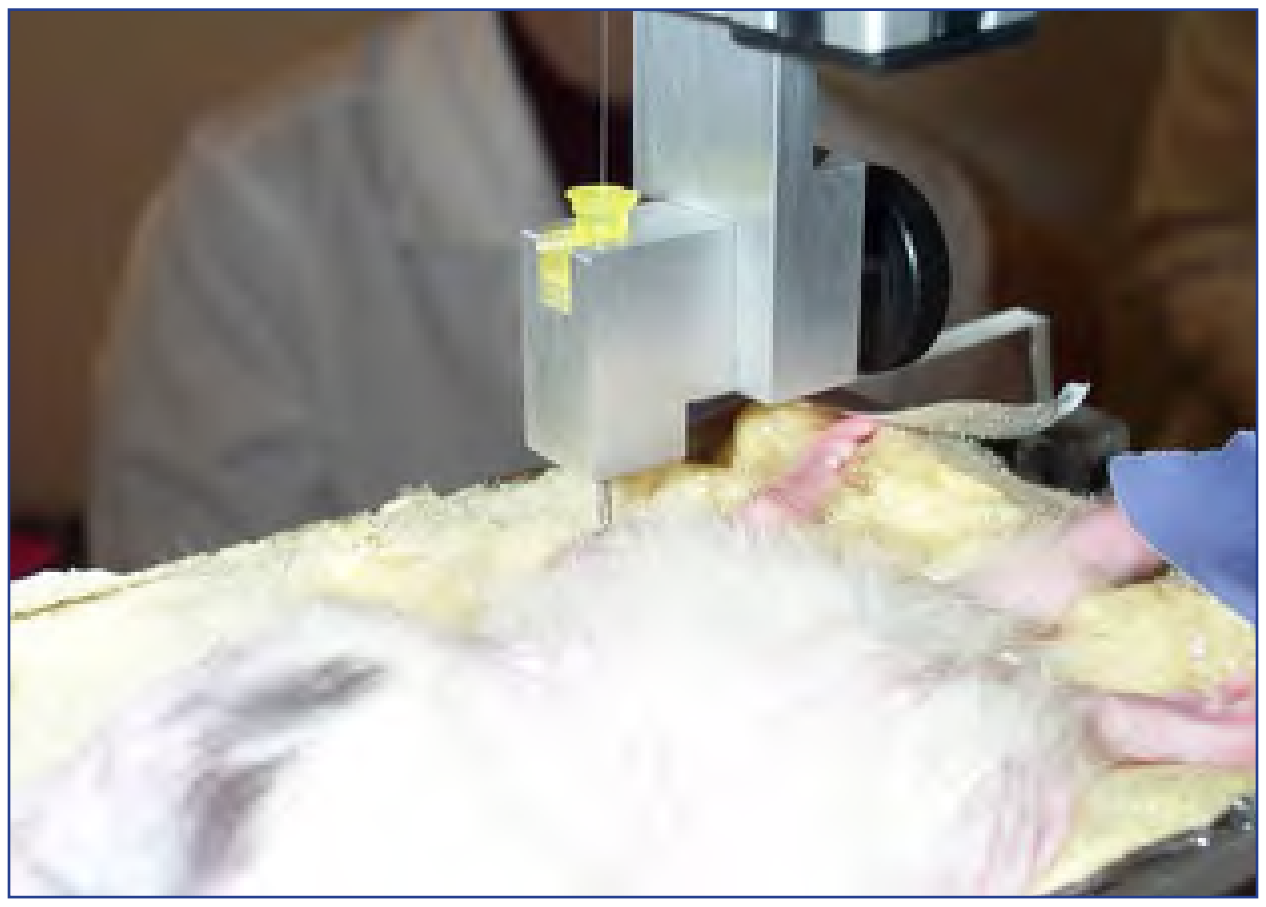

Fig. 49: Rodent experiment at MSKCC; robot is inserting measurement probe through cannula into tumor

Therefore, fiducial markers are used for the registration between image coordinates and robot coordinates. During the robot registration procedure, the cannula is replaced by a registration probe, which is guided to the markers using a force control mode. Force control is possible because the system contains a two-axis sensor (XY) beneath the rodent bed and a single-axis sensor (Z1) near the attachment mechanism for the registration probe and cannula.

The system was delivered to MSKCC in January 2005. Figure 49 shows a rodent experiment performed at MSKCC, where the robot placed a measurement probe at a set of points identified on a PET image. The robot system is expected to facilitate cancer research and further applications, such as biopsy and injection, using other imaging modalities, such as CT, MRI and SPECT.

*Small Animal Radiation Research Platform (SARRP):* This project addresses a significant need in radiation therapy research. At present, simple single beam/single fraction techniques are commonly used to irradiate laboratory animals. This technology is far removed from the advanced three-dimensional (3D) imaging, planning, and computer-controlled delivery technologies that are used for human treatment. The CISST ERC, in collaboration with the Department of Radiation Oncology and Molecular Radiation Sciences at the JHU, developed the Small Animal Radiation Research Platform (SARRP) to bridge this gap between laboratory radiation research and human treatment methods. The SARRP integrates cone-beam computed tomography (CBCT) imaging and conformal irradiation delivery to enable image-guided radiation research with small animals. This leverages the research infrastructure of the ERC, including the use of the CISST software libraries.

The system (Fig. 48 and 50) consists of a kilovoltage x-ray tube mounted on a manually rotated gantry. The tube provides a low-energy beam for CBCT imaging, with a flat panel image detector, and a high-energy beam for radiotherapy. The animal is placed on a four-axis robotic positioner with three translations and one rotation. CBCT imaging is performed by placing the x-ray tube in the horizontal position (as in Fig. 50 top), but without the collimator) and then rotating the animal while capturing 2D x-ray images. A standard reconstruction algorithm is used to obtain the 3D image. Radiation therapy can be delivered from any orientation of the gantry, using a variety of collimators, even as small as 0.5 mm in diameter. We developed a calibration method to enable accurate targeting of the x-ray beams. This method uses an x-ray camera, mounted on the robotic positioner, to measure the mechanical axis of rotation and the precise location of the x-ray beam at each gantry rotation. We chose a target on the center film and delivered a radiation beam, 1 mm diameter, from multiple gantry angles, 45 and 75 degrees from horizontal, and at 45 degree increments of the rotary stage. The center film shows the spot where all x-rays intersect.

The SARRP was installed in the School of Medicine in December 2006 and is used to support several research studies, including the response of normal tissue stem cells to focal radiation injuries, the development of positron emission tomography (PET) markers for early assessment of radiation induced toxicity in the lungs, and the study of molecularly targeted therapy in combination with radiation in pancreatic and prostate tumor models. A second system is currently being constructed to satisfy the high demand within JHU, while outside institutions have also expressed interest. We believe that the SARRP provides the infrastructure that will enable researchers to improve the treatment of cancer.

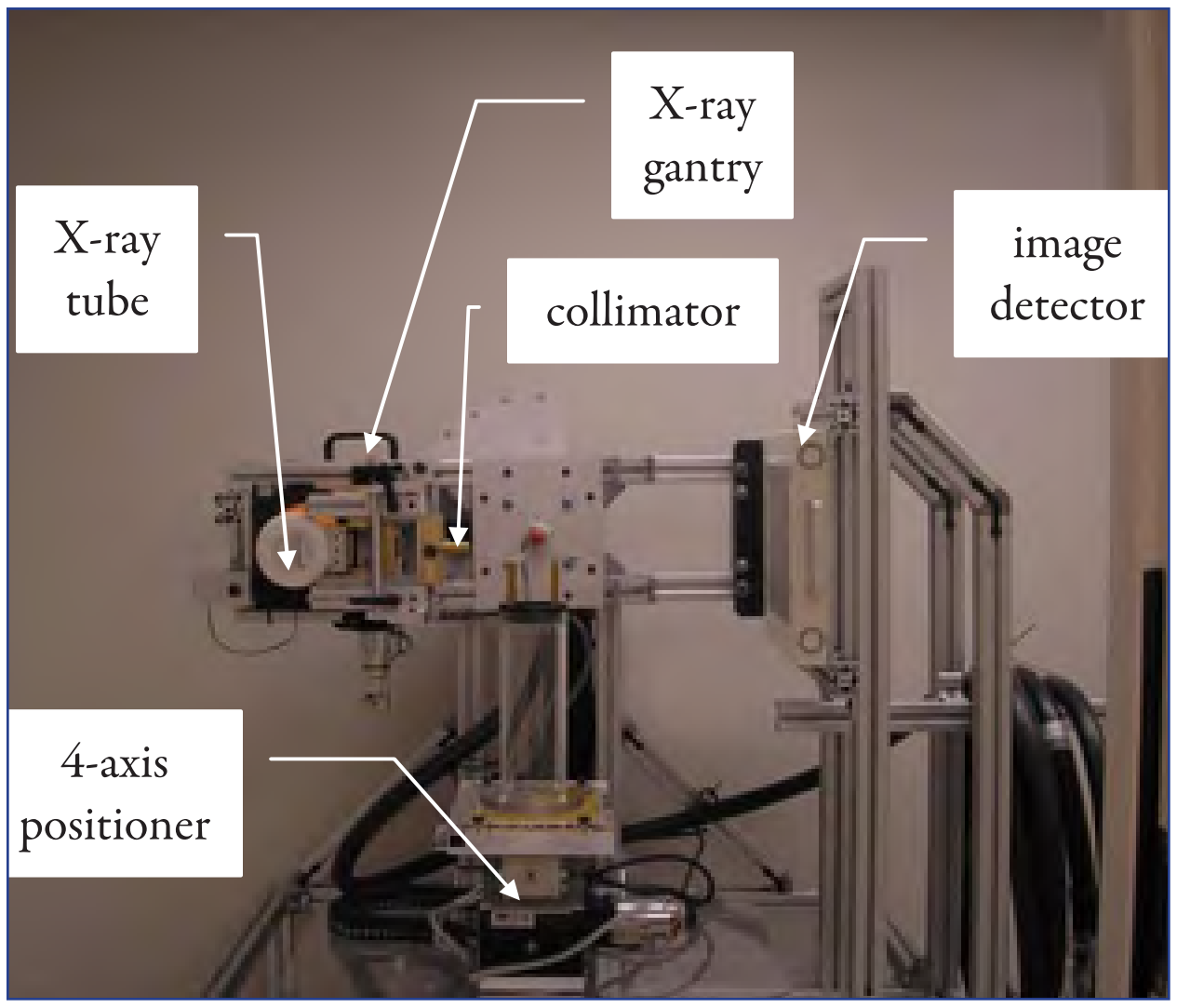

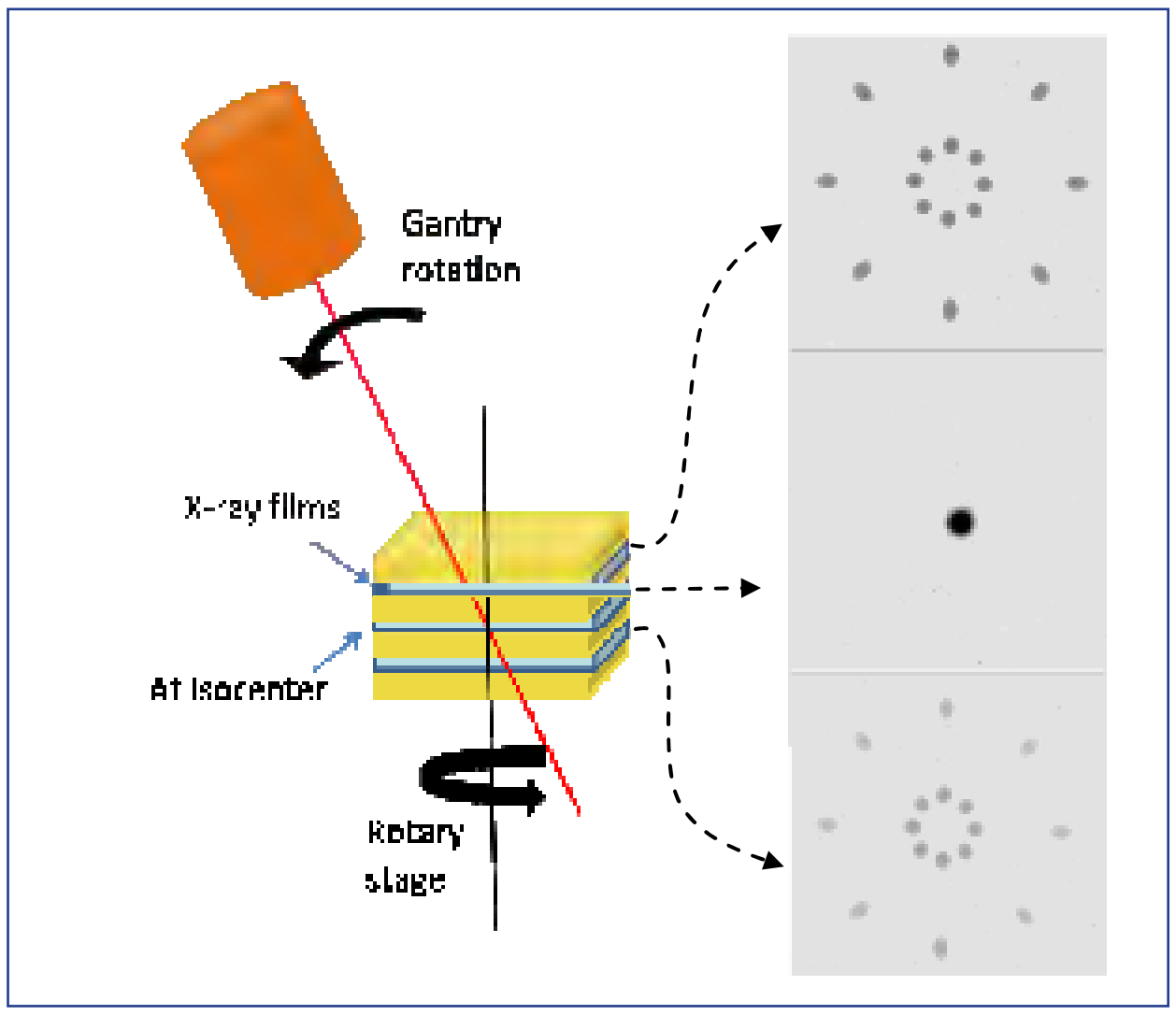

Fig. 50: Small Animal Radiation Research Platform. (Top) Basic system shown with radiation shields removed and with gantry in horizontal position (0°); (Bottom) Validation of beam accuracy using X-ray films at gantry angles of 45° and 75°, with 45° stage rotations.

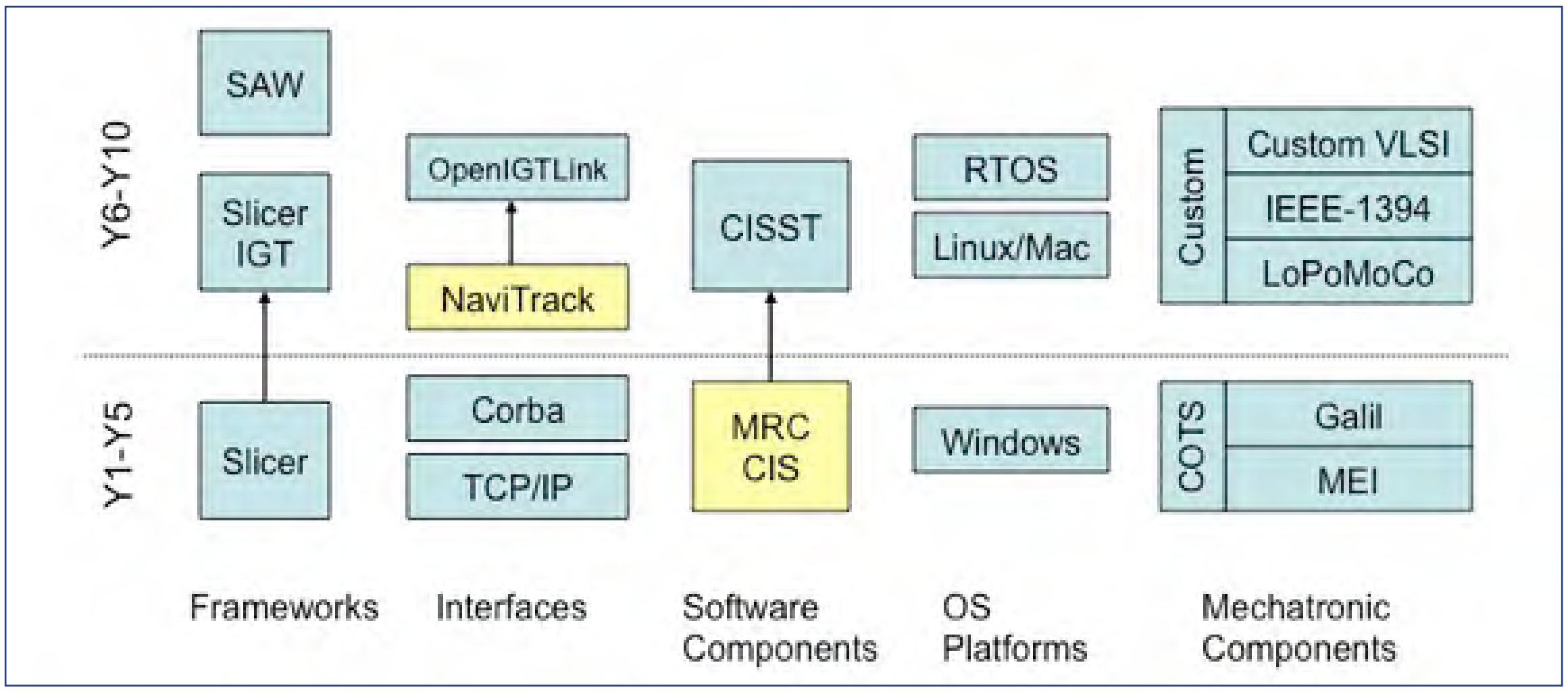

Fig. 51: CIS System Architecture: elements and evolution

## *Systems and Infrastructure*

The CISST ERC has developed a set of modular core hardware and software components. These components and related system development tools and libraries allow us to create new CIS systems quickly. We have great flexibility in tailoring and customizing these systems as new clinical needs emerge. In this section, we describe some of the key technologies we have developed, which are focused around the following themes: system architecture, software and mechatronic components, development process, and information systems.

### *System Architecture*

We define system architecture as “the fundamental organization of a system, embodied in its components and their relationships to each other, the environment, and the principles governing its design and evolution” (ANSI/IEEE 1471-2000). Thus, architecture relies on the components and development process, but focuses on the interfaces and high-level design patterns.

As shown in Fig. 51, our architecture initially revolved around the 3D Slicer open source software developed by the Surgical Planning Laboratory at BWH. Although the primary application for 3D Slicer is medical image visualization and associated surgical planning tasks, the ERC has been a primary driver of Slicer into image guided therapy (IGT), especially when involving robotic assistance. Although Slicer-IGT remains a key architectural resource, we are currently developing a Surgical Assistant Workstation (SAW) architecture that is the culmination of many years of experience developing computer-integrated surgery research testbeds. This was enabled by an ERC Supplement awarded to JHU and Intuitive Surgical Systems.

The SAW use cases include the integration of patient models and laparoscopic ultrasound within the stereo video display of a telesurgical robot (Fig. 52). The system includes a 3D user interface, where the "master" manipulators can be used to interact with graphical objects and menus. As a representative example, the Volume Viewer uses this functionality to enable the surgeon to interact with 3D datasets (e.g., CT or MRI images). The interface supports overlay of registered patient models with adjustable transparency onto the stereo display. This also allows for the manipulation of unregistered models (e.g., the ability to view CT cross-sections). Laparoscopic ultrasound images appear on a 2D plane that is correctly positioned within the 3D view. This registration is achieved by the Tool Tracking module which uses a combination of robot position feedback and visual tracking.

The SAW architecture includes the definition of a generic Robot API with specific implementations for each different robot. This abstraction layer allows researchers to develop portable software that is not dependent on a particular robot API. This will enable researchers to use the daVinci "master" manipulators to control other "patient side" manipulators.

System architecture also includes the middleware that connects the major subsystems (Fig. 51). Although TCP/IP sockets with custom protocols provide a practical solution, it is preferable to adopt a standard middleware solution. In 2000, researchers at JHU and BWH reported on a CORBA (Common Object Request Broker Architecture) interface between 3D Slicer, robots, and tracking systems. These experiences were mixed and thus subsequently researchers at Morgan State University performed a benchtop comparison of CORBA and SOAP (Simple Object Access Protocol). Their tests included an evaluation of latency, throughput, and memory usage for six representative message types. They concluded that CORBA was still a better choice than SOAP. In January 2008, the lack of a clear middleware solution for CIS systems led BWH, JHU and others to propose a new standard, OpenIGTLink, that is currently in development under the aegis of the National Alliance for Medical Image Computing (NA-MIC) consortium.

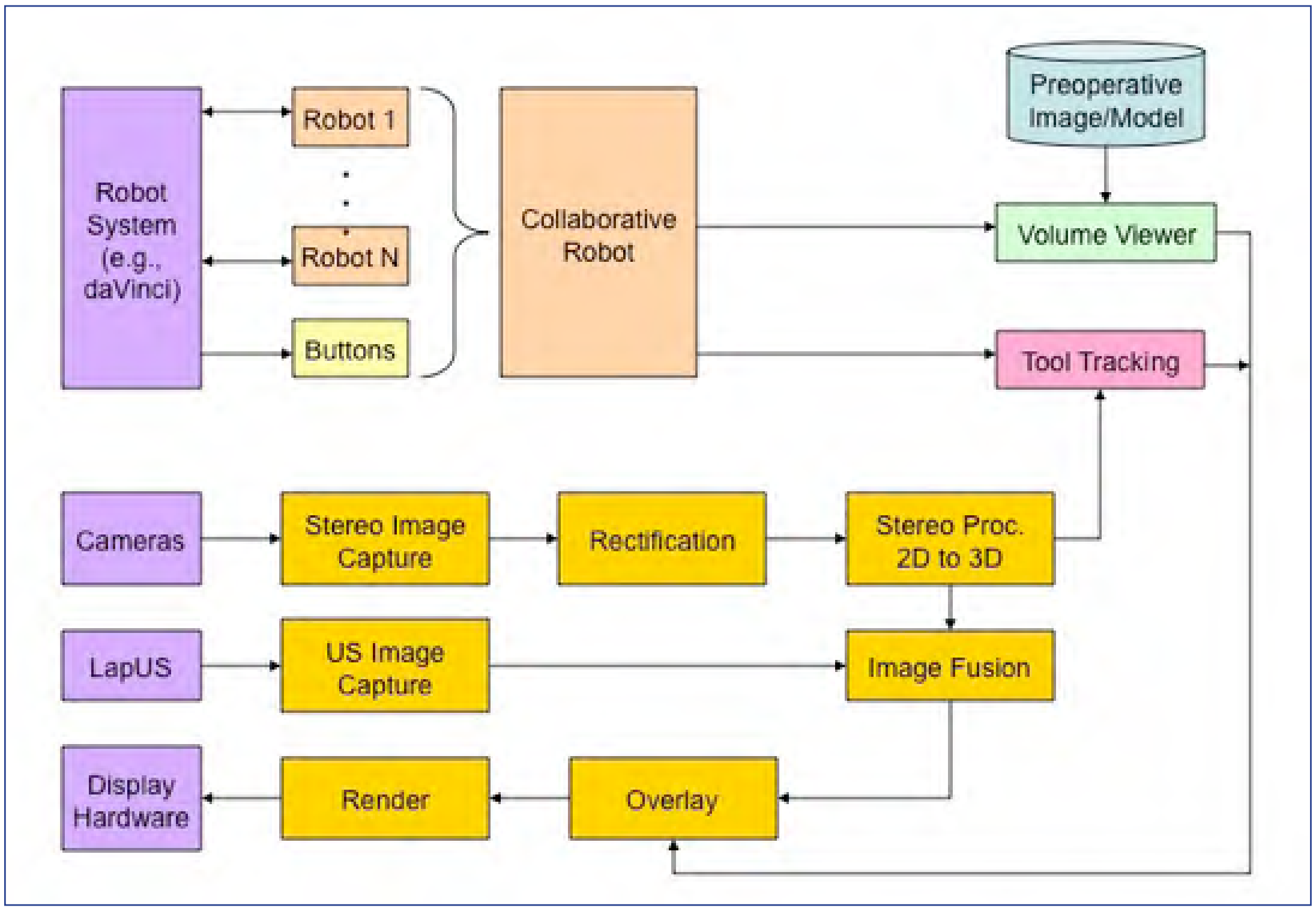

Fig. 52: Representative SAW data flow

time compatibility, and regression testing. For example, the original software libraries were targeted for Microsoft Windows, whereas the focus on portability enabled support of other operating systems and compilers.

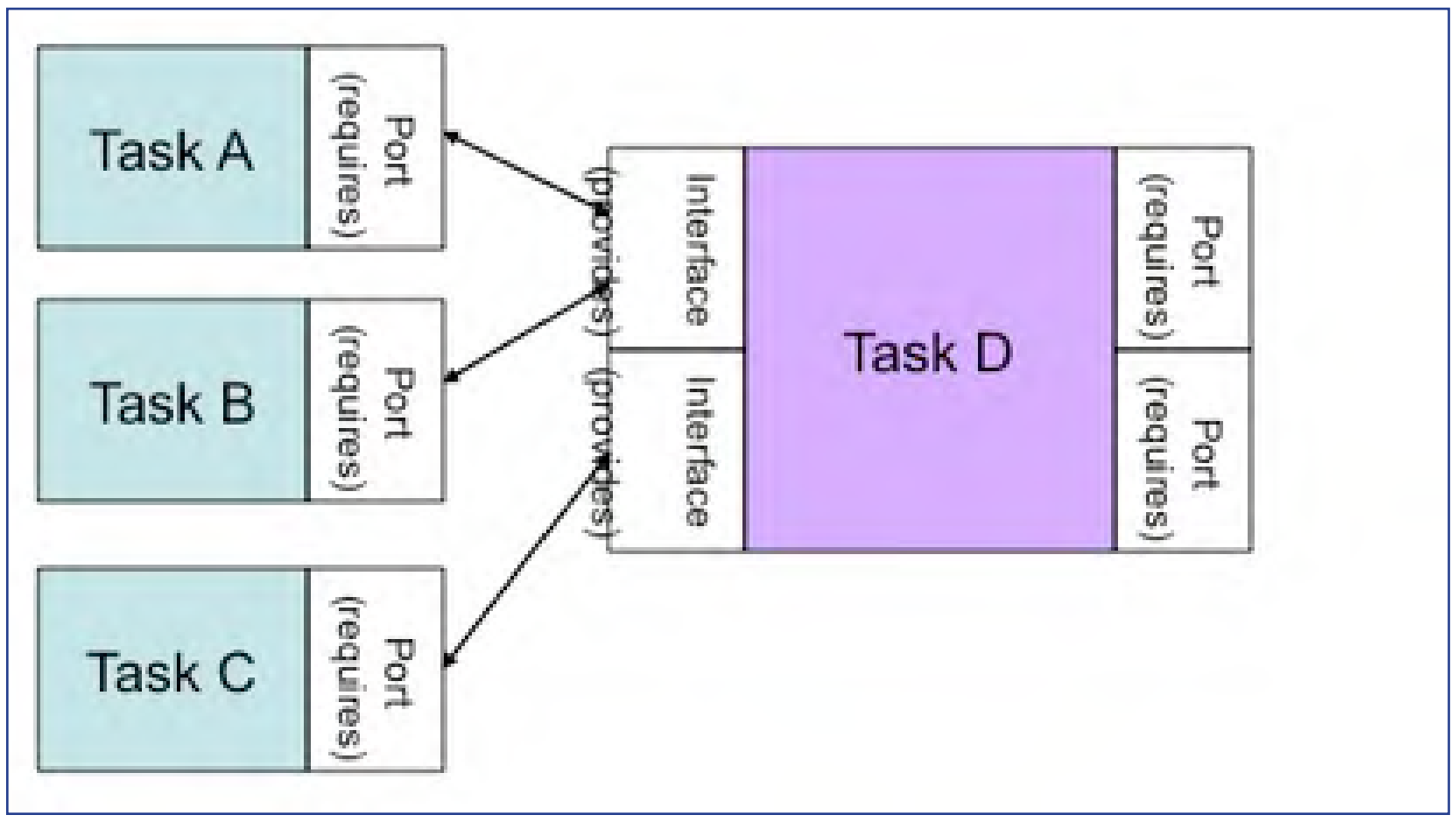

Fig. 53: Component-based design for task interfaces

*Software Components*

During the first five years, we focused on the development of a library for Computer-Integrated Surgery (CIS) application development, which included a common interface to different tracking systems (cisTracker) and a library for Modular Robot Control (MRC). Subsequently, we undertook a major redesign of the CIS/MRC software, now called the cisst package, with the intent of making it available under an open source license. The redesign effort focused on portability, maintainability, real-

A prime motivation for the development of the cisst package has been our increasing need to implement novel control algorithms for new interventional systems. This was not feasible with MRC because it relied on intelligent hardware to provide the low-level real-time control. Our vision for cisst was that it should support real-time processing so that low-level control could be performed by a software task coupled with non-intelligent (I/O) hardware. At the same time, cisst should still transparently allow any control function to be provided by intelligent hardware. We achieved this with a component-based design, where all interaction occurs via interfaces (Fig. 53). A task can expose multiple interfaces, which is useful if the underlying hardware has multiple capabilities or to support different access levels. We use the Command Pattern, where an interface provides services via command objects, to obtain loose coupling. During startup, a task queries the interface connected to each port to obtain pointers to the command objects that implement the required services. At runtime it is only necessary to invoke the command object's Execute method. Commands are strongly typed by the number of parameters and whether the parameters are read-only or write-only. These features, plus support for safe and efficient multi-threading, are provided by the cisstMultiTask library. This library relies on cisstOSAbstraction to provide portability between different operating systems.

The CISST package is available under an open source license. Several foundation libraries can be downloaded directly from our web site (www.cisst.org/cisst). These include logging, error and exception handling, serialization, class and object registries (cisstCommon), vectors and matrices (fixed size and dynamic), quaternions and transformations in 2D/3D (cisstVector), interfaces to thread-safe numerical methods such as those provided by LAPACK3E (cisstNumerical), and support for embedding the Python-based Interactive Research Environment (IRE) in C++ programs (cisstInteractive). Other libraries, in addition to cisstMultiTask and cisstOSAbstraction mentioned above, include an extensive collection of stereo vision algorithms (cisstStereoVision) and interfaces to various devices such as robots, tracking systems, haptic devices, and force sensors (cisstDevices). These will be available for public download once sufficient documentation and testing are completed.

*Mechatronic Components*

Application requirements can often be met by commercial off-the-shelf (COTS) solutions, but sometimes it is necessary to design custom hardware. For example, the snake robot required both speed and torque control of small DC brush motors, with milli-Ampere resolution of the measured motor current. There were no COTS solutions that satisfied these requirements, so we designed a Low Power Motor Controller (LoPoMoCo) to provide all I/O and power amplification necessary to control four small DC motors (Fig. 54). We subsequently reused this design to control the research da Vinci mechanical hardware, two master manipulators and two patient-side manipulators, acquired from Intuitive Surgical.

The LoPoMoCo satisfied our control requirements, but because it is physically located inside the PC it requires bulky and problematic cables to the sensors and actuators inside the robot. Although a typical solution is to distribute embedded microprocessors, and I/O devices, near the sensors and actuators, we prefer to implement all control on the PC because it provides a low-cost, high-performance processor and a familiar development environment. Thus, researchers do not need to learn the peculiarities of different embedded processors, nor do they need to procure the development tools. These factors, coupled with the emergence of multi-core processors and high-speed serial networks, led us to develop a new paradigm for system design, where we distribute just the I/O interfaces and perform all computations in a multi-core processor. We selected the IEEE-1394 (Firewire) and are developing boards that use a Field Programmable Gate Array (FPGA), rather than a microprocessor, to provide direct access to the I/O hardware. This reduces the latency and complexity of the system. We implemented and tested a prototype, using an FPGA development board and are currently designing a version appropriate for installation on the actuation units of the snake robot. This design will provide significant cabling reduction, replacing four 68-pin cables and four 9-pin cables with one Firewire cable and one power cable.

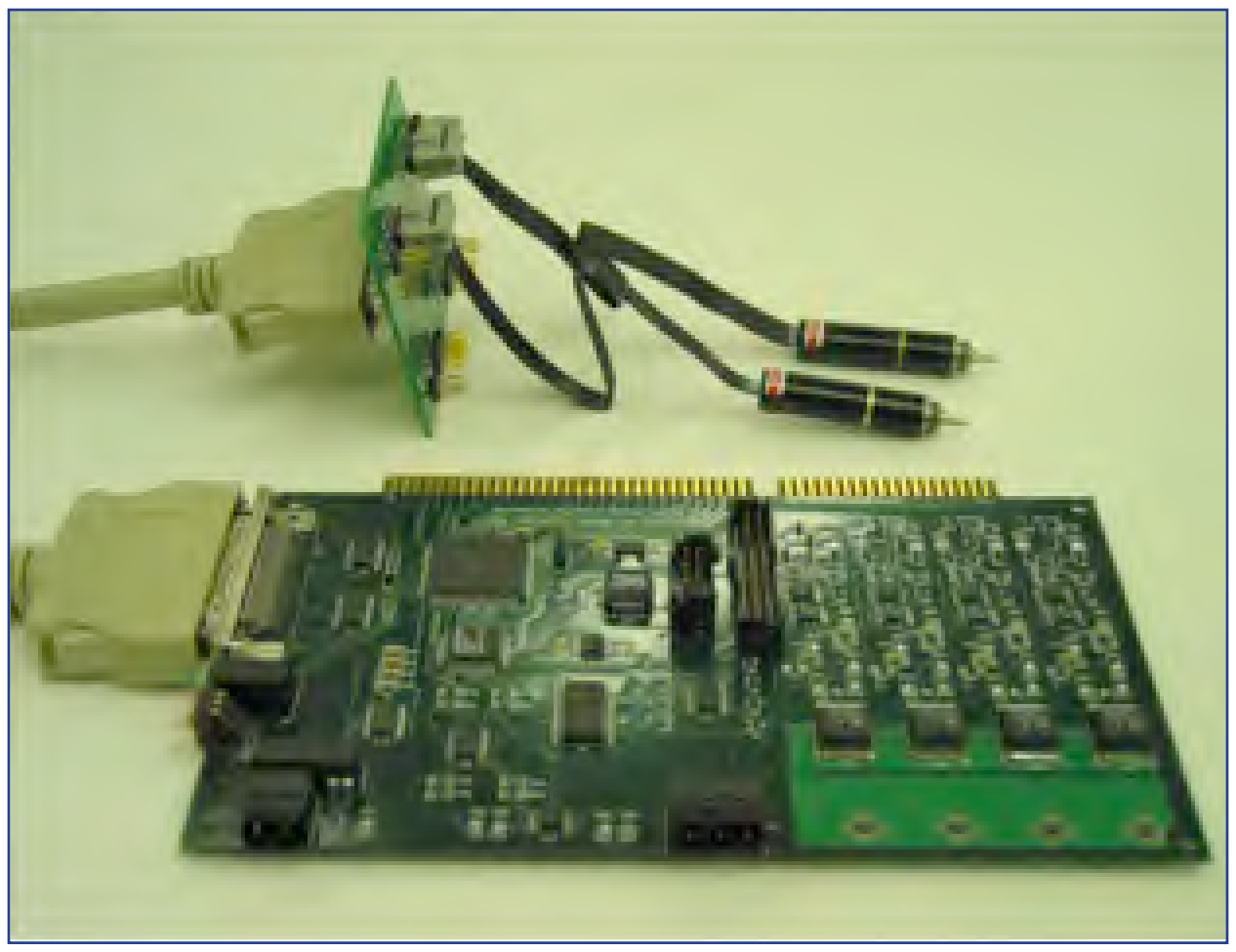

Fig. 54: Low Power Motor Controller (LoPoMoCo)

Physical size is often a challenge for CIS systems because they must operate in confined spaces such as alongside the surgeon, inside an imaging device, or inside the patient. This motivated an ERC diversity fellowship recipient to design and fabricate a custom VLSI chip that implements most of our controller functionality. Elements of the design have been presented at three conferences and appeared in one journal. A patent has been filed and the chip is currently undergoing benchtop testing.

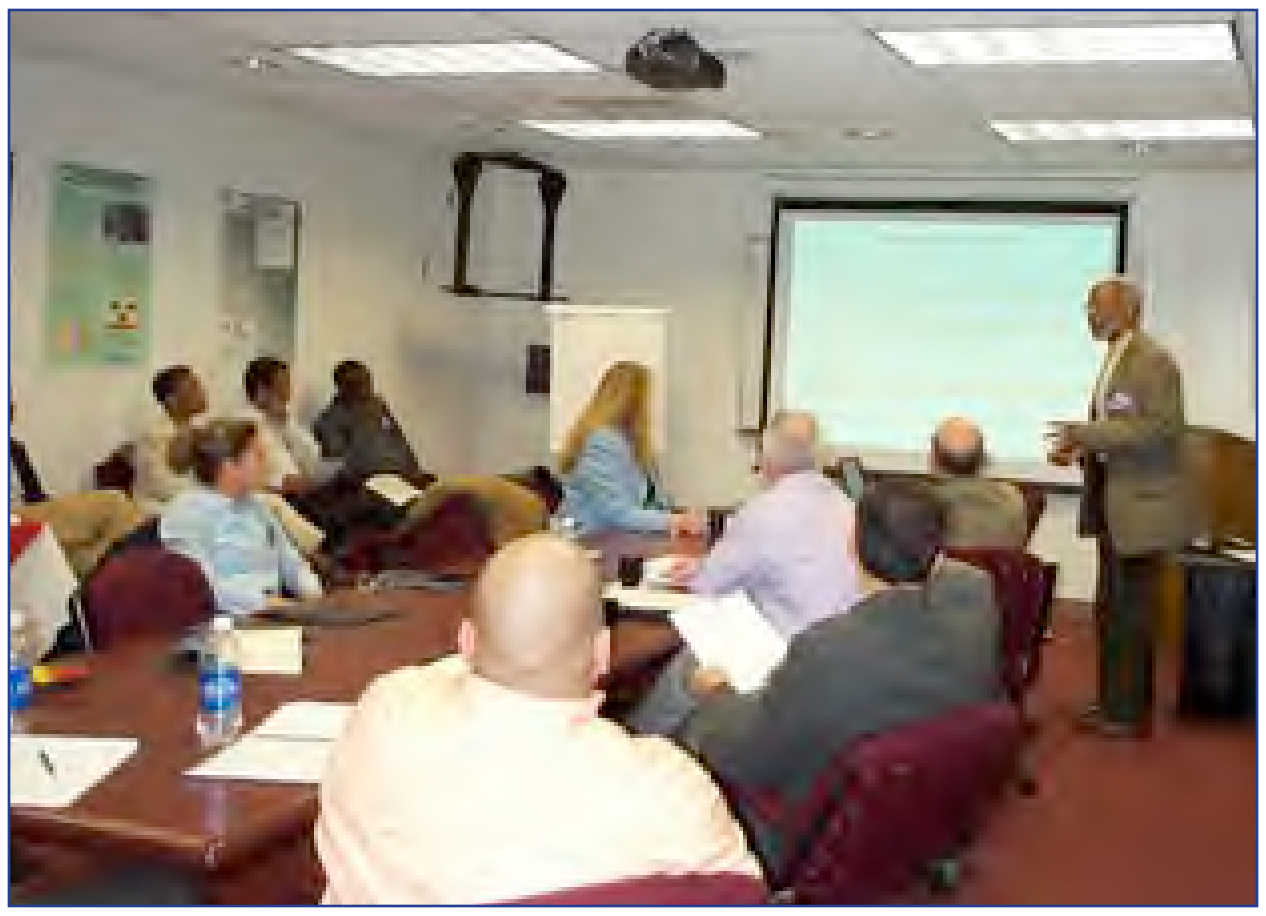

Fig. 55: Presentation of Research Information System to ERC directors, Morgan State University, April 2006

### *Development Process and Tools*

Although a formal software development process is generally not necessary for a specific research project, it is appropriate when the goal is to create a core development platform that can support many research projects and can ultimately be used in clinical trials. Our development process follows many of the best practices established by successful open source software packages such as VTK, ITK, and 3D

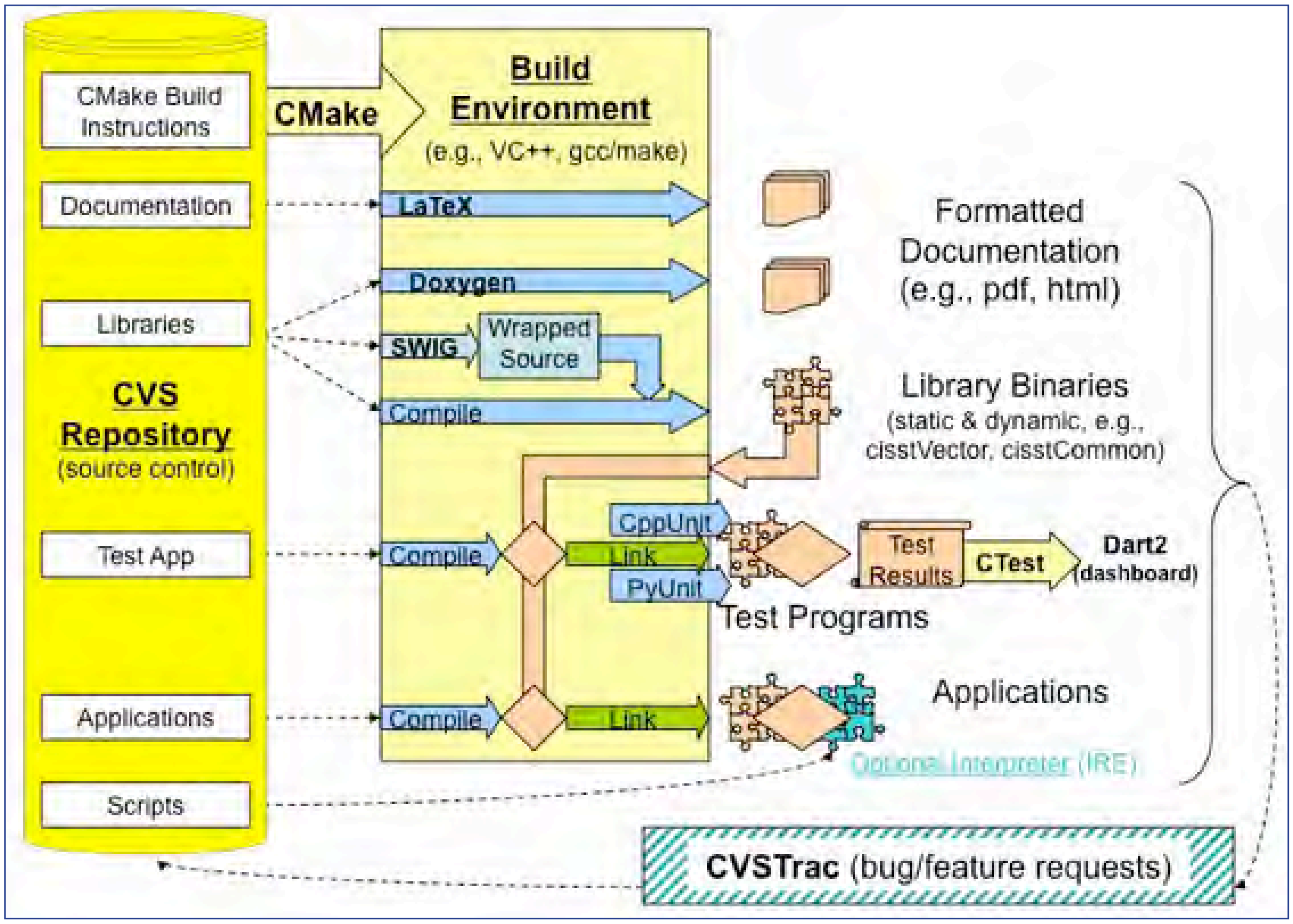

Fig. 56: Development process and tools

Slicer, while using many of the same tools (Fig. 56). We use Doxygen to automatically extract a reference manual from the source code and augment this with manually-created user documentation. Our other development tools include CVS (source code control), CMake (portable build tool), Swig (Python wrapping), CppUnit and PyUnit (unit testing for C++ and Python), Dart/CDash (a web-based dashboard for displaying results of automated testing), and CVSTrac (bug tracking and feature requests). These tools support a development process that produces well-tested, documented, and portable software.

### *Information Systems*

This project was conceived three years ago by Dr. LeeRoy Bronner at Morgan State University. The goal is to develop information design and delivery methodologies that communicate the current and on-going research information for all ERCs. The initial focus is on scientific poster patterns since posters are a well-developed and frequently used mechanism for communicating scientific information. In particular, all ERCs generate a large number of posters per year, most often in conjunction with a major event such as a site visit or industry meeting. The project extends the concept of a scientific poster to include multimedia such as video, audio, and animations. In April 2006, we reviewed the design and demonstrated prototype versions of the tools with the Directors, or designated alternate of several other ERCs (Fig. 55). More recently, we conducted several Joint Application Design (JAD) sessions with targeted user groups, including JHU and MSU students, to refine the concepts and user interface design. As a result, the system has evolved to incorporate more collaborative features, such as the ability to post comments on the research.

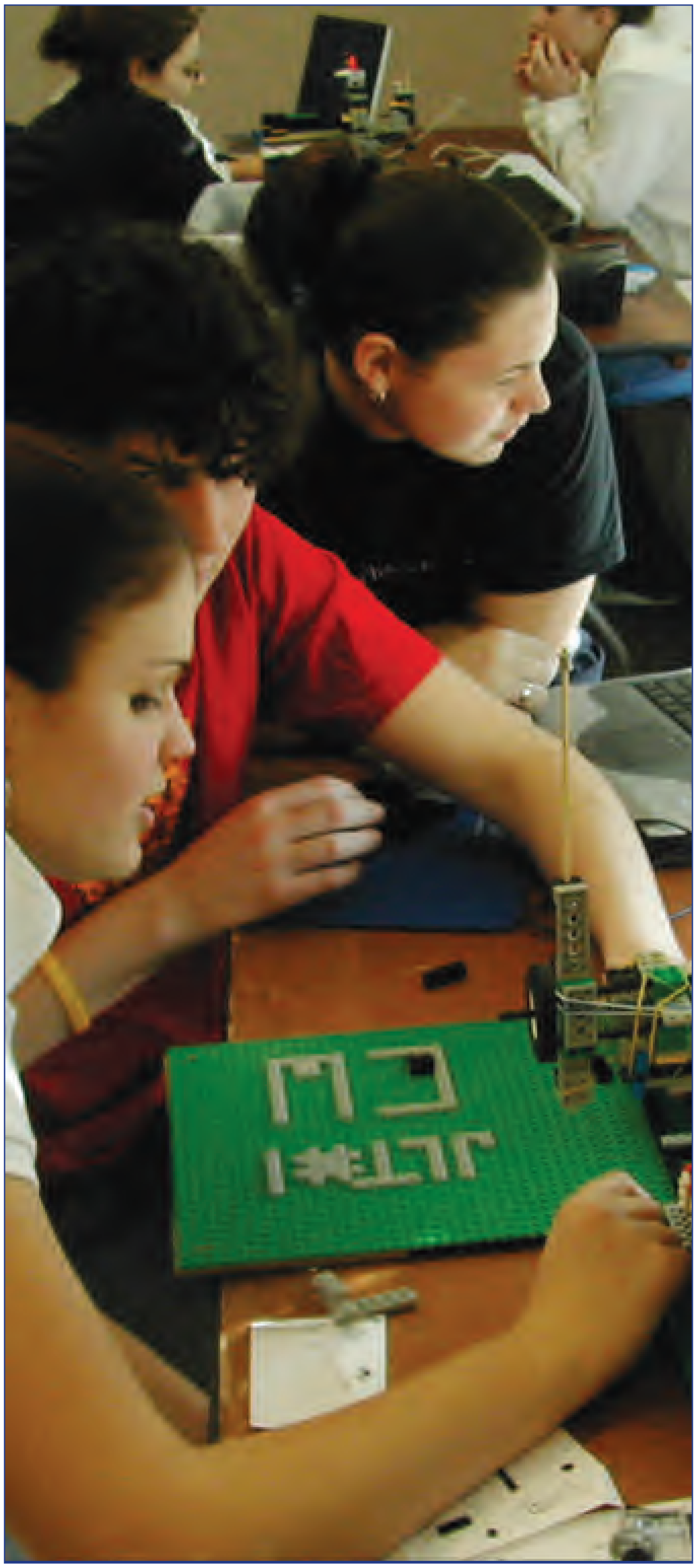

# Education & Outreach

The CISST ERC strives to develop students, teachers, and researchers into scholars and role models. To accomplish this, the Center's educational activities are organized into five communities: K-12, Community Colleges, Undergraduates, Graduates, and Practitioners. The CISST ERC has developed educational programs for each of these communities that focus on providing unique opportunities for women and minorities, and highlighting career paths in engineering and related fields. To strengthen the alignment of the research and educational missions of the Center, the education team developed several initiatives designed to foster innovative course dissemination, increase minority participation, encourage pre-college outreach in engineering, and create effective collaborations with ERC partners and their students.

### *K through 12*

**Research Experiences for Teachers:** The NSF's RET program enables local K-12 teachers to participate in world-class research during their summer breaks. A leading participant, the CISST ERC has created a national model that brings teachers into laboratories at JHU and other research facilities across the nation. The center's partners in this effort include other institutions of higher learning (including tribal colleges), school districts across the country, and various government agencies and associations.

The success of the RET program at the CISST ERC fostered a national effort to unite the various constituencies in STEM education for a Summit on Engineering Education in K-16 (SEEK-16). Corporate sponsors and in-kind contributions from the National Academy of Engineering and several universities contributed to make

SEEK-16 a big success. It has directly led many states, including Maryland, and several organizations to form teams to address science and engineering education issues which affect our greater community. One team, the ORAU team of Carnegie Mellon and Middle Tennessee State University, worked with members of Congress to include the Science and Engineering Education Pilot Program into the Energy Bill.

ERC educational programs have been incorporated into The Center for Educational Outreach (CEO). CEO currently runs three programs that directly involve teachers, K – 12 students and JHU/WSE faculty and students: Engineering Innovation; Future Inventors, Researchers, Scientists and Teachers; and Broader Impact from Graduate Students Transferring Engineering Principle.

**Engineering Innovation:** Engineering Innovations works to create a national pipeline of innovative and creative thinkers who will lead society's technological advancements, by engaging pre-college students in STEM education and inspiring them to consider further studies and careers in engineering. The academic program demystifies technology, emphasizes engineering's innovative and creative aspects, and exposes students to engineering's diverse educational and career opportunities. Working as part of a team, students learn the basics of engineering--conducting experiments, taking part in design challenges, interpreting data, applying technology, and meeting with research engineers who are pushing the boundaries of the field.

**Future Inventors, Researchers, Scientists and Teachers:** More than 30 teachers and over 140 students collaborated with mentors to develop concepts which demonstrated the trans-disciplinary nature of the research and invention process within the framework of a "Quality of Life' theme.

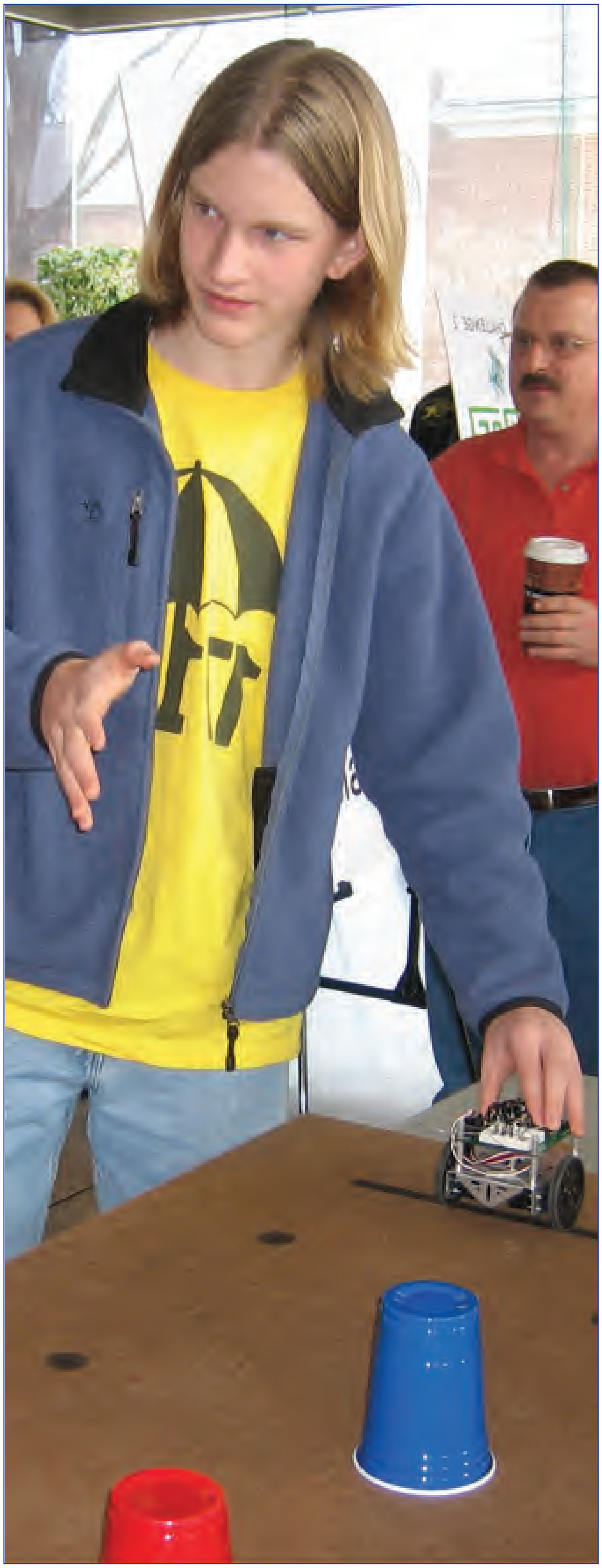

**Broader Impact from Graduate Students Transferring Engineering Principles (BIGSTEP):** The WSE, with Baltimore City and County schools and White Earth and Leech Lake Tribal Reservation schools, began a coalition to encourage under-represented students from inner cities and rural Native American communities to pursue studies in STEM Education.

Graduate student Fellows rotate through an internship with several distinct K-12 schools that serve disadvantaged children. Teachers who have participated in the JHU RET program work with the Fellows to develop skills to teach children with different learning styles, enhance the content knowledge of district teachers, and facilitate the creation of standards-aligned content based on cutting-edge research. Each project focuses on topics from Environmental Engineering and Geography (EEG). Fopefolu Folowesele, an ERC student, worked on developing ways to analyze and design electric motors (intended to communicate the benefits of electric vehicles to the environment). The involvement of the faculty director and (co-)PIs of this GK-12 grant contributied to the project allowing teachers to have direct access to the latest developments in various multi-disciplinary fields.

**Robotics Summer Camp:** Students learn about robotics construction and theory through a problem-solving application that teaches them basic programming, electronic theory, soldering, and mechatronics. The camps include a trip to JHU engineering labs so students can visualize and participate in current research being conducted in various engineering disciplines. Students put together BoeBot robots and program them to compete in a challenge; they redesign the robot to complete the task of herding ping pong balls into a goal. The camp also teaches website design.

**Robotic System Challenge:** The Robotic System Challenge engages a large number of very enthusiastic middle and high schools students in teams which compete against each other in four different challenges: 1) Petite Slalom; 2) Mystery Course; 3) Unleashing the Mad Scientist; and 4) Search & Destroy- Robotic Brain Tumor Surgery.

For the 2008 challenge, ERC faculty members welcomed the group to start off the day with a guest speaker, the Senior Assistant Director of Admissions for JHU, who spoke about the college admissions experience and gave valuable information about how to achieve a successful college experience.

Lab Tours to Pre-College Students - The ERC provides lab tours and demonstrations of research. These tours give students an opportunity to see research first-hand and learn how it relates to the real world.

### *Undergraduates*

**Research Experience for Undergraduates (SiteREU):** This ten-week summer program provides undergraduates from universities nationwide an opportunity to learn about Computer-Integrated Surgical Systems and research at the JHU Medical Institutions and the WSE. The 2007 program attracted a diverse array of students, with 55% being female and 18% under-represented minorities. With an increase in advertising and networking, especially to LSAMP and other minority serving institutions, we had an overwhelming number of qualified applications for the 2008 program.

Minor in Computer-Integrated Surgery: Students are able to graduate with a Minor in CIS. This Minor is published in several catalogs for JHU, as well as on the CS web page. The ECE Department at JHU has implemented a concentration in CIS for Computer Engineering (CE) majors. The concentration requires juniors and seniors to take courses that fall under the CIS minor requirements. A number of CIS courses are now part of the ECE list of recommended courses, especially as senior labs.

**Articulation Agreement:** This agreement allows students from Center partners CMU, MIT and MSU to enroll for one semester at one of the other institutions. Students pay the tuition of the home institution while attending the other university. We are in the final stage of setting up a new Articulation Agreement with Queens University, Canada. We expect this conduit between JHU and Queens to be heavily used since Prof. Fichtinger (who moved there in 2007) has many ongoing collaborations and grants with members of the Center.

**Surgery for Engineers (SFE):** Students receive hands-on surgical training while applying engineering problem-based concepts. The course is offered in the summer so students from our partner institutions can take advantage of the ERC Articulation Agreement.

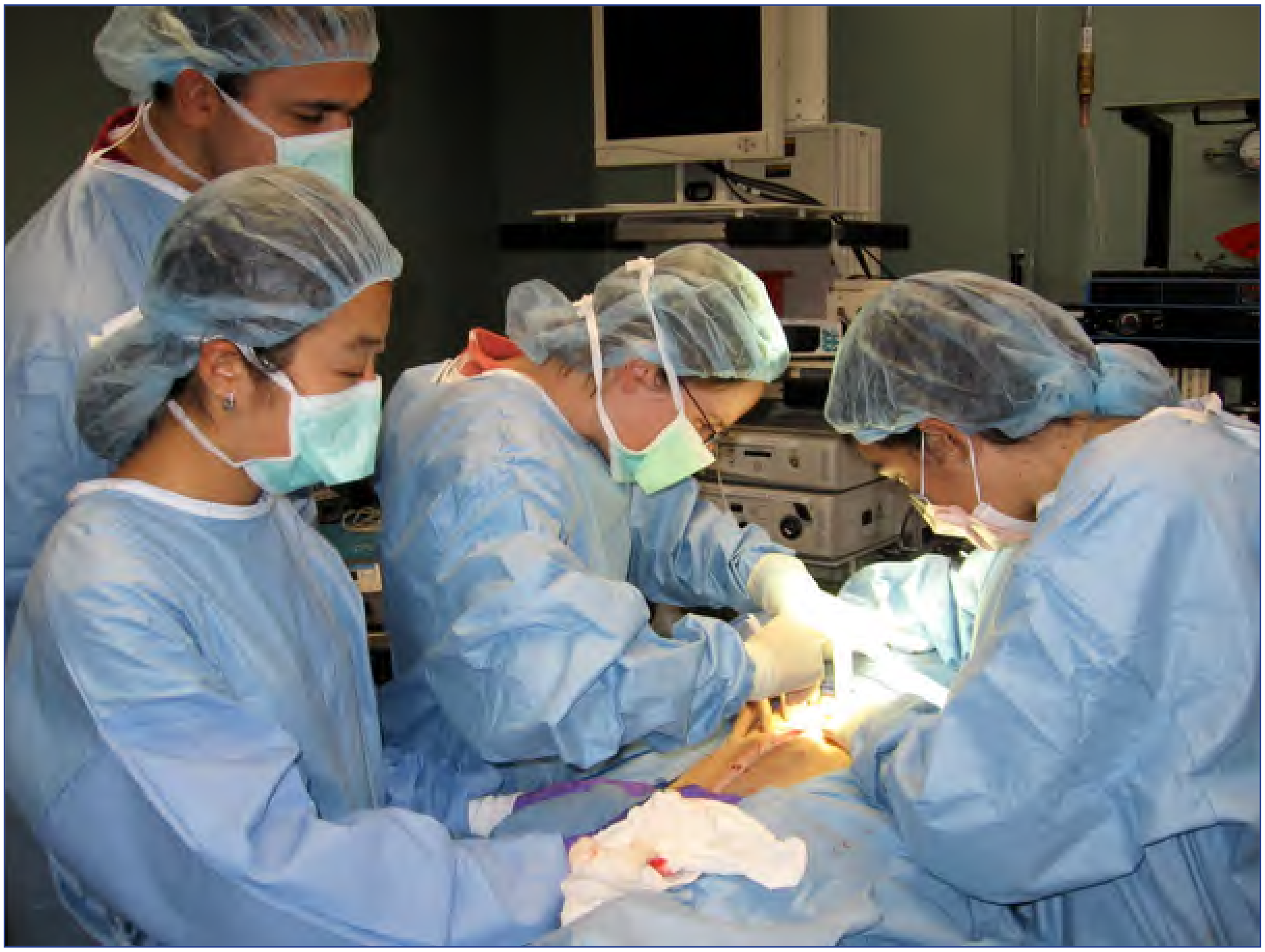

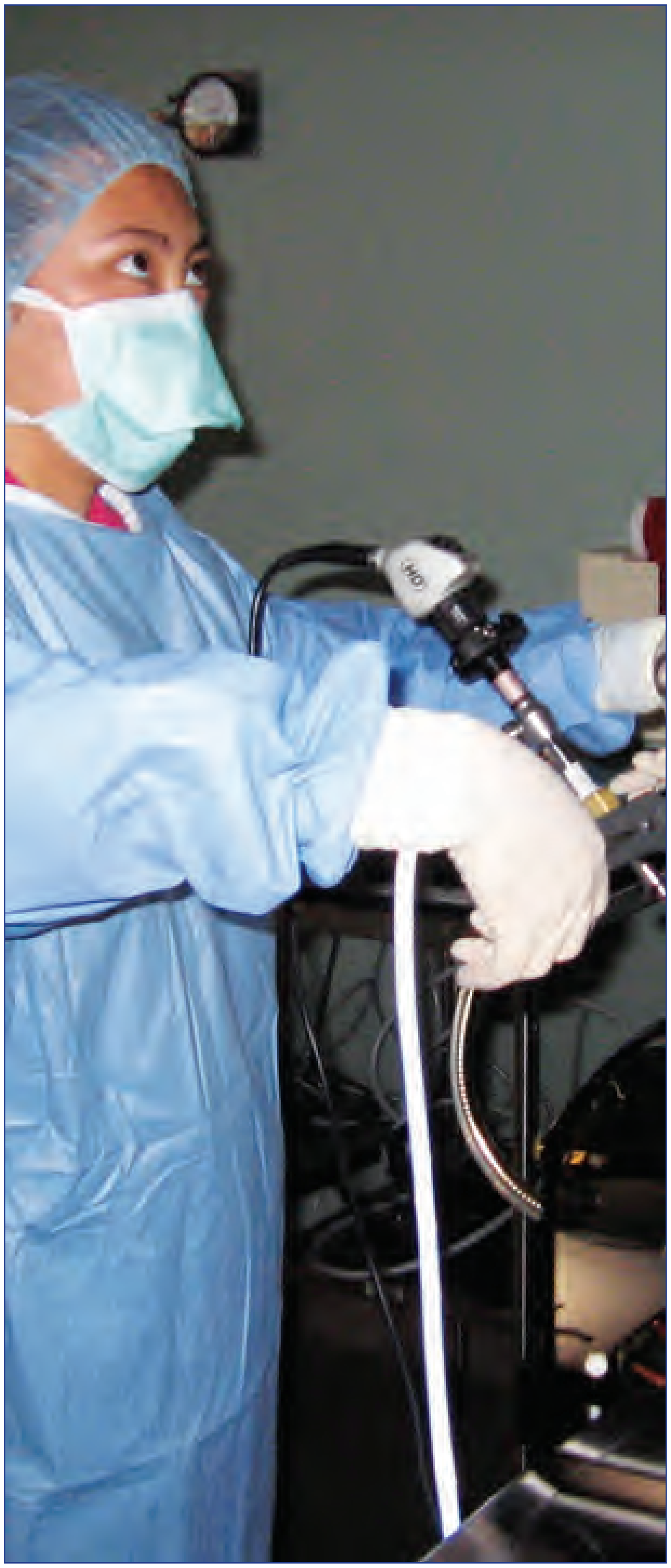

# Education & Outreach

**Computer-Integrated Surgery Student Research Society (CISSRS):** Founded in 1998 in response to a need for greater student support and involvement in computer-integrated surgery, CISSRS plays a highly active role in the Center's research and education activities. CISSRS students provide tours of the Center's labs, and participate in classroom visits and recruitment of incoming freshmen. CISSRS membership continues to increase and many activities have been planned and executed. Guest speakers from the medical institution and the JHU campus were invited to speak at meetings, and members have participated in Computer-Mania Day (see Pre-College Outreach), the WISE program, and the Robotic System Challenge.

*Graduate Students*

**Weekly Seminars:** The Center hosts a weekly seminar, featuring faculty and student presentations of current research, as well as outside speakers. Typically the speakers include faculty, clinical or industry invited guests as well as speakers from JHU professional offices such as the Career Center and the Office of Organizational Development and Diversity. All seminars are well attended.

Friday Student Seminars: Friday Student Seminars began in Fall 2006. Designed to engage ERC students in informal conversations with regard to their research topics and professional development, they address topics like overviews of research being conducted in different ERC labs; meeting the challenges of the 21st century engineer; how to search, apply and negotiate for an industry career; advice and resources for ERC students (new and old); and teaching and mentoring.

**Mentoring, Outreach and Service:** Graduate students are essential to all center education and outreach programs and faculty actively encourage them to assume leadership responsibilities. In their respective research labs, graduate students work with faculty to mentor and supervise undergraduate and teacher researchers. They serve on various center-wide committees, such as the Education, Diversity and Outreach Committee (EDOC). They also are integral to the center's recruiting process: they were heavily involved in the CISST ERC Visit Day and interface extensively with prospective students. The center highly values its graduate students, and we try very hard to provide a congenial social and academic environment that promotes excellence, including regular faculty-student interaction in the labs, seminars, and social events.

*"We have amazing 'star' students who've grown a lot by working with each other so closely on projects under the umbrella of the ERC."*

*~ Carol Reiley*

***Practitioners***

**Surgery for Engineers (SFE) for Industrial Affiliates:** The SFE course has been a great success for the academic members of the ERC, and it has received tremendous interest from our industrial affiliates as well. Since the main barrier for the industrial affiliates is scheduling, we are working with our clinical instructors to offer a compressed version of the course. We also offered the course as part of a Winter School on Medical Robotics which was marketed to students, clinicians and other professionals. The first Winter School was held in January 2009 and lasted for one week. Mornings were filled with technology oriented tutorials, the afternoon was dedicated to projects and SFE, and the evening included keynote talks by clinicians (see page A141).

*CISST ERC Affiliates, are entitled to the following benefits:*

- Non-exclusive, non-transferable, royalty-free grant of rights to all intellectual property developed by the ERC using NSF core grant funds
- An option to obtain a non-exclusive or exclusive license to technology developed as a deliverable product of research sponsored by that Affiliate
- Priority access to ERC faculty and students
- Discounted overhead rate applied to any research project associated with CISST research and sponsored by Affiliate; this rate applies to contracts entered into with JHU during Affiliate's participation and requires full payment for the research project in advance
- Reduced overhead rate on research and development contracts involving graduate student support; applicable to contracts entered into with JHU researchers
- Priority access, subject to CISST ERC researcher requirements, to testbed and laboratory facilities at a reduced cost
- Attendance at annual meetings at no cost
- Priority registration and reduced rate for employees attending CISST ERC short courses, workshops and seminars
- The opportunity to request on-location short courses provided by CISST ERC at fees to be negotiated to cover costs
- Option to locate Affiliate personnel in CSEB

# Industry

The Center recognizes the importance of having industrial partners who support our goals and objectives. During the past few years, we have witnessed a major increase in the level of collaboration with several of our industry partners, as measured through research projects, grants, and student funding. We have restructured the Industrial Affiliates Program to provide additional benefits for our industry partners, as well as increased opportunities for joint research projects. An additional goal is to develop broader participation by the Affiliates in the educational programs of the Center.

Our industrial partners play a vital role in helping us fulfill our NSF mandate. They create a critical pathway for clinical translation of research results and new technologies into practical and widely applied healthcare solutions. Member companies also help guide our research programs with their unique knowledge of the marketplace. They share insights on the needs of clinician users, healthcare system customers and other important constituencies. The Affiliates' financial contribution, in the form of membership fees and sponsored research, is an important resource for the CISST ERC. Industrial Affiliates benefit from access to leading edge research, opportunities for collaborative research, and interactions with researchers, clinicians and students.

Any corporation, company, partnership, sole proprietorship, or any other legally recognized business entity, may become an Affiliate of the CISST ERC (see sidebar).

The annual fee for an Affiliate is based upon the number of full-time employees within the Affiliate's corporate entity. A corporate entity can include a corporation, company, independent business unit or division, majority-owned venture, or subsidiary. In recognition of the fact that we are working with a greater number of small start-up companies, we have changed the Affiliate membership fee structure to enable greater participation.

| Number of Employees | Annual Fee |
| --- | --- |
| Fewer than 100 | $5,000 |
| Between 100 and 1,000 | $10,000 |
| More than 1000 | $15,000 |

A business unit with 10 or fewer employees or a company with fewer than 50 employees is eligible for a trial membership fee of $1000 for the first year. After the first year, dues will be assessed at $2500 per year.

An additional revision in the Affiliate membership fee policy enables a member to substitute an equipment donation valued at 10 times the membership fee, e.g. equipment valued at $100,000 can be substituted for $10,000 in cash.

To encourage more research collaborations, the Affiliates now have the right to a discount in the cost of a sponsored research project if they submit full payment at the beginning of the project. A new funding mechanism enables them to deposit funds in an "escrow account" at the CISST ERC. Before funds can be spent, the company must approve a statement of work and budget for each project. In addition to the advantage of a reduction in overall project costs, this type of funding mechanism enables Affiliates to have the flexibility of quickly beginning projects throughout the year. The Affiliates also now receive a reduced rate on research and development contracts involving graduate student support.

These companies have participated in ERC programs:

*American Shared Hospital Services*
*Foster-Miller*
*Intuitive Surgical*
*Medtronic Navigation*
*Siemens Corporate Research*
*Hologic*
*Philips Research North America*
*Acoustic Medical Systems*
*VISIONSENSE*
*Infinite Biomedical Technologies*
*Ikona Medical*
*SRI International*
*Northern Digital*
*GE Healthcare*
*Integrated Surgical Systems*

During the last few years, we have focused our efforts on deepening our relationships with current Affiliates. Our current partnerships with companies are broad, multi-faceted and based on mutual goals for technology development and deployment. These partnerships include a number of activities which foster a mutually beneficial, long-term relationship:

- Continuous meetings between CISST ERC faculty and corporate partners to cultivate a research roadmap
- Sponsored research projects that support the roadmap
- Donated equipment and funds for research
- Funded fellowships
- Student internships/permanent hires
- Joint proposal submissions
- Lectures given by guest corporate researchers; lectures given by CISST ERC faculty to corporate personnel
- Educational opportunities for corporate personnel
- Joint publications
- Joint presentations at conferences
- Corporate licensing of ERC-developed technologies
- Philanthropic support for K-12 education initiatives

In response to our SWOT analysis regarding a need for more frequent communication, we now disseminate a "CISST ERC News Bulletin." The Bulletin contains information on new research results, papers published, conference presentations, relevant upcoming internal and external meetings, student activities, new collaborations, grants and awards, open source software, funding opportunities, etc.

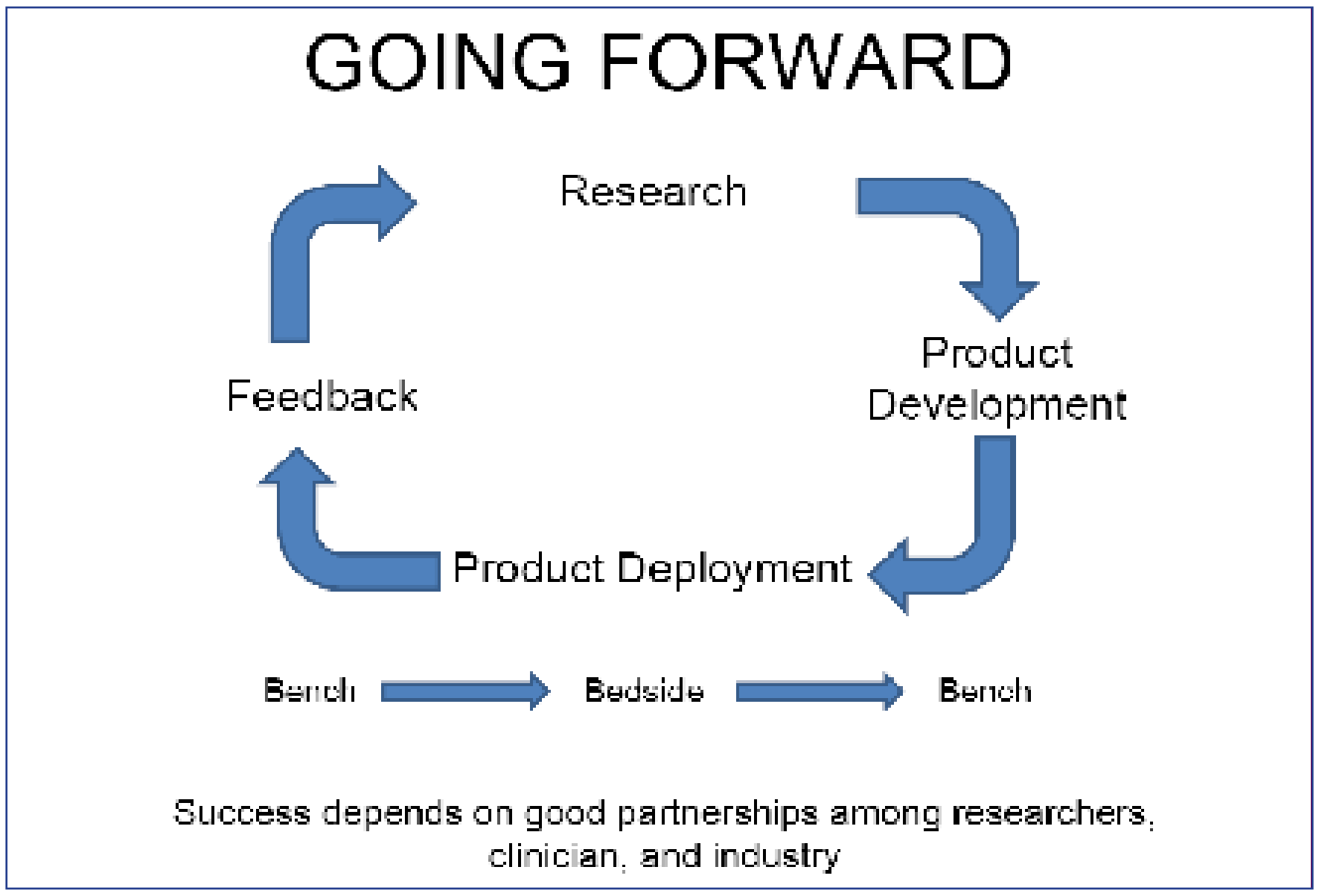

Fig. 57

We encourage our Affiliates to support student projects and hire students as interns and employees by sending them Student Poster Books and Student Resume Books.

As the CISST ERC is nearing graduation, it is evolving into the Institute for Integrating Imaging, Interventions, and Informatics in Medicine (I4M). Our I4M vision integrates imaging, interventional systems and technology, and informatics to transform medicine in the same way that CAD/CAM, and TQM have transformed conventional manufacturing. Our goal is a highly integrated, multi-divisional organization supporting basic engineering research, clinical research, and clinical deployment of systems.

Developing the systems needed to implement this vision will require innovation over a very broad range of technical specialties, as well as the adoption of technical innovations in clinical applications, and the deployment of new systems in the interventional suite. As a result, there has been a shift in the "center of gravity" in collaborations from engineering research motivated by clinical needs to clinical and pre-clinical research enabled by new capabilities.

During the early years of the CISST ERC, when the focus of research programs was on transformational research, industry contributed to the support of the development of fundamental knowledge, enabling technologies and engineered systems to advance computer-aided surgery. The goal was broad participation to address individual technical and scientific challenges.

Going forward, there is an emphasis on translational research. Using "test bed" systems, I4M will focus on partnerships with industry that promote further development, and accelerated commercialization and deployment of new systems. We will solicit participation by a growing number of targeted companies based on clinical applications. We will exploit and build our science, technology, engineering, and systems infrastructure to promote new opportunities and actively pursue joint funding with industry (STTR, SBIR, R01, etc.)

*Hologic – CISST Collaboration*

In 2004, the U.S. Surgeon General issued a report on the health and financial impact of osteoporosis. The report showed that in the U.S. there are approximately 1.5 million bone fractures per year and the direct care costs for osteoporotic fractures are $18 billion a year. Osteoporosis is a disease where 34 million Americans are at risk. 40% of Caucasian women will have an osteoporotic fracture and 13% of Caucasian men will suffer one as well. When someone endures a hip fracture, of which there are about 300,000 per year, 50% of them will be unable to return to independent living while 20% of them will die. Clearly there needs to be more done to both treat and prevent osteoporotic fractures.

Both Hologic and the CISST ERC joined together in September of 2005 to combat this issue and come up with a better diagnostic test for osteoporosis than the ones which were already available. The main goal was to create a better test than the widely accepted areal BMD (g/cm2) measurement of the hip with a dual-energy x-ray bone densitometer. With a low radiation dose and moderate costs, this test is excellent in many ways. However, more than 50% of fractures occur in people ho are not classified by areal BMD as osteoporotic. The biggest flaw in the areal BMD is that it is a two-dimensional measurement, which does not compare with the three-dimensional bone structure it is testing. While technology exists to create a three-dimensional measurement, the costs and high dosage of radiation needed make this test impractical.

Hologic and the CISST ERC focused their efforts in combining a few areal BMD images of the hip, taken at different angles with a bone densitometer, to produce a volumetric BMD patient specific model. Only three to five images are required to create this model, and the equipment needed is much less costly than a traditional CT scan while also using a significantly decreased amount of radiation. Since the original BMD image is used to help create the model, the model is completely additive to the information gathered in the first BMD test. The test has proven successful *in vitro*, with high correlation to CT. *In vivo* studies have been completed and are currently being analyzed. FDA approval for this test will be sought soon enough to have it approved before the end of the calendar year.

Osteoporosis is a debilitating disease that progresses silently over the course of many years. With the growing availability of accurate testing technology and effective therapies, the loss of independence and suffering associated with osteoporosis can no longer be considered an unavoidable consequence of aging. The challenge for all practitioners who see premenopausal patients is to provide testing that can lead to effective prevention and treatment. The joint venture of Hologic and the Johns Hopkins CISST ERC have found a diagnostic test which could help physicians give patients more successful prevention and treatment techniques so that they can live more independent lives.

*"Each party brings capability that expands opportunities for the other.... the collaboration has produced technology that is now in the commercialization phase and should be in the market in the near future.... The financial impact of both product sales and research funding on Acoustic MedSystems has been substantial."*

*~ Clif Burdette*

*Technology Transfer*

We have licensed several of our inventions to Affiliates and are in negotiations with others for additional intellectual property. Our "snake" robot invention, "Devices, Systems, and Methods for Minimally Invasive Surgery of the Throat and other portions of Mammalian Body," was licensed to Intuitive Surgical. This invention uses remotely actuated, flexible spines to provide high dexterity manipulation in very small spaces such as the throat and upper airway. Although the current prototype has 4.2 mm diameter end-effectors, the design's simplicity makes it readily scalable to smaller sizes.

The "Device, Method, and System for Needle Insertion inside a Medical Imager, from within Body Cavity" patent licensed to Siemens Corporation, includes a device, computerized system, and methods for entering a needle into the body inside a medical imager such as an MRI scanner CT, X-ray fluoroscopy, and ultrasound imaging, from within a body cavity (such as the rectum, vagina, or laparoscopically accessed cavity). A three degree-of-freedom mechanical device translates and rotates inside the cavity, inserts a needle into the body, and steers the needle to a target point selected by the user. The device is guided by real-time images from the medical imager. Networked computers process the medical images and enable the clinician to control the motion of the mechanical device that is operated remotely from outside the imager.

The "iMR" imaging system for tumor detection and treatment, also licensed to Siemens Corporation, can determine the three-dimensional position and orientation of an effector (a needle, probe, etc.) relative to a subject using single cross sectional imagers (e.g. from a CT or MRI scanner). This method for image-guided effector placement requires no immobilization of the patient or fiducial implantation.

| *Sectors* | | | | | | | | | | |
|---|---|---|---|---|---|---|---|---|---|---|
| Sector | Y1 | Y2 | Y3 | Y4 | Y5 | Y6 | Y7 | Y8 | Y9 | Y10 |
| Industry | 12 | 15 | 15 | 15 | 15 | 16 | 13 | 11 | 11 | 12 |
| Federal Government | 0 | 0 | 0 | 0 | 0 | 0 | 2 | 3 | 3 | 3 |
| Non-Profit | 0 | 0 | 0 | 0 | 0 | 0 | 0 | 2 | 4 | 4 |
| Private Foundation | 0 | 0 | 0 | 0 | 0 | 1 | 1 | 2 | 1 | 1 |
| Medical Facility | 0 | 0 | 0 | 0 | 0 | 0 | 1 | 0 | 0 | 0 |

Fig. 58

| *Patents* | | | | |
|---|---|---|---|---|
| Patent or License Title | Date Filed | Date Granted | Patent/ License Number | Licensed to |
| Friction Trasmission with Axial Loading | 2/20/1997 | | 60/038,115 | Exclusive License Executed |
| Method of employing Angle Images for Measuring Object Motion in Tagged Magnetic Resonance Imaging | 8/9/1999 | 9/17/2002 | 6,453,187 | Diagnosoft Inc |
| Method of employing Angle Images for Measuring Object Motion in Tagged Magnetic Resonance Imaging | 4/22/1999 | 5/10/2005 | 6,892,089 | Diagnosoft Inc |
| Ultrafast motion imaging using Harmonic Phase MRI (HARP-MRI) | 2/10/2000 | 7/22/2003 | 6,597,935 | Diagnosoft Inc |
| FAMOUS: A quantitative assessment of muscle tone in the wrist | 9/6/2001 | 7/8/2003 | 6,589,190 | |
| A Single Image Registration Method for Image Guided Interventions | 9/18/2000 | 5/29/2007 | 7,225,012 B2 | |
| Adjustable Remote Center of Motion Robotic Module | 2/6/2002 | 4/4/2006 | 7,021,173 | |
| Apparatus for Insertion of a Medical Device During a Medical Imaging Process | 4/22/2002 | | 60/374,376 | Siemens AG Medical Solutions & Senentinelle Medical |
| MRI Compatible Positioning Arm | 1/9/2003 | 2/22/2005 | 6,857,609 | Dumitro Magilo,Bard Urological Division , Siemens AG, Medical Sol and Nucleotron BV |
| Minimally Invasive Surgical Assistance System for Throat Surgery | 5/21/2003 | | 60/472,168 | Intutive Surgical |

Fig. 59

*"Intuitive has found that only a few universities, among dozens the company has interacted with, possess the broad engineering capability and clinical collaborations required to make an effort like this succeed....These traits have made CISST-ERC the most productive of Intuitive's worldwide collaborations."*

*~ Chris Hasser*

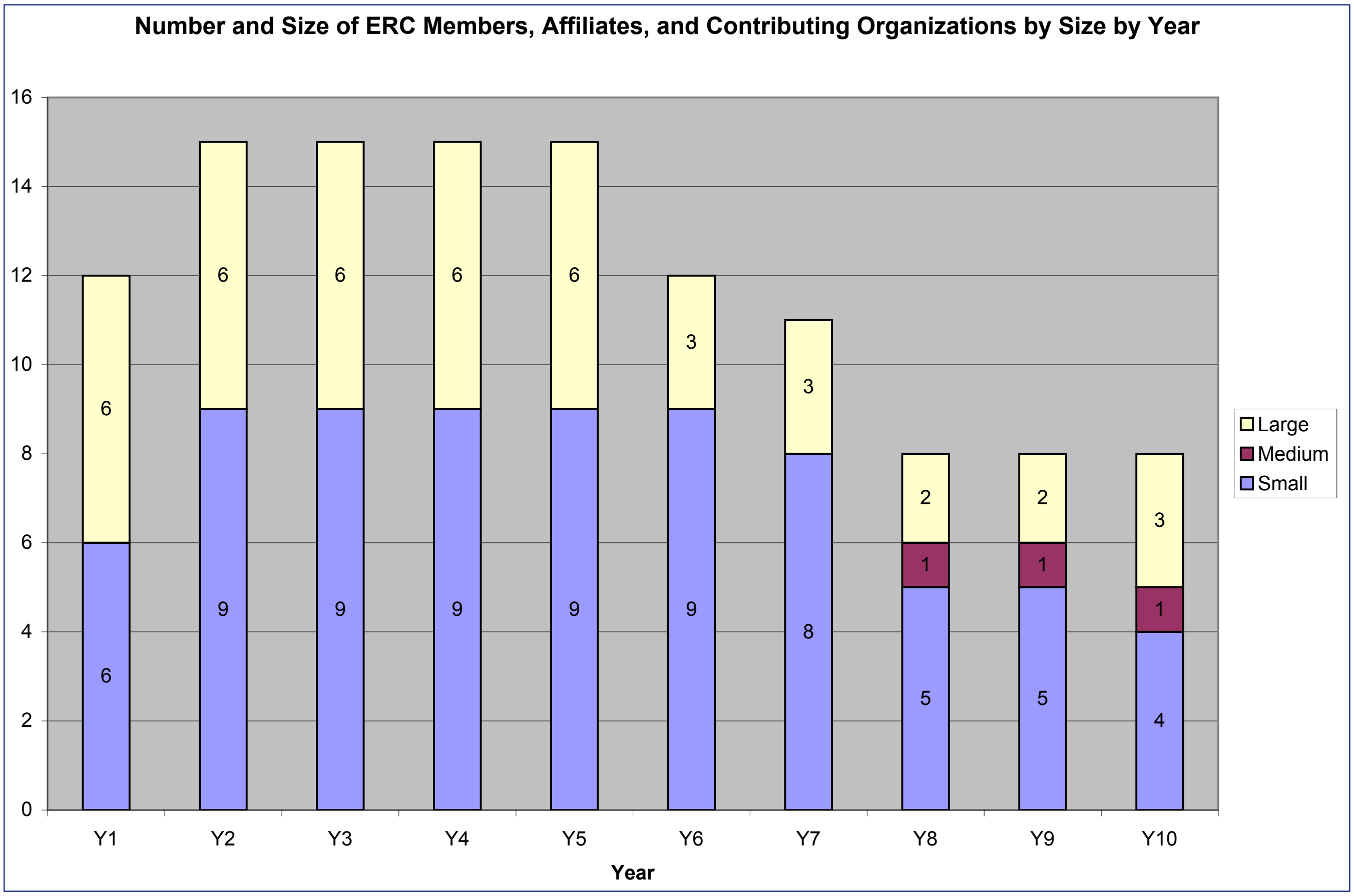

Fig. 60

*Industry*

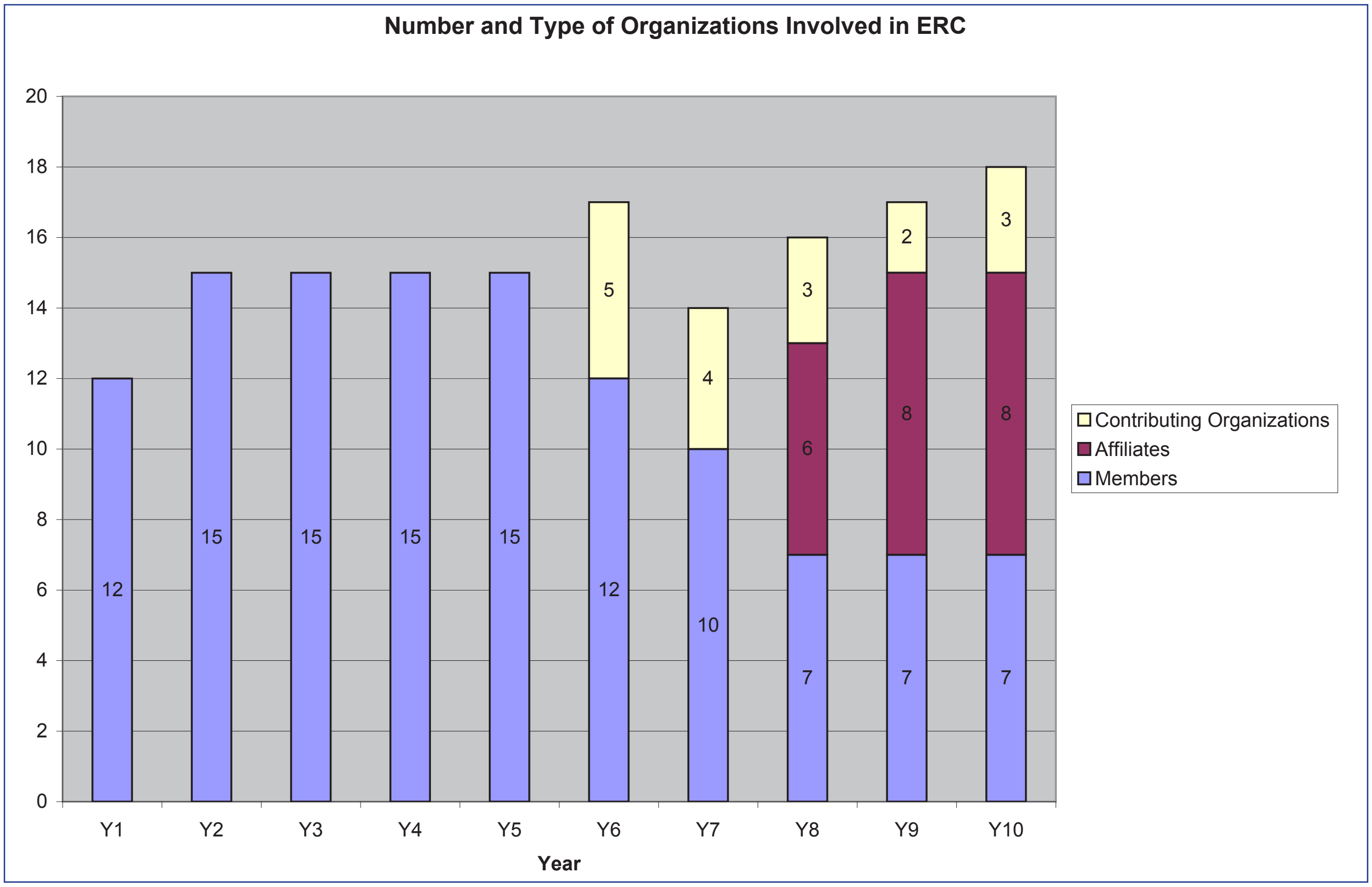

Fig. 61

*"There are new collaborations bubbling along all the time."*

*~ Claire Tempany*

| Adopting Company | Technology | Industrial Application<br>When (date) | Use in Company | Impact (e.g. cost savings, productivity gain, etc.) |
|---|---|---|---|---|
| Integrated Surgical Systems, Inc. | Bone registration and tracking using an optical system | May-01 | Internal research, clinical testing | Would have produced greater ease and ease-of-use in Robodoc surgery system |
| Intuitive Surgical | Laparoscopic Ultrasound for DaVinci Robot | 11/11/04-7/31/05; 8/15/06-8/14/09 (ongoing) | Advanced development; planned product | Improved visualization in surgery (e.g., laparoscopic partial nephrectomies) |
| Intuitive Surgical | Surgical Assistant Workstation | 9/1/2006 - 4/1/2009 (ongoing) | Advanced development | Significantly enhanced function for surgical system; enhanced efficiency in technology transfer between university research and ISI |
| Hologic | 3D Volumetric and Biomechanical Models of the Proximal Femur from DXA Projection Images | 10/1/2005-10/1/2010 (ongoing) | Advanced development; planned eventual product | Improved screening and monitoring for osteoporosis |
| Northern Digital | Characterization of EM tracker system | 6/2002-6/2005 | Feedback & evaluation on NDI product | Improved accuracyfor applications |
| Foster-Miller | Cooperatively controlled micromanipulation system for cell-level injections | 7/1/2002-12/31/2002 | | |
| Image Guide | Robotic system for image-guided needle placement | 2002-2007 | Basis for startup company (now defunct) | reduced radiation for surgeon; greater accuracy in minimally-invasive image guided therapy |

Fig. 62

| Adopting Company | Technology | Industrial Application When (date) | Use in Company | Impact (e.g. cost savings, productivity gain, etc.) |
|---|---|---|---|---|
| IMAGE GUIDE INC | Stoianovici et al. entitled "Friction Transmission with Axial Loading and a Radiolucent Surgical Needle Drive." (DM: 3171) US Application 09/026,669 filed 02/20/1998 | 2002 | Yes | Enabling Technology |
| IMAGE GUIDE INC | Susil, et al. entitled "Methods and Systems for Image-Guided Surgical Interventions" (DM: 3720) US Application 09/663,989 filed 09/18/2000 PCT Application filed 09/18/2001 | 2002 | Yes | Enabling Technology |
| IMAGE GUIDE INC | Fichtinger et al. entitled "Controllable Motorized Device for Percutaneous Needle Placement in Soft Tissue Target and Methods and Systems Related Thereto" (DM: 3752) US Application 09/943,751 filed 08/30/2001 PCT Application US01/27228 filed 08/30/2001 | 2002 | Yes | Enabling Technology |
| IMAGE GUIDE INC | Stoianovici et al. entitled "Laser Based CT & MR Registration" (DM: 3863) US Provisional Application 60/357,451 filed 02/15/2002 | 2002 | Yes | Enabling Technology |
| IMAGE GUIDE INC | Stoianovici et al. entitled "A Fluoro-serving Method For Robotic Targetting of Surgical Instrumentation" (DM: 3888) US Provisional Application 60/336,931 filed 11/08/2001 | 2002 | Yes | Enabling Technology |
| IMAGE GUIDE INC | Stoianovici et al. entitled "Adjustable Remote Center of Motion Robotic Module" (DM: 3980) | 2002 | Yes | Enabling Technology |
| SIEMENS | Fichtinger, Atalar, Krieger, Susil, Tanacs and Whitcomb, entitled "Device, Method, and System For Needle Insertion Inside a Medical Imager, From Within Body Cavity." (JHU Ref. 4033). PCT Application Serial Number US03/12253 filed April 22, 2003; US Patent Application Serial Number 10/512,150 filed October 22, 2004; EPO Patent Application Serial Number 03736475.9 filed October 22, 2004; Japan Patent Application Serial Number 2003-585588 filed October 22, 2004; Australia Patent Application Serial Number 20033237089 filed October 22, 2004; and US Patent Application Serial Number 60/782,705 filed March 16, 2006. | 2006 | Yes | Enabling Technology |
| SIEMENS | Stoianovici, Wyrobek, Mazilu, and Whitcomb, entitled "MRI Compatible Positioning Arm." (JHU Ref. 4166). US Patent 6,857,609 issued February 22, 2005. | 2006 | Yes | Enabling Technology |
| SIEMENS | Susil, and Taylor, entitled "A Single Image Registration Method for Image Guided Interventions." (JHU Ref. 3720). US Patent Application Serial Number 09/663,989 filed September 18, 2000. | 2006 | Yes | Enabling Technology |
| HeartLander Medical | HeartLander robot | 2005 | | foundation product of company |
| Johns Hopkins Department of Radiation Oncology | Robot assistant for transrectal ultrasound-guided prostate brachytherapy | Aug-07 | Clinical trial | Good clinical performance, but too early to draw conclusions |
| Johns Hopkins Department of Radiation Oncology | Small Animal Radiation Research Platform | Dec-06 | Internal research | Enabling technology, productivity gain |
| University of Arkansas Department of Radiation Oncology | Collimator for radiation therapy | Apr-08 | Internal research | Cost savings |
| Memorial Sloan Kettering Cancer Center | Image-guided robot for small animal research | Jan-05 | Internal research | Productivity gain |
| National Institutes of Health | CISST Software | Aug-07 | Internal research | Cost savings |
| Integrated Surgical Systems, Inc. | Robot-assisted bone cement removal in hip revision surgery | Jan-01 | Internal research | Improved accuracy; fewer fractures in revision THR |

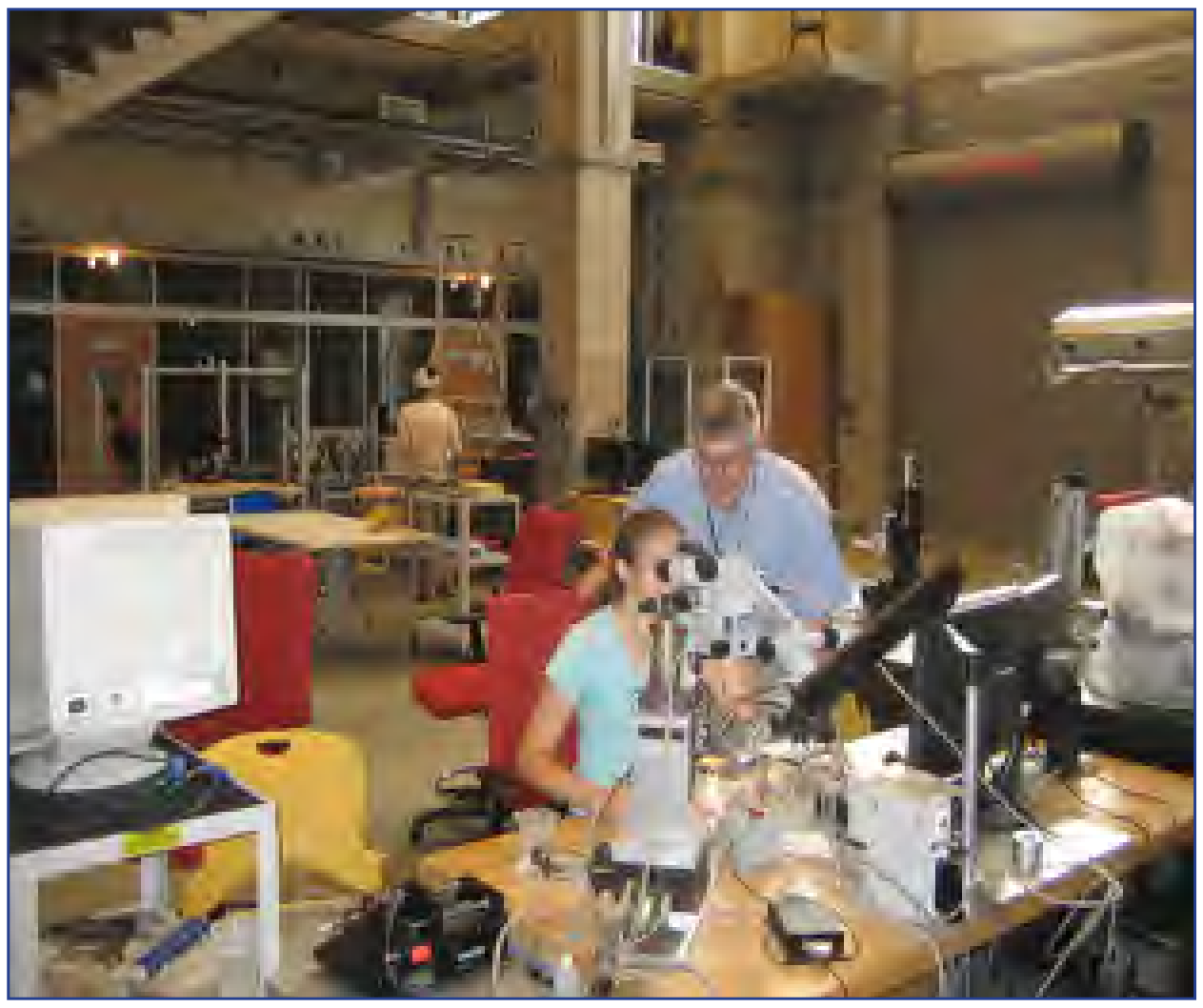

Fig: 63 and 64: CISST ERC laboratory space in the new Computational Science and Engineering Building on the Johns Hopkins University Homewood Campus. (Above) The high bay lab, showing a steady-hand microsurgery assistant robot in the foreground and two telesurgical robots in the background. (Below) A view of the Richard A. Swirnow Computer-Integrated Surgical and Interventional Systems Mock Operating Room from an open air walkway passing through the building, showing an experiment with a steady-hand robot assistant for neurosurgery.

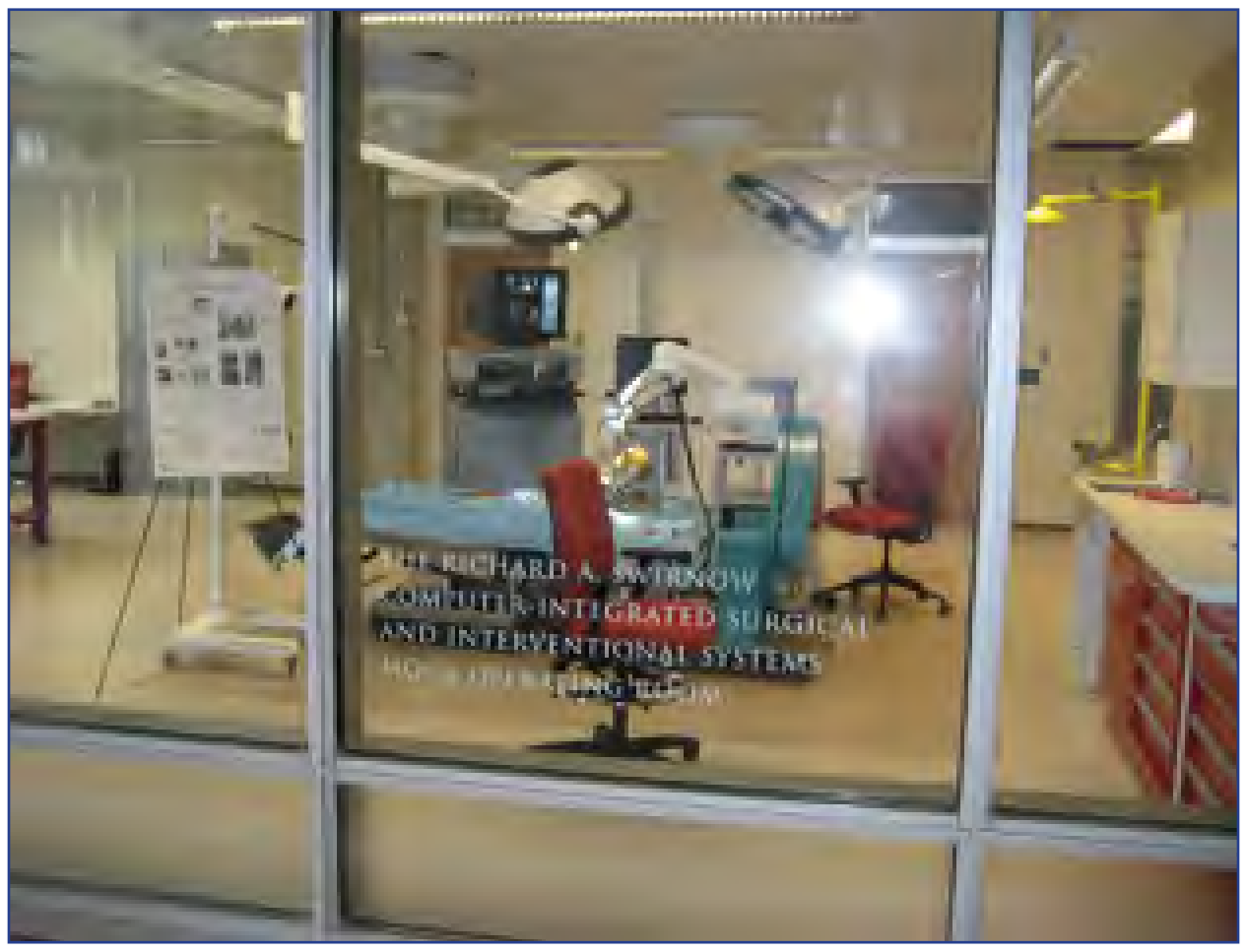

*Major Facilities and Capabilities Developed*

One major institutional impact of the CISST ERC has been the development of a world-class robotics program at JHU, with a major focus area of medical robotics, but also including a much broader range of exceptional research. This impact has included development of key faculty and staff (see below), physical and equipment infrastructure, and a web of collaborative relationships and research funding. Currently, there are 15 faculty and research faculty at JHU working robotics and computer vision, in addition to a number of others working in medical imaging and sensing. All of the robotics faculty have or have had one or more active projects involving ERC-related research.

In 2007, the robotics faculty created LCSR, concurrent with the opening of CSEB on Johns Hopkins' Homewood campus. CSEB was designed as a venue for interdisciplinary research and education and the school's strong ties to medicine and the life sciences. In addition to providing high visibility laboratory and office space and other infrastructure while promoting collaboration among key CISST faculty, LCSR also promotes closer communication between CISST research and that of other JHU Centers and Laboratories in CSEB, such as the Institute for Computational Medicine.

LCSR occupies 15,000 square feet on the bottom two floors of CSEB, and houses the CISST ERC's main laboratory and office space. Eight robotics faculty members -- including the ERC Director, Deputy Director, and two of the three research thrust leaders -- have their primary offices within LCSR, as well as the ERC administrative staff, engineering staff, and many of the ERC's post-docs and graduate students. Another five robotics faculty members also have laboratory space within LCSR. The laboratory facilities include a shared high bay laboratory with individual laboratories opening off it, a small machine shop, and the Richard A. Swirnow Computer-Integrated Surgical and Interventional Systems Mock Operating Room. The building also includes conference rooms and a small auditorium used for ERC seminars and other colloquia.

*Faculty, Staff and Students*

The ERC has many significant influences on the field of Computer–integrated surgery, including the development of personnel – students, faculty and staff whose careers are greatly impacted by their association with the ERC. These people leave for various academic, government and industry positions, spreading ERC influence. The ERC's impact has been felt at the lead institution, JHU, where it is directly responsible for hiring tenure track and research faculty:

• Dr. Greg Hager – The first ERC tenure track faculty hire, he is now Deputy Director and Associate Director of Research for the ERC, and was named an IEEE fellow since joining the center.

• Dr. Allison Okamura – The second ERC tenure track hire, she has since taken over as leader of the Surgical Assistants Thrust. Dr. Okamura was recently named the first Gilbert Decker Faculty Scholar for her achievements in the area of haptics. She has also won career, teaching, and diversity awards.

• Dr. Gabor Fichtinger – Hired as Engineering Director, Dr. Fichtinger became an Associate Research Professor and Surgical CAD/CAM Thrust Leader. He is now an Associate Professor at Queen's University in Toronto while continuing to lead Thrust 2. Dr. Fichtinger has won many awards for his mentoring and teaching.

• Dr. Peter Kazanzides – Hired as Engineering Director upon Dr. Fichtinger's promotion and is now an Associate Research Professor and Infrastructure Thrust leader. Dr. Kazanzides directs the engineering staff and makes a significant impact mentoring ERC students.

• Dr. Iulian Iordachita - Hired as a mechanical engineer, he is now a research scientist. He leads significant design effors in medical robots and systems while also mentoring students and junior staff.

As a result of bringing in talented faculty, the ERC has been able to produce some extremely talented students. These students go on to jobs in industry and academia, or pursue advanced degrees. Some of our students have been hired as faculty at JHU. Dr. Emad Boctor earned his Ph.D. working with Gabor Fichtinger and our industry affiliate Clif Burdette and was subsequently hired by the JHMI Department of Radiology as an instructor. He is also appointed in the JHU CS department as an Assistant Research Professor. Similarly, Dr. Rajesh Kumar, our first ERC student to graduate, went into industry working with one of our affiliates, Intuitive Surgical, before returning to the ERC as an Assistant Research Professor.

The ERC has assisted students and postdocs in launching academic careers at our affiliated institutions. Dr. Polina Goland, an Associate Professor at MIT, received her Ph.D. with support of the ERC. Dr. Nabil Simaan, a former ERC postdoc at JHU, is an Assistant Professor in the Department of Mechanical Engineering at Columbia University.

ERC students are well trained in the field, and have won numerous awards, including many best paper and poster awards at major conferences. Many students have been awarded NSF and other competitive government fellowships. One visiting student, Paweena U-Thainal, won the National Outstanding Student Award from the Ministry of Science and Technology of Thailand, her home country. She is now pursuing an advanced degree at Queen's University. Carmen Kut, a JHU undergraduate, was one of 20 students named to USA Today's All-USA College Academic First Team, partly due to her research activities with the ERC and the Department of Radiation Oncology at JHMI.

*Overview of our Diversity Strategic Plan*

A major goal of the CISST ERC is to embrace the cultural, gender, racial, and ethnic diversity of the U.S. in the composition of its leadership, faculty, staff, and associated students, as a means of providing talented people the opportunity to pursue a career in engineering research and education. Although this is an area in which one can never do enough, we are proud of our accomplishments in developing an effective program for enhancing diversity within our institutions.

To achieve this goal, the CISST ERC established a Diversity Committee in November 2003 to develop a specific Strategic Plan to increase the number of women and underrepresented minorities in the ERC, to develop a community that embraces its diverse composition and to promote and highlight the achievements of the members of the community. This Committee leads the ERC through a series of actions providing effective solutions to the diversity challenges within our Center. Through research, informal discussions, and professional consulting, we create an understanding of diversity challenges facing the center, our school, and engineering education as a whole. We educate and train our faculty, staff, and students on diversity issues, addressed specifically within the context of higher education, which has different standards and expectations than the typical workplace. We devise and implement specific measures to improve performance in all aspects of our strategic plan. Finally, we set up mechanisms to track and report progress and problems in meeting our goals. Evaluation is critical to demonstrating the impact of our diversity activities.

*"Diversity results in better engineering, and if the United States is to be successful, we need to draw on a broader talent pool and bring the best people on board."*

*~ Allison Okamura*

# Diversity

The original Diversity Committee was transformed into Education, Outreach and Diversity Committee (EODC) in 2006. Its mandate was expanded to oversee all the education and outreach activities in CISST, as well as developing and implementing a strategic plan for improving diversity in our ERC, with specific goals in the areas of recruiting, retention, community, and achievement. It is an ongoing task of the EODC to examine the policies and practices that influence the composition of our leadership, faculty, staff, and students. This will help us to understand how to improve our diversity composition, take concrete steps toward our objectives, and establish better mechanisms to monitor our performance. The EODC is co-chaired by Jerry Prince (former Associate Director for Research) and Ralph Etienne-Cummings, the Associate Director for Education and Outreach. The other members of the Committee are drawn from every level and every core institution in the Center, including faculty members from Carnegie Mellon University, Massachusetts Institute of Technology, and Morgan State University. The EODC make-up is rotated annually; however, Profs. Prince and Etienne-Cummings have stayed on as Co-Chairs since 2003. The EODC is a permanent component of the ERC and is the primary mechanism for developing and implementing our strategic plan for improving diversity.

The Strategic Plan developed by the earlier Diversity Committee and subsequently improved upon by the EODC can be divided into three elements:

1. Recruiting and Retention. The ERC will develop programs and provide resources to help retain a diverse population of graduate and undergraduate students in engineering. This includes outside students who visit the center for summer research projects or short tours.

2. Community. The ERC will provide a sense of community, such that all students, faculty and staff feel comfortable communicating their needs and effecting change. Mentoring will occur at many levels. We will continue to provide resources to CISSRS (the Computer-Integrated Surgery Student Research Society) to recruit and retain active members and organize activities of interest to students.

3. Achievement. The ERC will provide exceptional support to all students, faculty, and staff so that they can achieve their professional goals. As more women and under-represented minorities distinguish themselves, there will be more diverse leadership in the engineering profession and relevant role models for tomorrow's engineers.

We are proud to say that we have successfully achieved many of our aims while continuing to work on some of the illusive ones. The following are some examples of the activities we have used to implement the three prongs of our Strategic Plan:

***Recruiting and Retention***

The recruiting issue can be examined at several levels: recruiting graduate students to our center; recruiting engineering undergraduates to our schools; and recruiting a diverse population of students to enter engineering study in general by creating a pipeline reaching back to K–12. The Committee's top priority over the last several years has been to recruit a diverse population of students to our center as well as to JHU. They placed special emphasis on recruiting women and minority graduate students to JHU this year, through the following actions:

• We targeted women and minority universities on the NSF list in our REU program advertising.

• We upgraded our website for improved visibility of our students, emphasizing their involvement in the research activities of the center.

• Faculty representatives screened all applicants from their respective departments, specifically looking for promising women and minority candidates and invited them to visit on the ERC Visit Day.

• We established a relationship with the Howard University LSAMP program as well as other programs nationally to encourage students to participate in our REU programs and to apply for graduate school.

• We sent representatives to Historically Black Colleges and Universities (HBCU) to give research talks and encourage students to apply to JHU graduate programs. We also participated in Career Day for various minority serving organizations.

*Community*

The Committee recognizes that an organizational and social culture that supports diversity is required. Part of the EODC's charge has been to encourage and sponsor activities within the ERC that contribute to an overall sense of well-being and belonging such as the following:

• We host a series of weekly seminars and activities focusing on teamwork, communication, and diversity, including "kick-off" seminars to help orient new students.

• We publish an ERC "social" e-letter to provide an opportunity to inform others (primarily students) about personal events of interest as well as broadcast job and Fellowship opportunities.

• We conducted an ERC-wide "climate survey" to learn what works well and what needs improvement in the area of Diversity. Results were discussed in a mandatory seminar. Last year, the JHU Human Resources Department conducted an institution-wide climate survey that was, in part, patterned after our survey.

• The EODC has a permanent representative on the JHU Whiting School of Engineering (WSE) Diversity Council. This council advises the Dean on diversity and climate issues. Prof. Prince serves on the University-wide Academic Council which oversees all matters of scholarship, promotion, and tenure. These positions provide us with important input into the processes at both WSE and JHU levels. A new review category on diversity and climate was added to faculty annual reports due largely to EODC suggestion.

• Our student research organization, CISSRS, plays a major role in energizing the student population, and membership continues to increase dramatically. Among other things, CISSRS organized a weekly student seminar series.

• ERC women play a major role in exposing issues that concern women in engineering to the wider JHU community. The Women of Whiting, an organization of female students and professors, invites national speakers to JHU. ERC funds partially cover their related expenses.

*Achievement*

The ERC is committed to helping each faculty, student, and staff member better achieve his or her professional goals. Also, many of our outreach programs focus on the achievement of K-12 students and others. These activities promote a culture of achievement within the ERC which ideally should be pervasive and infective. There are numerous examples of the achievement of women and minorities within our ERC. Here are some specific achievements, as well as programs and strategies undertaken to help in the achievement goals of all students, faculty, and staff:

• Many of our graduate students have received the NSF Graduate Research Fellowship. Recent examples are Netta Gurari (ME), Carol Riley (CS), Jacob Vogelstein (BME) and Kathrin Tsai (ECE).

• Science Spectrum magazine named Dr. Ralph Etienne-Cummings a "Trailblazer." "The Science Spectrum Trailblazers are outstanding Hispanic, Asian American, Native American, and Black professionals in the science arena whose leadership and innovative thinking on the job and in the community extend throughout and beyond their industry."

• Professor Allison Okamura was named the first Gilbert Decker Faculty Scholar at Johns Hopkins Univeristy, in recognition of her exceptional achievements in the area of haptics.

• Professor Ralph Etienne-Cummings won Fulbright and Visiting African Fellowships to spend six months at the University of Cape Town (UCT), South Africa. He taught classes, advised students and conducted research with UCT faculty. Two of his UCT advisees have been admitted into the ECE Ph.D. program at JHU and will work on ERC related projects.

• Ndubuisi Ekekwe, a Diversity Fellow in the ERC, won the UK Computer Assisted Orthopaedic Surgery (CAOS) 2007 fellowship. Ndubuisi is advised by Drs. Ralph Etienne-Cummings and Peter Kazanzides.

• Dr. Allison Okamura received an NSF CAREER award.

• Our ERC has been active in K-12 activities:

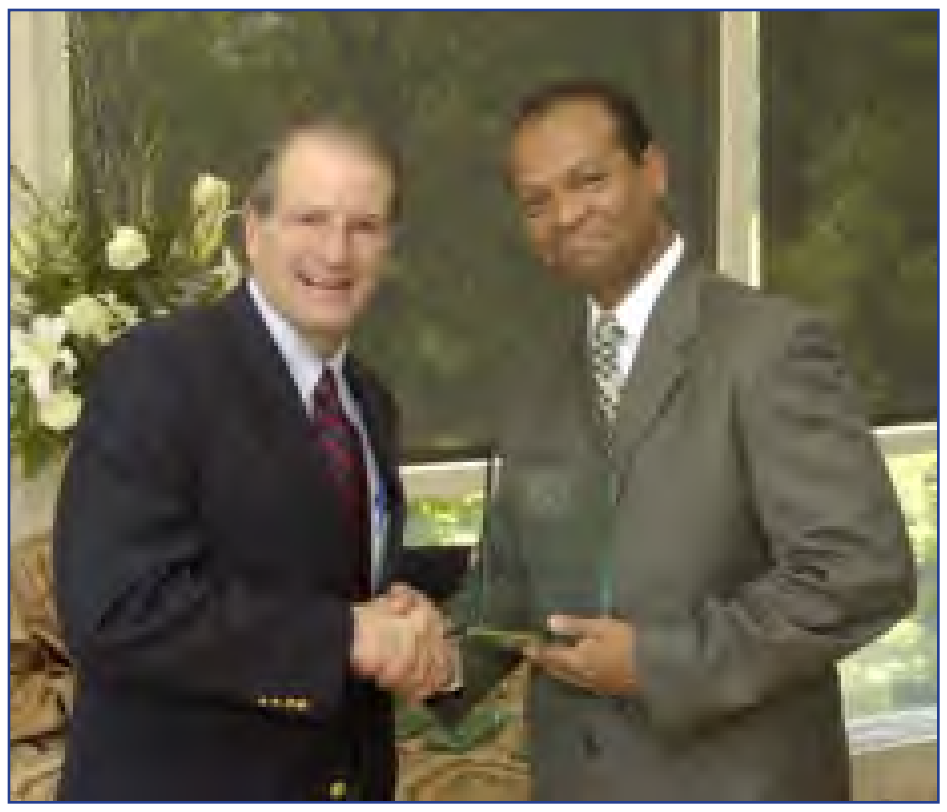

Fig. 65: JHU President Bill Brody congratulates Ralph Etienne-Cummings on his Diversity Leadership Award

* The Center for Women and Information Technology at UMBC hosted their fourth annual Computer Mania Day where over 800 middle school girls and their parents participated. The program provides a broad based introduction to the ways different careers make use of information technology. Women graduate students from the ERC conducted break-out sessions for the middle school girls.

* For the past five years, the ERC sponsored a middle-school Summer Robotics Camp to engage students in robotics, computers, and basic engineering principles. In 2005 we received a commendation for this program by Congressman Elijah Cummings.

* For the past three years, the ERC has organized The Robotic System Challenge for high school and middle school students. JHU Admissions staff provides mini-seminars on college admissions, the importance of doing well in high school, and what steps to take to ensure a successful college experience. In 2007, The Baltimore Sun

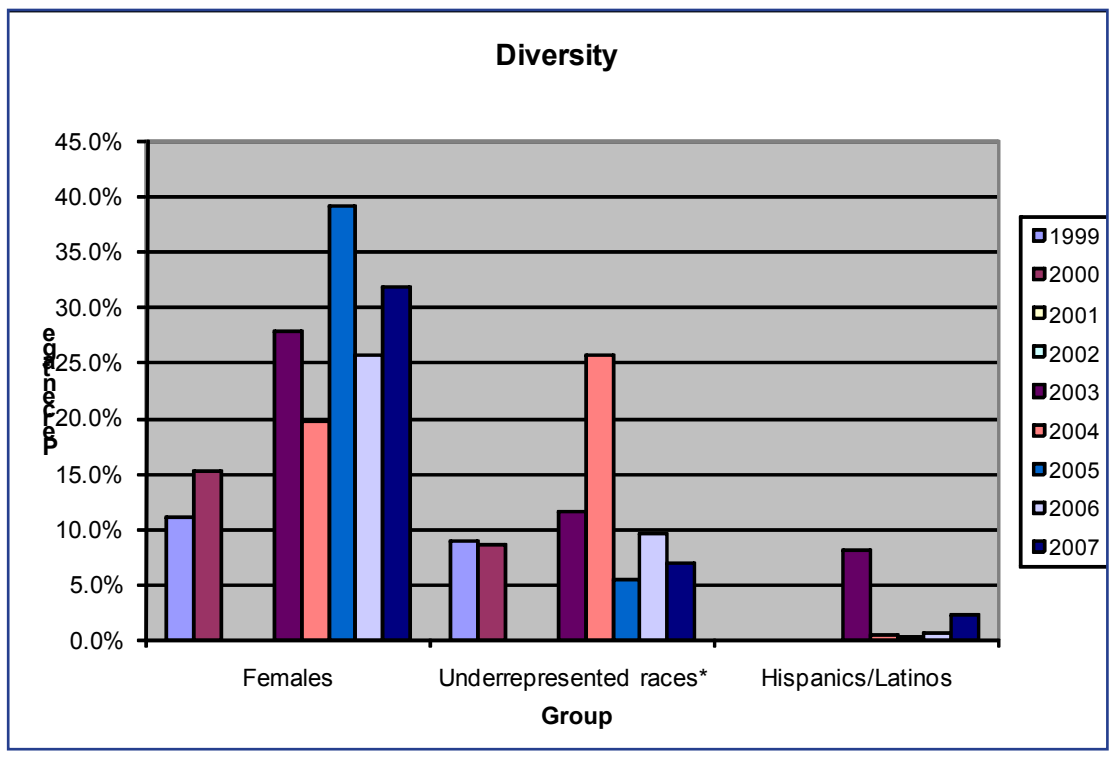

Fig. 66A: Demographic make-up of all constituents in the ERC CISST.

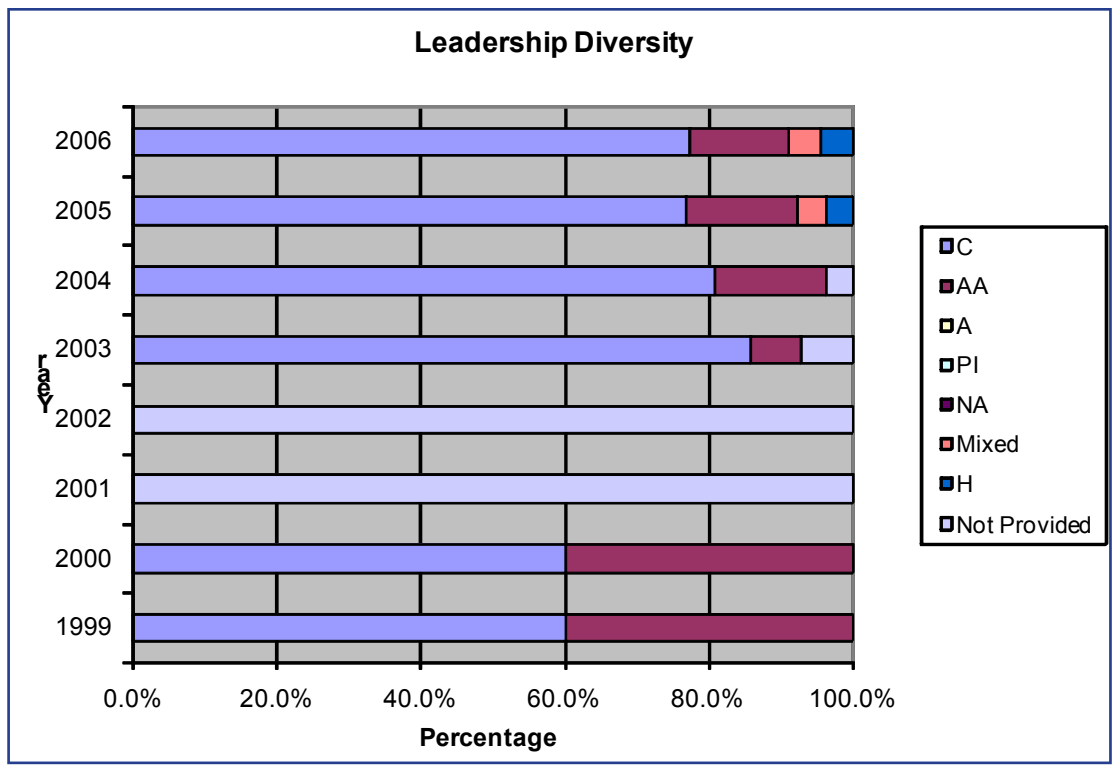

Fig. 66B: Demographic make-up of Leadership in the ERC CISST. No data exists for 2001 and 2002.

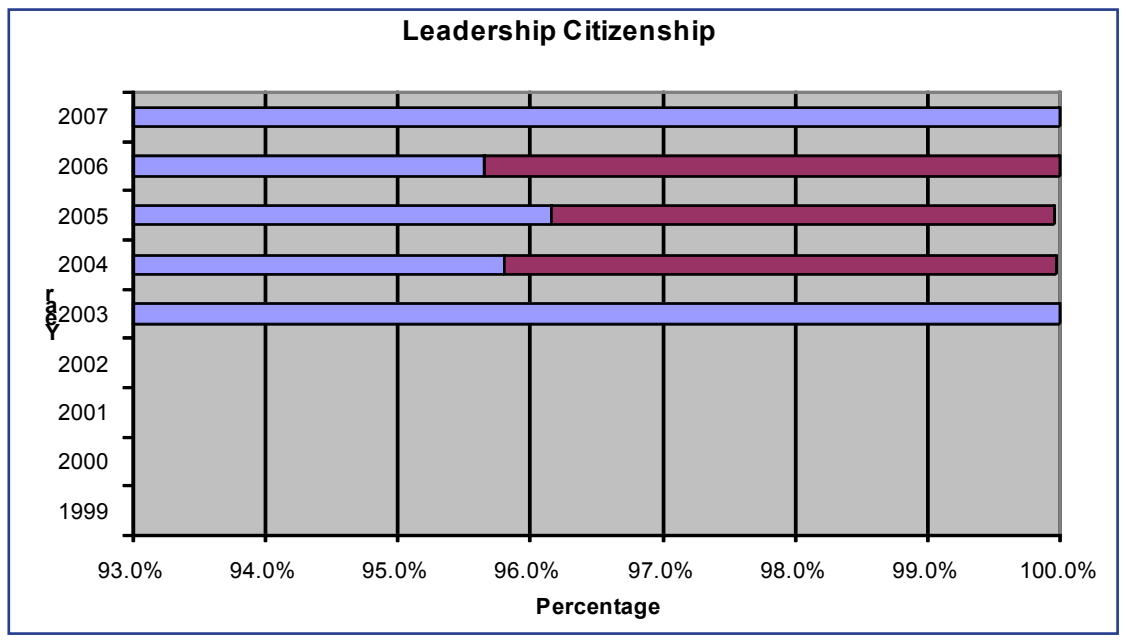

Fig. 66C: Citizenship of Leadership in the ERC CISST. No data exists before 2003.

published an article with many pictures to capture the enthusiasm, competitiveness and camaraderie of the day.

- The ERC staff members are encouraged to increase their skills and knowledge through Professional Development courses offered by Johns Hopkins University, leading to an increase in productivity and efficacy in the ERC.

### *Achievements and Outcomes*

The CISST ERC has made significant progress toward improving our diversity. We have been successful in recruiting five faculty members who are women and/or under-represented minorities. These faculty members participate at various levels of the ERC's leadership, e.g. Associate Professor Allison Okamura is a Thrust Leader and Professor Ralph Etienne-Cummings is an Associate Director. Both Profs. Okamura and Etienne-Cummings started as Assistant Professors in their respective departments, and achieved promotion partly as a result of their work with the ERC. We built strong research ties with Morgan State University (an HBCU), engage in collaborative research with female clinicians at JHU and BWH, and connect with Native Americans, African Americans, and Hispanics through several K – 16+ educational and outreach programs. Figure 67 shows the demographic trends of the ERC's constituents over the life of the Center. Figure 3 shows the diversity and citizenship of the students. To the best of our knowledge, both Figures 2 and 3 represent the CISST ERC's diversity numbers. However, it is sometimes difficult to accurately capture numbers of students participating on projects who are not on payroll, and many students choose not to report their demographics.

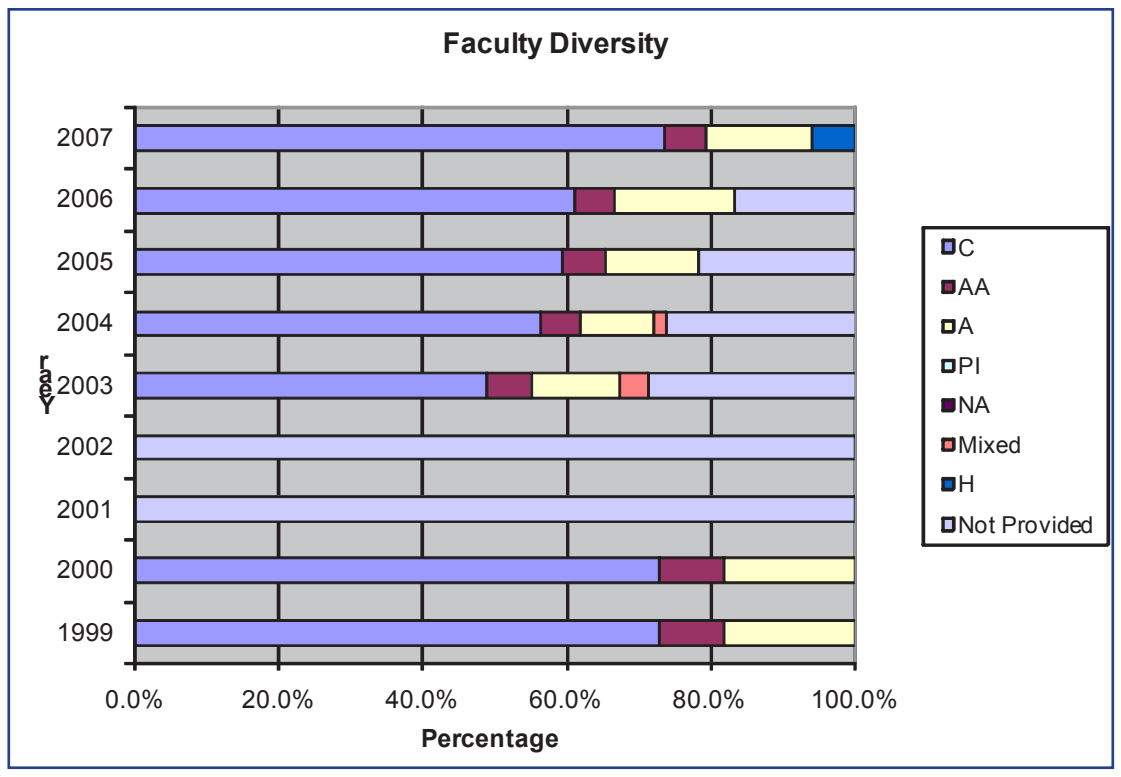

Fig. 66D: Demographic make-up of the Faculty in the ERC CISST. No data exists for 2001 and 2002.

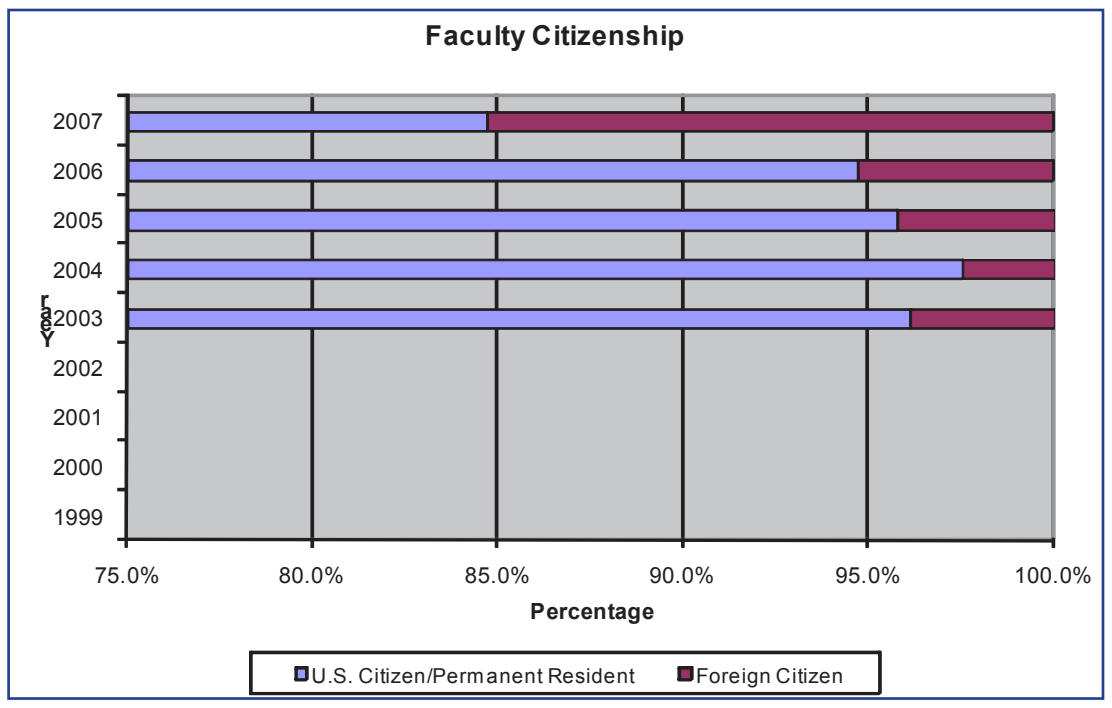

Fig. 66E: Citizenship of the Faculty in the ERC CISST. No data exists before 2003.

### *Infrastructure, Outreach, Collaboration*

Associate professor Allison Okamura, who has been a faculty member of the ERC since 2000, has benefited a great deal from its infrastructure, outreach and collaborative environment.

"The ERC's engineering staff support helped me develop complex robotic systems much more quickly than I could have on my own," she says. "And I was able to use the existing software framework to jump-start my research, as have many others here and at other universities. We're now building on that model through a separate NSF grant to make it more broadly applicable for robotics manipulation. It will likely be used by hundreds of labs."

Okamura credits ERC educational and diversity initiatives with helping her recruit and educate the best students, especially from groups that are underrepresented in engineering. "Greater diversity in the ERC has become a model that is changing the culture of the university as a whole," she says.

Regarding collaboration, Okamura says, "You need to bring multiple engineering disciplines to bear to build these complex systems, and the successful application of our work requires interaction with clinicians. There are more such collaborations through the ERC than I can count."

*A major goal of the CISST ERC is to embrace the cultural, gender, racial, and ethnic diversity of the U.S. in the composition of its leadership, faculty, staff, and associated students, as a means of providing talented people the opportunity to pursue a career in engineering research and education.*

# *Diversity*

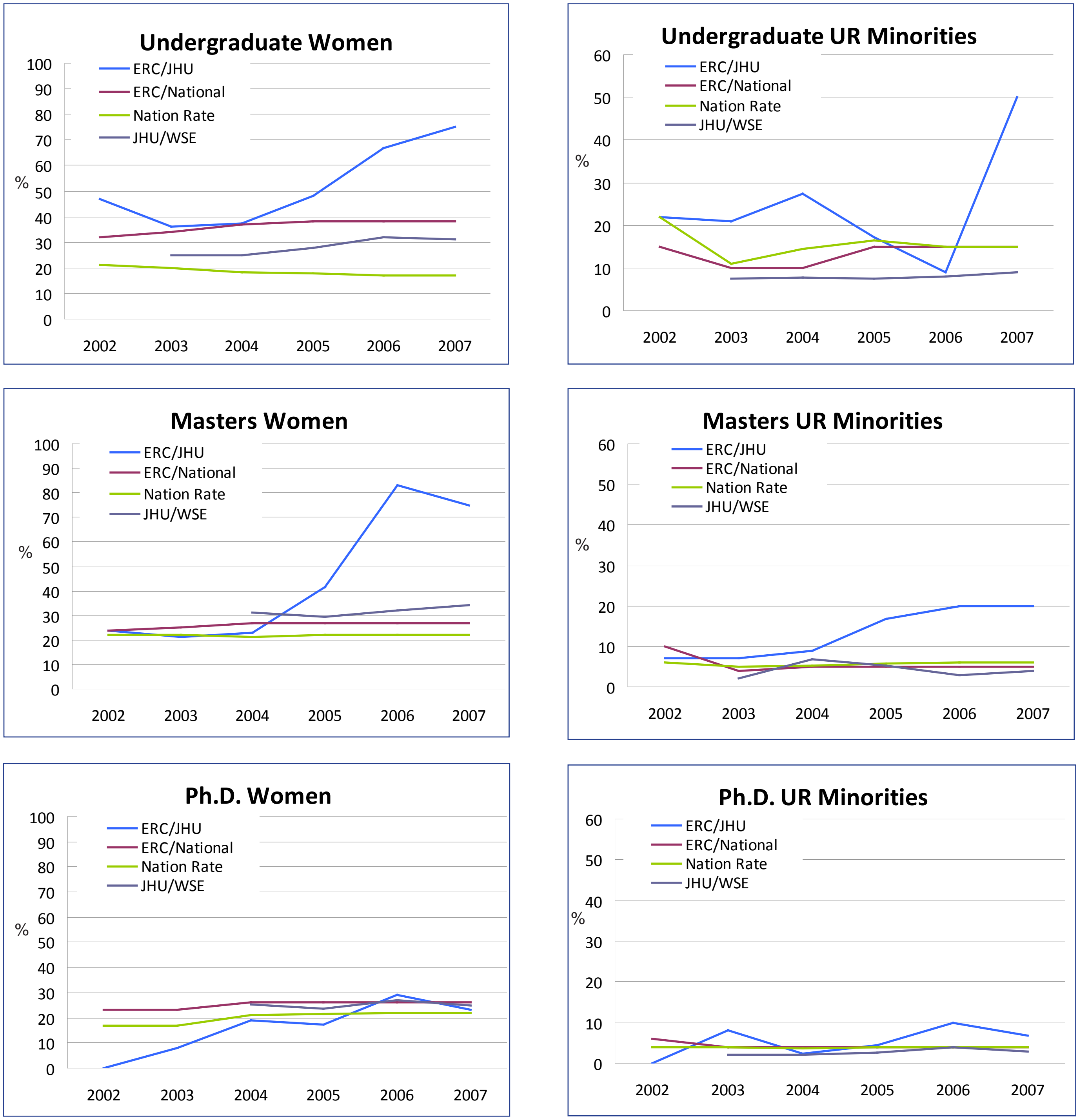

Fig. 67: Participation rate of female and under-represented (UR) minorities for undergraduates (UG), Masters and Doctoral students in the CISST ERC (blue), national average for ERCs (red), National engineering schools (green) and for JHU Whiting School of Engineering (purple).

The Engineering Research Center for Computer-Integrated Surgical Systems and Technology began on September 1, 1998 with an award from the National Science Foundation. Over the subsequent ten years the NSF provided a total of $33,069,922 in funding to the CISST ERC (Fig. 75). This figure includes core ERC funding as well as supplemental funding to support a variety of special purpose programs, both education and research related. In addition to funding provided by the NSF, Johns Hopkins University provided support to CISST ERC in the form of cost sharing. Total JHU cost share provided over the life of the Center totals $14,132,821 (this figure does not include cost share funding provided by core partner universities). The CISST ERC also obtained significant funding from other sources such as NIH, other NSF programs, other government agency grants and industry funding. Total support (expenditures) for the CISST ERC over its lifetime exceeds $64.5 million dollars. The CISST ERC continues to secure funding for our mission. Projects related to CISST ERC research objectives make up a significant portion of the funding received thus far through LCSR.

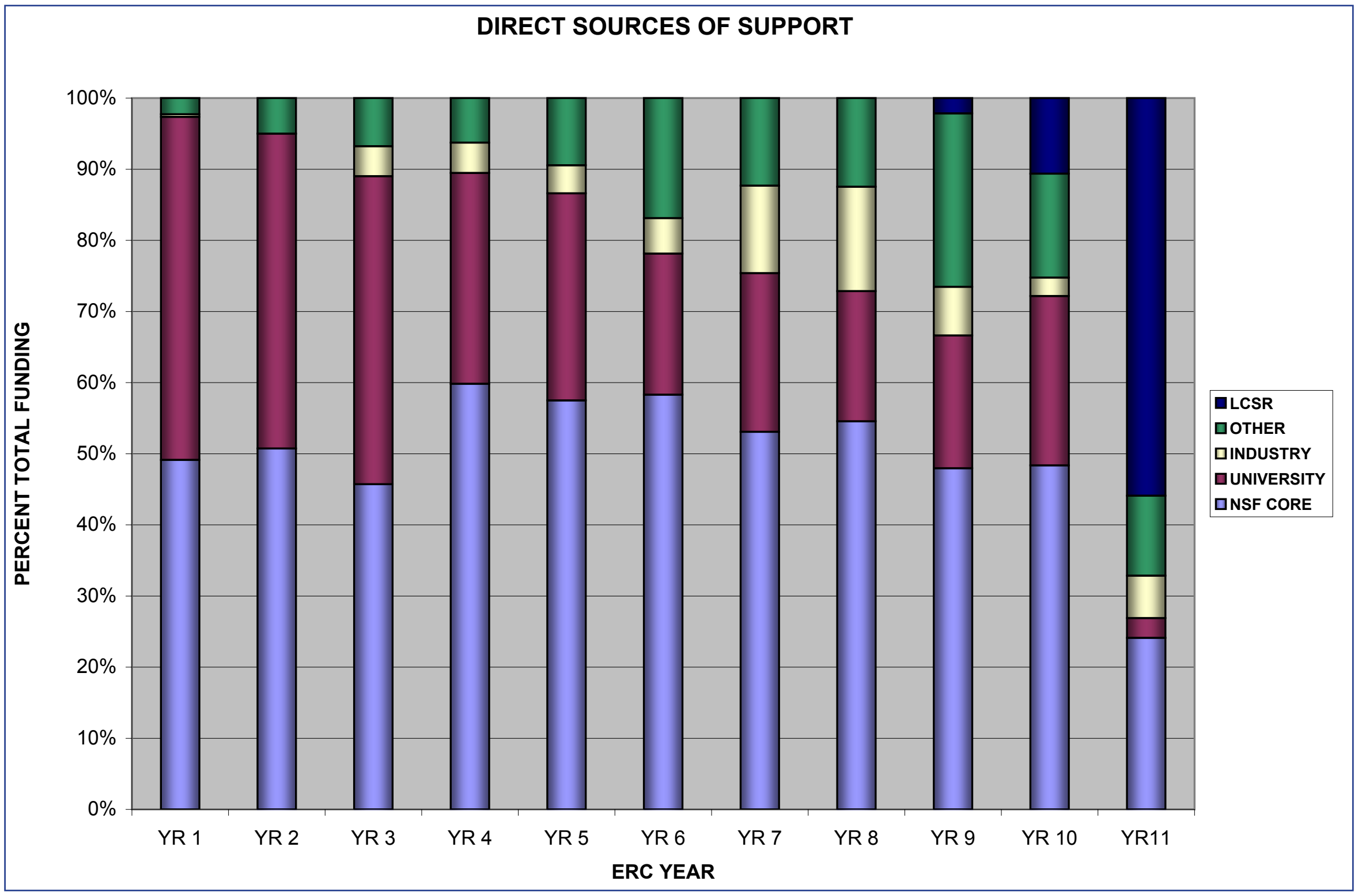

Fig. 68: Direct sources of support including ERC core funding for each year of the CISST ERC. ERC core includes core unrestricted funding as well as special purpose supplement funding. University figures include certified cost share expended at JHU, the lead institution, as well as at core partner institutions. Funding through LCSR is included as a separate unit since the CISST ERC will become a center within the LCSR as part of the ERC graduation strategy.

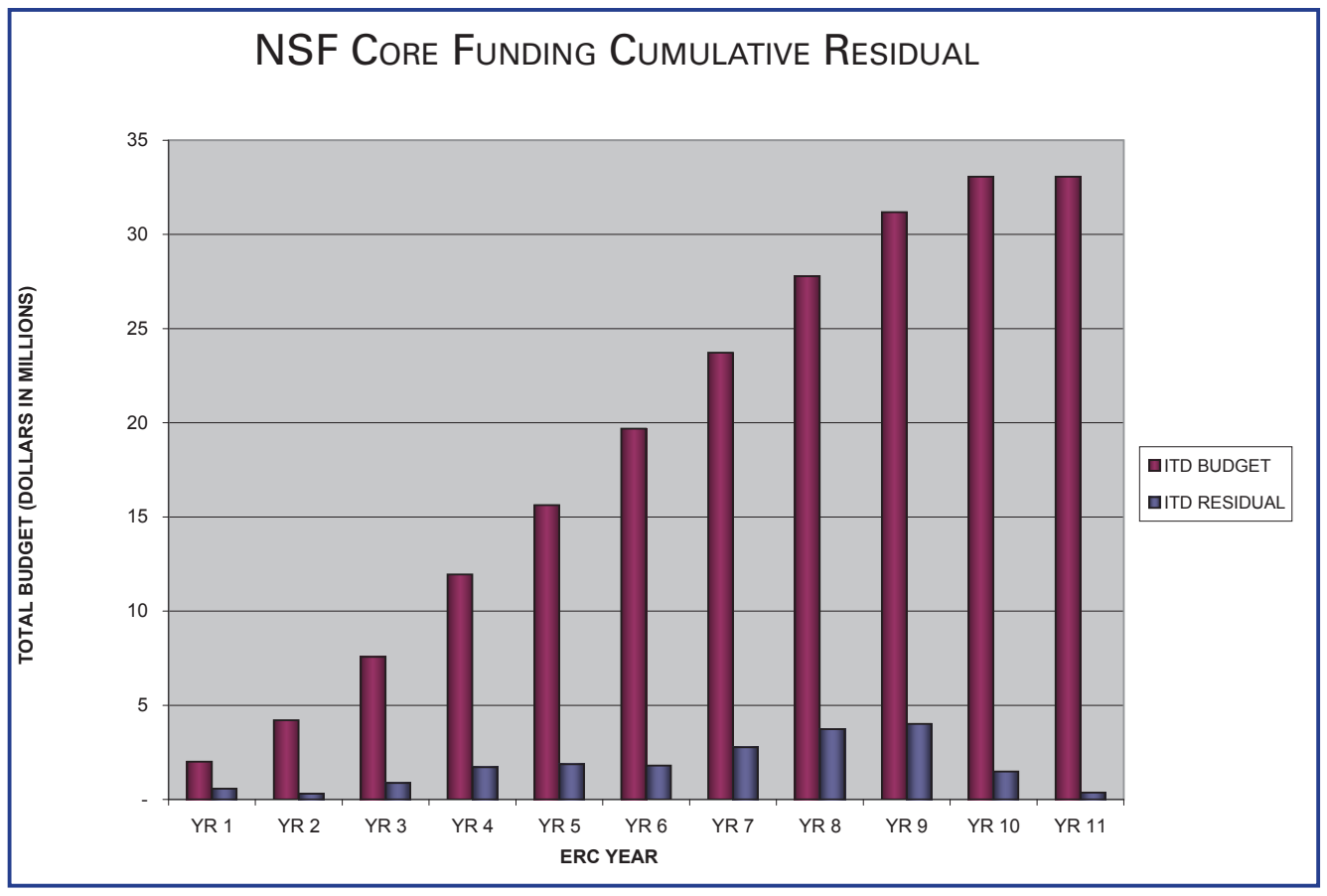

Fig. 69: Shows cumulative total funds and culmulative residual funds over the life of the Center. Numbers are based on ERC core and supplemental funding. Residual values include funds actually expended only, and do not include obligated funds. Residual for YR 11 includes residual funds attributable to the Surgical Assistant Workstation project which is on no cost extension through December 31, 2009.

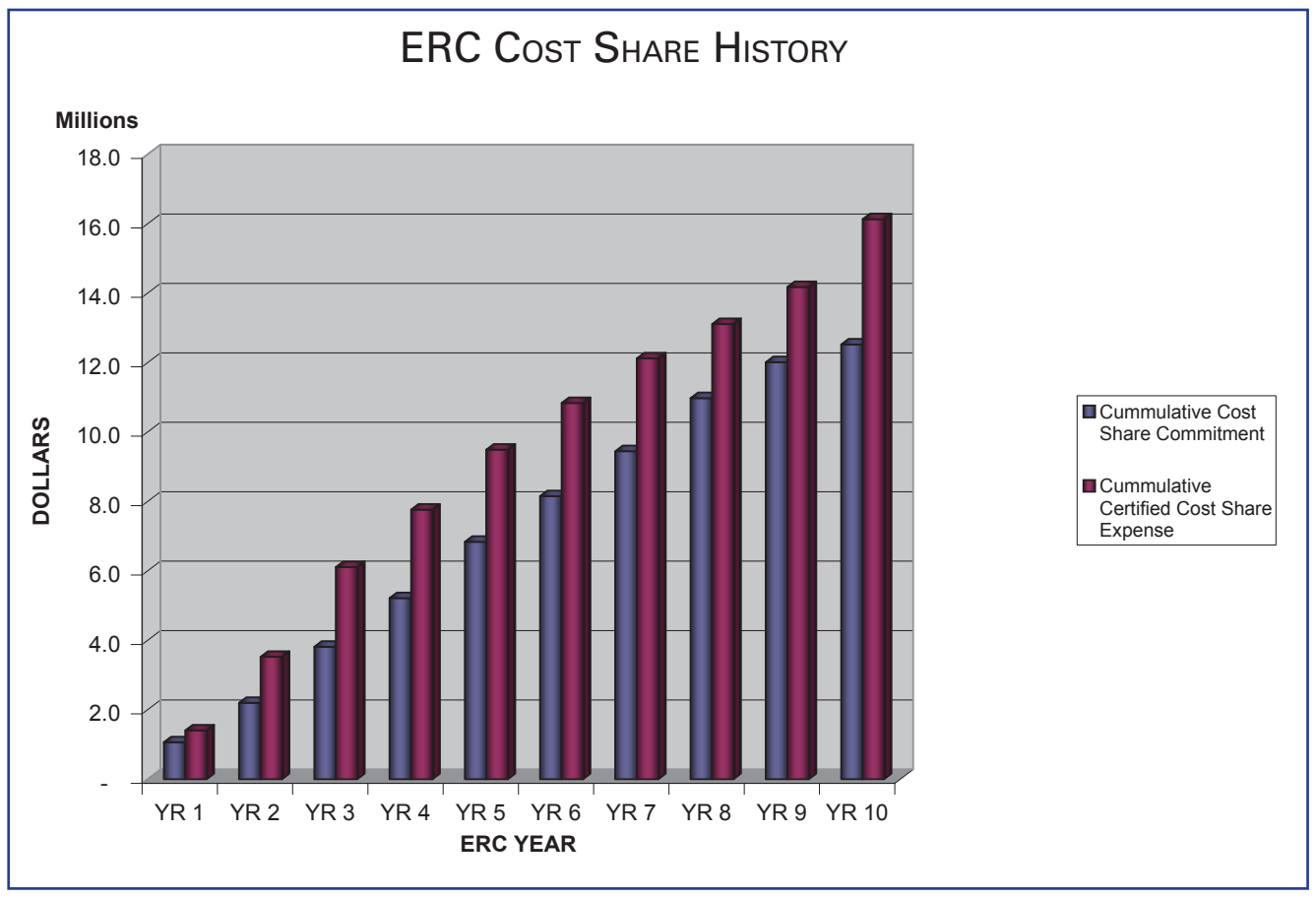

Fig. 70: Shows the history of CISST ERC cost share commitments and expenditures. Cost share expenditures include expenses incurred at the lead institution and all core partners. Total cost share expended over the life of the ERC was $16,131,626. This figure includes JHU tuition support of ERC students. Total ERC cost share commitment as stated in award letters was $ 12,522,096.

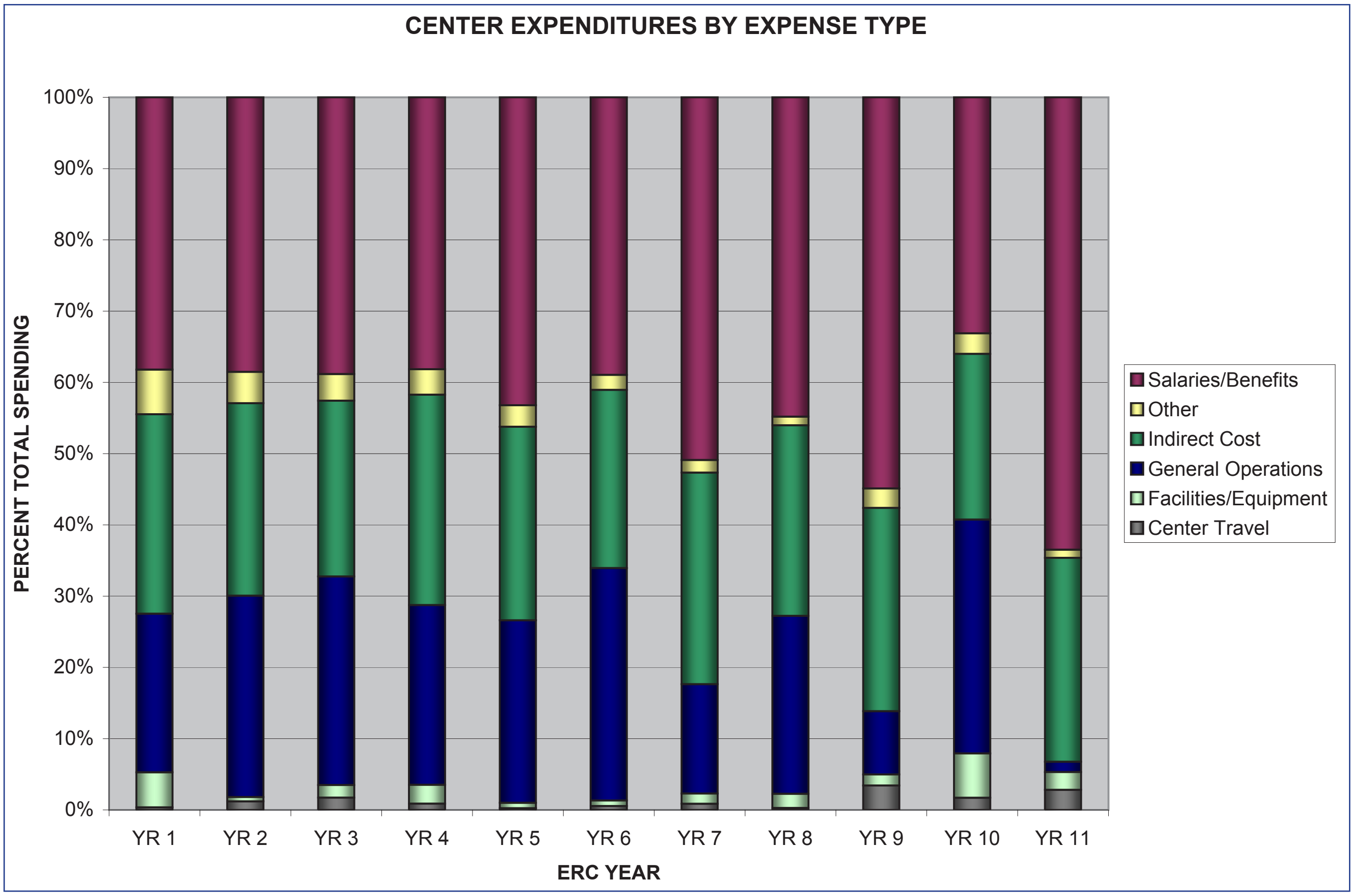

Fig. 71: Distribution of CISST ERC funding by type of expenditures. Salaries and benefits include all ERC salaries, including student salaries and tuition support.

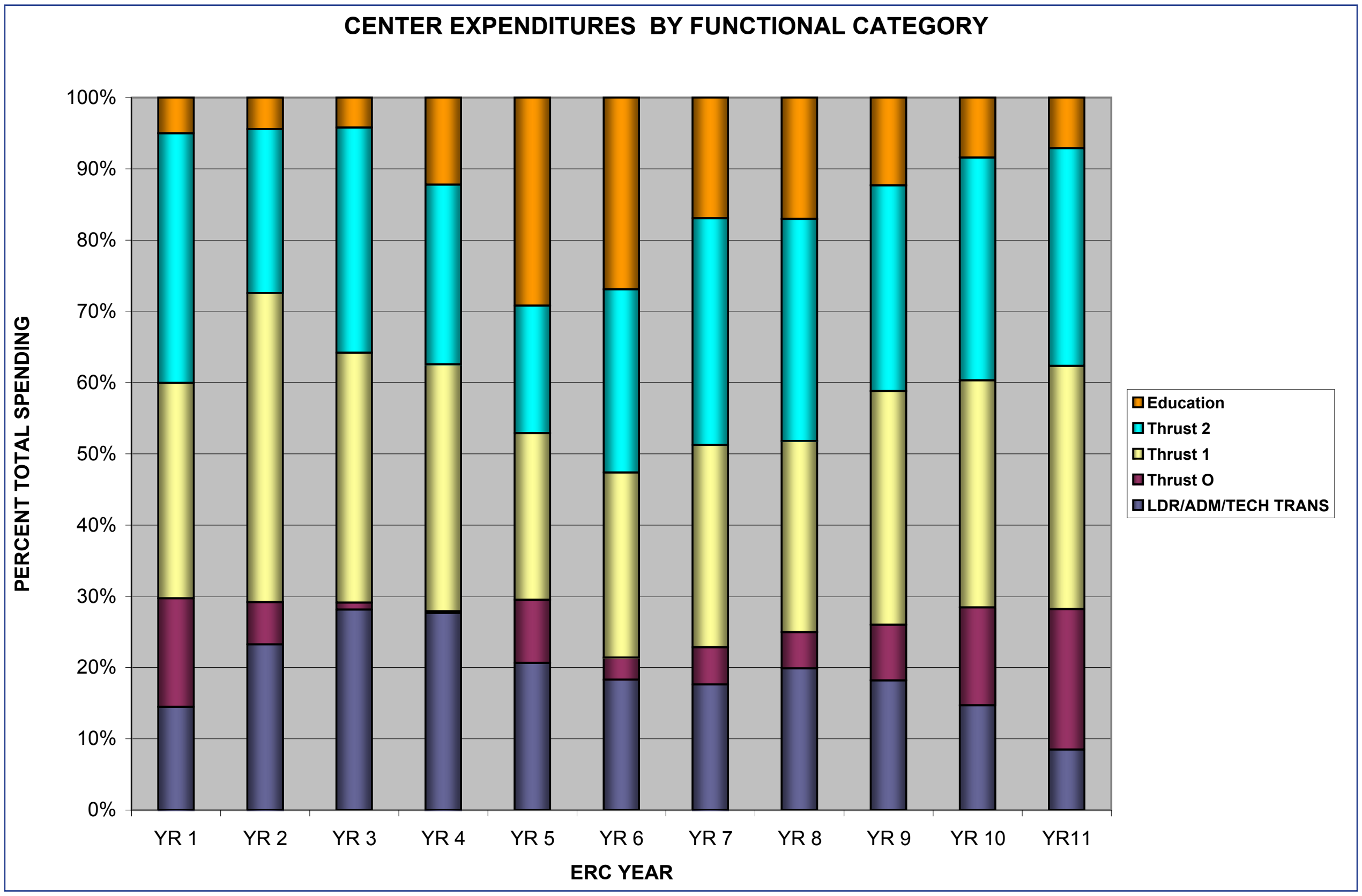

Fig. 72: shows annual funding allocation to the various ERC functional groups. Expenses associated with Leadership, Administration and Technology Transfer (LDR/ADM/TECH TRANS) are included as one group. The CISST ERC supported 3 research thrust areas: Infrastructure (Thrust 0), Surgical Assistants (Thrust 1) and Surgical CAD/CAM (Thrust 2). Funding to diversity programs is included under the Education category.

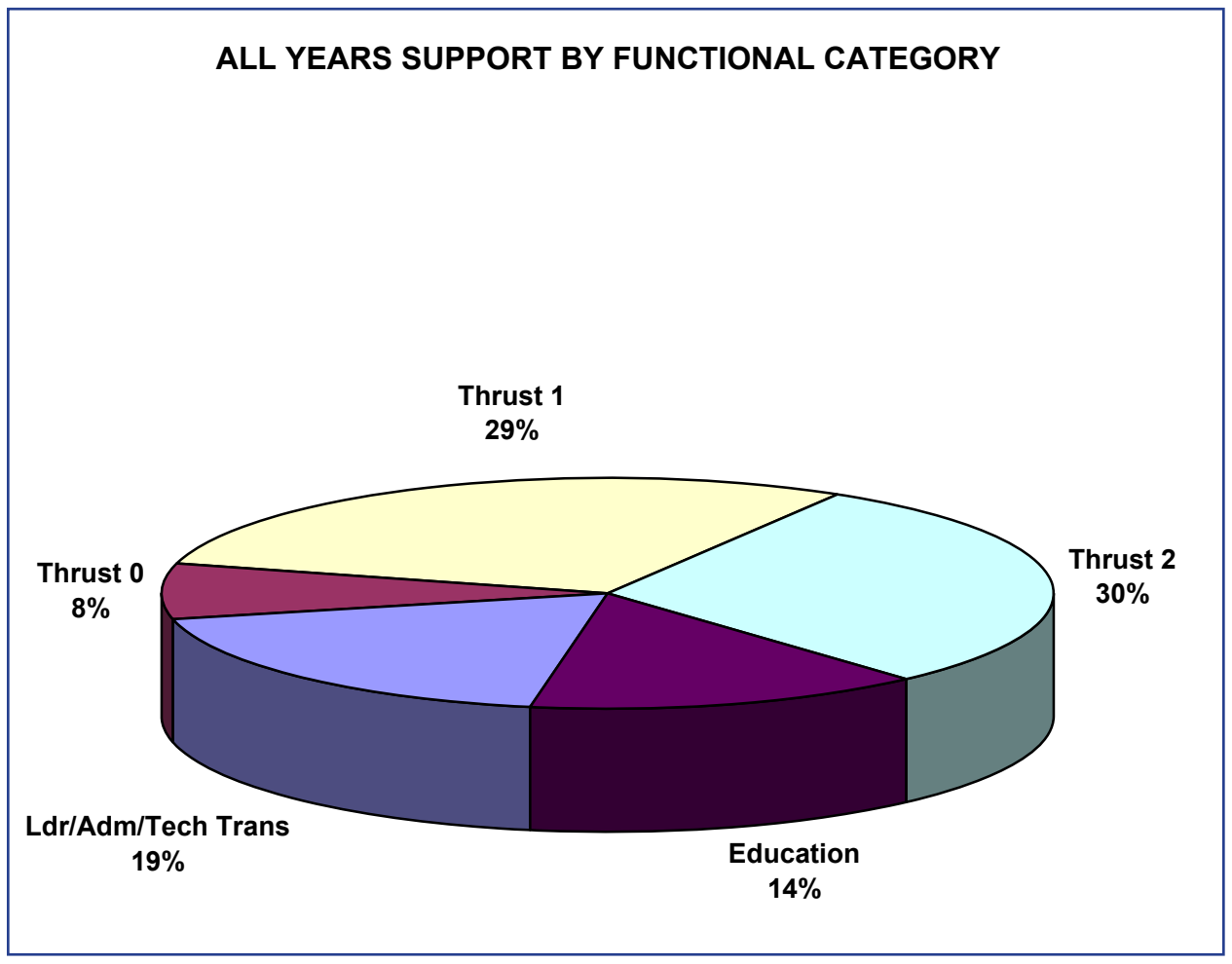

Fig. 73: Total support to each of the CISST ERC functional groups over the life of the Center. Categories are the same as described in Fig. 72, above. Leadership, Administration and Technology Transfer (LDR/ADM/TECH TRANS) are included as one group.

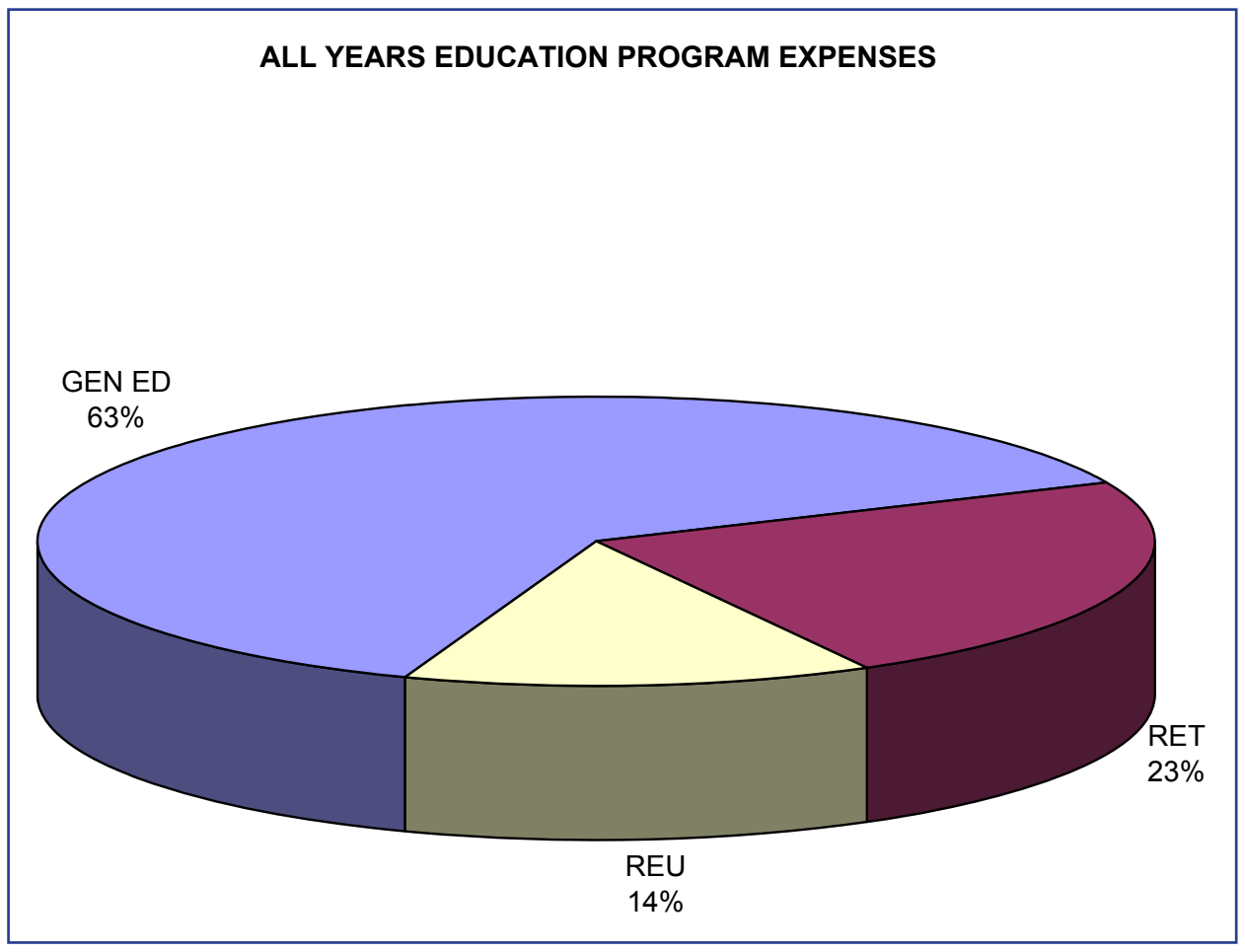

Fig. 74: A further breakdown of total funding to the CISST ERC Education program. REU is the Research Experience for Undergraduates program and RET is the Research Experience for Teachers program. General Education funding includes all other funding directed towards ERC education programs.

| ERC FUNDING HISTORY<br>Cooperative Agreement # EEC-9731748 | | | | |
|---|---|---|---|---|
| Amendment No. | Award Type | | Purpose | Award |
| Base Award | EEC9731748 | Base Award | Provide core funding for Center | 1,999,682 |
| #1 | EEC9731748 | Year 2 funding | Continuation of Center | 2,000,000 |
| #2 | EEC0040848 | Supplemental Funding | Continuation of Center | 200,000 |
| #3 | EEC0043218 | Year 3 funding | Continuation of Center | 2,694,000 |
| #4 | EEC0120146 | Supplement | First of 3-yr REU award | 55,700 |
| #5 | EEC0108901 | Supplement | RET | 150,000 |
| #6 | EEC0121940 | Supplemental Funding | Continuation of Center | 100,000 |
| #7 | EEC0121977 | Supplemental Funding | PER/Initiative | 63,498 |
| #8 | EEC0121937 | Supplemental Funding | PER for Dr. Atlar | 309,629 |
| #9a | EEC0133135 | Year 4 funding | Continuation of Center | 2,694,000 |
| #9b | EEC0133135 | Supplemental Funding | PER for Dr. Abts REUTI | 538,000 |
| #10 | EEC0209269 | Supplemental Funding | RET-Howard Com. College | 60,002 |
| #11 | EEC0209885 | Supplemental Funding | For HCC, MIT, CMU, BWH | 250,000 |
| #12 | EEC0225889 | Supplemental Funding | RET | 246,287 |
| #13a | EEC0231861 | Supplemental Funding | Second of 3-yr REU (#4) | 65,510 |
| #14a | EEC0232038 | Supplement | Continuation of Center | 485,833 |
| #14b | EEC0232038 | Supplemental Funding | RET-Howard Com. College | 29,999 |
| #14c | EEC0232038 | Year 5 funding | Continuation of Center | 3,586,357 |
| #15 | EEC0243376 | Supplemental Funding | RET | 50,195 |
| #16 | EEC0335920 | Supplemental Funding | Third of 3-yr REU (#4/13a) | 55,700 |
| #17 | EEC0343608 | Year 6 Renewal | Continuation of Center | 3,790,000 |
| #18 | EEC9731748 | Amend language | Cooperative Agreement | |
| #19 | EEC9731748 | Amend language | Cooperative Agreement | |
| #20 | EEC9731748 | Amend language | Cooperative Agreement | |
| #21 | EEC0413935 | Supplemental Funding | First of 3-yr REU award | 56,500 |
| #22 | EEC9731748 | Supplemental Funding | REU LSAMP | 147,300 |
| #23 | EEC0450508 | Supplemental Funding | Support for Dr. Heard | 57,225 |
| #24 | EEC0445537 | Year 7 funding | Continuation of Center | 3,980,000 |
| #25 | EEC9731748 | Amend language | Cooperative Agreement | |
| #26 | EEC0531014 | Supplemental Funding | Second of 3-yr REU (#21) | 56,500 |
| #27 | EEC9731748 | Amend language | Cooperative Agreement | |
| #28 | EEC0550059 | Year 8 funding | Continuation of Center | 3,990,000 |
| #29 | EEC0553048 | Supplement | Continuation of Center | 14,000 |
| #30 | EEC0631945 | Supplement | Third of 3-yr REU (#21, #26) | 55,935 |
| #31 | EEC0648952 | Year 9 funding | Continuation of Center | 2,646,567 |
| #32 | EEC0646678 | Supplemental Funding | Surgical assist workstation | 691,454 |
| #33 | EEC0637024 | Supplemental Funding | International Res. & Ed. In Engineering | 69,449 |
| #34 | EEC0745774 | YR 10 funding | Continuation of Center | 1,733,329 |
| #35 | EEC0751230 | Supplemental Funding | Intelligent Management | 147,271 |
| #36 | | | no cost extension | 0 |
| #37 | | | no cost extension Surgical Assistant Workstation | 0 |
| | | | **Grand Total Cooperative Agreement** | **$ 33,069,922** |

Fig. 75: The history of CISST ERC funding under Cooperative Agreement # EEC 9731748. In total the CISST ERC received $33,069,922 in annual and supplemental funds from the NSF.

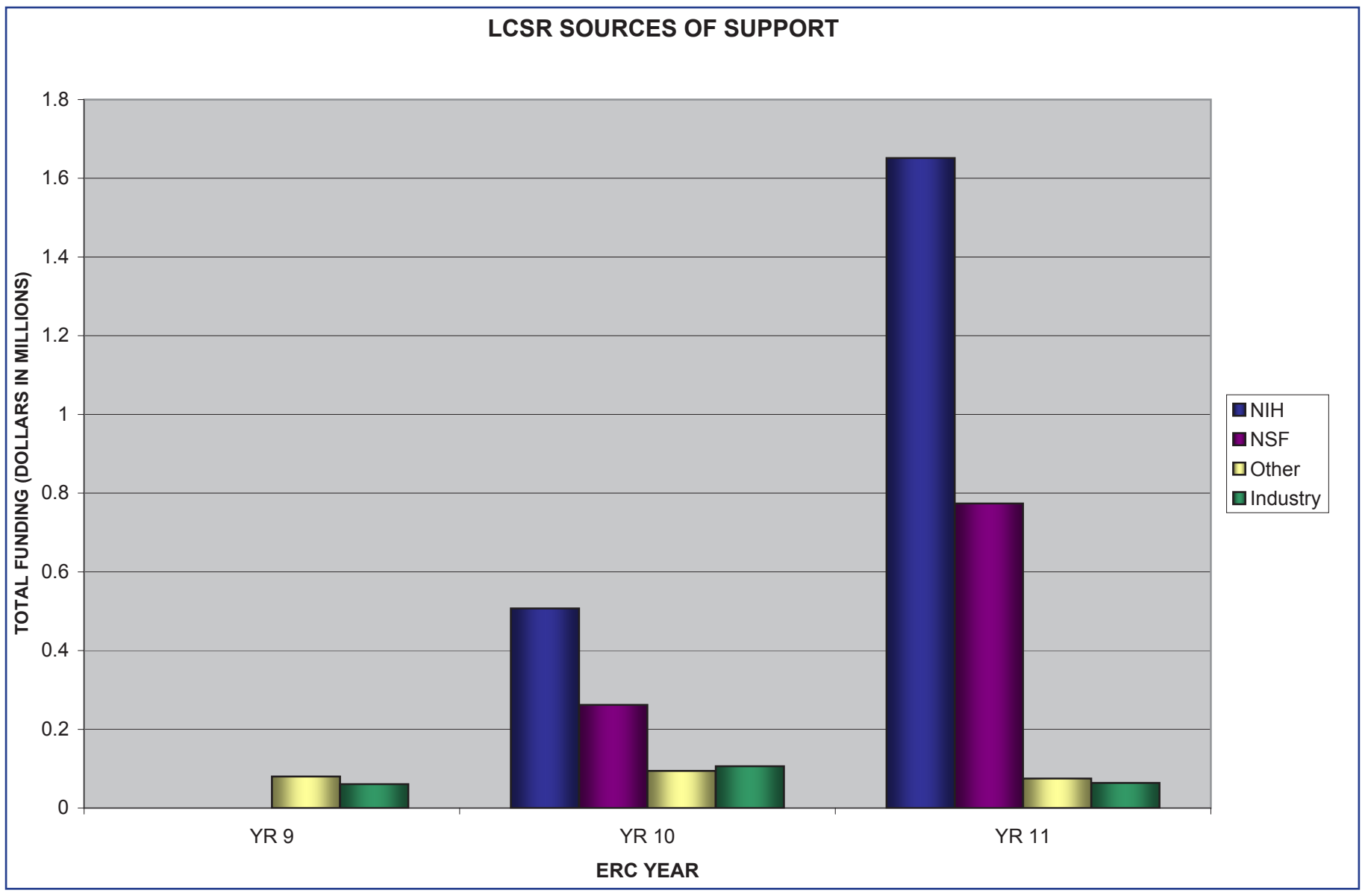

Fig. 76: Detail of sources of sponsored funding for the Laboratory for Computational Sensing and Robotics. The National Institutes of Health provide the bulk of funding at this time. Only LCSR funding that fits the ERC mission of Computer-Integrated Surgical Systems and Technology are included in this report.

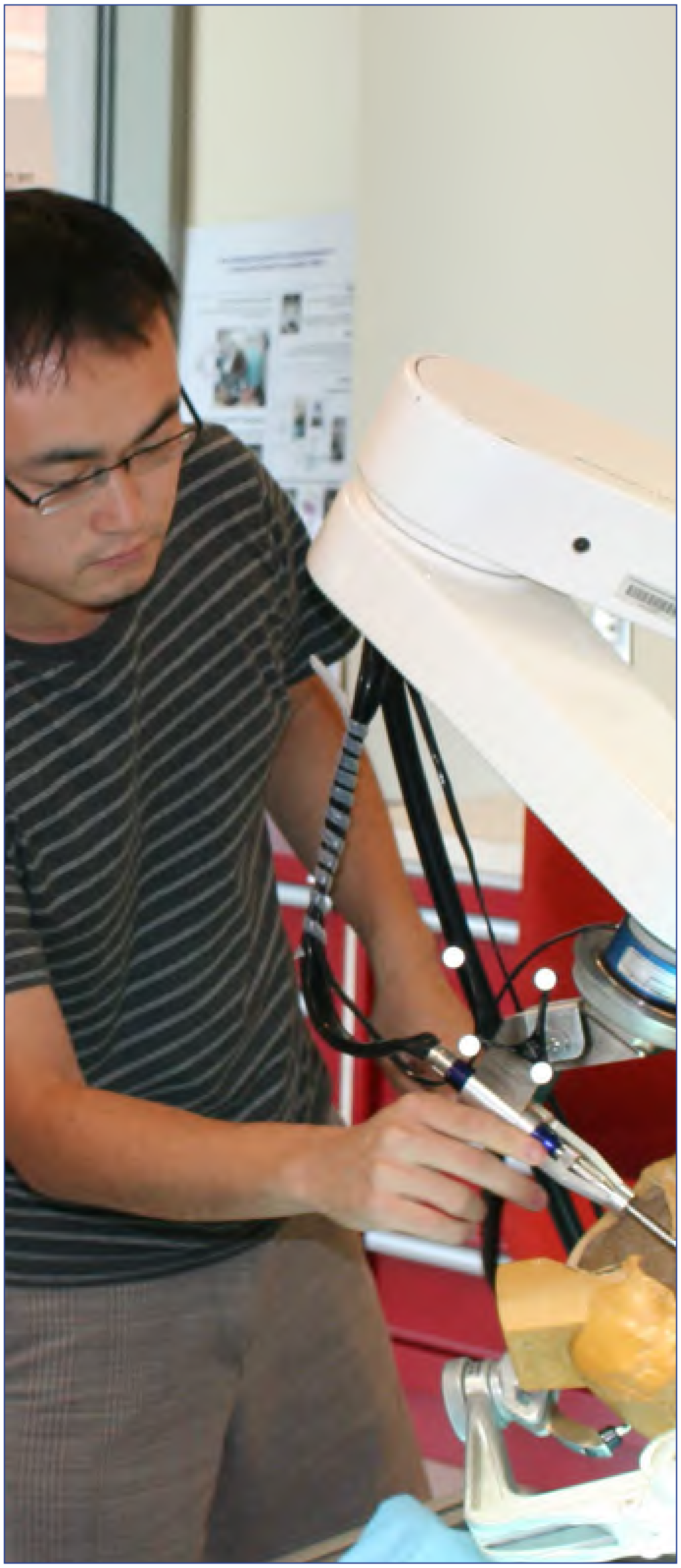

# Beyond Graduation

Since late 2003 we have been developing a strategy for maintaining our Center after NSF core funding has expired and for undertaking the transition from the current configuration of our organization and research program to one that will sustain us in the future. The strategy retains our vision of systems-level development of innovative surgical technology. It also seeks to maintain our engineering infrastructure, clinical testbeds, educational outreach, industrial liaison, and management support functions as a platform for a variety of targeted research initiatives.

***Goals of our graduation strategy***

Our strategic goal as we move forward through "graduation" as an NSF-funded ERC is to build a sustainable, effective organization and infrastructure that will enable us to achieve our CIS vision. In particular, we want to preserve and strengthen:

- Our technical excellence in basic research, technology, and systems focus leading to increased "transitional" research opportunities and deployed clinical systems.
- The web of relationships among faculty, clinicians, key industry partners, and collaborating institutions that have promoted our vision and will serve as the basis of our future evolution.
- The engineering and systems infrastructure that supports this work, especially the experience accrued by our enginnering staff.
- Our educational and diversity initiatives, which will be increasingly important in following JHU's educational mission and in gaining a "competitive edge" in pursuing major NSF, NIH, and Foundation funding opportunities.
- The unique administrative expertise required to manage multi-disciplinary and multi-institutional programs in CIS.

*Challenges*

• Developing sufficient "bridge" funding to sustain all elements of our program while additional "large" grant funding is secured from NIH and elsewhere and while transitional research further matures.

• Significantly strengthening the scope of our Industry program without fragmenting or burning out our faculty.

• Nurturing and sustaining multi-institutional education and outreach initiatives during the transition period.

*Opportunities and points of leverage*

Our graduation strategy seeks to address these challenges while embracing significant opportunities in our environment. Strong points that we will exploit include the following:

• The CISST ERC and its associated faculty have acquired worldwide reputation for research leadership in computer-integrated interventional medicine, especially in the area of medical robotics.

• We have learned how to work successfully with clinician "end users".

• We have established close working ties among our participating institutions and have demonstrated an ability to successfully compete for large multi-institutional grants such as NIH Bioengineering Research Partnerships (BRP).

• Many of the "Roadmap" initiatives at NIH, as well as current DARPA and DOD initiatives are aligned with the strategic direction and expertise of our center, and we have been developing these program sources.

• The importance of reducing surgical errors and enabling less invasive therapy is widely recognized. The ERC's technology has a natural role here and is synergistic with under-explored initiatives in medical informatics.

***Balancing Academics and Community Service***

As a pre-med undergraduate in the Hopkins biomedical engineering program, Carmen Kut founded the Johns Hopkins Student Research Group to find research opportunities between faculty and students. Through that, she learned of an opening for the position of lab supervisor with the ERC, in which position she became acquainted with its director, Russ Taylor.

Taylor connected her with Johns Hopkins Medicine's radiation oncology department, where she did research involving the tracking of patient movement during radiation therapy. Kut also did research with Taylor into tumor detection – comparing manual palpation with ultrasound elastography as ways to determine tumor location and size. She has already received a number of academic honors for her work. Such academic achievements, along with her strong commitment to community service (she founded Educational Perspectives, an international student group providing preventive health care and education worldwide), led to her being one of 20 students named to USA Today's All-USA College Academic First Team.

She plans to practice clinical medicine and do research in the future. "The ERC has been extremely influential in my career decisions," she says. "Dr. Taylor has been an amazing supervisor and mentor."

• There is strong interest within JHU in integrating large-scale information processing methods and statistical process control techniques with image-based interventional systems. This creates an opportunity to expand the ERC's mission.

• The new CSEB at JHU provides high visibility laboratory and office space, the Swirnow Mock Operating Room, and other infrastructure. Housing the LCSR, it encourages collaboration among 12 robotics faculty members, including the ERC's core leadership team, and promotes communication between CISST and other JHU Centers and Laboratories in CSEB, such as the Institute for Computational Medicine.

## *Strategy Summary*

The key to meeting these challenges and seizing opportunities is building upon and expanding our current ties with JHMI and other clinical collaborators. Over time, JHMI should become the lead partner in a multi-divisional, multi-institutional partnership, a natural progression as the technology of the ERC matures and transitional/clinical research assumes a more prominent role. It will also benefit our ability to attract industry and donor support for major initiatives. Major elements of our strategy include the following:

***Evolution of the ERC's core organization at JHU***

Upon graduation, the CISST ERC will become a Center within the LCSR. For administrative purposes, the ERC staff will transfer to LCSR, which will assume responsibility for facilities and financial administration, including ERC grants and contracts. We expect that the ERC will retain its own identity and logo within LCSR and that it will also retain significant resources that can be used to invest in small startup projects and as bridge funding to sustain selected activities.

***Sustaining our engineering and systems infrastructure***

Our goal is to sustain and enlarge our ability to design, develop, and validate systems for research and clinical application. We place special emphasis on our engineering and technical staff, essential both for research and as mentors for our students. Resources will come from several sources, including:

• Line items for specific tasks within research grants and contracts

• Grants and contracts specifically tied to infrastructure

• Participation in major JHU initiatives with a significant transitional research component

• Philanthropy

• Returns on research from WSE to LCSR.

This strategy is succeeding. Examples include:

• Currently funded non-infrastructure grants contain support for 4 staff engineers

• A currently funded NSF Major Research Instrumentation Development grant provides support for 2 engineers, as well as $839,000 for equipment

• Recent projects with Fraunhofer Gesellschaft have significant engineering elements and are the first step toward a significant strategic partnership focusing on transitional research

Beyond Graduation

• A generous gift from Richard Swirnow provides funds for developing infrastructure within our new Mock OR

• We have sufficient reserves from returns-on-research to bridge any unexpected shortfall in engineer salaries. In fact, we are currently planning to hire additional engineering staff in the coming year.

*Funding of research through grants and contracts*

Largely because of the intellectual, physical, and human infrastructure developed through NSF ERC funding, we are excellently positioned to sustain our research program. As this is written in July 2008, we have over $17.6M in active "non-core" ERC-related extramural research grants and contracts. A significant portion of this funding is in the form of large multi-institutional grants such as NIH BRP grants. Currently, the ERC has two such BRPs, one on microsurgery involving JHU and CMU and another on image-guided prostate therapy involving BWH, JHU, Burdette Medical, and Queen's University. We are currently developing BRP proposals on image-guided neurosurgery, high dexterity minimally invasive surgery, and information-driven radiation oncology.

*Development of major initiatives within JHU and with collaborators*

Strong interest and support exists within JHU in a new multi-divisional focus, "Integrating Imaging and Information in Interventional Medicine" or "I4M", an initiative which grew out of the ERC-based collaboration between the WSE and the School of Medicine (SoM). I4M extends the technical focus

*Moving Innovations from the Bench to the Bedside*

Michael Chote, chief of the Handelsman Division of Surgical Oncology at Johns Hopkins Medicine, was one of the early ERC collaborators. A cancer surgeon, he is interested in using robotic systems for the delivery of image-guided cancer therapies, especially needle-based techniques.

"The problem with current technology for placing a needle into a target mass and destroying it is that it requires precise manipulation," he says. "It's encumbered by having to put it in freehand, with crude imaging." He has helped develop systems and devices to the point where they now have patents, collaborations with industry, and NIH funding and STTR grants. "It's been very fruitful," Choti continues. "We're now developing a robotic intraoperative ultrasonography device to work with Intuitive Surgical's robotic da Vinci Surgical System to help guide probes to treat tumors in the liver, pancreas and other organs."

As an offshoot of the ERC program, Choti hopes to build an image-guided cancer therapy center at the Johns Hopkins School of Medicine. As he puts it, "I want to get this out to help patients."

Noting how the ERC has made it possible to recruit and mentor outstanding talents who will go on to become leaders in the field, Choti also points out that a number of other surgeons and surgical fellows have had the opportunity to work on ERC-related projects.

He concludes, "There will be a durable legacy from ERC programs, with independent funding and lifelong collaborations."

***Ready for the Real World***

Though she came to Hopkins planning to major in physics, undergraduate Kathryn Hayes took some courses taught by Russ Taylor and decided she wanted to focus on computer-integrated surgery. Switching to computer science, she began working in the ERC with the open source image visualization software, Slicer. As a senior she went to MIT as an exchange student, working in what is now the Computer Science and Artificial Intelligence Laboratory, or CSAIL. This experience led eventually to her being hired at Brigham and Women's Hospital after completing her Master's project: developing a visual interface for a neurosurgical drilling robot with Assistant Research Professor Peter Kazanzides. Today, she is a senior software engineer in the hospital's Surgical Planning Laboratory.

Hayes says, "The ERC gave me the skills and knowledge to seamlessly transition from school into the workforce. I was up on the state of the art, and I knew many of the people in research and industry. What I learned was applicable in the real world." She also learned a lot as an active member – and president at one point – of the Computer-Integrated Surgery Student Research Society, which not only promotes research opportunities for students but also does pre-college outreach and mentoring.

# *Beyond Graduation*

to include a strong emphasis on methods to organize enormous quantities of heterogeneous data, interpret this data to extract useful information, and use the information to control and manage treatment processes. This enlarged scope has led to increased involvement from other JHU divisions, including the Krieger School of Arts and Science, the School of Hygiene and Public Health, and the JHU Applied Physics Lab. Specific organizational structures for promoting I4M are still being defined, with most of the activity being focused on informational symposia and coordination of large NIH proposals such as the radiation oncology BRP. These activities are already beginning to have an impact within JHU.

A second initiative is our emerging relationship with Fraunhofer Gesellschaft (FhG). JHU and FhG have established a formal partnership on technological innovations in interventional medicine, with special focus on I4M-related areas. Our institutions have started three pilot projects in the area of endoscopic surgery with $2M in combined FhG and JHU funding over an 18 month period. This activity will provide useful technology and preliminary results for joint grant proposals, and we expect it will lead to a much larger FhG/JHU effort to promote transitional research and technology transfer.

Finally, the potential exists to create a multi-institutional consortium in the Baltimore/Washington, D.C. area, similar to CIMIT in Boston. Such a consortium would provide a venue for regional collaborations and facilitate large-scale initiatives with CIMIT and similar organizations. We have begun preliminary conversations concerning such a consortium as part of our I4M activities.

*Sustaining our education and diversity initiatives*

The Center will continue to develop a diverse, superbly trained, highly educated, diverse workforce in the area of Computer Integrated Surgery. We will accomplish this by focusing on education and outreach from K – post doctoral students and developing training programs for educators and field practitioners.

For K-12, we will work with our teachers and other K-12 partners to obtain funding for existing programs. We will aid the work of Brian Bruneau, a former RET participant, by providing support for the Summer Robotics Camp. He will assist with the System Robotics Challenge, which is supported by CISSRS and industrial partners. We will again submit proposals to NSF and NASA to expand and fully support these efforts.

The Center plans to take the following steps to maintain programs for undergraduate (UG), graduate (G), post-docs (PD), and field practioners:

• Offer CIS research experiences for undergraduates. A site REU has been obtained for this purpose.

• Continue to target HBCUs, LSAMP, and McNair to diversify the student body.

• Enroll students in the CIS minors and concentrations and begin enrollment in the forthcoming Robotics Minor.

• Offer special courses like SFE to train students, surgeons, and affiliates in CIS. Our first bi-annual Winter School, funded by the NSF and the IEEE,

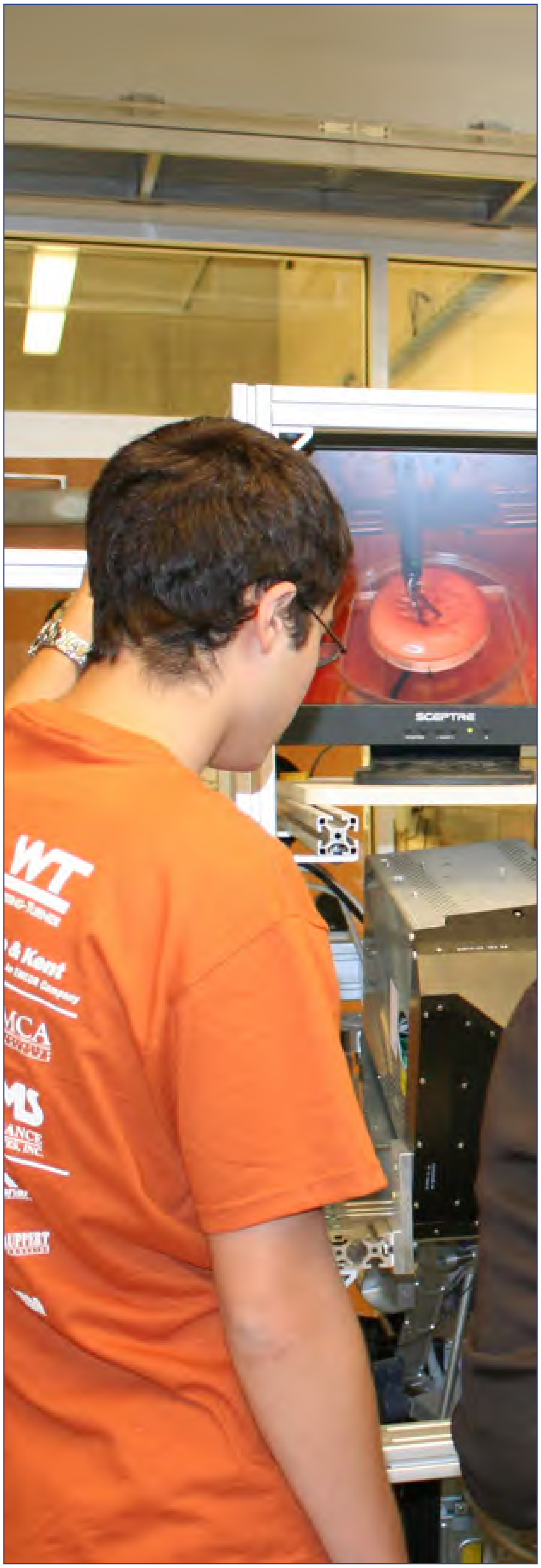

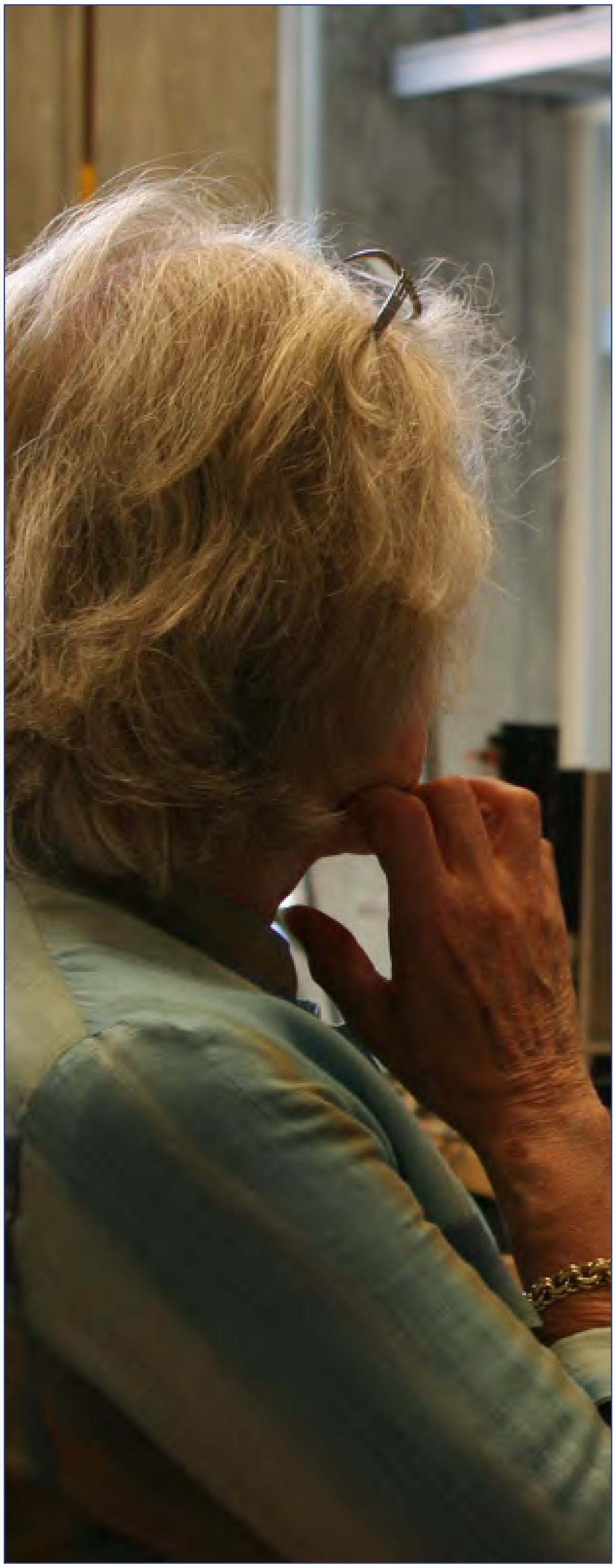

# *Beyond Graduation*

included SFE as a component, and an SFE version for affiliates is in the works.

• Develop Masters concentrations programs in CIS, including part-time, evening, and weekend programs for field practitioners.

• Organize CIS scholarships/fellowships by applying for training grants. We currently have a GK-12 award, and we have submitted a T32 Training Grant proposal to NIH.

The Center plans to leverage the above activities to generate a self-sustaining revenue stream. Grant funding will support the UG, G, and PD students during their studies. Specialty courses and part-time program degrees will generate income to support some of the ERC's education and outreach programs. The Center will begin to charge a lab fee for courses such as SFE (unless fully funded by educational grants). We hope to entice clinicians and industry affiliates to take SFE to learn and train on cutting-edge ERC systems and technologies for immediate use in their practices. The Center will offer space, equipment and technical expertise for its industrial affiliates to train clinicians and researchers on their technology, a common practice among other JHMI centers. The latter two items, if appropriately priced and marketed, can be a significant revenue generator for the Graduated ERC. The Center will also explore developing a spectrum of CIS educational tools, books, and DVDs that can be sold to support other education and outreach activities.

The Center's substantial program aimed at diversifying the participants in CIS at JHU will continue post graduation. We are well represented on the JHU/WSE Diversity Committee. A proposal to the Dean is in development to fund our efforts in Recruitment and Retention of a well represented staff, student, and faculty workforce; in developing a community to promote scholarly and social interactions among our constituents and in promoting and celebrating their achievements.

***Sustaining our industry initiatives***

We have found that the most valuable aspect of our Industry Affiliates program has been its role in facilitating active collaboration with affiliates on specific projects of mutual interest, often with extramural government funding. As we move forward, our goal is to extend and intensify these collaborations. Accordingly, our current plan is to retain the ERC Industry Affiliates program, with dues going to support the administrative and communication costs of the program.

***Looking Ahead***

From the beginning, we regarded the NSF support as seed money. Now, the seeds have been planted, results are vindicating our research strategy, and a whole network of ERC alumni is in place at universities, medical institutions, corporations and government agencies. We are confident that we will be able to sustain and amplify our focus on clinical applications of novel, high-performance CIS systems.

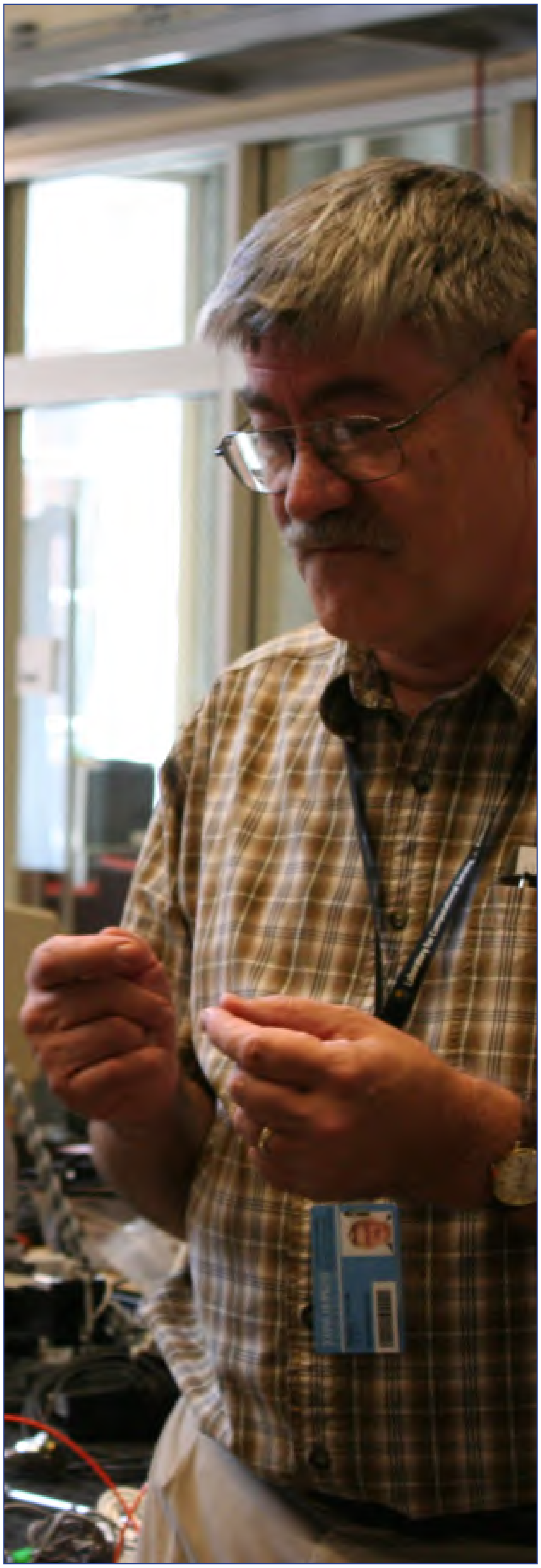

# *Appendix*

# *Lessons Learned*

*ERC Vision*

• A clear vision of the core goals and intellectual structure of the ERC is absolutely essential for success. It is crucial to understand what the ERC is trying to accomplish, why its goals might be important, and how the various research elements fit together. The vision can evolve somewhat over the years, but it is what must ultimately provide intellectual coherence to the entire enterprise.

In the case of the CISST ERC, our core vision has remained remarkably stable over the years, even if our strategy for pursuing it has evolved. It has been the "glue" that has bound together a very diverse group of engineering researchers and clinical practitioners, as well as industry and institutional collaborators.

*Structuring and Running the Research Program*

• There are many ways to structure the thrusts of an ERC. For us, the most effective means was to create two major thrust areas, Surgical CAD/CAM and Surgical Assistants. These correspond to the two aspects of our core vision, focusing on information-enhanced closed-loop interventional medicine. A third thrust, which addressed the ERC's Infrastructure, provided the essential networking capabilities to share our results with the other thrusts.

• Research plans need to strongly tie to a process of developing programs that become independent. Some key questions to ask about any project are: 1) What do you want to do? 2) Why is this important and to whom is it important? 3) How will you know you have succeeded? 4) What follow-up steps will be needed to exploit success? 5) Where will the resources for these follow-up steps come from? In our experience, focusing on these questions, with particular emphasis on questions four and five, has tended to produce more effective research, with stronger teams and greater overall impact.

• Similarly, we have found that successful projects can result from either a "clinical pull" or an "engineering push". A key element for either approach is a physician champion who understands the clinical needs and can help to steer the technology in the right direction.

• NSF ERCs provide resources that are not available in traditional NSF grants and programs. There is an obvious difference in the size of the total award, though this may be less important for individual faculty members since the money must be stretched over many projects. Other aspects are more important.

o ERC funding provides a mechanism for developing and supporting significant engineering infrastructure, especially in the form of a professional engineering staff, and in development of long term testbed systems. We have found that the faculty, staff, and equipment resources available in our Infrastructure thrust were essential in developing large interdisciplinary research projects. Engineering research is much easier if it can be built upon robust modular hardware and software components that embody earlier research results.

o The infrastructure and environment provided by ERCs can help junior faculty jump-start their careers. In addition to the physical and staff infrastructure mentioned above,

# Lessons Learned

this support extends to the interdisciplinary intellectual environment, students, and external exposure. This is an important point to make in recruiting. Since the main output of any university-based program is ultimately people, it is also an important point for any ERC to keep constantly in mind. One should ask how ERC investments will help make our junior faculty and research scientists more productive.

• The research program needs to maintain flexibility. Often the biggest gains are not the planned research, but something new that comes up. We found that the relative long-term stability and flexibility of ERC funding is a terrific asset in enabling faculty to pursue high-risk or opportunistic research directions consistent with our overall vision and strategy.

• Collaborative partners need to be closely managed. In particular, the program should again focus on pushing collaborations to develop their own funding, which tends to tie the group together.

• Measuring success is crucial for any research program, and ERCs need to measure system-level performance as well as that of individual technology. Developing methods for doing this should be part of a research program. Within our ERC, we put considerable effort into developing objective measures of validating system performance with actual end users. For us, this activity is essential both for optimizing designs and for gaining eventual clinical acceptance.

*Center management, leadership, and culture*

• Building a strong administrative organization from the very beginning is very important. Catching up from a slow start is very difficult, especially given the heavy requirements for reporting, annual site visits, and other administrative tasks associated with NSF Centers.

• Multi-cultural teams take a long time to develop, but they have high payoff and become self-sustaining if properly managed. Again, it is important to have a clear vision of what you are trying to achieve, otherwise you can easily spread yourself too thin.

• In creating a multi-cultural environment, it is crucial to get everyone to invest the time to understand the problems and mindset of other participants. For example, we found that it was really crucial for the engineering staff, faculty, and students to develop a thorough understanding of medical requirements and constraints through OR visits, frequent discussions with clinicians, and courses such as Surgery for Engineers.

• Great attention should be paid to the influence of physical facilities on the ERC culture. A shared space that allows faculty, students, and staff from different disciplines to be near each other is an effective way to encourage and enhance collaborations. From the very beginning, we emphasized shared lab space wherever possible. The positive results from this experience were very important in leading JHU to dedicate a large portion of the CSEB on the Homewood campus to LCSR, which houses a substantial portion of the ERC. In the year since the CSEB opened, we have already begun to see numerous benefits to what was already a very strong collaborative environment. Similarly, we have found that maintaining laboratory space on both our medical and engineering campuses, with substantial movement of people between them, has been crucial for fostering the engineer-clinician collaboration that is a key part of our strategy.

*Industry and Practitioners*

• Industrial strategy needs to identify clearly areas where the ERC can make a difference and work closely with companies who can contribute in those areas. Again, this comes back to the relationship of the Industry strategy to the overall vision of the ERC.

• Structuring an industry program to match the developmental abilities of our industrial partners is an important aspect of the ERC. Trying to build a very large program in an emerging field such as medical robotics can be counter-productive. In our case, we found that the most effective strategy was to focus our industry program on developing specific collaborations between companies and ERC faculty and staff. In many of these cases, we were able to develop SBIR or STTR funding in order to develop the technology.

• It is important to realize that the Industry offers much more than just funding. Often, expertise and access to technology is far more important. This is especially true in an area such as medical robotics or systems since there are many constraints associated with sterility, safety, regulatory approvals, etc.

• We also found that working directly with end users (i.e., clinicians) provided much of the problem focus and "market" feedback that industry may provide directly for ERCs working in more established sectors. We discovered that the most effective path to an industrial company was through a clinician who was already a customer or collaborator with that company. This is quite possibly the case for many ERCs in emerging biomedical or biotechnology areas.

*Education*

• To identify and understand the problems in computer integrated surgery, curriculum must offer cross-disciplinary training at all levels. Participants in traditionally distinct disciplines must be exposed to ideas and techniques outside their domain. The ERC achieved this by creating unprecedented courses, such as Surgery for Engineers and Engineering for Surgeons.

• The learning experience is not limited to the academic material presented in the classroom. The majority of learning is achieved from projects conducted both as a part of courses and as research. Such project oriented learning is significantly enhanced when students work with experienced engineering staff members. When said staff member has industrial experience, the impact is even higher. This was achieved in the ERC because of its physical and intellectual infrastructure: engineering staff members are located with students and faculty in research labs and co-supervise research projects.

• Cross-institutional interaction in education is hard to achieve when the partner institutions are geographically distant from each other. Furthermore, a mismatch in academic requirements, calendars, and program timing make student exchanges difficult, even when there are joint projects between faculty advisors. Hence, the infrastructure for teleconferencing, web-based courses and cross-institutional seminars can be used to bridge the distance. Also, offering specialty unique, cross-field courses in the summer is useful in allowing students to matriculate at partner institutions. This is, of course, predicated on the existence of an effective Articulation Agreement between the institutions to facilitate these exchanges.

*Diversity and outreach*

• Having a balanced and diverse community within the center is important to maximize the research, education and outreach impact. This community is achieved by having in place a substantial strategic plan with the following goals: 1) recruit and retain diverse community members, 2) coalesce the community by providing mentoring, academic/social events and information/announcements, and 3) celebrate achievements. With a representative Education, Outreach and Diversity Committee and financial resources to execute the strategic plan, effective programs can be established that can have landscape altering implications well beyond the ERC.

• Timing is very critical in developing the strategic plan for diversity. The sooner the plan can be developed and the infrastructure to implement the plan established, the more effective the plan will be. The effectiveness of the ERC's plan peaked three to four years after its enactment, when the core funding was beginning to roll off.

• The involvement of local school teachers in our outreach programs has been effective in promoting science, technology, engineering and math education through robotics and computer integrated surgery. The ERC's early RET programs introduced JHU faculty to the Baltimore area K – 12 teachers, which may not have happened otherwise. Similarly, our partner institutions have also benefited from our RET programs. The teachers have been involved in various programs, such as our Summer Robotics Camp, Systems Robotics Challenge, and BIGSTEP GK – 12 that we have offered over the last 8 years

*In the case of the CISST ERC, our core vision has remained remarkably stable over the years, even if our strategy for pursuing it has evolved.*

# *Post-Doctoral Mentoring Plan*

*We attract both domestic and international Post-Doctoral Fellows. Once here, they are quickly assimilated into the community, given a voice in ERC-wide issues and provided opportunities for professional development.*

There are several postdoctoral scholars who have been supported by the CISST ERC turning its tenure. In general, postdocs are associated with one or more faculty mentors who guide the career development of the postdoc. The specifics of each interaction vary somewhat from mentor to mentor, but the following guidelines are generally adhered to:

***Weekly meetings with the PI***

The PI and the postdoc meet on a weekly basis for at least one hour to discuss both technical and professional development issues. Much of the work in the CISST ERC is team-based; postdocs usually also participate in team meetings and may also take a leadership role in these teams.

***Professional development seminars:***

Since the Fall 2006 semester, CISST ERC, in collaboration with the larger robotics group at JHU, has held Student Seminars designed to engage students and postdocs informally, in conversations with regard to their research topics and professional development. Example seminar topics are: Mentoring research students, Interdisciplinary collaboration, Intellectual property, Diversity and sexual harassment, Grant proposal writing, Obtaining a faculty or industry position, Research responsibility and scientific misconduct, Data management and retention, and Authorship and publication practices.

***Teaching opportunities***

JHU has a unique opportunity for graduate students and postdocs to gain independent

# Post-Doctoral Mentoring Plan

teaching experience in a short time span: the Intersession period. Intersession is a mini-semester that offers optional three-week courses for academic exploration, experiential learning, and personal enrichment. Several students have taught during intersession. Postdocs also often participate in normal lecture courses during the semester by giving guest lectures or helping with course development.

***Networking opportunities***

Postdocs are always invited to meet one-on-one with visitors to JHU, many of whom are doing work related to computer-integrated surgical systems. These meetings are useful not only for technical exchange, but also for the postdoc to find out about current or future job openings at the visitor's institution. Postdocs are also encouraged to attend conferences. Although most postdocs do not have a new conference paper completed in the first six months, the mentor often sponsors postdocs to attend at least one conference in the first six months in order to keep up with the field and meet researchers at other institutions.

*We have been very successful at placing out Post-Doctoral Fellows into professoriate at top universities, like the University of Pennsylvania and Washington University, and coveted industrial and national labs, like Seimens and JHU/APL.*

***Proposal writing***

Postdocs are often recruited to assist with the development of proposals related to the postdocs' interests. Indeed, one career path for postdocs at JHU is to develop their own funded program, and to remain as research scientists or research faculty on their own funding.

***Mentoring***

Postdocs in the CISST ERC invariably work with and mentor graduate students, undergraduate students, and REU students. This provides excellent training for a subsequent faculty position.

# *Nuggets*

*"With all the protégés of Russ Taylor, all the people he has led into the science and through the educational process.... his word is being disseminated across the country. It's a great achievement."*

*~ Claire Tempany*

# *Discovery*

# Needle Steering

*Allison Okamura, PI (JHU), Noah Cowan (JHU), Greg Chirikjian (JHU), Ken Goldberg (University of California at Berkeley), Gabor Fichtinger (JHU/Queen's University)*

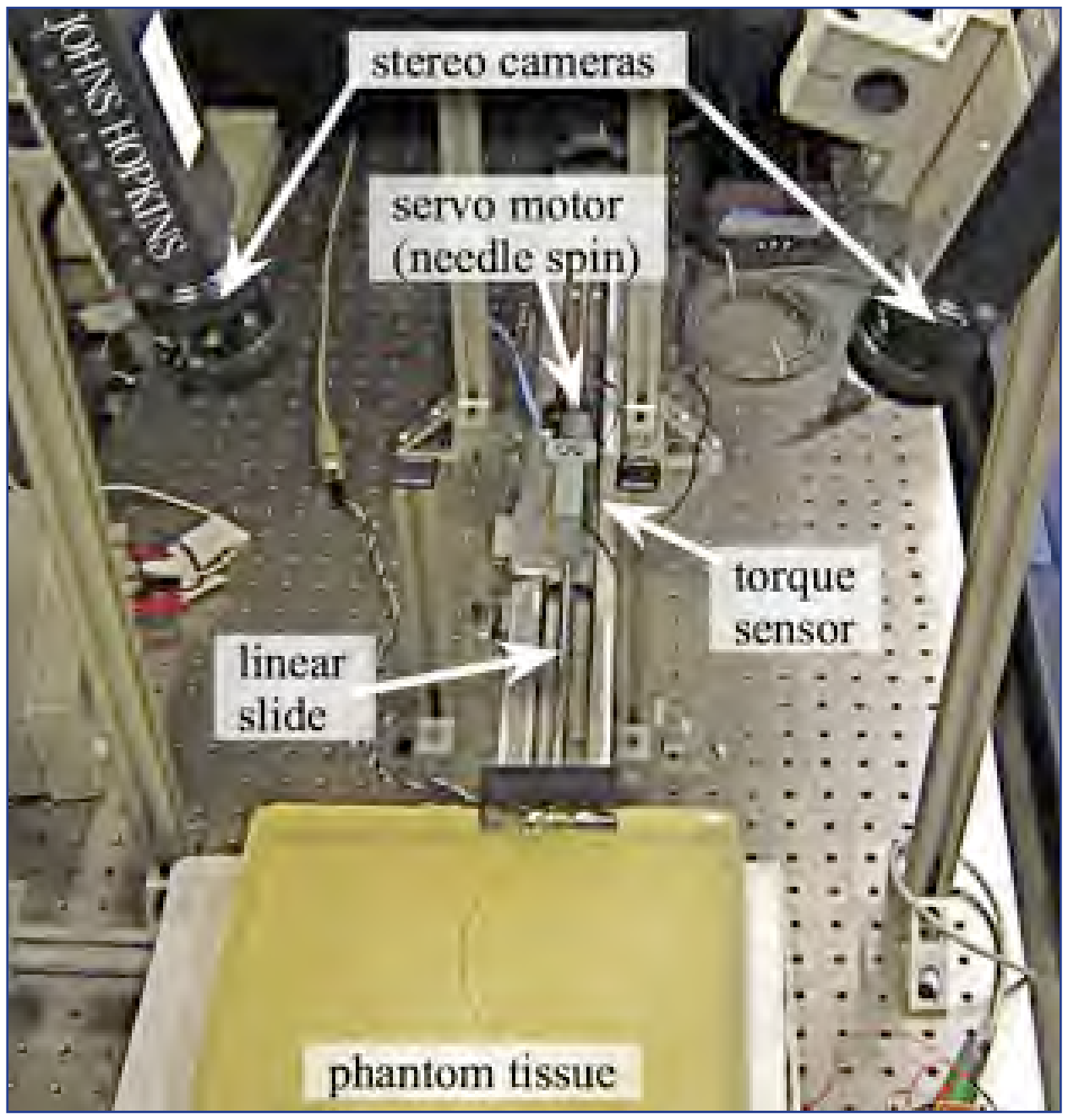

Fig. A1: A needle steering robot can insert needles autonomously or be teleoperated by a surgeon or interventional radiologist.

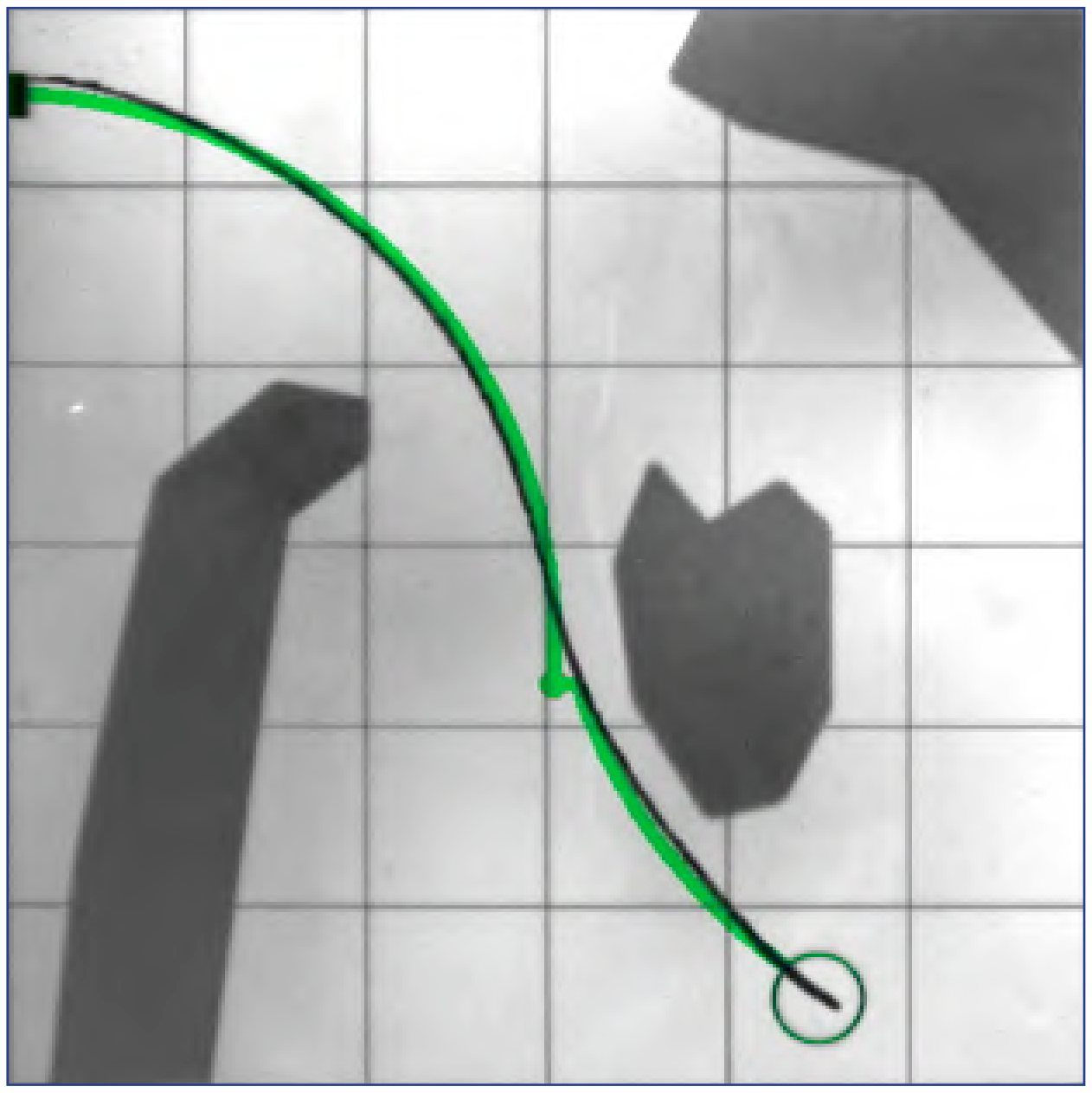

Fig. A2: A needle (black line) closely follows a planned trajectory (green line) around obstacles in transparent artificial tissue.

Minimally invasive surgical techniques are highly successful in improving patient care, reducing risk of infection, and decreasing recovery times. Needle steering robots aim to further reduce invasiveness by developing techniques to insert thin, flexible needles into the human body and steer them from the outside. New results in needle and tissue modeling, robot motion planning, and image-based control have enabled steering of flexible needles inside soft tissue. This can improve both targeting accuracy and the ability to steer around delicate areas or impenetrable anatomic structures.

Our method uses simple, flexible needles, with asymmetric bevel tips or pre-bent tips, which cause the needle to adjust when it is inserted into tissue. By pushing the needle forward from the outside and spinning it around its main axis, a robot can control the needle to acquire targets in a three-dimensional space, while avoiding obstacles, with minimal trauma to the tissue. We have developed a needle steering robot that inserts the needle either autonomously or via human input to a teleoperator (Figure A1). High-level motion planning algorithms are used to determine an optimal path into deformable tissue, one which acquires the target while avoiding pre-defined regions of workspace (Figure A2). Image-based controllers use computer vision tracking and feedback control to maintain the needle's motion along the desired path.

By enhancing the physicians' abilities to accurately maneuver inside the human body, needle steering could potentially improve a range of procedures from chemotherapy and radiotherapy to biopsy collection and tumor ablation, all without additional trauma to the patient. The results of this project could significantly improve public health by lowering treatment costs, infection rates, and patient recovery times. By increasing the dexterity and accuracy of minimally invasive procedures, anticipated results will not only improve outcomes of existing procedures, but also enable percutaneous procedures for many conditions that currently require open surgery. This is a collaborative project between Johns Hopkins University, University of California at Berkeley, and Queen's University.

*Acknowledgements: This is a collaborative project between JHU, University of California at Berkeley, and Queen's University.*

# ASSIST - Automated System for Surgical Instrument and Sponge Tracking

*R. Taylor, M.R. Marohn, J.A. Miragliotta, N. Rivera, R. Mountain, L. Assumpcao, A.A. Williams, A.B. Cooper, D.L. Lewis, R.C. Benson.*

The CISST ERC, the JHU Department of Surgery, the JHU School of Engineering, and the JHU Applied Physics Laboratory are collaborating to develop a system that automates the error-prone manual counting procedure used during surgery to ensure that no surgical item is left inside a patient. Our immediate goal is to develop a prototype that can serve as proof of concept.

A dangerous medical error that can occur during surgery is unintentionally leaving a surgical instrument or sponge inside a patient. Commonly known as a "retained foreign object," this error can lead to inflammation, obstruction, perforation, sepsis, and sometimes death. The problem is thought to be avoidable when stringent manual counting guidelines are followed by Operating Room (OR) personnel. While these guidelines are very effective in reducing the risk, the problem persists. Some estimates report that the incidence can be as high as one in 1500 surgeries.

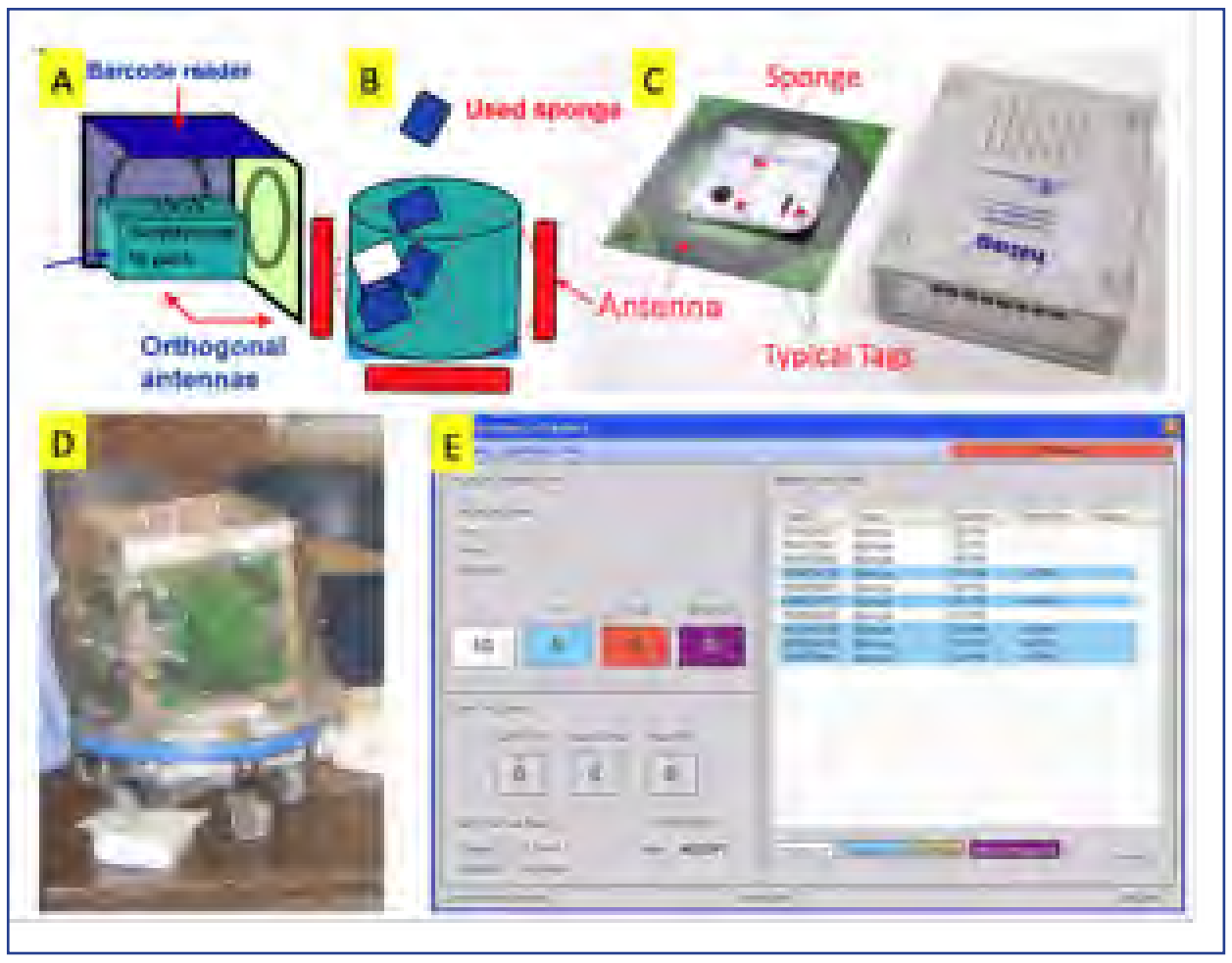

Fig. A3: ASSIST System -- A) check in station; B) bucket; C) RFID components and sponge; D) current prototype kick bucket; E) intraoperative display.

Human error is not the only drawback of manual counting. During sponge counting, nurses are unable to provide support for the surgeon as they focus on accurately counting sponges. Each sponge count lasts a couple of minutes, with at least three counts per surgical procedure, taking up a significant portion of a nurse's time during surgery. When a miscount is found there is a significant increase in the OR time since an x-ray of the patient is often required.

We are developing ASSIST, an automated system for surgical instrument and sponge tracking that increases the safety of surgical procedures. ASSIST uses Radio Frequency Identification (RFID) technology to detect and identify each surgical item at various stages during surgery. The use of low frequency RFID enables reliable detection of tags even when in the vicinity of metallic objects such as surgical tools, soaked in body fluids, or inside a patient's body. A check-in station verifies the content of the package and registers each tagged item in an inventory database, and a check-out station uses a smart bucket, where the used sponges are discarded, to account for used sponges (Fig. A3).

At the time of writing (June, 2008), we have a fully working prototype of ASSIST. Our initial investigation shows that the system can account for 100% of tagged sponges during surgery. This high level of reliability is attained by RFID verification at a check-in station, and detecting used sponges with multiple orthogonal antennas at a check-out kick-bucket. The measured check-in time for a ten-sponge packet is just two seconds, while regular check-out time is between one to five seconds. Preliminary tests also show that we can reliably detect missing sponges inside in vivo porcine model.

An abstract of the system was presented at a medical conference, a full paper was presented at an academic conference, and a utility patent was filed to the USPTO. Future endeavors aim to develop a system for detecting every type of surgical instrument that can be retained in a surgical procedure. We plan to continue to assess the technology and partner with the appropriate external entities that can quickly bring ASSIST to the OR.

*Acknowledgements: This work has been funded through the CISST ERC core grant and JHU Internal funds.*

# The Perk Station: Percutaneous Intervention Training Suite

*Gabor Fichtinger*
*John Carrino (JHU Radiology)*
*Paweena U-Thainual (Queen's)*
*Iulian Iordachita (JHU), Siddharth Vikal (Queen's)*

Image-guided percutaneous (through the skin) needle-based surgery has become part of routine clinical practice in performing procedures such as biopsies, injections, and therapeutic implants. Trainees usually perform needle interventions under the supervision of a senior physician, which is a slow training process that lacks an objective and quantitative assessment of the surgical skill and performance. To address these issues we are developing the Perk Station, an inexpensive, simple and easily reproducible surgical navigation workstation for laboratory practice with non-biohazardous specimens.

The Perk Station (Fig. A4) comprises image overlay, laser overlay, and standard tracked freehand navigation in a single suite. The image overlay consists of a flat display and a half-silvered mirror mounted on a gantry. When the physician looks at the patient through the mirror, the CT/MR image appears to be floating inside the body with the correct size and position, as if the physician had 2D 'X-ray vision'. The laser overlay uses two laser planes; one transverse plane and one oblique sagittal plane. The intersection of these two laser planes marks the needle insertion path. A stand-alone laptop computer is used for image transfer, surgical plan, and appropriate rendering. The image overlay is mounted on one side with the laser overlay and tracked navigation system on the opposite side, so the user can swap between guidance techniques by turning the system around. The surgical planning and control interface is based on the 3D Slicer, open source medical image computing and visualization software.

To promote transferability, the complete design of the Perk Station, including hardware blueprints, phantom blueprints, and software source code, will be made publicly available as open source. Simple design and low costs allow interested parties to replicate the hardware and install the software. CT/MRI data and pre-made surgical plans will also be provided, so users can operate the Perk Station without having access to medical imaging facilities.

The Perk Station is designed to be a replicable and adaptable tool for teaching computer-assisted surgery at all levels, from high-school science classes to clinical residency. Small, portable, and light weight, the Perk Station will fit inside a suitcase when disassembled. It promises to serve the education and outreach mission of the CISST ERC.

The Perk Station is fully designed and awaits manufacturing. The physical embodiment will be presented at the SMIT 2008 conference. The system will debut in undergraduate teaching in fall 2008 at Queen's University.

*Acknowledgements: This work has been funded by Queen's University Teaching and Learning Enhancement Grant, U.S. National Institutes of Health 1R01CA118371-01A2, and the National Alliance for Medical Image Computing (NAMIC), funded by the National Institutes of Health through the NIH Roadmap for Medical Research, U54 EB005149.*

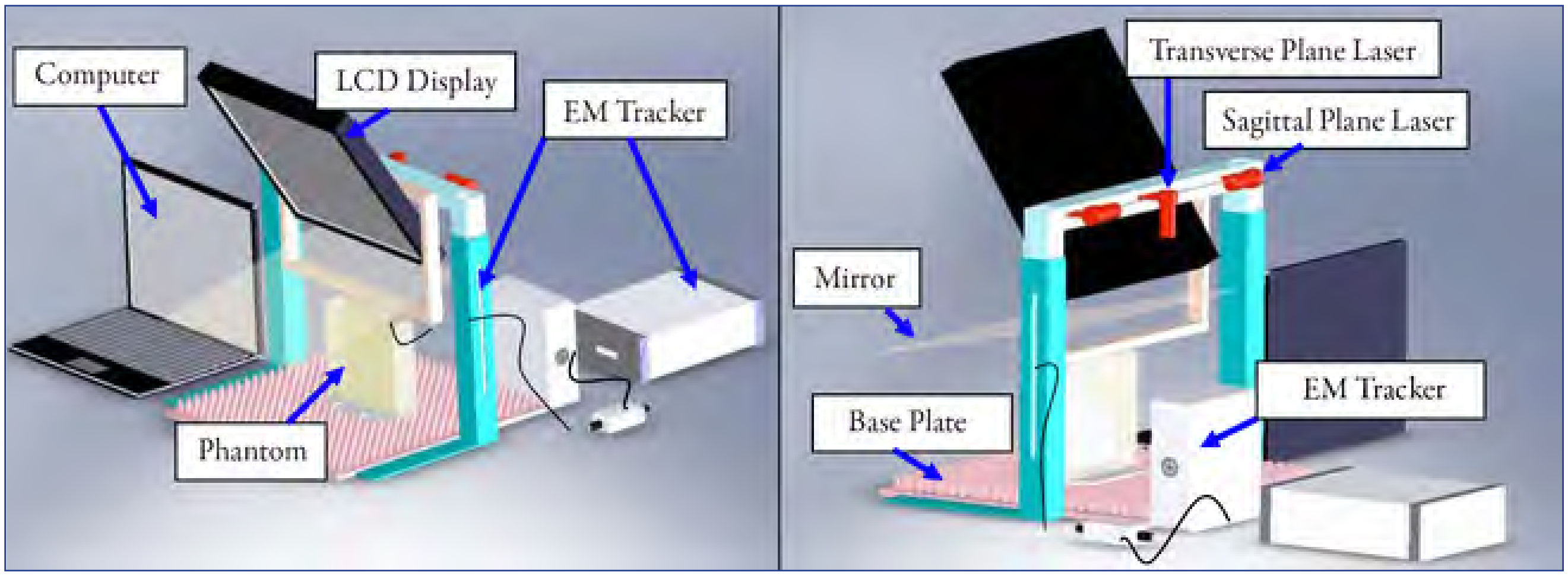

Fig. A4: CAD design of the Perk Station, w/ image overlay (left) and laser overlay & tracked navigation (right)

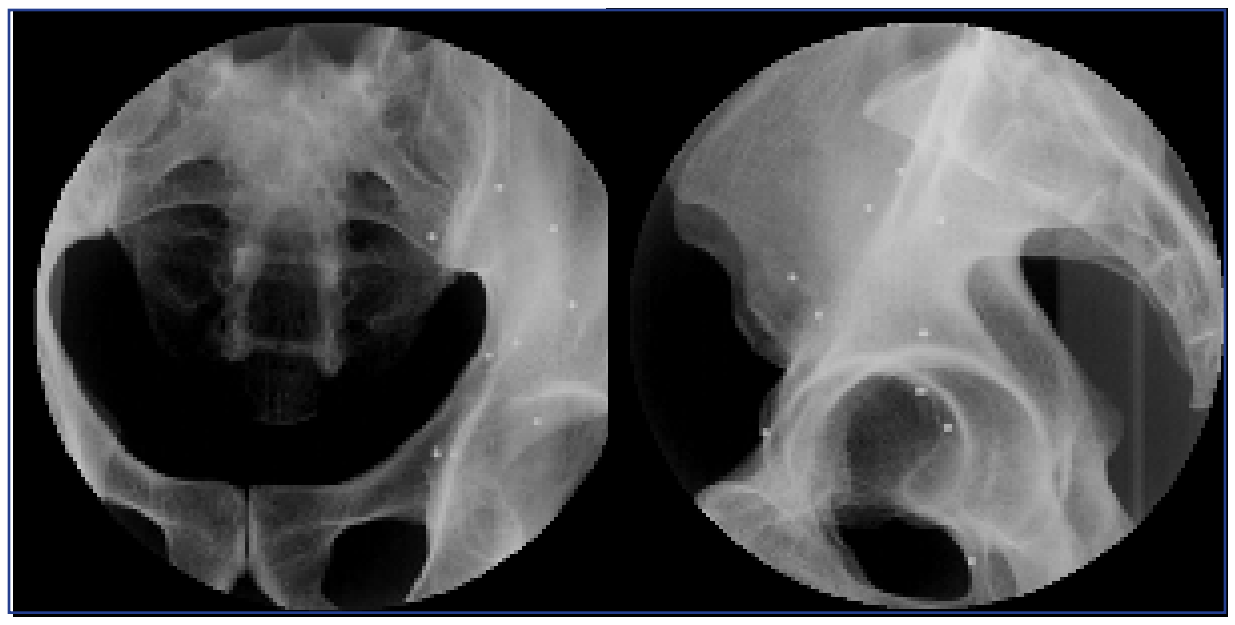

Fig. A5: X-Ray images of a dry pelvis bone taken with a mobile C-arm. Note image truncation.

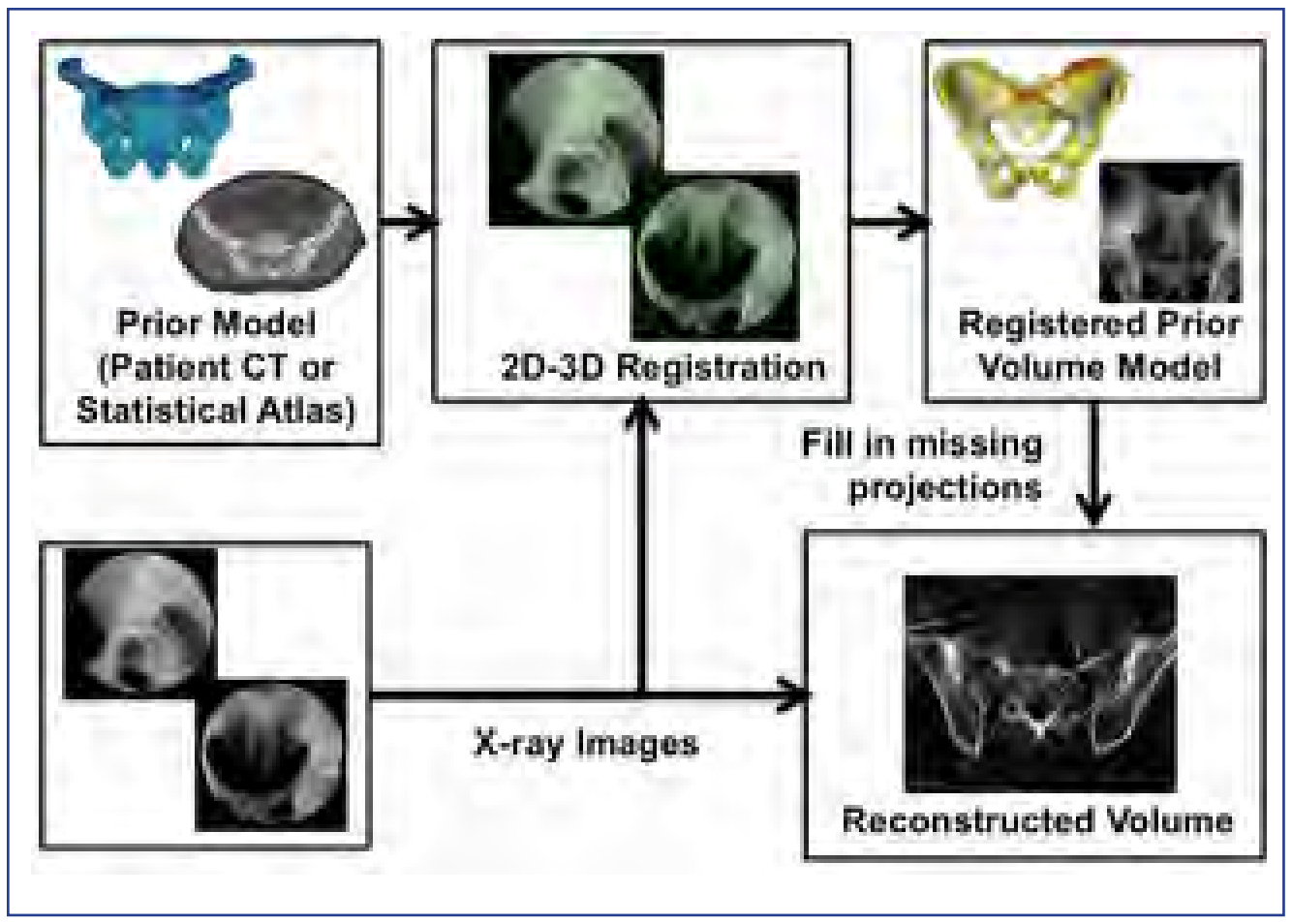

Fig. A6: Hybrid tomographic reconstruction from limited x-ray projection images and a prior CT scan of the patient or statistical atlas of patient anatomy.

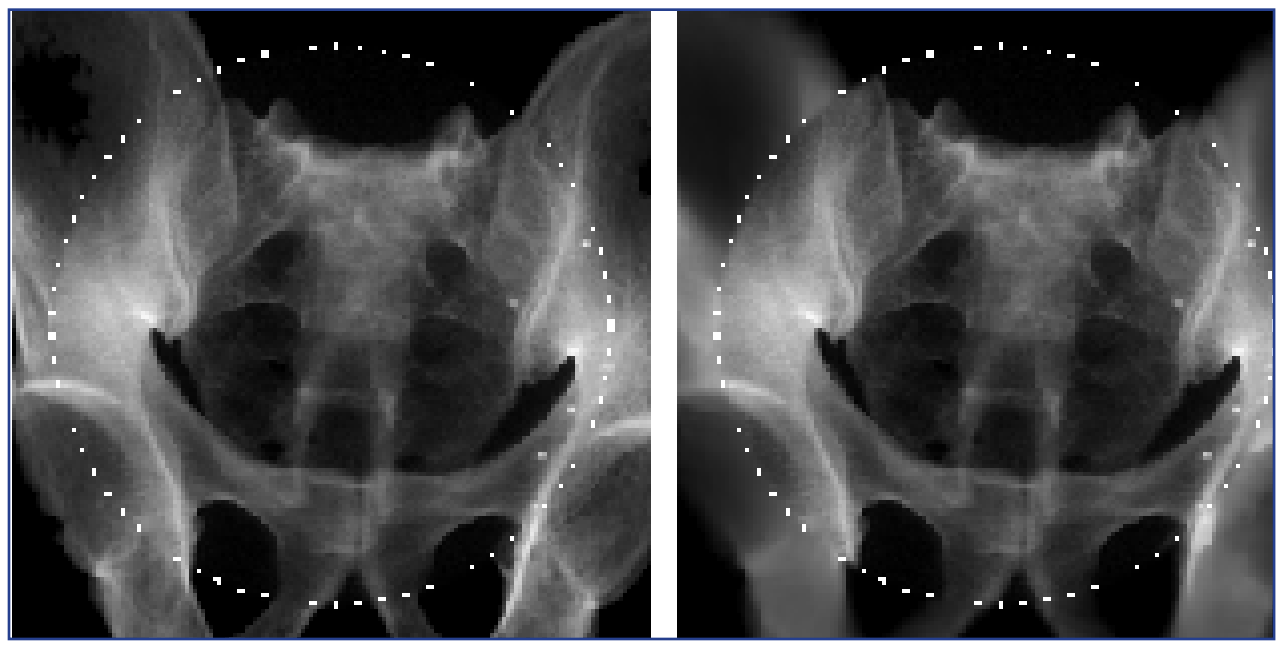

Fig. A7: Fusion of pelvis x-rays (in circle), registered to prior CT (left) and statistical atlas (right) outside the circle. The projections outside the circle compensate for image truncation.

Researchers at the CISST ERC are developing a method to produce three-dimensional (3D) reconstructions of bone from a limited collection of x-rays, that would give orthopedic surgeons 3D information about the patient during surgery using the mobile fluoroscopic x-ray "c-arms" commonly available in orthopaedic operating rooms.

Three-dimensional reconstruction from x-ray projections is at the heart of x-ray computed tomography (CT) and is a well-studied problem. Standard reconstruction methods require the knowledge of x-ray projection intensity for every ray through every reconstructed volume element (voxel). "Cone beam" reconstruction methods of 3D volumes from multiple 2D x-ray projection images are known and are widely available on high-end x-ray systems (such as those found in angiography suites). Only a relative handful of smaller mobile x-ray c-arms, typically found in orthopaedic operating room systems, have such capabilities, and even in these cases there are many limitations such as, it is often impractical to rotate the c-arm far enough around the patient to provide the required x-ray projections. The relatively small detectors in mobile systems also cause the view of the patient's anatomy to be incomplete (Fig. A5) especially in wide parts of the body, including the pelvic and spinal regions. The repeated acquisition of large numbers of intraoperative x-rays can be undesirable both for the patient and operating room personnel. The fluoroscopic detectors and structural components of most mobile c-arms introduce image distortions and other effects that complicate tomographic reconstruction.

We have developed a hybrid reconstruction tomographic reconstruction method that combines intraoperative x-ray images with prior information about the patient in the form of a CT scan or

# Hybrid Tomographic Reconstructions From Prior Models and a Limited Number of Projection Images

*O. Sadowsky, J. Lee, K. Ramamurthi,*
*L. Ellingsen, G. Chintalapani, R. Taylor, J. Prince*

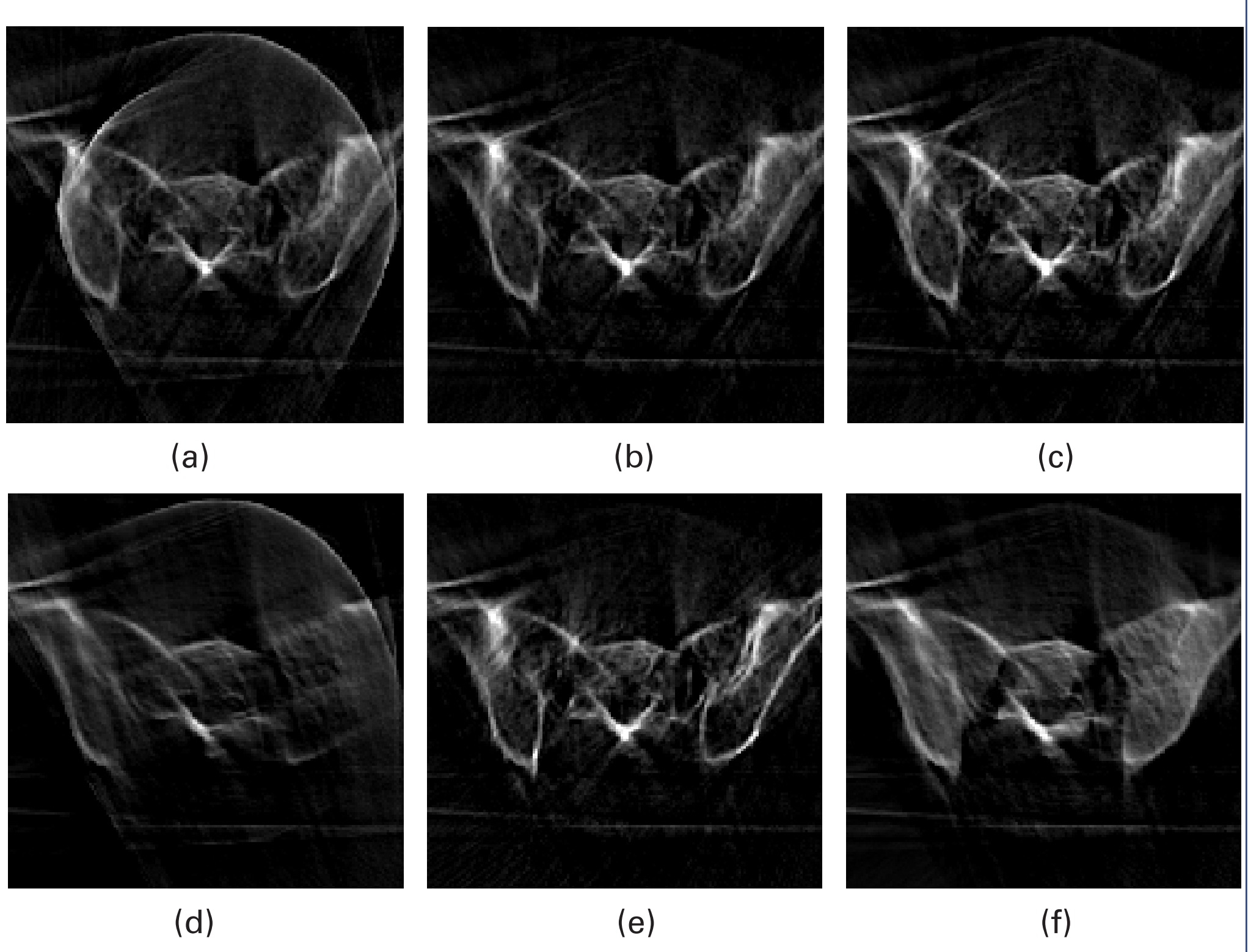

Fig. A8
Hybrid reconstruction of x-ray and prior models. Top row: using a 200° x-ray "short scan." Bottom: using 110° of x-ray scan and 90° of prior model. (a) Cross section reconstructed from truncated x-rays only. (b) Reconstruction from image fusion with prior CT. (c) Reconstruction from fusion with statistical atlas. (d) Reconstruction from 110° x-rays without prior. (e) Reconstruction using 90° projections of prior CT. (f) Reconstruction using 90° projections of deformable atlas.

statistical "atlas" constructed from multiple patients (Fig. A6). The steps of this reconstruction method are as follows. First, a "scan" of x-ray images of the patient from multiple projection angles is obtained. Second, these images are corrected to account for image distortions and other geometric and calibrated intensity-response characteristics of the c-arm. Then the 2D projection images are registered to either the prior CT scan (rigid registration) or the atlas (non-rigid registration). In the current implementation, after the atlas is aligned to the available data the pixel intensities of the projected atlas images are matched to those of the x-ray images. This allows a smooth fusion of both modalities to be obtained (Fig. A8). To compensate for image truncation or missing x-ray views, the data from x-rays and atlas are blended together as inputs to a conventional cone-beam reconstruction algorithm.

Results of this method on a cadaveric human pelvis phantom using a typical mobile fluoroscopic c-arm (OEC 9600) are shown (Fig. A7), using both a prior CT model and a shape/density atlas constructed from CT images of 110 normal adult male subjects. As this is written (May 2008) our immediate plans are to apply this method in cadaver trials to reconstruct intraoperative changes such as cement injections in bone augmentation and to extend the method to other anatomic structures.

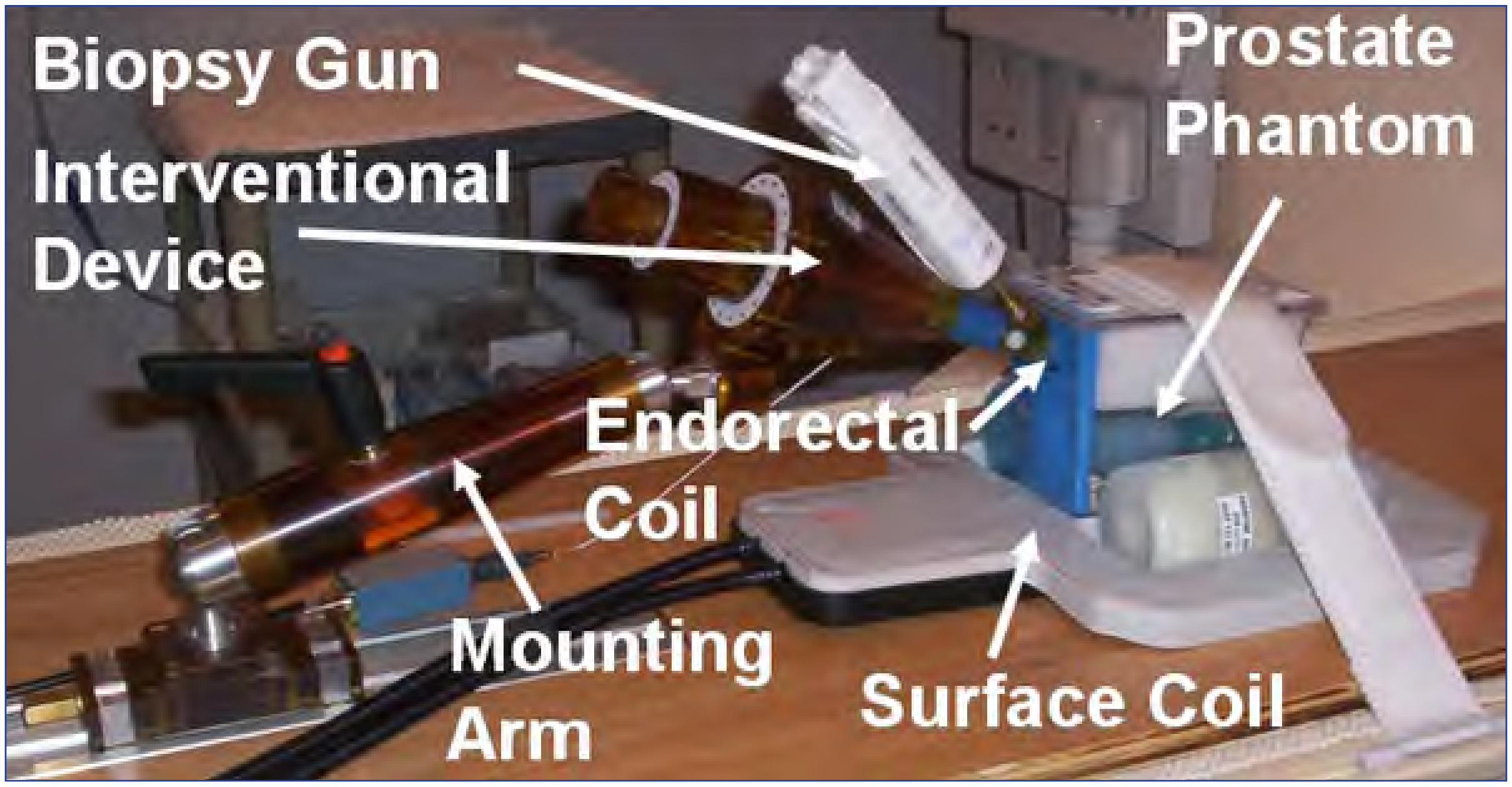

Fig. A9: MRI-guided transrectal imaging and biopsy device with integral endorectal imaging coil placed in a prostate phantom.

Researchers at the CISST ERC have developed a robotic device to place needles precisely into the prostate under Magnetic Resonance Imaging (MRI) guidance. The MRI provides an excellent soft-tissue contrast and has the potential to significantly improve image-guided prostate interventions, which are currently performed with ultrasound [1]. Transrectal MRI guided prostate interventions, such as biopsies and gold marker placements inside a high-field MR scanner, have been reported in initial clinical trials utilizing active [2] and passive fiducial tracking [3]. The CISST ERC recently completed initial phantom and clinical trials of a next-generation version of the system [2]. This new system, shown in Figure A9, employs innovative probe mechanics and a novel hybrid tracking method. The primary goals of the new system are to shorten procedure time and to significantly simplify deployment of the system on different scanners, without any compromise on previously achieved needle placement accuracy [2, 4, 5].

Figure A9 shows the new device placed in a standard prostate phantom. The device guides the needle tip of a standard automatic MR compatible biopsy gun to a predetermined target in the prostate. The device contains an endorectal probe with an integrated single loop imaging coil. A steerable needle channel is joined into the probe. The three degrees of freedom (DOF) to reach a target in the prostate are rotation of the probe, angulation change of the steerable needle channel, and insertion of the needle. The interventional device incorporates a hybrid tracking method comprised of passive fiducial marker tracking and joint encoders. At the beginning of the procedure, the position of the interventional device is obtained from MR images by segmenting the fiducial markers placed on the device.

Two clinical procedures have been performed on a 3T Philips Intera MRI scanner. One procedure encompassed combined biopsy and gold marker placements, while the second employed biopsy only. Figure 3 shows the results of the combined procedure. Four targets were selected on axial T2 weighted FSE images (Fig. A10, top row).

The targeting accuracy of three biopsy needle placements was assessed using proton weighted axial TSE needle confirmation images (Figure 4, second row). The mean in-plane targeting error for the biopsies was 1.1 mm with a maximum error of 1.8 mm.

These phantom studies and clinical procedures demonstrated accurate and fast needle targeting of the complete clinical target volume. The errors and procedure time compare favorably to the reported results achieved with our active tracking method

# MRI-Guided Transrectal Prostate Intervention System

*Axel Krieger, Peter Guion, Csaba Csoma, Iulian Iordachita, Anurag K. Singh, Aradhana Kaushal, Gabor Fichtinger, Louis L. Whitcomb*

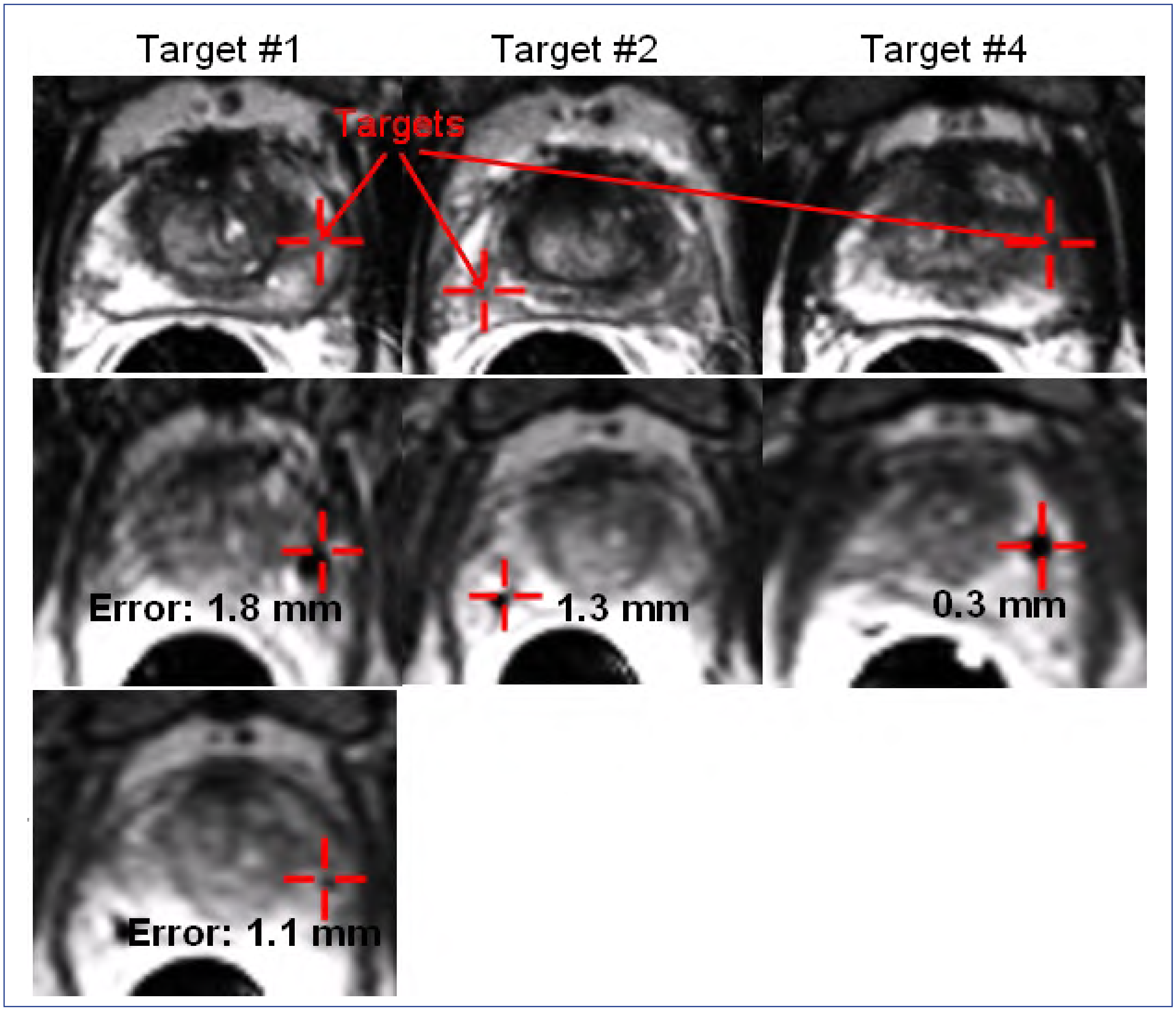

Fig. A10: Targeting images, needle visualization images, and gold marker image of clinical procedure using the transrectal prostate intervention system. Top image row: Suspicious targets (red cross hairs) were selected on axial TSE T2-weighted images. Second image row: The needle tip void was visualized in axial TSE proton density images. The desired targets match the actual position of the needle. Error number: The number indicates the in-plane targeting error for the needle placement. Third image row: Axial TSE proton density image showing the location of the marker placed at target location number 1. The marker void is visible close to the target.

in clinical trials (average error 1.8 mm and average procedure times of 76 minutes) [2, 4, 5]. The hybrid tracking method allows this system to be used on any MRI scanner without extensive systems integration and calibration.

JHU has applied for two patents of invention on this novel technology, and is successfully licensing these patents for commercialization.

*Acknowledgements: This project was initiated with support from the NSF CISST ERC that enabled our team to obtain preliminary results. These preliminary data enabled us to obtain support for the project from the National Institutes of Health under Grant 1 R01 EB02963.*

References:

[1] Yu KK. Radiol Clin North Am 2000;38:59.
[2] Krieger A. Trans on Biomed Eng, 52(2):306-313, 2005.
[3] Beyersdorff D. Radiology, 234(2):576-581, 2005.
[4] Susil RC, J Urol, 175(1):113-20, 2006.
[5] Susil RC, Radiology. 228(3):886-894, 2003.

Seven graduate students from the CISST ERC had the opportunity, through the CISST ERC International Research, Education, and Engineering program (IREE), to spend a summer abroad to collaborate with four laboratories at the Technical University of Munich (TUM). These include the Computer Aided Medical Procedures and Augmented Reality Laboratory (CAMP), the Minimally-Invasive Interdisciplinary Therapy and Intervention Lab (MITI), Micro Technology and Medical Device Technology (MIMED), and Institute of Automatic Control Engineering. These

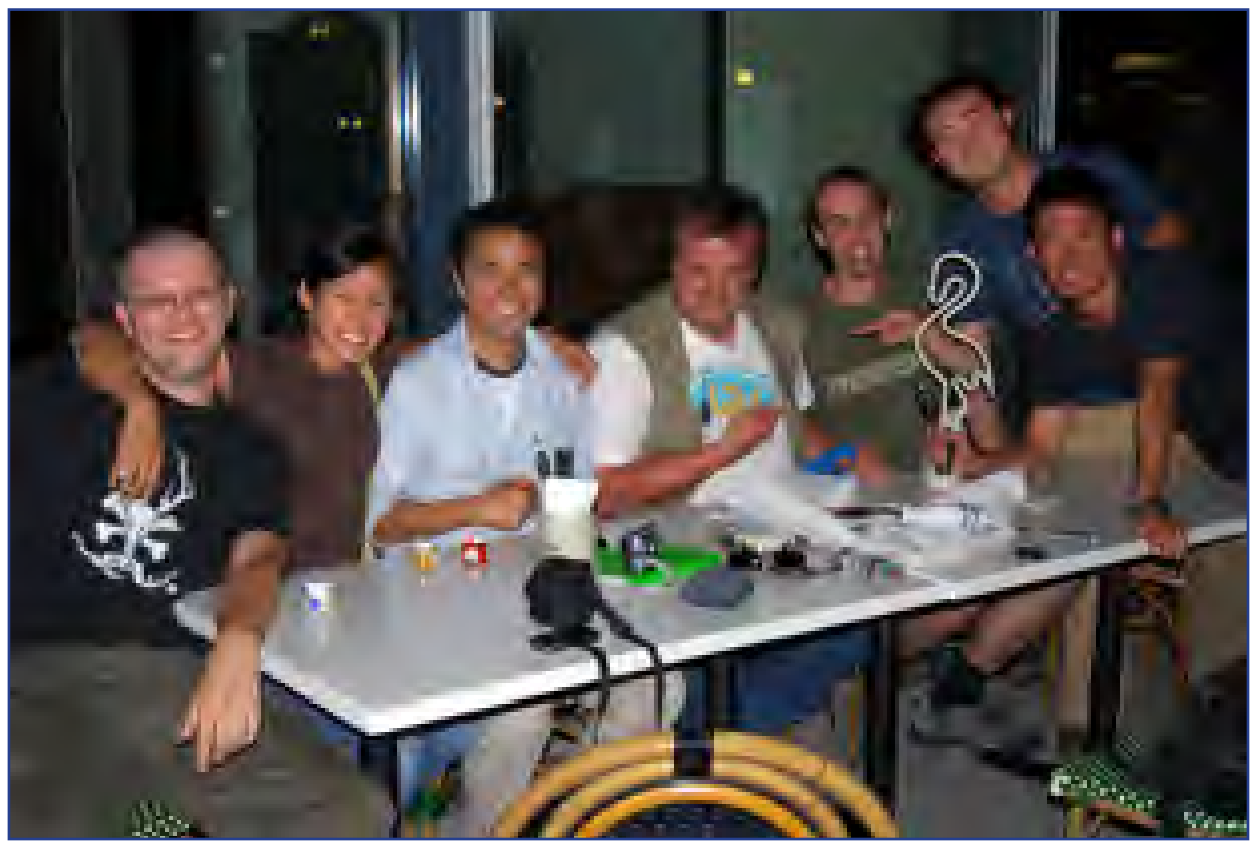

Fig. A11

collaborations' goal was to take the first steps toward a more formal joint exchange program with TUM. The visit, which was organized by Professor Hager (CISST ERC) and Professor Burschka (TUM), included both scientific and cultural elements. Brief descriptions of three of the scientific projects are as follows:

***Surgical Skill Evaluation:*** The goal of this project was to collect synchronized video data from laparoscopic cholecystectomy procedures. To accomplish this, we designed a four-camera live surgery recording system. The first two cameras were Allied Vision Technologies (AVT) Guppy cameras constituting a stereo video system that focuses on the patient's chest and abdominal area. Sterile color markers are attached to laparoscopic tools to assist in tool tracking, identification, and orientation extraction. The third camera is the mono-endoscope already used in laparoscopic cholecystectomies. This provides information about the motions of the tools and tissue inside the abdominal cavity. The fourth camera has a complete view of the operating room. By the end of our research experience, our team successfully recorded a test surgery and two live cholecystectomies. Arrangements were made with the CAMP lab at TUM to continue recording data for further collaboration.

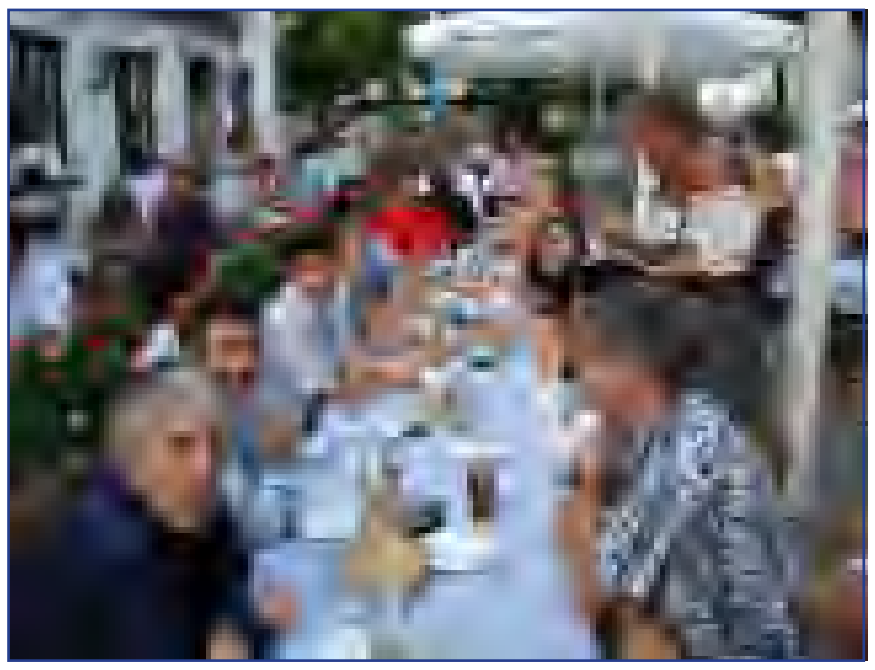

Fig. A12

***A daVinci Tool Control:*** This project was created to develop a stand-alone actuation and control unit for the daVinci Surgical System's

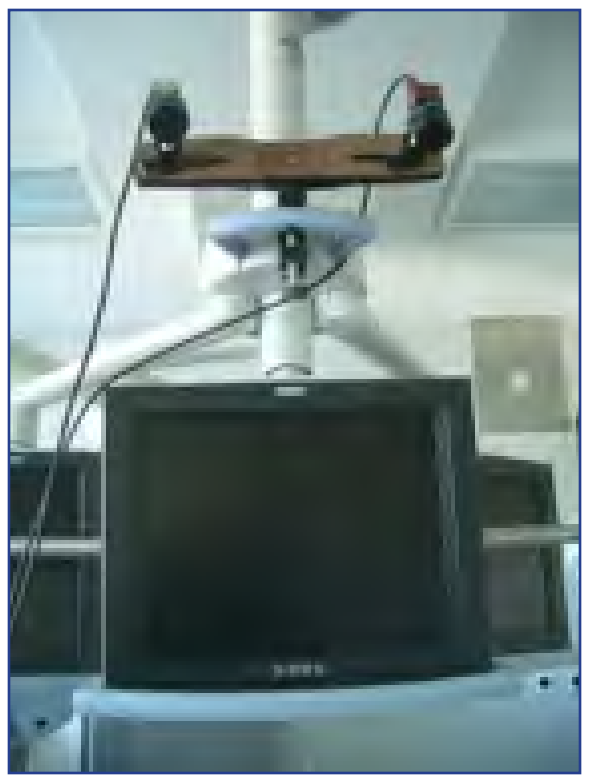

Fig. A13

# CISST ERC IREE: Collaboration with the Technical University of Munich

EndoWrist instruments. The developed device serves as a platform for the development of new daVinci surgical instruments and surgical procedures using existing instruments without the master robots. The joint effort from MIMED (TUM) and CISST ERC (JHU) resulted in an inexpensive software and hardware solution that is easily replicated, and can be used in a surgical environment for instrumentation research. We hope that this platform will lead to future collaboration between the two research centers and other surgical robotics research teams.

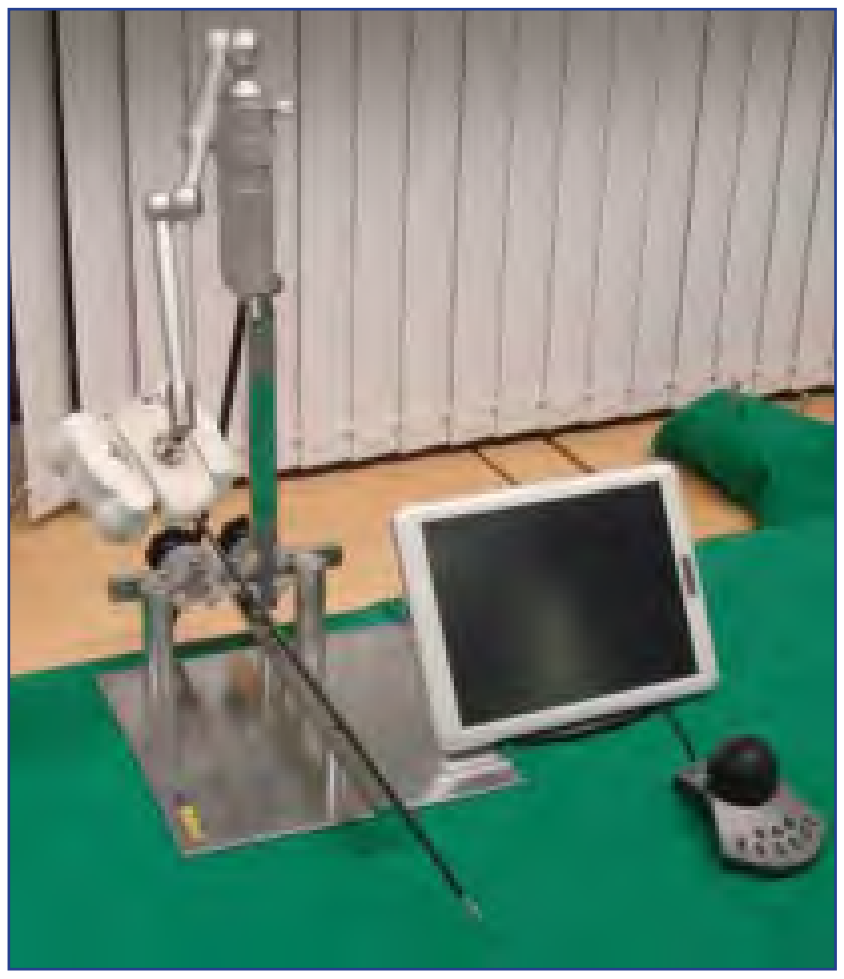

Fig. A15

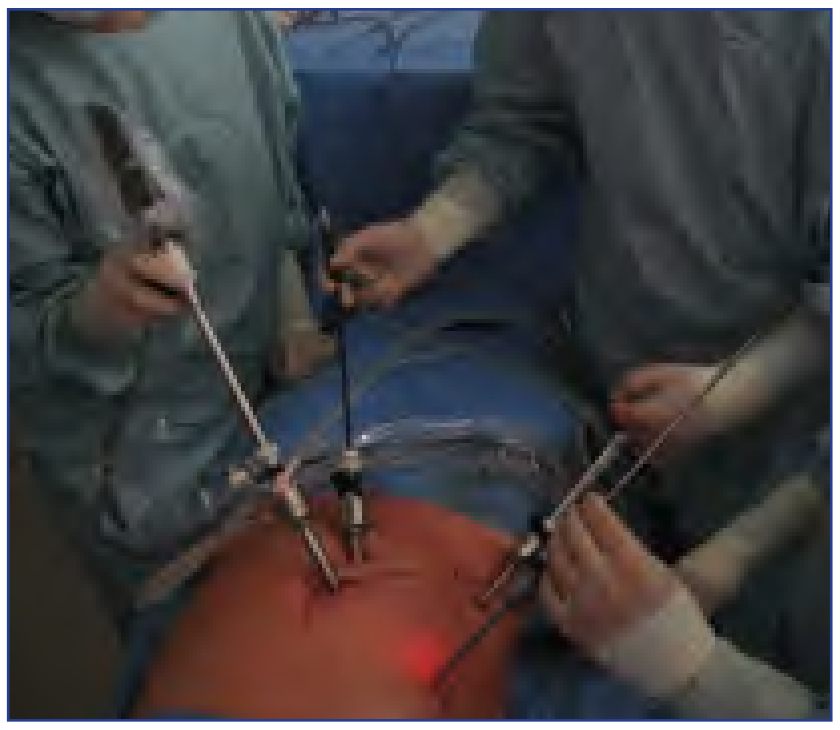

Fig. A14

***Parameter Estimation During Teleoperation:*** The goal of this work was to estimate the mechanical properties of soft tissues during teleoperation. This knowledge will be used to return realistic force feedback to human operators so that they feel the interaction force as if they are directly touching the remote environment. Moreover, the obtained environment model can be used for surgical simulation, evaluation of environment states during an operation, and visual sensory substitution to help surgeons detect tumor locations. To identify unknown parameters of the environment, we collaborated with TUM researchers to develop a means of logic-based switching for rapidly convergent parameter estimation. We have solved theoretical problems of the multi-estimator and conducted experiments to test the performance of several estimation techniques.

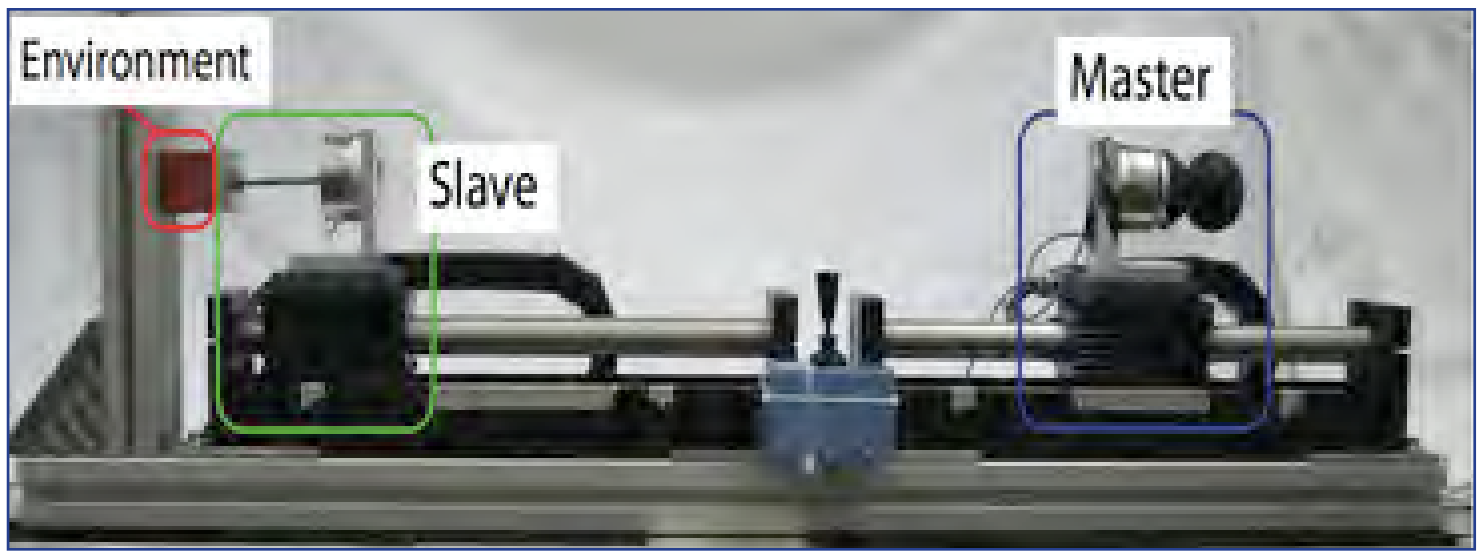

Fig. A16

*The Winter School provided us with a superb opportunity to showcase the accomplishments of the CISST ERC over its first ten years while promoting the education of the next generation of researchers in this emerging industry.*

*~Russ Taylor*

# *Education*

# *Winter School on Medical Robotics and Computer-Integrated Interventional Systems*

This past January, the NSF Computer-Integrated Surgical Systems Technology Engineering Research Center (CISST ERC) hosted the first Winter School on Medical Robotics and Computer-Integrated Interventional Systems at Johns Hopkins University. During this week-long program, 19 internationally recognized faculty and 38 students from around the world participated in an intensive short course that provided an introduction to the basic technology, systems, and research themes in the emerging field of medical robotics and computer-integrated interventional systems.

The first day of the course was combined with the 10-year anniversary celebration event for the CISST ERC, and included research overviews, laboratory demonstrations, and invited lectures from Dr. Edward Miller, the Dean of the Johns Hopkins School of Medicine, Dr. Jonathan Lewin, the Chairman of the Johns Hopkins Radiology Department, and Dr. Gary Guthart, President of Intuitive Surgical, Inc., as well as an industry/clinician panel.

For the balance of the week, the students and faculty attended morning tutorial lectures. In the afternoons, 25 of the students took a compressed version of Johns Hopkins' Surgery for Engineers course, taught by Dr. Michael Marohn at JHU's Minimally-Invasive Surgery Training Center, while the rest participated in a mini-symposium featuring additional "advanced topics" talks by the faculty. In the evenings, all participants regrouped for poster presentations, dinner, and after-dinner lectures.

Further information may be found on the School's Wed Site:

http://www.cisst.org/wiki/MRCIIS

*Acknowledgements: Partial funding for this event was provided by the National Science Foundation under Grant EEC-0838813 and by the IEEE Robotics and Automation Society.*

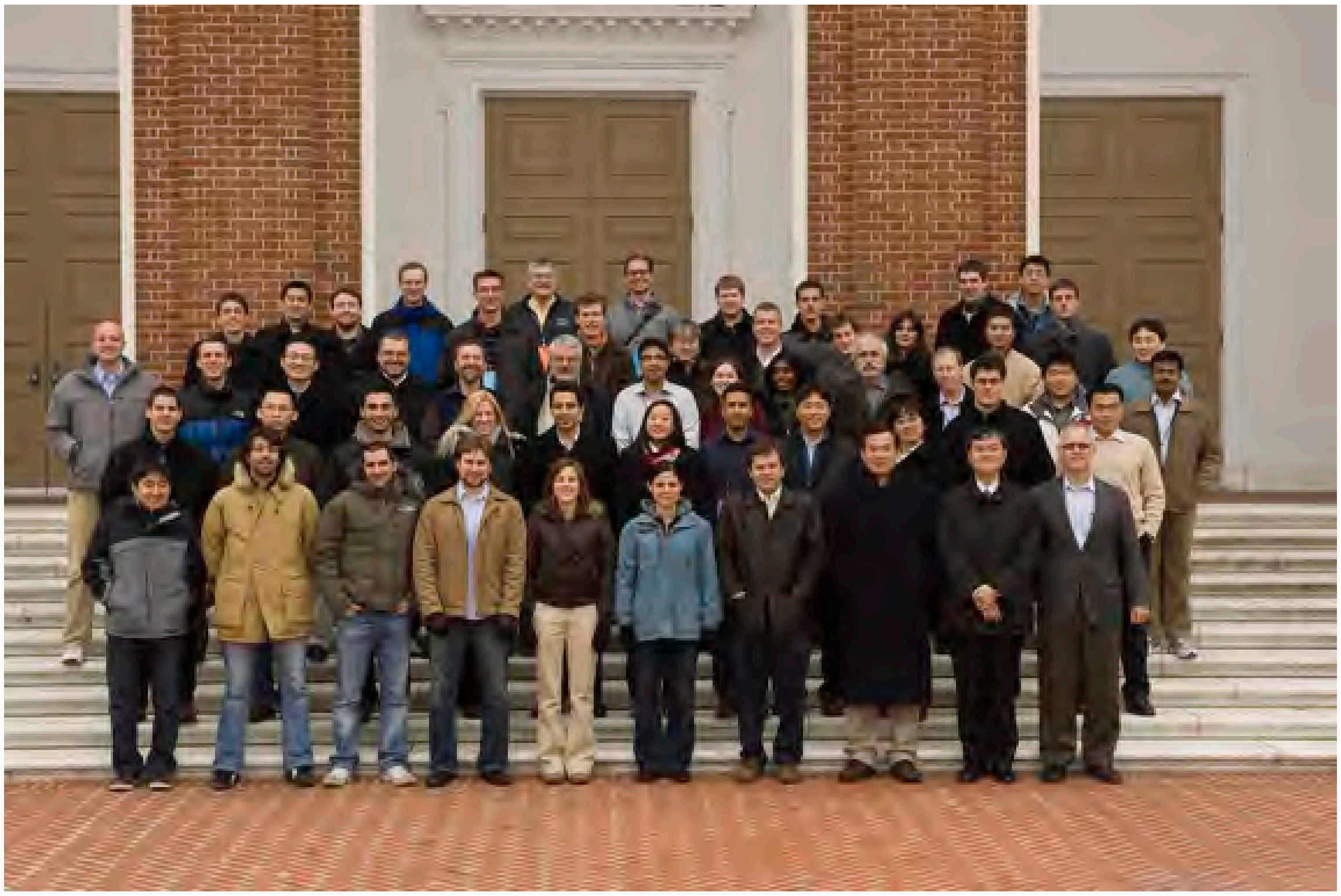

Fig. A17: Participants in the first Winter School on Medical Robotics and Computer-Integrated Interventional Systems

# Project Based Learning: The Feeling of Color
## A Haptic Feedback Device for the Visually Disabled

*Electronics Design Laboratory*
*Electrical and Computer Engineering*
*Students: Elen Tsai and Helen Schwerdt*

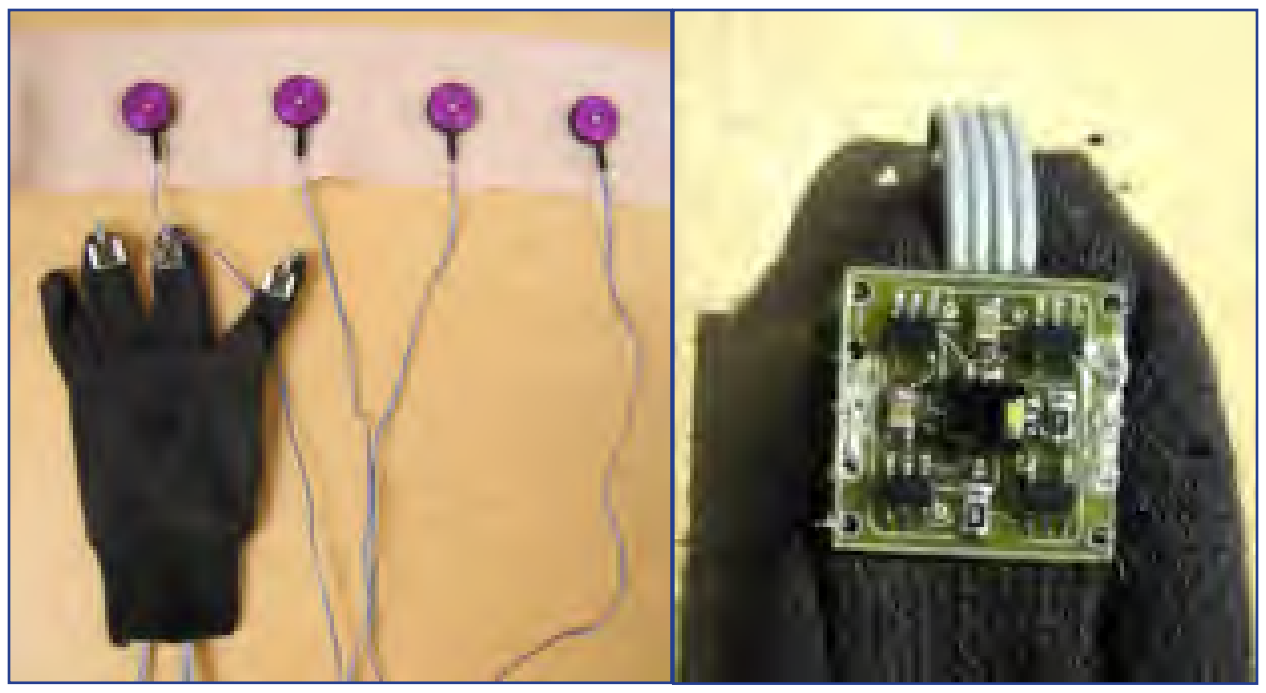

Fig. A18: Concept of the ColorGlove for providing the visually impaired information about the color of objects that they touch.

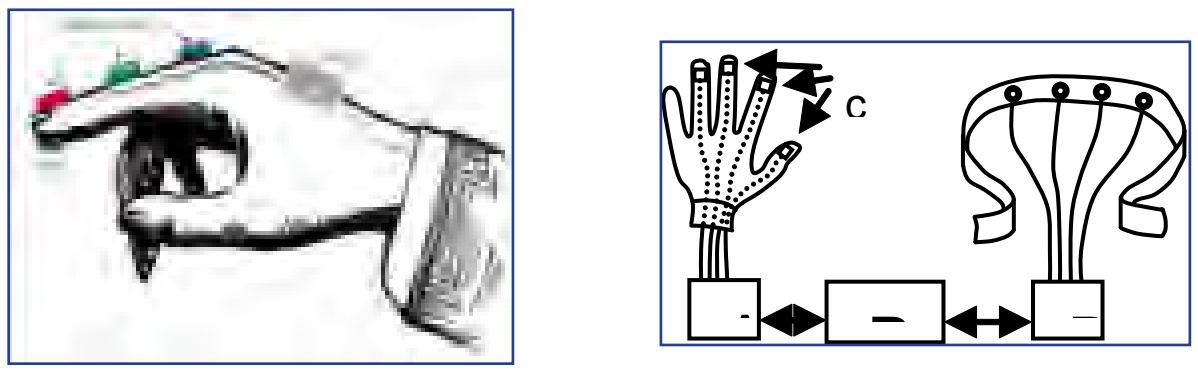

Fig. A19: Prototypes of the ColorGlove with 4 tactile feedback "tractors". The first three fingers are instrumented with color sensors, also shown enlarged to the right. The color information from the sensors is communicated by the combination of poisition, amplitude and frequency of the tactors.

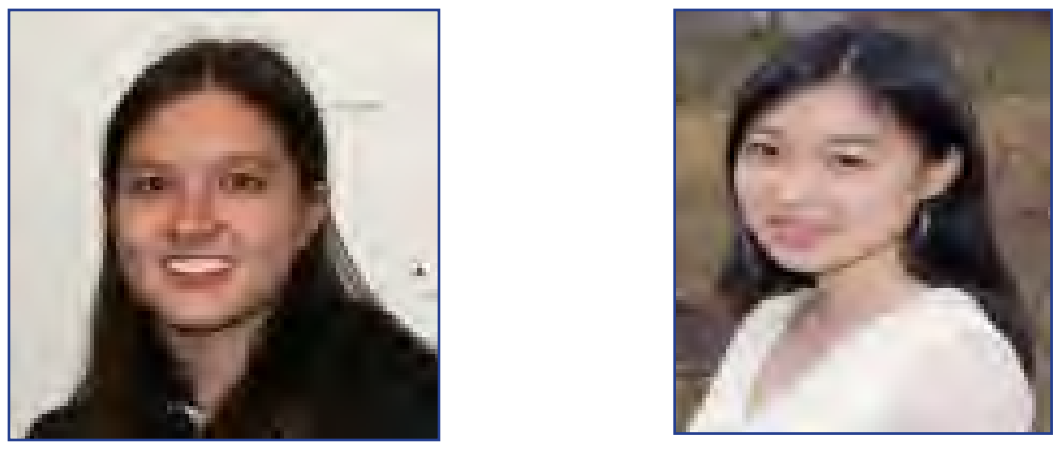

Fig. A20: Students working on the project. Left: Helen Schwerdt, Senior in Biomedical Engineering. Right: Elen Tsai, Senior in Electrical and Computer Engineering.

Professor Ralph Etienne-Cummings runs a project course in the spring of each year that combines electronics design with product development. Students are required to conceptualize a needed product, conduct research to see if said product exists, analyze what is available in the market place and any related literature, and then develop their own product to address the gaps. Typically, the devices that are developed are biomedically oriented. Ms. Tsai and Ms. Schwerdt worked on a glove that could potentially give a blind person the ability to sense color. A final prototype is now being developed which will be evaluated through a properly designed experimental protocol using human subjects. If successful, and after obtaining input from the end users, we expect that this device could be of significant value to improving the lives of the visually impaired.

The ColorGlove was designed to provide a means of color perception to the blind or visually impaired. The glove implements a sensor using a phototransistor, which responds to light intensity and wavelength corresponding to a specific LED color (red, green, or blue) being activated and the reflected light, dependent on the surface color. Feedback is provided to the user by means of cell phone vibrators which vibrate at an intensity that correlates with the intensity of the color at the surface of the sensors. The vibrators are assembled between the three finger joints, where each vibrator corresponds to a red, green, or blue LED. In order to effectively perceive color, the user learns to distinguish colors based on the amount of red, green, and blue on the object of interest.

# Pre-College Outreach Programs:
## The Robotics Systems Challenge and The Summer Robotic Camp

The CISST ERC organized and sponsored two successful pre-college outreach programs, positively impacting both middle and high school students from across the state of Maryland.

For the third year, the CISST ERC Student Leadership Council and CISSRS organized and held the Robotic Systems Challenge on April 5th, 2008. This event took place in the JHU Gymnasium with twenty-two competing teams and fifty-seven participants. There were four challenges that students could complete. These were chosen in advance and indicated on the required registration form. For each of the challenges, a robot was built and programmed by the teams to be used in the competition. Every challenge was different with each team using Parallax Boe-Bot and LEGO Mindstorm kits.

The first challenge was the Petite Slalom featuring a board of gates that the robots had to maneuver through. After seeing the course, the teams had fifteen minutes to program their robots. The second challenge was Search and Destroy the Brain Tumor using Boe-Bots. The robots used the LED's in the kit to randomly detect "tumors" on a board. Once the tumor was found, the robot had to demonstrate the detection by either flashing its lights or beeping. The third challenge was a Mystery Course where the premise was not revealed until just before the start. The robots had to navigate through a maze using "whiskers" to detect walls and other obstacles so that it could complete the course. The fourth challenge was unique because the teams prepared a robot in advance which will be used in society. Written and oral reports were also a part of the judging for each team. During lunch, a student teacher from the JHU Engineering Innovation program used a slide show to discuss the different kinds of engineering that are so important in the world today. After lunch the teams were awarded prizes and certificates for first, second, and third place.

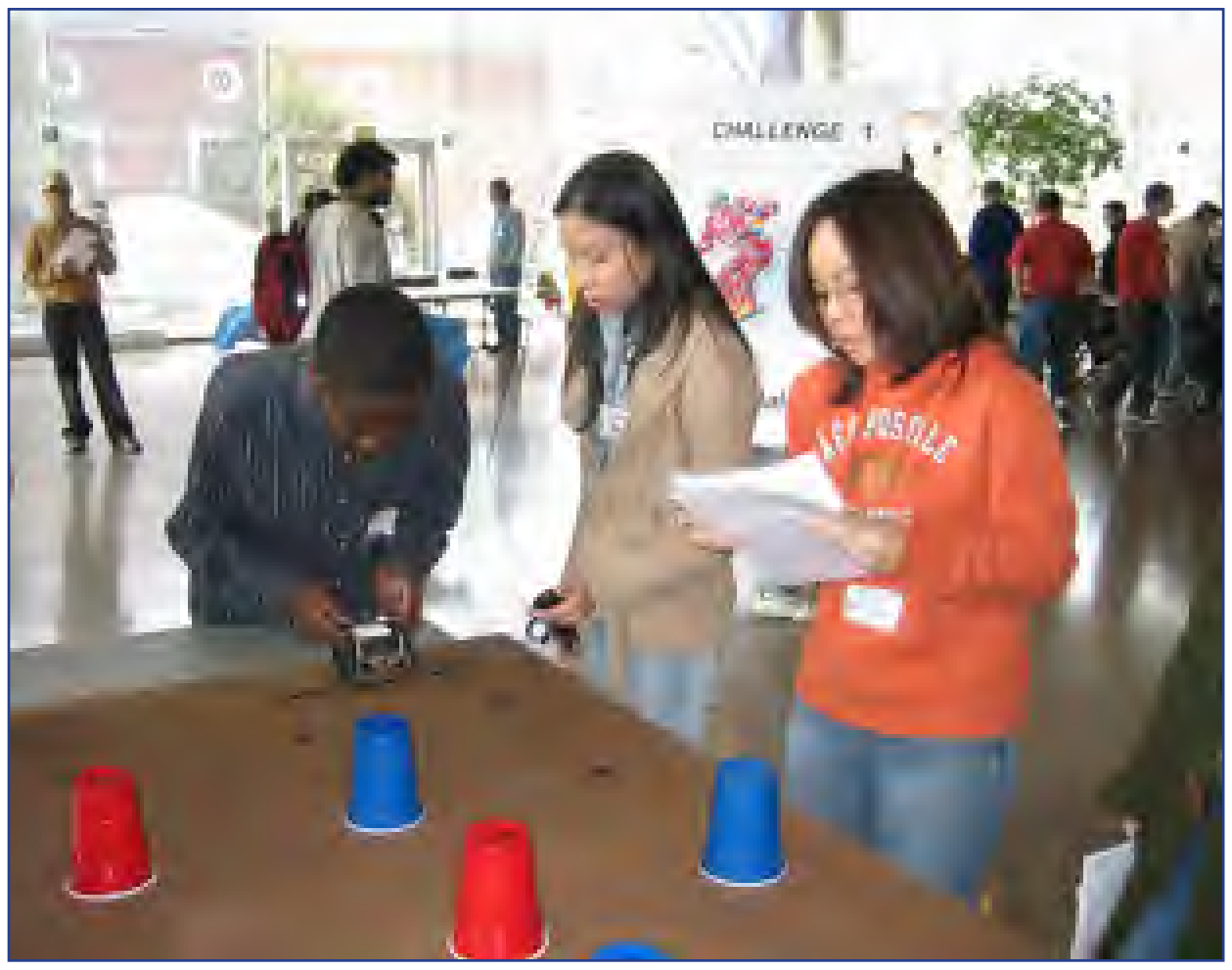

Fig. A21: Students take part in the Robotics System Challenge

The next program, the Summer Robotics Camp, was held for the fifth consecutive summer during June 16-27, 2008. The camp was held two separate weeks with thirteen students attending the first week and fourteen appearing the second week. The student participants built blinkie kits with blinking LED's on day one. The second and third days were spent building robots that reacted to sound. The students also took place in a tour of the ERC labs. On the fourth day, students designed robots that were built from the basic stamp one kit and programmed them to run a route and blink lights.

# Mock Operating Room

One of the greatest challenges in computer-assisted interventions is the translation of innovative research into clinical practice. A prominent factor is the disparity between the engineering laboratory, where the research is performed, and the operating room or interventional suite, where the device must be integrated within the clinical workflow and environment. To address this gap, the CISST ERC has created the Richard A. Swirnow Computer-Integrated Surgical and Interventional Systems Mock Operating Room on the Engineering campus (Figs. A22 & A23).

The Mock OR provides a realistic environment for pre-clinical system integration and validation testing. The Mock OR contains two ceiling-mounted operating room lights, one equipment boom, and a standard OR table, all donated by the Steris Corporation. It also provides a home for major medical equipment that was previously donated to the CISST ERC by its industrial affiliates. The equipment includes a General Electric OEC x-ray fluoroscope, a Medtronic Stealthstation navigation system, a Northern Digital Polaris tracking system, and an ISS Neuromate robot. Additional equipment acquisitions, such as an Intuitive Surgical daVinci robot, are expected in the near future. The room is quite expansive, containing several benches which provide laboratory space for projects that benefit from the Mock OR facilities. In the last semester, several students conducted their research projects in the Mock OR.

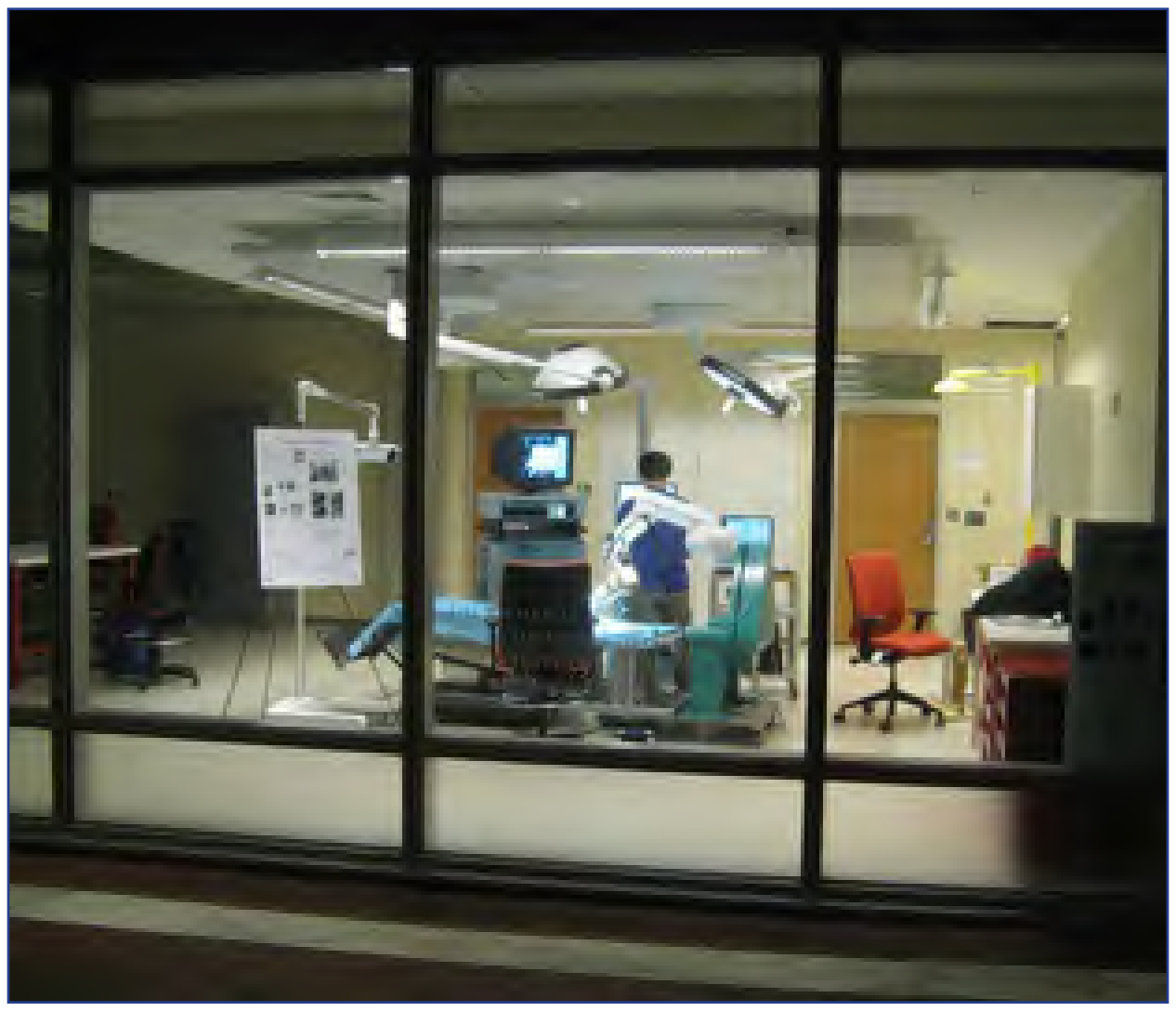

Fig. A22: Mock OR viewed from walkway

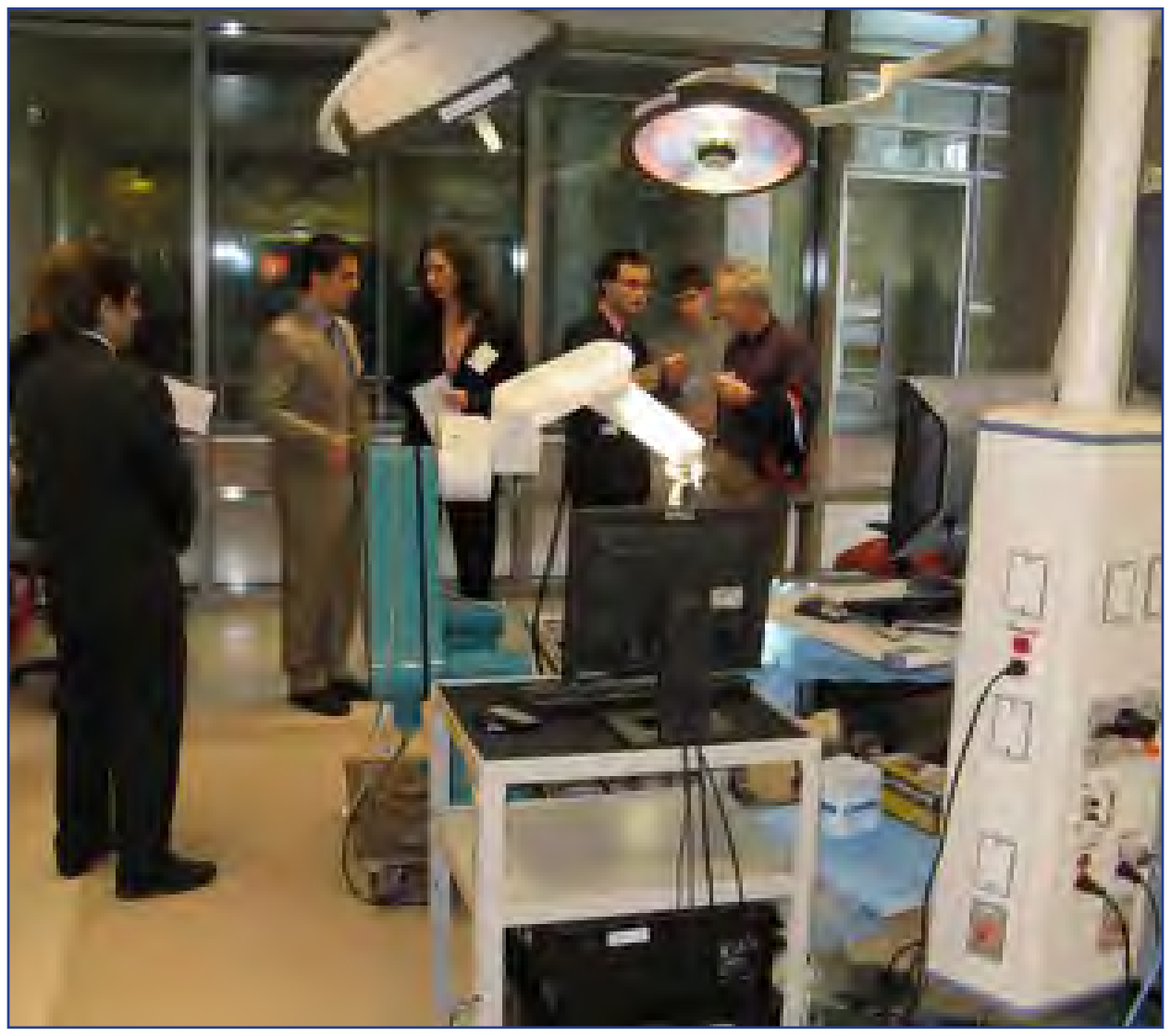

Fig. A23: Demonstration during Open House event

The Mock OR's construction was completed in March 2008, just prior to the University Open House for the new CSEB where the Mock OR and other CISST ERC labs are housed. During the Open House event, two students demonstrated the neurosurgical robot system (Fig. A20). This system is an illustrative example of the type of research that is enabled by the Mock OR, created by integrating the Neuromate robot and the Medtronic Stealthstation. Although this integration could have been done in any facility, the Mock OR environment allowed the researchers to also consider clinical feasibility such as placement of this equipment within an OR environment as well as the ergonomics of the user interface. We expect the Mock OR to bring together researchers from different fields due to the diversity of equipment, the large amount of floor space, and the ability to provide a familiar environment for our medical collaborators. Finally, the Mock OR showcases the CISST ERC research to the entire JHU community and its visitors; it has large glass windows and is located in a high-traffic walkway near the visitor center, making it a popular stop for campus tours.

We anticipate that by providing a realistic OR environment, the Mock OR will enable faster translation of research from the laboratory to the operating room, thereby benefiting patients, the overarching goal of the CISST ERC.

*Acknowledgments: Creation of this facility was supported by the Whiting School of Engineering, a donation from Rachel and Richard Swirnow, and an equipment donation from the Steris Corporation.*

# Small Animal Radiation Research Platform (SARRP)

Preclinical research using well-characterized small animal models has provided tremendous benefits to medical research, enabling low cost, large scale trials with high statistical significance of observed effects. Although many mouse models of human cancer are currently available, existing clinical imaging and therapeutic systems are ill suited to such small subjects, and the equipment is also in high demand and seldom available for lengthy laboratory trials. At present, simple single beam/single fraction techniques are commonly used to irradiate laboratory animals. This technology is far removed from the advanced three-dimensional (3D) imaging, planning and computer-controlled delivery technologies that are used for human treatment. The Computer-Integrated Surgical Systems and Technology (CISST) ERC, in collaboration with the Department of Radiation Oncology and Molecular Radiation Sciences at the Johns Hopkins University School of Medicine, developed the Small Animal Radiation Research Platform (SARRP) to bridge this technological gap between laboratory radiation research and human treatment methods. The SARRP integrates cone-beam computed tomography (CBCT) imaging and conformal irradiation delivery to enable image-guided radiation research with small animals. It leverages the research infrastructure of the CISST ERC, including the use of the CISST software libraries.

The system (Fig. A24) consists of a kilovoltage x-ray tube mounted on a manually rotated gantry. The tube provides a low-energy beam for CBCT imaging (with a flat panel image detector) and a high-energy beam for radiotherapy. The animal is placed on a four-axis robotic positioner (three translations and one rotation). CBCT imaging is performed by placing the x-ray tube in the horizontal position (as in Fig. A24, but without the collimator) and then rotating the animal while capturing 2D x-ray images. A standard reconstruction algorithm is used to obtain the 3D image. Radiation therapy can be delivered from any orientation of the gantry, using a variety of collimators (as small as 0.5 mm in diameter). We developed a calibration method to enable accurate targeting of the x-ray beams. This method uses an x-ray camera, mounted on the robotic positioner, to measure the mechanical axis of rotation and the precise location of the x-ray beam at each gantry rotation. Figure A25 shows the result of a validation experiment performed with three x-ray films in a vertical stack. We chose a target on the center film and delivered a radiation beam (1 mm diameter) from multiple gantry angles (45 and 75 degrees from horizontal) and at 45 degree increments of the rotary stage. The center film shows the spot where all x-rays intersect.

The SARRP has been installed in the School of Medicine since December 2006 and used to support several research studies, including the response of normal tissue stem cells to focal radiation injuries, the development of positron emission tomography (PET) markers for early assessment of radiation induced toxicity in the lungs, and the study of molecularly targeted therapy in combination with radiation in pancreatic and prostate tumor models. A second system is currently being constructed to satisfy the high demand within JHU; outside institutions have also expressed interest. We believe that the SARRP provides the infrastructure that will enable researchers to improve the treatment of cancer.

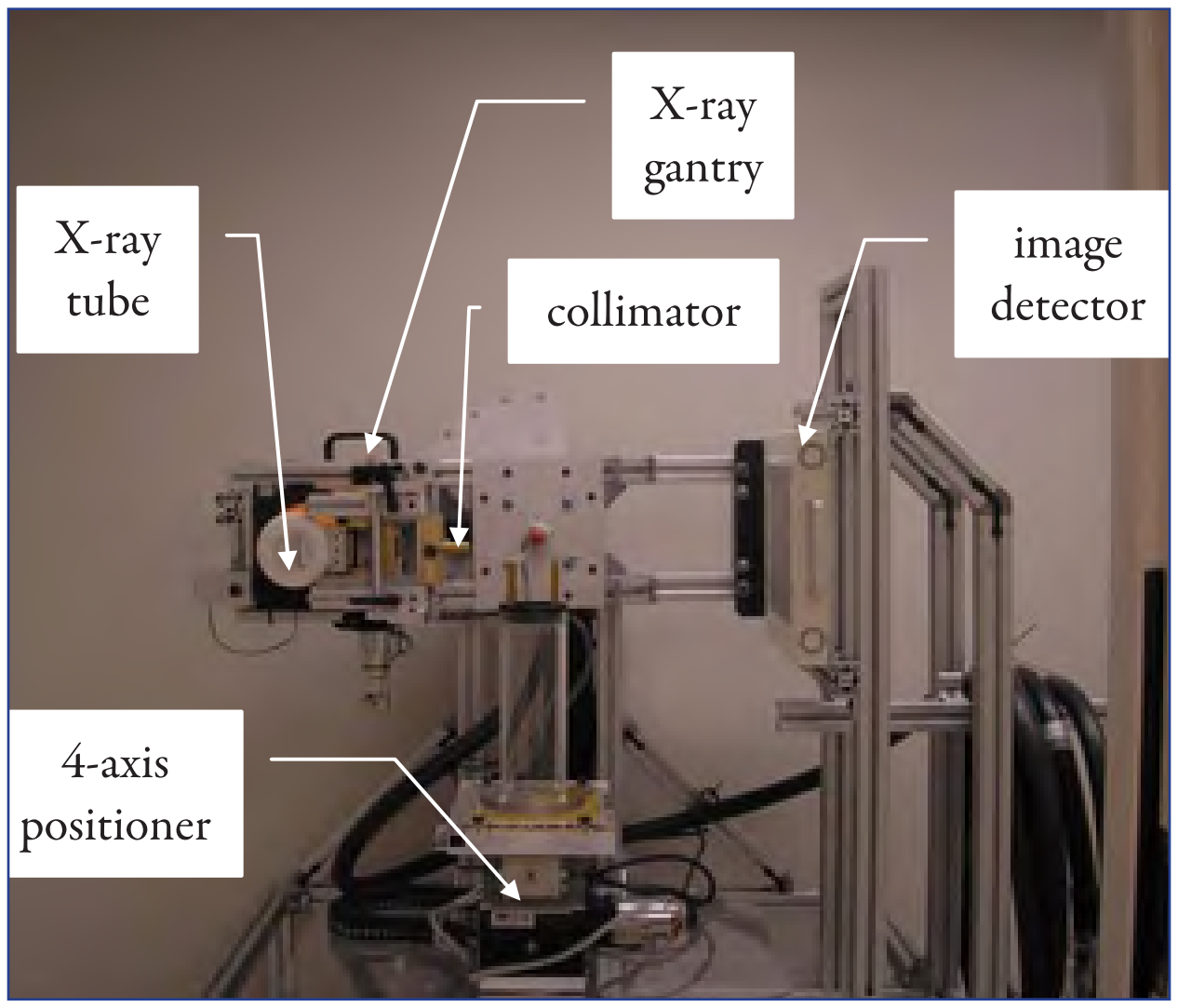

Fig. A24: Small Animal Radiation Research Platform (SARRP), shown with gantry in horizontal position (0°).

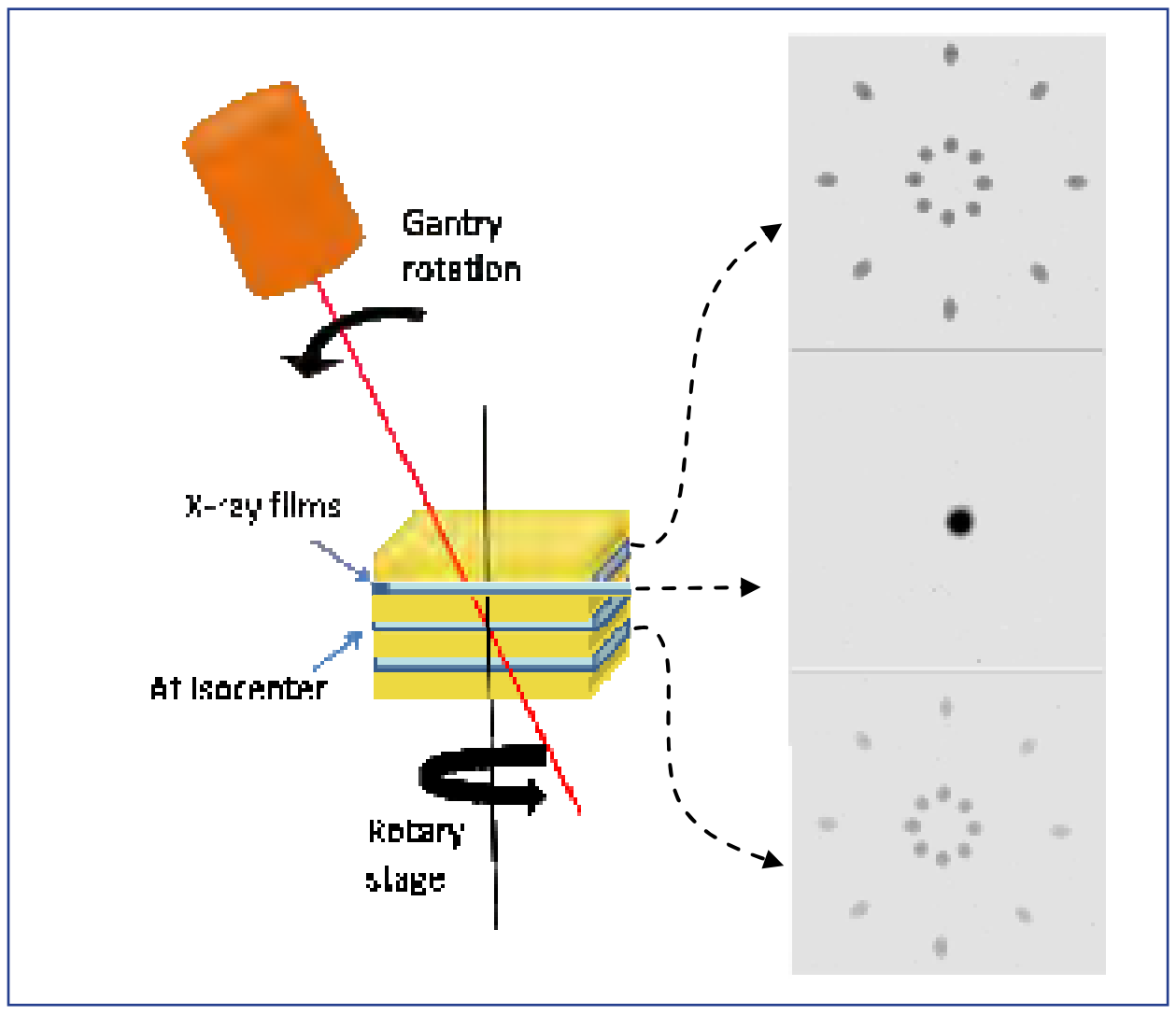

Fig. A25: Validation using X-ray films at gantry angles of 45° and 75°, with 45° stage rotations.

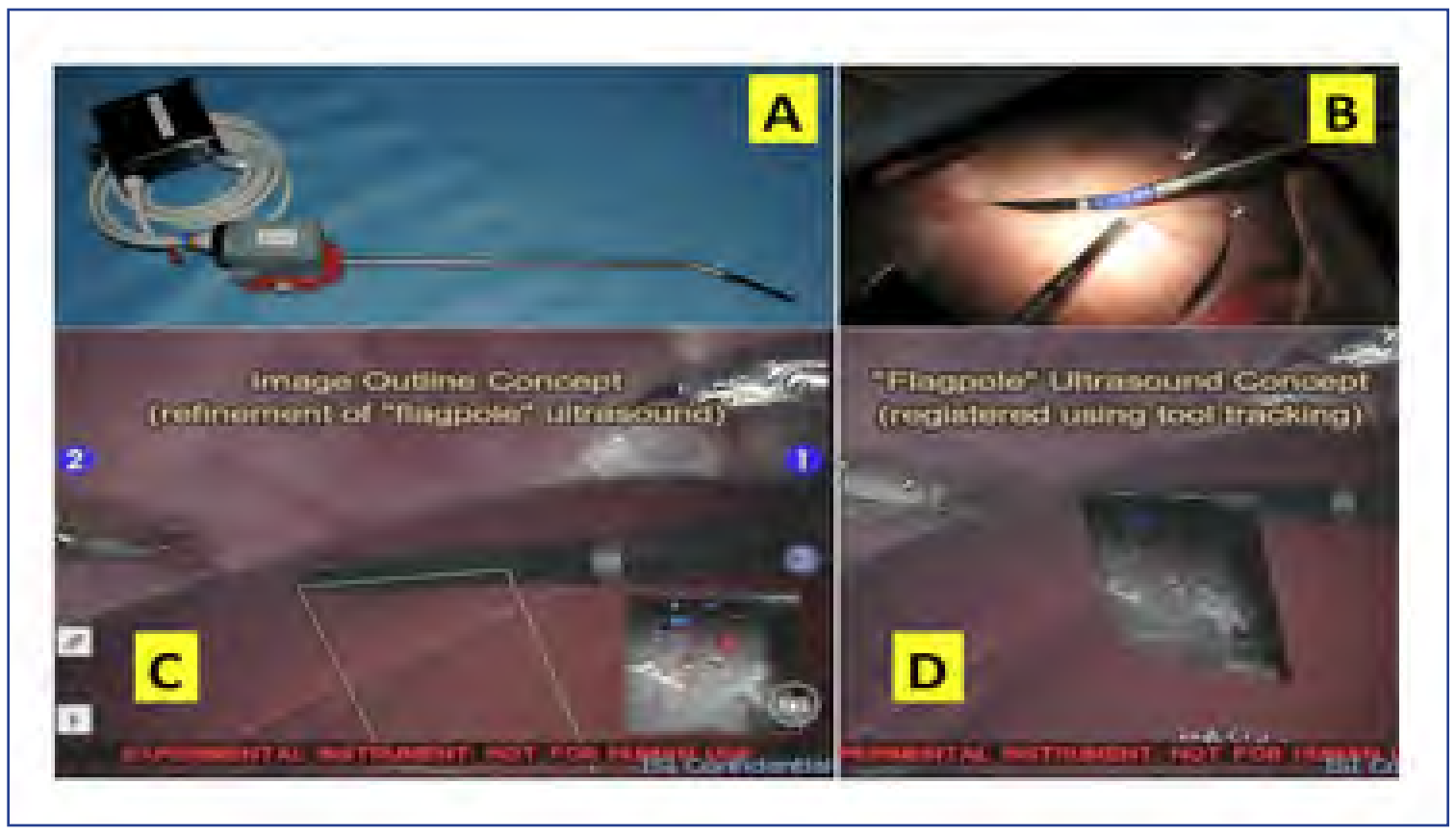

Fig. A26: daVinci Ultrasound system: A) dexterous laparoscopic ultrasound tool; B) intraoperative ultrasound scene; C, D) views through the daVinci stereo visualiza-tion system showing several video overlay concepts.

In a NIH-funded Phase II STTR project based upon CISST technology, the CISST ERC, JHU Department of Surgery, and Intuitive Surgical Systems (ISI) are working together to develop an integrated laparoscopic ultrasound capability for ISI's da Vinci surgical robot. Our immediate goals are to provide a laparoscopic ultrasound capability comparable to that currently available in open surgery and to enhance that capability through information fusion with the daVinci visualization environment.

The current daVinci paradigm is essentially that of open surgery translated into a laparoscopic environment. The surgeon's visualization is limited to structures visible in the stereo-endoscopic view. The integrated laparoscopic ultrasound capability will provide the surgeon with a direct means for visualizing deeper structures within organs, overcoming the lack of tactile capability in minimally invasive surgery. This will also create an important step in the evolution of the daVinci from simply replicating 20th Century laparoscopic environment surgical techniques into a true surgical assistant. Liver surgery was selected as the initial clinical platform to develop and test this system, as this is where ultrasonography is most often clinically applied. The potential clinical applications are expansive including diagnostic and therapeutic procedures on the kidney, pancreas, bile ducts, intestinal tract, retroperitoneum, and mediastinum.

Although intraoperative ultrasound has shown promise in minimally invasive laparoscopic procedures, its potential is limited by several difficulties presented for surgeons. These include: 1) manipulating rigid or articulated laparoscopic ultrasound probes; 2) relating images observed on ultrasound consoles to the corresponding physical positions of the probe as well as the observed positions of other objects such as surgical instruments; and 3) integrating these images with other information to form a comprehensive appreciation of the surgical environment. How-

# High dexterity robotic laparoscopic ultrasound

R. Taylor, P. Kazanzides, A. Deguet, B. Vagvolgyi, E. Boctor, J. Leven, C. Schnei-der, M. Choti, L. Assumpcao, C. Hasser (ISI), S. DiMaio (ISI), Greg Dachs (ISI), Wenyi Zhao (ISI) and Tao Zhao (ISI).

ever, the daVinci system does give surgeons dexterous laparoscopic tools, intuitive motion, and a high-fidelity 3D vision system which allow them to achieve outcomes comparable to or better than open surgery, with the benefits of minimally invasive surgery.

As of May 2008, an ISI-developed articulated laparoscopic probe has been integrated with a "S" model daVinci (Fig. A26). A number of visualization paradigms have been prototyped using the "Surgical Assistant Workstation (SAW)" software being developed with NSF funding by the ERC and ISI. This laparoscopic ultrasound capability is a principal "use case" in the SAW development.  Work is underway to develop advanced manipulation capabilities using the SAW infrastructure. This will allow the infrastructure to facilitate real time 2D and 3D ultrasound image acquisition, facilitate surgical tasks such as scanning solid organs for tumors, and aid in guiding the placement of biopsy needles and ablation probes.  Preliminary surgeon evaluations for kidney and liver cancer procedures have been performed on the system at ISI, as well as procedures for gynecological applications. More formal evaluations comparing robotic system aided surgeon performance to freehand ultrasound are planned for the summer.  Concurrently, we have begun to evaluate the potential for using the probe in advanced ultrasound imaging techniques such as ultrasound elastography using a copy of the probe at JHU.

*Acknowledgements: CISST ERC, ISI & JHU Internal funds; NIH Phase II STTR 2 R42 RR019159*

*This will allow...real time 2D and 3D ultrasound image acquisition, facilitate surgical tasks such as scanning solid organs for tumors, and aid in guiding the placement of biopsy needles and ablation probes.*

# *Web Tables*

| Table 1: Quantifiable Outputs | | | | | | | |
|---|---|---|---|---|---|---|---|
| Outputs | Early Cumulative Total* | Sep 01, 2004 - Aug 31, 2005 | Sep 01, 2005 - Aug 31, 2006 | Sep 01, 2006 - Aug 31, 2007 | Sep 01, 2007 - Aug 31, 2008 | Sep 01, 2008 - Aug 31, 2009 | All Years |
| **Publications That Result from Center Support** | | | | | | | |
| In Peer-Reviewed Technical Journals | 35 | 11 | 19 | 16 | 24 | 9 | **114** |
| In Peer-Reviewed Conference Proceedings | 124 | 37 | 70 | 53 | 92 | 28 | **404** |
| In Trade Journals | 0 | 0 | 0 | 0 | 2 | 0 | **2** |
| With Multiple Authors: | 155 | 48 | 80 | 63 | 144 | 43 | **533** |
| Co-authored with ERC Students | 87 | 37 | 77 | 51 | 107 | 26 | **385** |
| Co-authored with Industry | 11 | 2 | 6 | 3 | 3 | 1 | **26** |
| With Authors from Multiple Engineering Disciplines | 31 | 15 | 37 | 28 | 90 | 5 | **206** |
| With Authors from Both Engineering and non-Engineering Fields | 53 | 27 | 24 | 20 | 51 | 15 | **190** |
| with authors from multiple institutions | 2 | 3 | 16 | 24 | 42 | 27 | **114** |
| **Publications That Result from Associated Projects in the Strategic Plan** | | | | | | | |
| In Peer-Reviewed Technical Journals | 0 | 2 | 0 | 0 | 0 | 5 | **7** |
| In Peer-Reviewed Conference Proceedings | 0 | 7 | 0 | 0 | 1 | 11 | **19** |
| **Publications Resulting From Sponsored Projects** | | | | | | | |
| In Peer Reviewed Technical Journals | 0 | 0 | N/A | N/A | 0 | 0 | **0** |
| In Peer Reviewed Conference Proceedings | 0 | 0 | N/A | N/A | 0 | 0 | **0** |
| **Participating Industrial and Practitioner Organizations** | | | | | | | |
| Members | 42 | 10 | 6 | 7 | 7 | 7 | **79** ** |
| Affiliates | 0 | 2 | 8 | 8 | 8 | 8 | **34** ** |
| Contributing Organizations | 5 | 7 | 5 | 2 | 3 | 8 | **30** ** |
| **ERC Technology Transfer** | | | | | | | |
| Inventions Disclosed (submitted to agencies by | 20 | 3 | 3 | 16 | 3 | 0 | **45** |
| Patent Applications Filed | 12 | 5 | 2 | 12 | 8 | 0 | **39** |
| Patents Awarded | 4 | 0 | 0 | 0 | 2 | 0 | **6** |
| Licenses Issued | 4 | 1 | 0 | 1 | 0 | 0 | **6** |
| Spin-off Companies Started | 0 | 0 | 0 | 0 | 0 | 0 | **0** |
| Estimated Number of Spin-off Company Employees | 5 | 0 | 0 | 0 | 0 | 0 | **5** |
| Building Codes Impacts | 0 | 0 | 0 | 0 | 0 | 0 | **0** |
| Technology Standards Impacts | 0 | 0 | 0 | 0 | 0 | 0 | **0** |
| New Surgical and other Medical Procedures Adopted | 0 | 0 | 0 | 0 | 0 | 0 | **0** |
| **Degrees to ERC Students** | | | | | | | |
| Bachelor's Degrees Granted | 40 | 6 | 2 | 4 | 2 | 0 | **54** |
| Master's Degrees Granted | 10 | 12 | 6 | 9 | 1 | 13 | **51** |
| Doctoral Degrees Granted | 9 | 8 | 5 | 10 | 6 | 8 | **46** |
| **ERC Graduates Hired by** | | | | | | | |
| Industry: | 9 | 1 | 3 | 5 | 2 | 2 | **22** |
| ERC Member Firms | 6 | 0 | 0 | 3 | 1 | 1 | **11** |
| Other U.S. Firms | 3 | 1 | 3 | 2 | 1 | 1 | **11** |
| Other Foreign Firms | 0 | 0 | 0 | 0 | 0 | 0 | **0** |
| Government | 2 | 2 | 0 | 1 | 0 | 0 | **5** |
| Academic Institutions | 3 | 2 | 7 | 5 | 2 | 3 | **22** |
| Other | 0 | 2 | 0 | 0 | 2 | 0 | **4** |
| Undecided/Still Looking/Unknown | 0 | 8 | 3 | 0 | 3 | 3 | **17** |
| **ERC Influence on Curriculum** | | | | | | | |
| New courses based on ERC research that have been approved by the curriculum committee and are currently offered**** | 2 | 0 | 1 | 5 | 3 | 0 | **11** |
| Currently offered, on-going courses with ERC content | 0 | 0 | 0 | 0 | 0 | 4 | **4** |
| New Textbook Chapter Based on ERC Research | 0 | 0 | 0 | 3 | 3 | 0 | **6** |
| New Textbooks Based on ERC Research | 0 | 1 | 0 | 5 | 0 | 0 | **6** |
| Free-Standing Course Modules or Instructional CDs | 0 | 0 | 0 | 0 | 0 | 0 | **0** |
| New full degree programs based on ERC research | 0 | 0 | 0 | 0 | 0 | 0 | **0** |
| New degree minors or minor emphases based on ERC | 2 | 1 | 0 | 0 | 1 | 0 | **4** |
| New certificate programs based on ERC research | 0 | 0 | 0 | 0 | 0 | 0 | **0** |
| **Active Information Dissemination/Educational Outreach** | | | | | | | |
| Workshops, Short Courses, and Webinars *** | 3 | 1 | 5 | 3 | 5 | 1 | **18** |
| Number of participants that attended activity | N/A | N/A | N/A | N/A | N/A | 51 | **51** |
| Seminars, Colloquia, Invited Talks, etc. | 131 | 101 | 88 | 109 | 53 | 29 | **511** |
| ERC Sponsored Educational Outreach Events for K-12 | 0 | 0 | 0 | 0 | 0 | 1 | **1** |
| Number of students that attended activity | 0 | 0 | 0 | 0 | 0 | 60 | **60** |
| Number of teachers that attended activity | 0 | 0 | 0 | 0 | 0 | 0 | **0** |
| ERC Sponsored Educational Outreach Events for Community College or Undergraduate students | 0 | 0 | 0 | 0 | 0 | 0 | **0** |
| Number of students that attended activity | 0 | 0 | 0 | 0 | 0 | 0 | **0** |
| Number of faculty that attended activity | 0 | 0 | 0 | 0 | 0 | 0 | **0** |
| **Personnel Exchanges** | | | | | | | |
| Student Internships in Industry | 3 | 1 | 1 | 2 | 3 | 3 | **13** |
| Faculty Working at Member Firm | 3 | 0 | 0 | 0 | 0 | 0 | **3** |
| Member Firm Personnel Working at ERC | 0 | 0 | 0 | 0 | 0 | 0 | **0** |

* - For Centers in operation for more than five years.
** - Cumulative count of Individual Firms/Organizations may not equal the sum across all years.

*** - For years prior to 2009, the values include ‘Workshops and short courses to industry’ and ‘Workshops and short courses to non-industry groups’

**** - New courses currently offered and approved by the curriculum committee are only counted in the first year that they are offered so there is no multiple counting of these courses.

Fig. A27

| Table 1a: Average Metrics Benchmarked Against All Active ERC's | | | | | |
|---|---|---|---|---|---|
| Metric | Average All Active ERC's FY2008 | Average Bioengineering Sector FY2007 | Average Bioengineering Sector FY2008 | Average for Class of 1998 - FY 2008 | Center for Computer-Integrated Surgical Systems and Technology Total |
| | (20 ERC's) | (7 ERC's) | (7 ERC's) | (4 ERC's) | FY2009 |
| **Industrial Member Firms** | 16 | 8 | 8 | 31 | 7 |
| Small | 34% | 35% | 42% | 17% | 57% |
| Medium | 15% | 9% | 13% | 21% | 14% |
| Large | 50% | 56% | 45% | 62% | 29% |
| **Non-Industrial Member Firms** | 1 | 0 | 0 | 1 | 0 |
| **Affiliate Organizations** | 1 | 3 | 3 | 6 | 8 |
| **Contributing Organizations** | 3 | 8 | 8 | 13 | 8 |
| **Industrial Membership Fees Received** | $200,636.00 | $83,500.00 | $134,176.00 | $194,559.00 | $0.00 |
| | | | | | |
| **Sources of Support [1]** | $4,161,451.00 | $6,272,084.00 | $3,473,709.00 | $5,737,375.00 | $924,879.00 |
| NSF | 61% | 44% | 63% | 43% | 14% |
| Industry | 12% | 4% | 6% | 21% | 30% |
| Other Federal | 2% | 3% | 4% | 5% | 30% |
| Academic | 21% | 41% | 23% | 22% | 14% |
| State | 4% | 3% | 4% | 9% | 0% |
| Other | 0% | 5% | 0% | 0% | 12% |
| **Associated Project Support** | $2,036,284.00 | $1,219,228.00 | $1,612,870.00 | $1,008,904.00 | $2,482,781.00 |
| **ERC Personnel & Educational Participants[2, 3]** | 971 | 921 | 762 | 451 | 96 |
| Leadership Team [7] | 9 | 12 | 8 | 9 | 8 |
| Faculty [2,4] | 33 | 38 | 29 | 44 | 33 |
| Graduate Students [2] | 105 | 167 | 84 | 116 | 35 |
| Undergraduate Students [2] | 137 | 347 | 160 | 42 | 5 |
| REU Students | 11 | 21 | 7 | 9 | 0 |
| K-12 Teachers [3] | 50 | 24 | 19 | 5 | 0 |
| K-12 Students [3] | 568 | 205 | 389 | 74 | 60 |
| Faculty that attended ERC Sponsored Educational Outreach Events [3] | 0 | 0 | 0 | 0 | 0 |
| Community College or Undergraduate students that attended ERC Sponsored Educational Outreach Events [3] | 0 | 0 | 0 | 0 | 0 |
| % Women [5,6] | 32% | 43% | 39% | 33% | 18% |
| % Underrepresented Racial Minorities [5,6] | 14% | 12% | 13% | 14% | 10% |
| % Hispanic [5,6] | 11% | 5% | 5% | 8% | 1% |
| | | | | | |
| **Publications** | **Average** | **Average** | **Average** | **Average** | **Total** |
| In Peer Reviewed Technical Journals | 28 | 15 | 23 | 40 | 9 |
| In Peer Reviewed Conference Proceedings | 35 | 15 | 25 | 38 | 28 |
| Multiple Authors: Co-Authored With ERC Students | 45 | 18 | 36 | 71 | 26 |
| Multiple Authors: Co-Authored With Industry | 4 | 1 | 3 | 3 | 1 |
| | | | | | |
| **Intellectual Property** | **Average** | **Average** | **Average** | **Average** | **Total** |
| Invention Disclosures | 5 | 10 | 9 | 11 | 0 |
| Patent Applications | 5 | 12 | 10 | 8 | 0 |
| Patents Awarded | 1 | 2 | 2 | 3 | 0 |
| Licenses (patents, software) | 7 | 1 | 1 | 34 | 0 |
| **Education and Outreach Outputs** | **Average** | **Average** | **Average** | **Average** | **Total** |
| New Courses Developed | 7 | 13 | 6 | 7 | 0 |
| Currently offered, on-going courses with ERC conte | 0 | 0 | 0 | 0 | 4 |
| New Full Degree Programs | 0 | 0 | 0 | 0 | 0 |
| New degree minors or minor emphases | 0 | 0 | 0 | 0 | 0 |
| New certificate programs based on ERC research | 0 | 0 | 0 | 0 | 0 |

1 Includes new support (unrestricted cash, restricted cash, and in-kind donations) from table 9 only. Residual funds carried over from previous years are not included in benchmarking figures.
2 Includes total ERC Personnel from table 7.
3 Includes participant values from Table 1 Quantifiable Outputs.
4 Includes Directors, Education Program Leaders, Thrust Leaders, Senior Faculty, Junior Faculty, and Visiting Faculty from table 7.
5 These data do not include K-12 Student or Teacher Participants in the percentage calculations. Demographic data are not collected for K-12 Student or Teacher Participants. We only collect the total number of K-12 Student and Teacher Participants.
6 The percentage calculations are based on the following categories of Personnel only:
Faculty, Graduate Students, Undergraduate Students, REU Students, Directors, Thrust Leaders,
Research Thrust Management & Strategic Planning, Administrative Director, and Industrial Liasion Officer.
7 Includes Directors, Thrust Leaders, Education Program Leaders, Research Thrust Management & Strategic Planning, Administrative Director, and Industrial Liasion Officer.

Fig. A28

**Table 2: Research Program Organization and Effort**

| Cluster/Thrust | Engineering and Systems Infrastructure | Cluster/Thrust Leader | Peter Kazanzides | | | |
|---|---|---|---|---|---|---|
| Personnel: 0 Faculty Members, 0 Undergraduates, 0 Graduate Students, 0 Post Docs, 0 Other Personnel | | | | | | |
| **Project** | **Leader** | **Investigators (name, department, academic institution)** | **Disciplines Involved** | **Number of Students and Post Docs** | **Current Award Year Budget** | **Proposed Award Year Budget** |
| **Center-controlled Projects** | | | | | | |
| Intelligent Management | Kazanzides | Peter Kazanzides<br>Computer Science<br>Johns Hopkins University | Computer science | U=0<br>G=0<br>P=0 | $126,268 | $0 |
| Surgical Assistant Workstation | Taylor | Peter Kazanzides<br>Computer Science<br>Johns Hopkins University<br>Rajesh Kumar<br>Computer Science<br>Johns Hopkins<br>Russell Taylor<br>Engineering Research Center<br>Johns Hopkins University<br>Simon DiMaio<br>Radiology<br>Intuitive Surgical | Computer science, Health/medical technologies | U=0<br>G=0<br>P=0 | $90,873 | $0 |
| Task 0.1A: 2033/0111 JHU Infrastructure | Peter Kazanzides | Gabor Fichtinger<br>Engineering Research Center<br>Johns Hopkins University<br>Iulian Iordachita<br>CISST<br>JHu<br>Peter Kazanzides<br>Computer Science<br>Johns Hopkins University<br>Russell Taylor<br>Engineering Research Center<br>Johns Hopkins University | Computer science, Mechanical engineering | U=0<br>G=0<br>P=0 | $3,203 | $0 |
| Task 0.1E: 2034 MSU Infrastructure | Peter Kazanzides | LeeRoy Bronner<br>Computer Science<br>Morgan State University<br>Peter Kazanzides<br>Computer Science<br>Johns Hopkins University | Computer science | U=0<br>G=0<br>P=0 | $0 | $0 |
| | | | | **Subtotal** | **$220,344** | **$0** |
| **Sponsored Projects** | | | | | | |
| Bone Augmentation | Kazanzides | Mehran Armand<br>APL<br>Peter Kazanzides<br>Computer Science<br>Johns Hopkins University<br>Russell Taylor<br>Engineering Research Center<br>Johns Hopkins University | Computer science, Bioengineering and biomedical engineering | U=0<br>G=0<br>P=0 | $40,113 | $0 |
| SARRP Small animal radiation (NIH) | Peter Kazanzides | Iulian Iordachita<br>CISST<br>JHU<br>John Wong<br>Radiology & Oncology<br>Johns Hopkins Medical Institution<br>Peter Kazanzides<br>Computer Science<br>Johns Hopkins University | Computer science, Mechanical engineering, Medicine (e.g., dentistry, optometry, osteopathic, veterinary) | U=0<br>G=0<br>P=0 | $25,053 | $0 |
| SGER Subcontract from Morgan State University | Peter Kazanzides | LeeRoy Bronner<br>Computer Science<br>Morgan State University<br>Peter Kazanzides<br>Computer Science<br>Johns Hopkins University | Computer science | U=0<br>G=0<br>P=0 | $0 | $0 |
| Small animal radiation research program | John Wong | John Wong<br>Radiology & Oncology<br>Johns Hopkins Medical Institution<br>Peter Kazanzides<br>Computer Science<br>Johns Hopkins University | Computer science, Medicine (e.g., dentistry, optometry, osteopathic, veterinary) | U=0<br>G=0<br>P=0 | $0 | $0 |
| | | | | **Subtotal** | **$65,166** | **$0** |
| **Associated Projects** | | | | | | |
| CA Orthopaedic Surgery | Taylor | Mehran Armand<br>APL<br>Peter Kazanzides<br>Computer Science<br>Johns Hopkins University<br>Russell Taylor<br>Engineering Research Center<br>Johns Hopkins University | Computer science, Bioengineering and biomedical engineering | U=0<br>G=0<br>P=0 | $22,706 | $0 |
| MRI:Development of Infrastructure for Integrated Sending, Modeling and Manipulation with Robotic and Human Machine systems | Allison Okamura | Allison Okamura<br>Mechanical Engineering<br>Johns Hopkins University<br>Iulian Iordachita<br>CISST<br>JHU<br>Peter Kazanzides<br>Computer Science<br>Johns Hopkins University | Computer science, Mechanical engineering | U=0<br>G=0<br>P=0 | $500,191 | $0 |
| | | | | **Subtotal** | **$522,897** | **$0** |
| **Grand Total for Engineering and Systems Infrastructure** | | | | | **$808,407** | **$0** |

Fig. A29: Personnel data not reported for no-cost extension.

| **LEGEND:** |
|---|
| **U - Number of Undergraduate Students** |
| **G - Number of Graduate Students** |
| **P - Number of Postdoctoral Fellows** |

Web Tables

| Cluster/Thrust | Surgical Assistants | Cluster/Thrust Leader | Allison Okamura | | | |
|---|---|---|---|---|---|---|
| Personnel: 0 Faculty Members, 0 Undergraduates, 0 Graduate Students, 0 Post Docs, 0 Other Personnel | | | | | | |
| **Project** | **Leader** | **Investigators (name, department, academic institution)** | **Disciplines Involved** | **Number of Students and Post Docs** | **Current Award Year Budget** | **Proposed Award Year Budget** |
| **Center-controlled Projects** | | | | | | |
| Task 1.1 Systems and Applications (all) | Russell Taylor | Allison Okamura<br>Mechanical Engineering<br>Johns Hopkins University<br>Cam Riviere<br>Robotics Institute<br>Carnegie Mellon University<br>Darius Burschka<br>Computer Science<br>Johns Hopkins University<br>George Jallo<br>Neurology<br>JHMI<br>Greg Hager<br>Computer Science<br>Johns Hopkins University<br>Iulian Iordachita<br>CISST<br>JHu<br>James Handa, MD<br>Ophthalmology Department<br>Johns Hopkins Medical Institution<br>Jin Kang<br>ECE<br>JHu<br>John Wong<br>Radiology & Oncology<br>Johns Hopkins Medical Institution<br>Li-Ming Su<br>Urology<br>University of Florida<br>Louis Whitcomb<br>Mechanical Engineering<br>Johns Hopkins University<br>Michael Marohn, MD<br>Surgery<br>Johns Hopkins Medical Institution | Computer science, Computer/systems engineering, Electrical, electronics, communications engineering, Engineering sciences, mechanics, physics, Mechanical engineering, Medicine (e.g., dentistry, optometry, osteopathic, veterinary) | U=0<br>G=0<br>P=0 | $156,325 | $0 |
| | | | | **Subtotal** | **$156,325** | **$0** |
| **Sponsored Projects** | | | | | | |
| Task 1.1 (2126)Dexterous, compact telesurgical robot for throat & airway | Russell Taylor | Nabil Simaan<br>Mechanical Engineering<br>Columbia University<br>Russell Taylor<br>Engineering Research Center<br>Johns Hopkins University | Computer science | U=0<br>G=0<br>P=0 | $0 | $0 |
| Task 1.1 (2156) Robot Assisted LAPUS Ph II STTR | Russell Taylor | Emad Boctor<br>Radiology<br>JHMI<br>Greg Hager<br>Computer Science<br>Johns Hopkins University<br>Michael Choti, MD<br>Oncology Center<br>Johns Hopkins Medical Institution<br>Michael Marohn, MD<br>Surgery<br>Johns Hopkins Medical Institution<br>Russell Taylor<br>Engineering Research Center<br>Johns Hopkins University<br>Simon DiMaio<br>Radiology<br>Intuitive Surgical | Computer science, Health/medical technologies, Medicine (e.g., dentistry, optometry, osteopathic, veterinary) | U=0<br>G=0<br>P=0 | $49,656 | $0 |
| Task 1.3 (2135) Structural induction for manipulation & interactive devices | Gregory D. Hager | Greg Hager<br>Computer Science<br>Johns Hopkins University<br>Izhak Shafran<br>Electrical and Computer Engineering<br>Johns Hopkins University<br>Sanjeev Khudanpur<br>Electrical and Computer Engineering<br>Johns Hopkins University | Computer science | U=0<br>G=0<br>P=0 | $51,742 | $0 |
| Task 1.3 (2137) Direct video-CT registration | Gregory D. Hager | Greg Hager<br>Computer Science<br>Johns Hopkins University<br>Masaru Ishii<br>Otolaryngology<br>JHMI<br>Russell Taylor<br>Engineering Research Center<br>Johns Hopkins University | Computer science, Medicine (e.g., dentistry, optometry, osteopathic, veterinary) | U=0<br>G=0<br>P=0 | $4,731 | $0 |
| | | | | **Subtotal** | **$106,129** | **$0** |
| **Associated Projects** | | | | | | |

Fig. A29 (continued)

| | | | | | | |
|---|---|---|---|---|---|---|
| 101370.Active Cannulas for BioSensing and Surgery | Noah Cowan | Allison Okamura<br>Mechanical Engineering<br>Johns Hopkins University<br>Noah Cowan<br>Mechanical Engineering<br>Johns Hopkins University | Mechanical engineering | U=0<br>G=0<br>P=0 | $0 | $0 |
| 101674.Toward Quantitative Disease Assessment | Gregory D. Hager | Rajesh Kumar<br>Computer Science<br>Johns Hopkins<br>Themistocles Dassopoulos, M.D.<br>Gastroenterology<br>Washington University | Computer science, Medicine (e.g., dentistry, optometry, osteopathic, veterinary) | U=0<br>G=0<br>P=0 | $107,239 | $0 |
| 101781.Manipulating and Perceiving Simultaneously | Gregory D. Hager | Allison Okamura<br>Mechanical Engineering<br>Johns Hopkins University<br>Noah Cowan<br>Mechanical Engineering<br>Johns Hopkins University | Mechanical engineering | U=0<br>G=0<br>P=0 | $75,115 | $0 |
| 101979. Active Motion Compensation | Gregory D. Hager | Greg Hager<br>Computer Science<br>Johns Hopkins University | Computer science | U=0<br>G=0<br>P=0 | $694 | $0 |
| 103394.Quantitative Endoscopic Measurement of Anatomy | Hager | Greg Hager<br>Computer Science<br>Johns Hopkins University | Computer science | U=0<br>G=0<br>P=0 | $79,748 | $0 |
| 103480.Intelligent Image Feature Matching | Hager | Greg Hager<br>Computer Science<br>Johns Hopkins University | Computer science | U=0<br>G=0<br>P=0 | $14,113 | $0 |
| 103547.Microsurgical Assistant System | Taylor | Greg Hager<br>Computer Science<br>Johns Hopkins University<br>Iulian Iordachita<br>CISST<br>JHu<br>James Handa, MD<br>Ophthalmology Department<br>Johns Hopkins Medical Institution<br>Peter Gehlbach<br>Ophthalmology<br>JHMI<br>Peter Kazanzides<br>Computer Science<br>Johns Hopkins University<br>Rajesh Kumar<br>Computer Science<br>Johns Hopkins<br>Russell Taylor<br>Engineering Research Center<br>Johns Hopkins University | Computer science, Mechanical engineering, Medicine (e.g., dentistry, optometry, osteopathic, veterinary) | U=0<br>G=0<br>P=0 | $505,596 | $0 |
| 104298.Navigation and Visualization | Kumar | Rajesh Kumar<br>Computer Science<br>Johns Hopkins | Computer science | U=0<br>G=0<br>P=0 | $6,261 | $0 |
| 105169.LARS loan | Taylor | Russell Taylor<br>Engineering Research Center<br>Johns Hopkins University | Computer science | U=0<br>G=0<br>P=0 | $5,654 | $0 |
| Task 1.2 R21-Image Guidance for Active Handheld Microsurgical Tools | Cam Riviere | Cam Riviere<br>Robotics Institute<br>Carnegie Mellon University<br>Greg Hager<br>Computer Science<br>Johns Hopkins University | Computer science, Computer/systems engineering | U=0<br>G=0<br>P=0 | $34,422 | $0 |
| Ultrasound based guidance | Hassan Rivaz | Emad Boctor<br>Radiology<br>JHMI | Computer science | U=0<br>G=0<br>P=0 | $13,856 | $0 |
| | | | | **Subtotal** | **$842,698** | **$0** |
| **Grand Total for Surgical Assistants** | | | | | **$1,105,152** | **$0** |

Fig. A29 (continued)

| |
|---|
| **LEGEND:** |
| **U - Number of Undergraduate Students** |
| **G - Number of Graduate Students** |
| **P - Number of Postdoctoral Fellows** |

Web Tables

| Cluster/Thrust | Surgical CAD/CAM | Cluster/Thrust Leader | Gabor Fichtinger | | | |
|---|---|---|---|---|---|---|
| Personnel: 0 Faculty Members, 0 Undergraduates, 0 Graduate Students, 0 Post Docs, 0 Other Personnel | | | | | | |
| **Project** | **Leader** | **Investigators (name, department, academic institution)** | **Disciplines Involved** | **Number of Students and Post Docs** | **Current Award Year Budget** | **Proposed Award Year Budget** |
| **Center-controlled Projects** | | | | | | |
| Task 2.1: Systems and Applications (all) | Gabor Fichtinger and Ron Kikinis | Allison Okamura<br>Mechanical Engineering<br>Johns Hopkins University<br>Gabor Fichtinger<br>Engineering Research Center<br>Johns Hopkins University<br>Greg Chirikjian<br>Mechanical Engineering<br>Johns Hopkins University<br>Greg Hager<br>Computer Science<br>Johns Hopkins University<br>Louis Whitcomb<br>Mechanical Engineering<br>Johns Hopkins University<br>Peter Kazanzides<br>Computer Science<br>Johns Hopkins University<br>Ron Kikinis<br>Department of Radiology<br>Brigham & Women's Hospital<br>Russell Taylor<br>Engineering Research Center<br>Johns Hopkins University<br>Simon DiMaio<br>Radiology<br>Intuitive Surgical | Computer science, Mechanical engineering, Health/medical technologies | U=0<br>G=0<br>P=0 | $8,289 | $0 |
| Task 2.2: Image Analysis and Data Fusion (all) | Eric Grimson | Carl-Freidrick Westin<br>Radiology<br>Brigham Womens hospital<br>Eric Grimson<br>AI Lab<br>Massachusetts Insititute of Technology<br>Gabor Fichtinger<br>Engineering Research Center<br>Johns Hopkins University<br>Jerry Prince<br>Electrical and Computer Engineering<br>Johns Hopkins University<br>Polina Golland<br>Computer Science<br>MIT<br>William Wells<br>Associate Professor<br>MIT and Brigham and Women's Hospital | Computer science, Electrical, electronics, communications engineering, Health/medical technologies | U=0<br>G=0<br>P=0 | $3,843 | $0 |
| Task 2.4: Percutaneous Delivery Devices (all) | Gabor Fichtinger | Allison Okamura<br>Mechanical Engineering<br>Johns Hopkins University<br>Gabor Fichtinger<br>Engineering Research Center<br>Johns Hopkins University<br>Greg Chirikjian<br>Mechanical Engineering<br>Johns Hopkins University<br>Louis Whitcomb<br>Mechanical Engineering<br>Johns Hopkins University<br>Noah Cowan<br>Mechanical Engineering<br>Johns Hopkins University<br>Ron Kikinis<br>Department of Radiology<br>Brigham & Women's Hospital<br>Russell Taylor<br>Engineering Research Center<br>Johns Hopkins University<br>Simon DiMaio<br>Radiology<br>Intuitive Surgical | Computer science, Mechanical engineering, Health/medical technologies | U=0<br>G=0<br>P=0 | $48,748 | $0 |
| Task 2.5: 2D-3D Registration & Reconstruction (all) | Jerry Prince | Jerry Prince<br>Electrical and Computer Engineering<br>Johns Hopkins University<br>Russell Taylor<br>Engineering Research Center<br>Johns Hopkins University<br>William Wells<br>Associate Professor<br>MIT and Brigham and Women's Hospital | Computer science, Electrical, electronics, communications engineering | U=0<br>G=0<br>P=0 | $19,888 | $0 |
| | | | | **Subtotal** | **$80,768** | **$0** |

Fig. A29 (continued)

| | | | | | | |
|---|---|---|---|---|---|---|
| **Sponsored Projects** | | | | | | |
| 100124.Robotic Prostate Biopsy | Gabor Fichtinger | Gabor Fichtinger<br>Engineering Research Center<br>Johns Hopkins University | Computer science | U=0<br>G=0<br>P=0 | $3,976 | $0 |
| 903270. Transrectal prostate therapy robot | Louis Whitcomb | Gabor Fichtinger<br>Engineering Research Center<br>Johns Hopkins University<br>Iulian Iordachita<br>CISST<br>JHu<br>Louis Whitcomb<br>Mechanical Engineering<br>Johns Hopkins University | Computer science, Mechanical engineering | U=0<br>G=0<br>P=0 | $111,431 | $0 |
| 903271.Prior Knowledge in Reconstruction | Jerry Prince | Jerry Prince<br>Electrical and Computer Engineering<br>Johns Hopkins University<br>Russell Taylor<br>Engineering Research Center<br>Johns Hopkins University | Computer science, Electrical, electronics, communications engineering | U=0<br>G=0<br>P=0 | $2,865 | $0 |
| 905214.Intra-operative dosimetry in prostate brachytherapy | Gabor Fichtinger | Gabor Fichtinger<br>Engineering Research Center<br>Johns Hopkins University | Computer science | U=0<br>G=0<br>P=0 | $0 | $0 |
| 906838.Robot assisted prostate intervention w/ MRI | Jerry Prince | Gabor Fichtinger<br>Engineering Research Center<br>Johns Hopkins University<br>Iulian Iordachita<br>CISST<br>JHu<br>Jerry Prince<br>Electrical and Computer Engineering<br>Johns Hopkins University<br>Russell Taylor<br>Engineering Research Center<br>Johns Hopkins University | Computer science, Electrical, electronics, communications engineering, Mechanical engineering | U=0<br>G=0<br>P=0 | $77,791 | $0 |
| 907744.Ultrasound elasticity image guidance for external beam partial breast irradiation | Gabor Fichtinger | Emad Boctor<br>Radiology<br>JHMI<br>Gabor Fichtinger<br>Engineering Research Center<br>Johns Hopkins University<br>Greg Hager<br>Computer Science<br>Johns Hopkins University | Computer science | U=0<br>G=0<br>P=0 | $24,112 | $0 |
| CA Orthopaedic Surgery | Mehran Armand | Mehran Armand<br>APL<br>Russell Taylor<br>Engineering Research Center<br>Johns Hopkins University | Computer science, Bioengineering and biomedical engineering | U=0<br>G=0<br>P=0 | $88,035 | $0 |
| Needle Placement | Gabor Fichtinger | Danny Song, MD<br>Radiation Oncology<br>Johns Hopkins Medical Institution<br>Elwood Armour<br>Radiation Oncology<br>Johns Hopkins Medical Institution<br>Gabor Fichtinger<br>Engineering Research Center<br>Johns Hopkins University<br>Theodore DeWeese, MD<br>Radiation Oncology<br>Johns Hopkins Medical Institution | Biochemistry and biophysics, Computer science, Medicine (e.g., dentistry, optometry, osteopathic, veterinary) | U=0<br>G=0<br>P=0 | $54 | $0 |
| Task 2.1.Z-2 SBIR Phase II RUF | Jerry Prince | Danny Song, MD<br>Radiation Oncology<br>Johns Hopkins Medical Institution<br>Elwood Armour<br>Radiation Oncology<br>Johns Hopkins Medical Institution<br>Gabor Fichtinger<br>Engineering Research Center<br>Johns Hopkins University<br>Jerry Prince<br>Electrical and Computer Engineering<br>Johns Hopkins University<br>Theodore DeWeese, MD<br>Radiation Oncology<br>Johns Hopkins Medical Institution | Biochemistry and biophysics, Computer science, Electrical, electronics, communications engineering, Medicine (e.g., dentistry, optometry, osteopathic, veterinary) | U=0<br>G=0<br>P=0 | $80,690 | $0 |
| Task 2.3 (2130) 3D volumetric models of the proximal femur | Russell Taylor | Russell Taylor<br>Engineering Research Center<br>Johns Hopkins University | Computer science | U=0<br>G=0<br>P=0 | $38,383 | $0 |
| | | | | **Subtotal** | **$427,337** | **$0** |
| **Associated Projects** | | | | | | |
| 101668.Steering Flexible Needles in Soft Tissue | Allison Okamura | Danny Song, MD<br>Radiation Oncology<br>Johns Hopkins Medical Institution<br>Greg Chirikjian<br>Mechanical Engineering<br>Johns Hopkins University<br>Noah Cowan<br>Mechanical Engineering<br>Johns Hopkins University | Mechanical engineering, Medicine (e.g., dentistry, optometry, osteopathic, veterinary) | U=0<br>G=0<br>P=0 | $461,999 | $0 |
| 101757. Image Overlay for MR | Iulian Iordachita | Iulian Iordachita<br>CISST<br>JHu | Mechanical engineering | U=0<br>G=0<br>P=0 | $2,247 | $0 |
| | | | | **Subtotal** | **$464,246** | **$0** |
| **Grand Total for Surgical CAD/CAM** | | | | | **$972,351** | **$0** |

| **Table 2: Research Program Organization and Effort Totals** | **Current Award Year Budget** | **Proposed Award Year Budget** |
|---|---|---|
| **Total, Center-controlled Projects** | **$457,437** | **$0** |
| **Total, Sponsored Projects** | **$598,632** | **$0** |
| **Total, Associated Projects** | **$1,829,841** | **$0** |
| **Grand Total, All Projects** | **$2,885,910** | **$0** |

| **LEGEND:** |
|---|
| **U - Number of Undergraduate Students** |
| **G - Number of Graduate Students** |
| **P - Number of Postdoctoral Fellows** |

Fig. A29 (continued)

# *Web Tables*

## Table 2a: Disciplines Involved in Research Projects

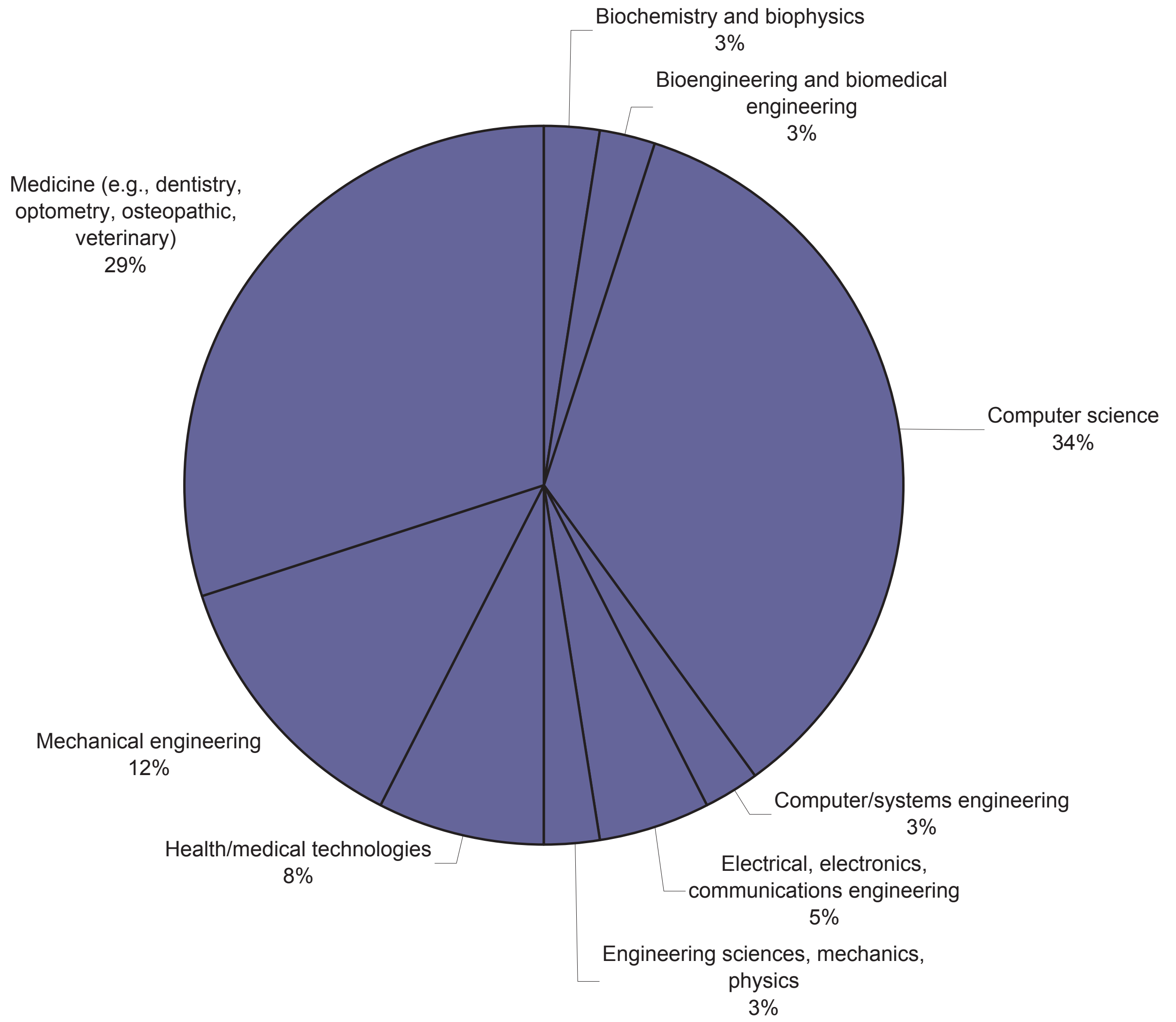

Fig. A30

| Table 3a: Educational Impact | | | | | | | | | | | | | | |
|---|---|---|---|---|---|---|---|---|---|---|---|---|---|---|
| | Total from Table 1: Quantifiable Outputs | With engineered systems focus | | With multidisciplinary content | | Team taught by faculty from more than 1 department | | Undergraduate level | | Graduate level | | Used at more than 1 ERC institution | | Cumulative Total for All Years |
| | | Current Year | % | Current Year | % | Current Year | % | Current Year | % | Current Year | % | Current Year | % | |
| New courses currently offered* | 0 | 0 | 0 | 0 | 0 | 0 | 0 | 0 | 0 | 0 | 0 | 0 | 0 | 11 |
| Currently offered, ongoing courses with ERC content** | 4 | 4 | 100 | 2 | 50 | 1 | 25 | 4 | 100 | 3 | 75 | 0 | 0 | 4 |
| Workshops, Short Courses, and Webinars | 1 | 0 | 0 | 1 | 100 | 1 | 100 | 0 | 0 | 1 | 100 | 0 | 0 | 1 |
| New textbooks based on ERC research | 0 | 0 | 0 | 0 | 0 | 0 | 0 | 0 | 0 | 0 | 0 | 0 | 0 | 6 |

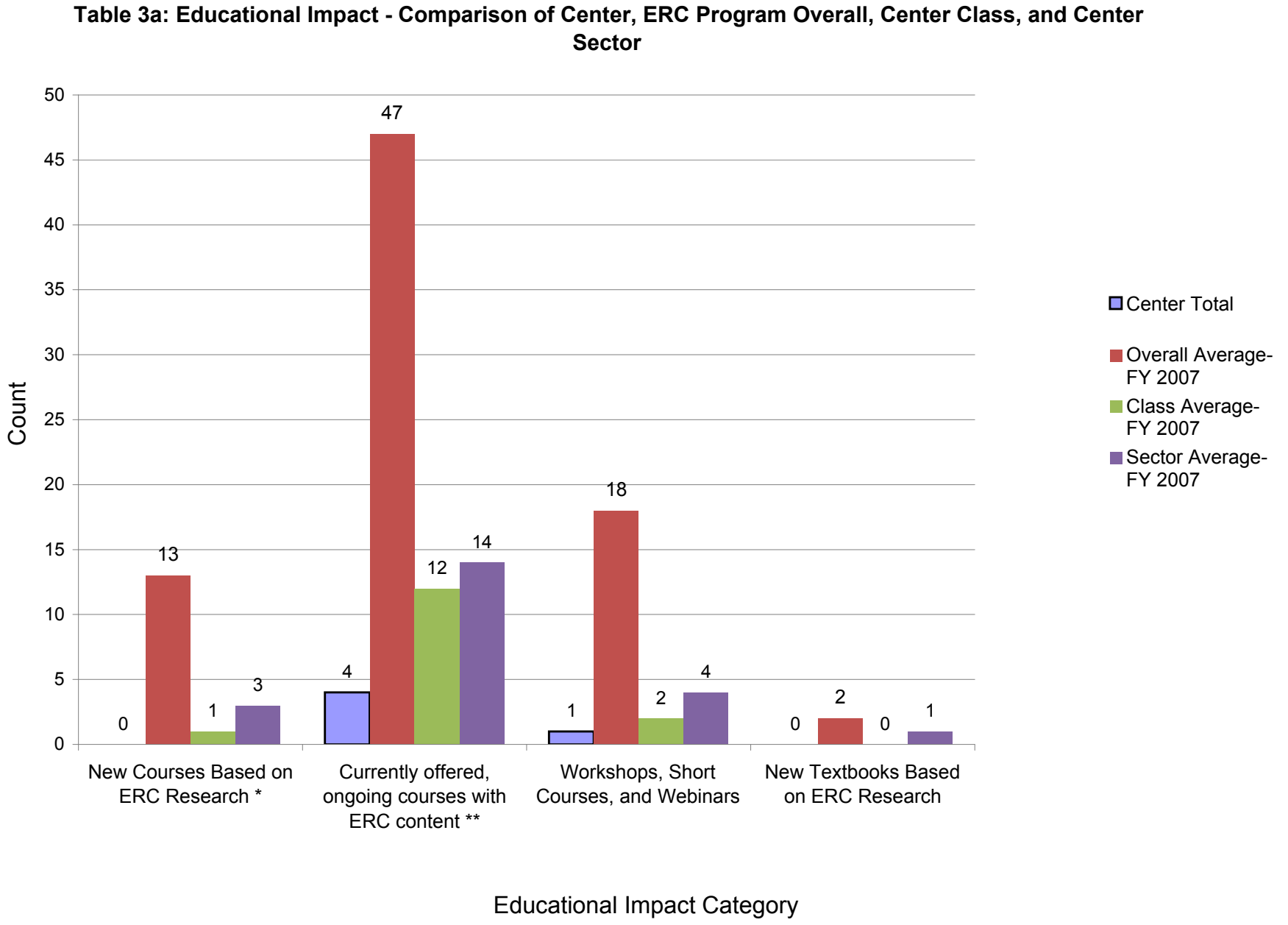

* - New courses currently offered and approved by the curriculum committee are only counted in the first year that they are offered so there is no multiple counting of these courses.
** - The cumulative totals for "Currently offered, ongoing courses with ERC content" may count the same course more than once. This is due to the fact that a single course can be modified in multiple years and therefore will be included in the cumulative total multiple times.

Fig. A31 (above)

| Table 3b - Ratio of Graduates to Undergraduates* | | | | | | | |
|---|---|---|---|---|---|---|---|
| Center Grouping | Academic Year Undergraduates | Graduates | Ratio Grad/UG | REU Students | Total College Students | Pre-College Students | Total (Undergraduates + Graduates + Pre-College) |
| Average All Active ERC's 2008 | 19 | 52 | 2.7 | 10 | 81 | 6 | 77 |
| Average Bioengineering Sector FY 2008 | 16 | 53 | 3.3 | 7 | 76 | 7 | 76 |
| Average for Class of 1998 - FY 2008 | 26 | 97 | 3.7 | 9 | 132 | 1 | 124 |
| Center for Computer-Integrated Surgical Systems and Technology FY 2008 | 4 | 31 | 7.8 | 7 | 42 | 0 | 35 |

* - This table was previously named Table 3c and has been revised starting in FY2008. Values collected from Table 7 are benchmarked against all Active ERCs, each specific sector, and each specific class in the first 3 rows of the table. ERC-specific totals are presented in the last row.

Fig. A32

Web Tables

**Table 4: Industrial/Practitioner Members, Affiliated and Contributing Organizations, and Funders of Associated Projects**

| Summary: |
|---|
| 7 - Industrial/Practitioner Members |
| 8 - Affiliate Organizations |
| 8 - Contributing Organizations |

| Section 1: Industrial/Practitioner Members - 7 Industrial/Practitioner Members | | | | | | | |
|---|---|---|---|---|---|---|---|
| **Organization** | **Sector** | **Type of Support** | **Type of Involvement** | **Domestic / Foreign** | **Size (Industry Only)** | **New Member (Yes/No)** | **Total # of Sponsored Projects** |
| **7 Industrial/Practitioner Members** | | | | | | | |
| Industrial/Practitioner Members That Have Already Provided Current Year Support | | | | | | | |
| **Hologic** | Industry | Restricted cash - grants, contracts, and donations targeted for specific Center directed projects | Participation in Joint Research Projects<br>Technology Transfer | Domestic | Medium (500-1000 employees) | No | 1 |
| **Intuitive Surgical** | Industry | Restricted cash - grants, contracts, and donations targeted for specific Center directed projects<br><br>In-kind Equipment, Materials, or Supplies | Member of Center's Industrial Advisory Board<br>Technology Transfer | Domestic | Small (<500 employees) | No | 0 |
| **Philips U.S.A.** | Industry | Membership cash - fees for unrestricted use<br><br>Restricted cash - grants, contracts, and donations targeted for specific Center directed projects | Technology Transfer | Domestic | Large (>1000 employees) | No | 0 |
| Industrial/Practitioner Members That Will Provide Support by the End of the Current Reporting Year | | | | | | | |
| **American Shared Hospital Services** | Industry | Other Support | None Listed | Domestic | Small (<500 employees) | No | 0 |
| **Foster Miller** | Industry | Other Support | None Listed | Domestic | Small (<500 employees) | No | 0 |
| **Medtronics SNT** | Industry | Other Support | None Listed | Domestic | Small (<500 employees) | No | 0 |
| **Siemens** | Industry | Other Support | Technology Transfer | Domestic | Large (>1000 employees) | No | 0 |

Fig. A33

| Section 2: Affiliate Organizations - 8 Affiliate Organizations | | | | | | |
|---|---|---|---|---|---|---|
| **Organization** | **Sector** | **Type of Support** | **Type of Involvement** | **Domestic/Foreign** | **Size (Industry Only)** | **Total # of Sponsored Projects** |
| **8 Affiliate Organizations** | | | | | | |
| Affiliate Organizations That Have Already Provided Current Year Support | | | | | | |
| **Acoustic Medical Systems** | Industry | Restricted cash - grants, contracts, and donations targeted for specific Center directed projects | Member of Center's Industrial Advisory Board<br>Technology Transfer | Domestic | Small (<500 employees) | 1 |
| **Brigham & Women's Hospital** | Non-Profit | Restricted cash - grants, contracts, and donations targeted for specific Center directed projects | Participation in Joint Research Projects | Domestic | N/A | 1 |
| **Department of Defense** | Federal Government | Restricted cash - grants, contracts, and donations targeted for specific Center directed projects | Participation in Joint Research Projects | Domestic | N/A | 2 |
| **Johns Hopkins University** | Non-Profit | Restricted cash - grants, contracts, and donations targeted for specific Center directed projects | Participation in Joint Research Projects | Domestic | N/A | 2 |
| Affiliate Organizations That Will Provide Support by the End of the Current Reporting Year | | | | | | |
| **Georgetown University** | Non-Profit | Other Support | None Listed | Domestic | N/A | 0 |
| **Infinite Biomedical Technologies (IBT)** | Industry | Other Support | None Listed | Domestic | Small (<500 employees) | 0 |
| **Memorial Sloan Kettering Foundation** | Private Foundation | Restricted cash - grants, contracts, and donations targeted for specific Center directed projects | Participation in Joint Research Projects | Domestic | N/A | 1 |
| **Morgan State University** | Non-Profit | None Listed | None Listed | Domestic | N/A | 1 |

| Section 3: Contributing Organizations - 8 Contributing Organizations | | | | | |
|---|---|---|---|---|---|
| **Organization** | **Sector** | **Type of Involvement** | **Domestic/Foreign** | **Size (Industry Only)** | **Total # Sponsored Projects** |
| **8 Contributing Organizations** | | | | | |
| Contributing Organizations That Have Already Provided Current Year Support | | | | | |
| **BCRF** | Non-Profit | Participation in Joint Research Projects | Domestic | N/A | 1 |
| **Burdette Medical Systems** | Industry | None Listed | Domestic | Small (<500 employees) | 0 |
| **IEEE** | Non-Profit | Participation in Education Projects | Domestic | N/A | 0 |
| **NIH** | Federal Government | Participation in Joint Research Projects | Domestic | N/A | 7 |
| **NSF** | Federal Government | Participation in Joint Research Projects<br>Participation in Education Projects | Domestic | N/A | 1 |
| Contributing Organizations That Will Provide Support by the End of the Current Reporting Year | | | | | |
| **EADS Deutschland** | Industry | Participation in Joint Research Projects | Foreign | Large (>1000 employees) | 0 |
| **GE Healthcare** | Industry | None Listed | Domestic | Large (>1000 employees) | 0 |
| **Invenios** | Industry | None Listed | Domestic | Small (<500 employees) | 0 |

Fig. A33 (continued)

| Section 4: Funders of Associated Projects - 8 Funders of Associated Projects | | | | | | | |
|---|---|---|---|---|---|---|---|
| **Organization** | **Sector** | **Type of Involvement** | **Sponsor's Role** | **Domestic/Foreign** | **Size (Industry Only)** | **Member (Yes/No)** | **Total # of Associated Projects** |
| **Intuitive Surgical** | Industry | Member of Center's Industrial Advisory Board<br>Technology Transfer | | Domestic | Small (<500 employees) | Yes | 1 |
| **Philips U.S.A.** | Industry | Technology Transfer | | Domestic | Large (>1000 employees) | Yes | 1 |
| **Department of Defense** | Federal Government | Participation in Joint Research Projects | | Domestic | N/A | No | 1 |
| **Johns Hopkins University** | Non-Profit | Participation in Joint Research Projects | | Domestic | N/A | No | 2 |
| **NIH** | Federal Government | Participation in Joint Research Projects | | Domestic | N/A | No | 5 |
| **NSF** | Federal Government | Participation in Joint Research Projects<br>Participation in Education Projects | | Domestic | N/A | No | 2 |
| **EADS Deutschland** | Industry | Participation in Joint Research Projects | | Foreign | Large (>1000 employees) | No | 1 |
| **Ikona Medical** | Industry | Participation in Joint Research Projects | Principally Research/Technology Transfer | Domestic | Small (<500 employees) | No | 1 |

| Section 5: Summary | | | | | |
|---|---|---|---|---|---|
| **Sector** | **Industrial/Practitioner Members** | **Percent Foreign** | **Percent Small** | **Percent Medium** | **Percent Large** |
| Federal Government | 0 | 0% | N/A | N/A | N/A |
| Industry | 7 | 7% | 64% | 7% | 29% |
| Private Foundation | 0 | 0% | N/A | N/A | N/A |
| Non-Profit | 0 | 0% | N/A | N/A | N/A |
| **Total** | **7** | **4%** | **9%** | **1%** | **4%** |

Fig. A33 (continued)

| Table 5: Lifetime Industrial/Practitioner Membership History | | |
|---|---|---|
| **Organization** | **Award Year of Membership** | **Technology Transfer Activities** |
| **American Shared Hospital Services** | 1998-1999, 2000-2001, 2001-2002, 2002-2003, 2005-2006, 2006-2007, 2007-2008 | None Listed |
| **Foster Miller** | 2003-2004, 2004-2005, 2005-2006, 2006-2007, 2007-2008 | None Listed |
| **Hologic** | 2004-2005, 2005-2006, 2006-2007, 2007-2008, 2008-2009 | Joint Project<br>Graduate Hired by Industry<br>Other Tech Transfer |
| **Intuitive Surgical** | 2000-2001, 2001-2002, 2002-2003, 2003-2004, 2004-2005, 2005-2006, 2006-2007, 2007-2008, 2008-2009 | Joint Project<br>Licensed Technology (other than software)<br>Graduate Hired by Industry<br>Student on Site at Industry<br>Test Bed |
| **Medtronics SNT** | 2001-2002, 2002-2003, 2003-2004, 2004-2005, 2005-2006, 2006-2007 | None Listed |
| **Philips U.S.A.** | 2004-2005, 2005-2006, 2006-2007, 2007-2008, 2008-2009 | Joint Project<br>Graduate Hired by Industry |
| **Siemens** | 1999-2000, 2000-2001, 2001-2002, 2002-2003, 2003-2004, 2004-2005, 2005-2006, 2006-2007 | Graduate Hired by Industry<br>Student on Site at Industry |
| **Medtronic SNT** | 2001-2002, 2002-2003, 2003-2004, 2004-2005 | Graduate Hired by Industry |
| **Northern Digital** | 1999-2000, 2000-2001, 2001-2002, 2002-2003, 2003-2004, 2004-2005 | Joint Project |
| **Computerized Medical Systems, Image Guidance Division** | 1998-1999, 1999-2000, 2000-2001, 2001-2002, 2002-2003, 2003-2004 | Individual on Campus from Industry<br>Joint Project<br>Test Bed |
| **Foster-Miller** | 2003-2004 | None Listed |
| **GE Medical Systems** | 2000-2001, 2001-2002, 2002-2003, 2003-2004 | None Listed |
| **ImageGuide, Inc.** | 2002-2003, 2003-2004 | Faculty on Site at Industry<br>Graduate Hired by Industry |
| **Johnson and Johnson** | 2000-2001, 2001-2002, 2002-2003, 2003-2004 | None Listed |
| **Computer Motion** | 2000-2001, 2001-2002, 2002-2003 | None Listed |
| **Integrated Surgical Systems** | 1999-2000, 2000-2001, 2001-2002, 2002-2003, 2003-2004 | Joint Project<br>Graduate Hired by Industry |
| **Bovis LendLease** | 1998-1999, 2002-2003 | None Listed |
| **Burdette Medical Systems** | 1998-1999, 1999-2000, 2000-2001, 2001-2002, 2002-2003 | Joint Project |
| **Carl Zeiss** | 1999-2000, 2000-2001 | None Listed |
| **Johnson & Johnson** | 2000-2001, 2001-2002, 2002-2003 | None Listed |
| **Orthosoft** | 2000-2001, 2001-2002, 2002-2003 | None Listed |
| **Toshiba** | 1999-2000, 2000-2001, 2001-2002, 2002-2003 | None Listed |
| **American Shared Hospital Systems** | 1998-1999, 1999-2000, 2000-2001 | None Listed |
| **Medtronic Surgical Navigation Technology** | 2001-2002 | Graduate Hired by Industry |
| **Visualization Technology** | 1999-2000, 2000-2001, 2001-2002 | Student on Site at Industry |

Fig. A34

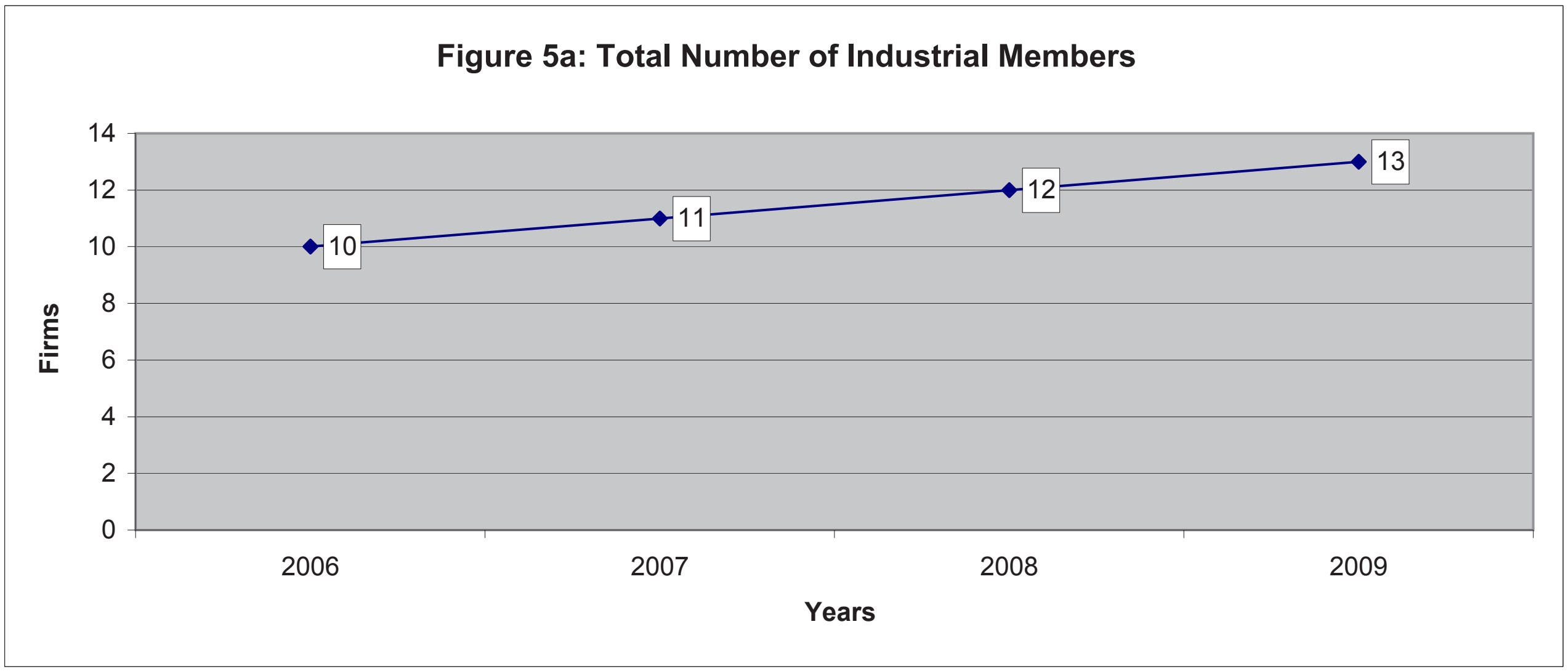

Fig. A35

| Table 5b: Industrial/Practitioner Members by Year | Sep 01, 2005 - Aug 31, 2006 | Sep 01, 2006 - Aug 31, 2007 | Sep 01, 2007 - Aug 31, 2008 | Sep 01, 2008 - Aug 31, 2009 |
|---|---|---|---|---|
| Industrial/Practitioner Members | 6 | 7 | 7 | 7 |
| Affiliated Organizations | 8 | 8 | 8 | 8 |
| Contributing Organizations | 2 | 2 | 3 | 8 |
| **Total Participating Organizations** | **16** | **17** | **18** | **23** |
| | | | | |
| Number of Member-sponsored projects | 0 | 3 | 1 | 3 |
| Number of non-member-sponsored projects | 0 | 20 | 29 | 28 |
| Total Number of Sponsored Projects | **18** | **23** | **30** | **31** |
| | | | | |
| Membership Fees | $10,000.00 | $0.00 | $0.00 | $0.00 |
| sponsored projects total dollar amount | $1,323,853.00 | $2,296,914.00 | $1,238,748.00 | $598,632.00 |
| Associated Projects total dollar amount | $1,388,751.00 | $452,317.00 | $969,865.00 | $1,901,038.00 |
| In-kind total dollar amount | $0.00 | $12,210,000.00 | $0.00 | $0.00 |
| **Total Dollar Amount, Industrial/Practitioner Support to Center** | $2,722,604.00 | $2,749,231.00 | $2,208,613.00 | $2,499,670.00 |

Fig. A35

| Table 6: Institutions Executing the ERC's Research, Technology Transfer, and Education Programs | | | | | | | | |
|---|---|---|---|---|---|---|---|---|
| Institutions | | | | Participants in ERC Activities | | | | |
| | | | | Personnel Involved in Research and Curric | | REU Students by Source Institutions | K-14 Personnel | |
| Name and Type | Total | Female Serving | Minority Serving | Faculty | Students | | Teachers | Students |
| **I. Lead** | **1** | **0** | **0** | **21** | **32** | **0** | **0** | **0** |
| Johns Hopkins University | | | | 21 | 32 | 0 | 0 | 0 |
| **II. Core Partners** | **3** | **0** | **0** | **5** | **0** | **0** | **0** | **0** |
| Carnegie Mellon University | | | | 1 | 0 | 0 | 0 | 0 |
| Massachusetts Institute of Technology | | | | 2 | 0 | 0 | 0 | 0 |
| Brigham and Women's Hospital | | | | 2 | 0 | 0 | 0 | 0 |
| **III. Collaborating (Outreach)** | **6** | **0** | **1** | **3** | **3** | **0** | **0** | **0** |
| Columbia University ,New York NY | | | | 0 | 1 | 0 | 0 | 0 |
| Harvard University ,Boston MA | | | | 0 | 0 | 0 | 0 | 0 |
| Morgan State University ,Baltimore MD | | | ✓ | 0 | 2 | 0 | 0 | 0 |
| Queen's University ,Kingston, ON FO | | | | 1 | 0 | 0 | 0 | 0 |
| University of California, Berkeley ,Berkeley CA | | | | 1 | 0 | 0 | 0 | 0 |
| University of Pennsylvania ,Philadelphia PA | | | | 1 | 0 | 0 | 0 | 0 |
| **IV. Non-ERC Institutions Providing REU Students** | **6** | **0** | **0** | **0** | **0** | **7** | **0** | **0** |
| University of Pittsburgh ,Pittsburgh PA | | | | 0 | 0 | 1 | 0 | 0 |
| University of Texas at Austin ,Austin TX | | | | 0 | 0 | 2 | 0 | 0 |
| University of Michigan | | | | 0 | 0 | 1 | 0 | 0 |
| University of Maryland, Baltimore County ,Baltimore MD | | | | 0 | 0 | 1 | 0 | 0 |
| Virginia Commonwealth Unversity | | | | 0 | 0 | 1 | 0 | 0 |
| Florida State University | | | | 0 | 0 | 1 | 0 | 0 |
| **V. NSF Diversity Program Awardees** | **3** | **0** | **2** | **0** | **0** | **0** | **0** | **0** |
| **Alliances for Graduate Education and the Professoriate (AGEP)** | **0** | **0** | **0** | **0** | **0** | **0** | **0** | **0** |
| No AGEP Awardees were entered. | | | | | | | | |
| **Centers of Research Excellence in Science and Technology (CREST)** | **0** | **0** | **0** | **0** | **0** | **0** | **0** | **0** |
| No CREST Awardees were entered. | | | | | | | | |
| **Louis Stokes Alliances for Minority Participation (LSAMP)** | **3** | **0** | **2** | **0** | **0** | **0** | **0** | **0** |
| Florida Agricultural And Mechanical University, Tallahassee (Florida Alliance for Minority Participation Project) | | | ✓ | 0 | 0 | 0 | 0 | 0 |
| University Of Massachusetts Amherst, Amherst( Northeast Louis Stokes Alliance for Minority Participation) | | | | 0 | 0 | 0 | 0 | 0 |
| University Of Puerto Rico-Rio Piedras Campus, Rio Piedras (Puerto Rico-Louis Stokes Alliance for Minority Participation) | | | ✓ | 0 | 0 | 0 | 0 | 0 |
| **Tribal Colleges and Universities Program (TCUP)** | **0** | **0** | **0** | **0** | **0** | **0** | **0** | **0** |
| No TCUP Awardees were entered. | | | | | | | | |
| **Other NSF Diversity Program Awardees** | **0** | **0** | **0** | **0** | **0** | **0** | **0** | **0** |
| No Institutions were entered. | | | | | | | | |
| **VI. K-14 Institutions** | **0** | **0** | **0** | **0** | **0** | **0** | **0** | **0** |
| No Institutions were entered. | | | | | | | | |
| **Total** | **19** | **0** | **3** | **29** | **35** | **7** | **0** | **0** |

Fig. A36 (Above)

Fig. A37 (Opposite page): (Top) Map details the U.S. locations of the lead, core partner and outreach institutions and institutions of REU and RET Outreach participants. This map is based on data already required in the ERCWeb database. (Bottom) Map details the foreign locations of lead, core partner and outreach institutions, and institutions of REU and RET Outreach participants. This map is based on data already required in the ERCWeb database. All of the maps were taken from ERCYR 10 data.

## Domestic Location of Lead, Core Partner, Outreach, and REU and RET Participants' Institutions for the Center for Computer-Integrated Surgical Systems and Technology

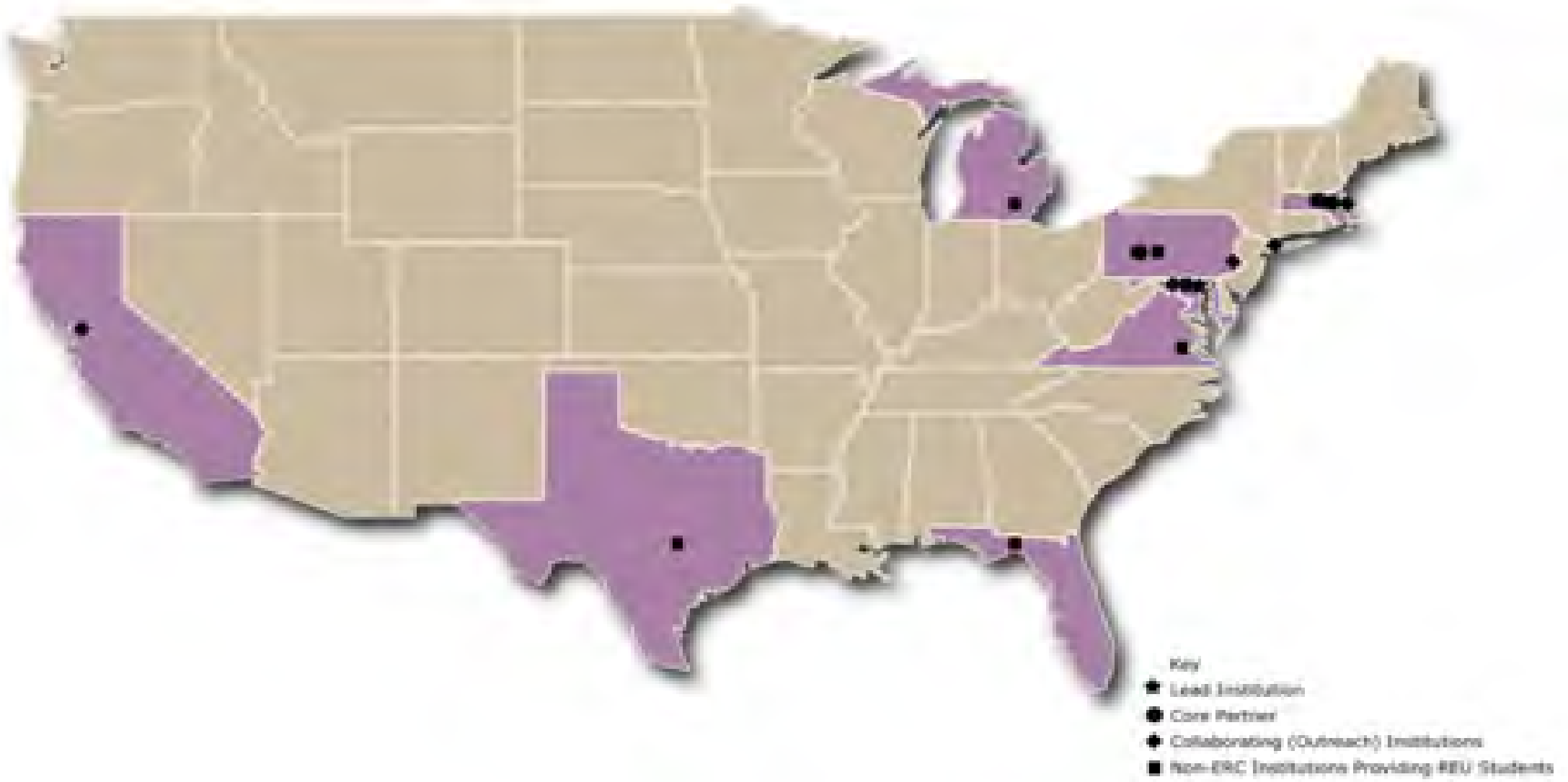

## Foreign Location of Lead, Core Partner, Outreach, and REU and RET Participants' Institutions for the Center for Computer-Integrated Surgical Systems and Technology

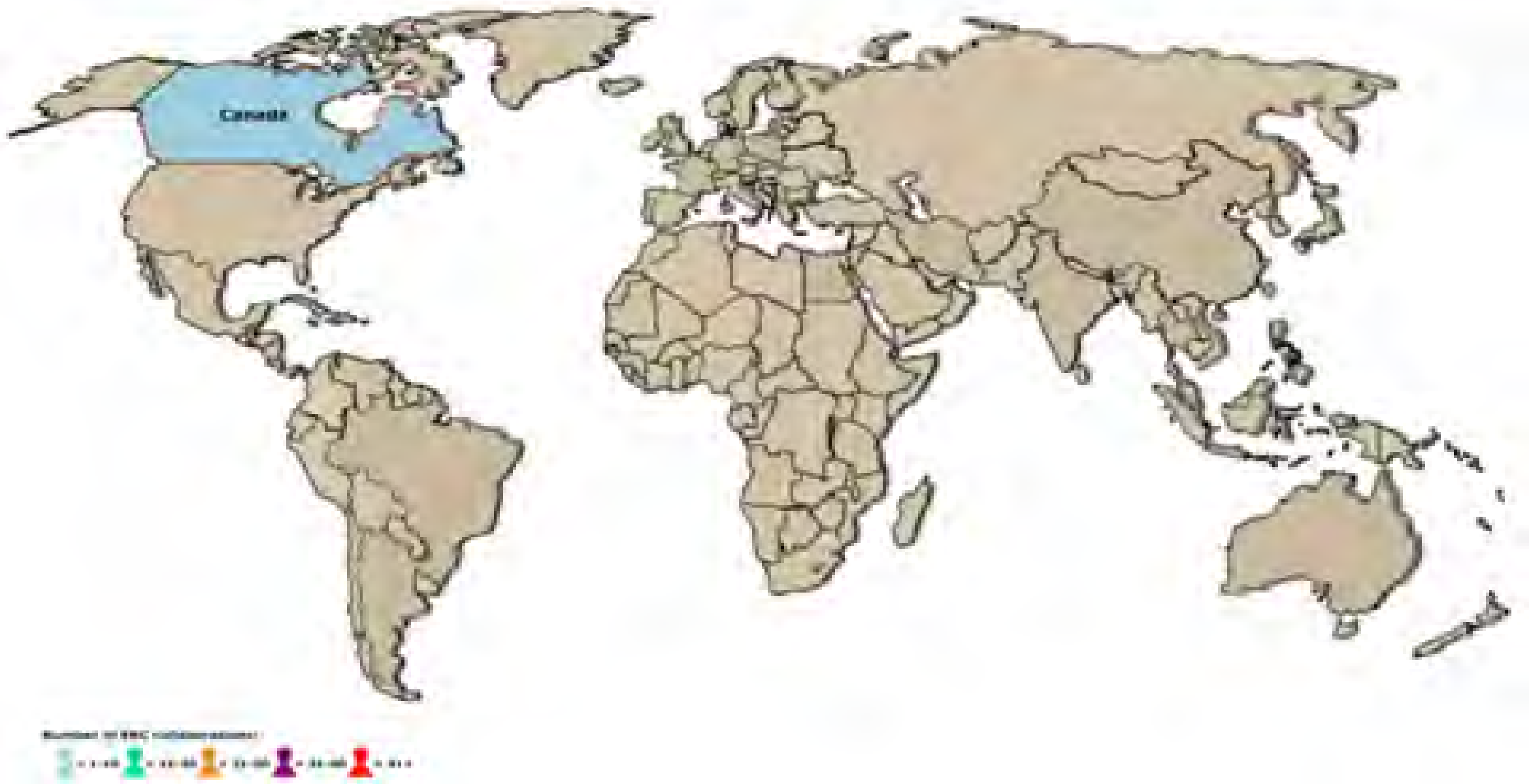

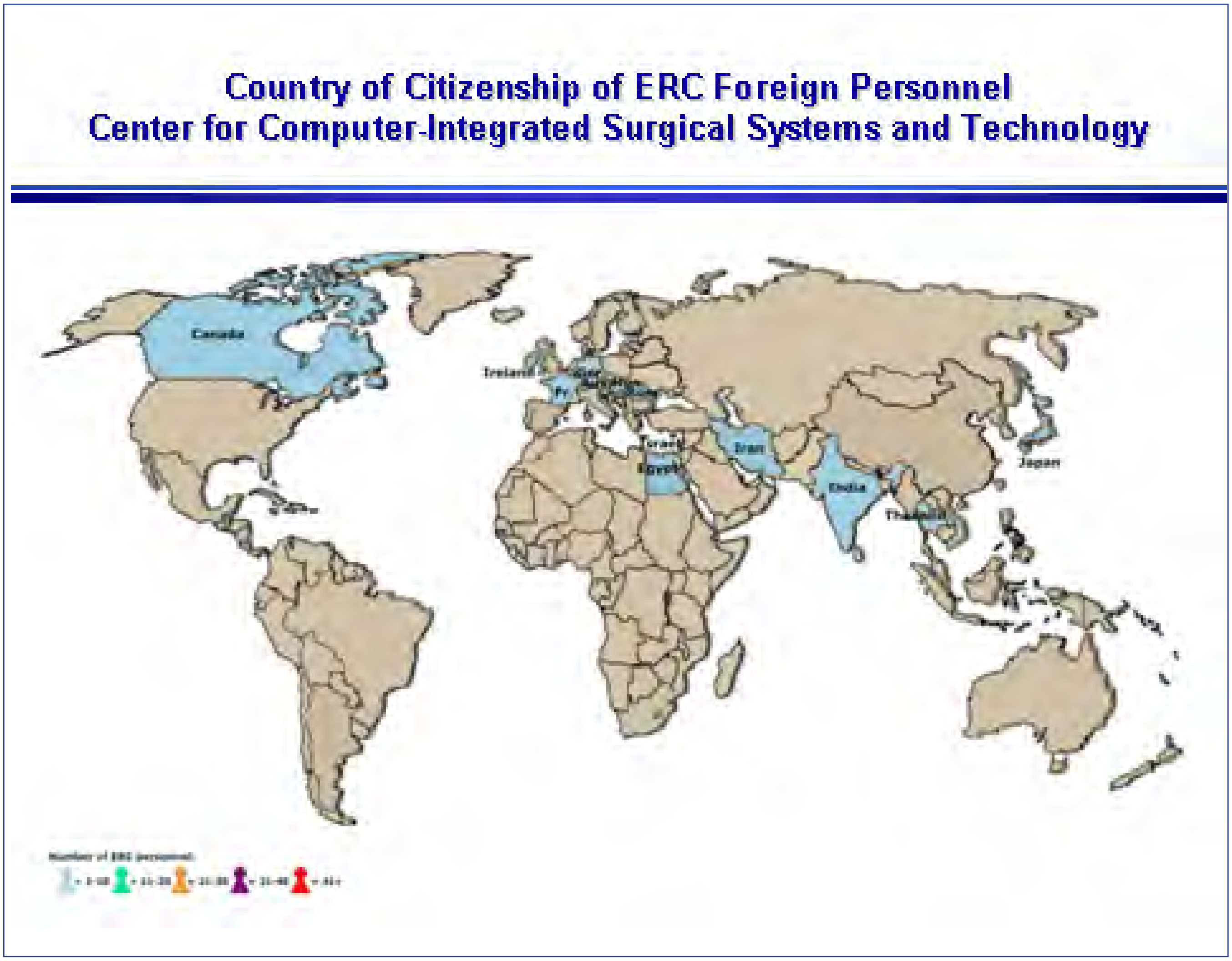

Fig. A38: Map detailing the countries of citizenship for ERC Foreign Personnel. This map is based on data already required in the ERCWeb database. All of the maps are were taken from ERCYR 10 data.

Web Tables

**Table 7: ERC Personnel**

| Personnel Type | Total | Gender | | | Citizenship Status | | | | | | | | | Ethnicity: Hispanic | | | Disability |
|---|---|---|---|---|---|---|---|---|---|---|---|---|---|---|---|---|---|
| | | | | | Race: U.S. citizens and permanent residents only | | | | | | | | | | | | |
| | | Male | Female | Gender Not Reported | AI/AN | NH/PI | B/AA | W | A | More than one race reported, minority | More than one race reported, non-minority | Not Provided | Other Non-U.S. | US/Perm | Temp | Not Reported | |
| **Total - All Institutions** | | | | | | | | | | | | | | | | | |
| **Total *** | **96** | **74** | **22** | **0** | **0** | **0** | **5** | **47** | **16** | **1** | **2** | **1** | **23** | **2** | **0** | **0** | **0** |
| **Leadership/Administration** | | | | | | | | | | | | | | | | | |
| Directors | 2 | 2 | 0 | 0 | 0 | 0 | 1 | 1 | 0 | 0 | 0 | 0 | 0 | 0 | 0 | 0 | 0 |
| Thrust Leaders | 3 | 1 | 2 | 0 | 0 | 0 | 0 | 2 | 0 | 0 | 1 | 0 | 0 | 0 | 0 | 0 | 0 |
| Research Thrust Management and Strategic Planning | 1 | 1 | 0 | 0 | 0 | 0 | 0 | 1 | 0 | 0 | 0 | 0 | 0 | 0 | 0 | 0 | 0 |
| Industrial Liaison Officer (ILO) | 1 | 1 | 0 | 0 | 0 | 0 | 0 | 1 | 0 | 0 | 0 | 0 | 0 | 0 | 0 | 0 | 0 |
| Education Program Leaders | 0 | 0 | 0 | 0 | 0 | 0 | 0 | 0 | 0 | 0 | 0 | 0 | 0 | 0 | 0 | 0 | 0 |
| Administrative Director | 1 | 0 | 1 | 0 | 0 | 0 | 0 | 1 | 0 | 0 | 0 | 0 | 0 | 1 | 0 | 0 | 0 |
| Staff | 6 | 1 | 5 | 0 | 0 | 0 | 0 | 6 | 0 | 0 | 0 | 0 | 0 | 1 | 0 | 0 | 0 |
| **Subtotal** | **14** | **6** | **8** | **0** | **0** | **0** | **1** | **12** | **0** | **0** | **1** | **0** | **0** | **2** | **0** | **0** | **0** |
| **Research under Strategic Research Plan** | | | | | | | | | | | | | | | | | |
| Senior Faculty | 17 | 16 | 1 | 0 | 0 | 0 | 2 | 12 | 2 | 0 | 0 | 0 | 1 | 0 | 0 | 0 | 0 |
| Junior Faculty | 10 | 9 | 1 | 0 | 0 | 0 | 0 | 5 | 4 | 0 | 0 | 0 | 0 | 0 | 0 | 0 | 0 |
| Research Staff | 5 | 5 | 0 | 0 | 0 | 0 | 0 | 1 | 0 | 0 | 1 | 0 | 3 | 0 | 0 | 0 | 0 |
| Visiting Faculty | 1 | 1 | 0 | 0 | 0 | 0 | 1 | 0 | 0 | 0 | 0 | 0 | 0 | 0 | 0 | 0 | 0 |
| Industry Researchers | 4 | 4 | 0 | 0 | 0 | 0 | 0 | 4 | 0 | 0 | 0 | 0 | 0 | 0 | 0 | 0 | 0 |
| Post Docs | 3 | 2 | 1 | 0 | 0 | 0 | 0 | 1 | 0 | 0 | 0 | 0 | 2 | 0 | 0 | 0 | 0 |
| Doctoral Students | 31 | 27 | 4 | 0 | 0 | 0 | 1 | 10 | 5 | 0 | 0 | 1 | 14 | 0 | 0 | 0 | 0 |
| Master's Students | 4 | 1 | 3 | 0 | 0 | 0 | 0 | 1 | 2 | 0 | 0 | 0 | 1 | 0 | 0 | 0 | 0 |
| Undergraduate Students | 5 | 3 | 2 | 0 | 0 | 0 | 0 | 1 | 3 | 1 | 0 | 0 | 0 | 0 | 0 | 0 | 0 |
| **Subtotal** | **80** | **68** | **12** | **0** | **0** | **0** | **4** | **35** | **16** | **1** | **1** | **1** | **21** | **0** | **0** | **0** | **0** |
| **Curriculum Development and Outreach** | | | | | | | | | | | | | | | | | |
| Senior Faculty | 0 | 0 | 0 | 0 | 0 | 0 | 0 | 0 | 0 | 0 | 0 | 0 | 0 | 0 | 0 | 0 | 0 |
| Junior Faculty | 0 | 0 | 0 | 0 | 0 | 0 | 0 | 0 | 0 | 0 | 0 | 0 | 0 | 0 | 0 | 0 | 0 |
| Research Staff | 0 | 0 | 0 | 0 | 0 | 0 | 0 | 0 | 0 | 0 | 0 | 0 | 0 | 0 | 0 | 0 | 0 |
| Visiting Faculty | 0 | 0 | 0 | 0 | 0 | 0 | 0 | 0 | 0 | 0 | 0 | 0 | 0 | 0 | 0 | 0 | 0 |
| Industry Researchers | 0 | 0 | 0 | 0 | 0 | 0 | 0 | 0 | 0 | 0 | 0 | 0 | 0 | 0 | 0 | 0 | 0 |
| Post Docs | 0 | 0 | 0 | 0 | 0 | 0 | 0 | 0 | 0 | 0 | 0 | 0 | 0 | 0 | 0 | 0 | 0 |
| Doctoral Students | 0 | 0 | 0 | 0 | 0 | 0 | 0 | 0 | 0 | 0 | 0 | 0 | 0 | 0 | 0 | 0 | 0 |
| Master's Students | 0 | 0 | 0 | 0 | 0 | 0 | 0 | 0 | 0 | 0 | 0 | 0 | 0 | 0 | 0 | 0 | 0 |
| Undergraduate Students | 0 | 0 | 0 | 0 | 0 | 0 | 0 | 0 | 0 | 0 | 0 | 0 | 0 | 0 | 0 | 0 | 0 |
| **Subtotal** | **0** | **0** | **0** | **0** | **0** | **0** | **0** | **0** | **0** | **0** | **0** | **0** | **0** | **0** | **0** | **0** | **0** |
| **ERC REU Student** | | | | | | | | | | | | | | | | | |
| NSF REU Site Award | 0 | 0 | 0 | 0 | 0 | 0 | 0 | 0 | 0 | 0 | 0 | 0 | 0 | 0 | 0 | 0 | 0 |
| ERC's Own REU | 0 | 0 | 0 | 0 | 0 | 0 | 0 | 0 | 0 | 0 | 0 | 0 | 0 | 0 | 0 | 0 | 0 |
| Other Visiting College Students | 2 | 0 | 2 | 0 | 0 | 0 | 0 | 0 | 0 | 0 | 0 | 0 | 2 | 0 | 0 | 0 | 0 |
| **Subtotal** | **2** | **0** | **2** | **0** | **0** | **0** | **0** | **0** | **0** | **0** | **0** | **0** | **2** | **0** | **0** | **0** | **0** |
| **Pre-College (K-12)** | | | | | | | | | | | | | | | | | |
| Teachers (RET) | 0 | 0 | 0 | 0 | 0 | 0 | 0 | 0 | 0 | 0 | 0 | 0 | 0 | 0 | 0 | 0 | 0 |
| Teachers (non-RET) | 0 | 0 | 0 | 0 | 0 | 0 | 0 | 0 | 0 | 0 | 0 | 0 | 0 | 0 | 0 | 0 | 0 |
| **Subtotal** | **0** | **0** | **0** | **0** | **0** | **0** | **0** | **0** | **0** | **0** | **0** | **0** | **0** | **0** | **0** | **0** | **0** |
| **Total *** | **91** | **69** | **22** | **0** | **0** | **0** | **3** | **44** | **16** | **1** | **2** | **1** | **23** | **2** | **0** | **0** | **0** |
| **Leadership/Administration** | | | | | | | | | | | | | | | | | |
| Directors | 2 | 2 | 0 | 0 | 0 | 0 | 1 | 1 | 0 | 0 | 0 | 0 | 0 | 0 | 0 | 0 | 0 |
| Thrust Leaders | 3 | 1 | 2 | 0 | 0 | 0 | 0 | 2 | 0 | 0 | 1 | 0 | 0 | 0 | 0 | 0 | 0 |
| Research Thrust Management and Strategic Planning | 1 | 1 | 0 | 0 | 0 | 0 | 0 | 1 | 0 | 0 | 0 | 0 | 0 | 0 | 0 | 0 | 0 |
| Industrial Liaison Officer (ILO) | 1 | 1 | 0 | 0 | 0 | 0 | 0 | 1 | 0 | 0 | 0 | 0 | 0 | 0 | 0 | 0 | 0 |
| Education Program Leaders | 0 | 0 | 0 | 0 | 0 | 0 | 0 | 0 | 0 | 0 | 0 | 0 | 0 | 0 | 0 | 0 | 0 |
| Administrative Director | 1 | 0 | 1 | 0 | 0 | 0 | 0 | 1 | 0 | 0 | 0 | 0 | 0 | 1 | 0 | 0 | 0 |
| Staff | 6 | 1 | 5 | 0 | 0 | 0 | 0 | 6 | 0 | 0 | 0 | 0 | 0 | 1 | 0 | 0 | 0 |
| **Subtotal** | **14** | **6** | **8** | **0** | **0** | **0** | **1** | **12** | **0** | **0** | **1** | **0** | **0** | **2** | **0** | **0** | **0** |
| **Research under Strategic Research Plan** | | | | | | | | | | | | | | | | | |
| Senior Faculty | 15 | 14 | 1 | 0 | 0 | 0 | 1 | 11 | 2 | 0 | 0 | 0 | 1 | 0 | 0 | 0 | 0 |
| Junior Faculty | 9 | 8 | 1 | 0 | 0 | 0 | 0 | 4 | 4 | 0 | 0 | 0 | 0 | 0 | 0 | 0 | 0 |
| Research Staff | 4 | 4 | 0 | 0 | 0 | 0 | 0 | 0 | 0 | 0 | 1 | 0 | 3 | 0 | 0 | 0 | 0 |
| Visiting Faculty | 1 | 1 | 0 | 0 | 0 | 0 | 1 | 0 | 0 | 0 | 0 | 0 | 0 | 0 | 0 | 0 | 0 |
| Industry Researchers | 4 | 4 | 0 | 0 | 0 | 0 | 0 | 4 | 0 | 0 | 0 | 0 | 0 | 0 | 0 | 0 | 0 |
| Post Docs | 3 | 2 | 1 | 0 | 0 | 0 | 0 | 1 | 0 | 0 | 0 | 0 | 2 | 0 | 0 | 0 | 0 |
| Doctoral Students | 30 | 26 | 4 | 0 | 0 | 0 | 0 | 10 | 5 | 0 | 0 | 1 | 14 | 0 | 0 | 0 | 0 |
| Master's Students | 4 | 1 | 3 | 0 | 0 | 0 | 0 | 1 | 2 | 0 | 0 | 0 | 1 | 0 | 0 | 0 | 0 |
| Undergraduate Students | 5 | 3 | 2 | 0 | 0 | 0 | 0 | 1 | 3 | 1 | 0 | 0 | 0 | 0 | 0 | 0 | 0 |
| **Subtotal** | **75** | **63** | **12** | **0** | **0** | **0** | **2** | **32** | **16** | **1** | **1** | **1** | **21** | **0** | **0** | **0** | **0** |
| **Curriculum Development and Outreach** | | | | | | | | | | | | | | | | | |
| Senior Faculty | 0 | 0 | 0 | 0 | 0 | 0 | 0 | 0 | 0 | 0 | 0 | 0 | 0 | 0 | 0 | 0 | 0 |
| Junior Faculty | 0 | 0 | 0 | 0 | 0 | 0 | 0 | 0 | 0 | 0 | 0 | 0 | 0 | 0 | 0 | 0 | 0 |
| Research Staff | 0 | 0 | 0 | 0 | 0 | 0 | 0 | 0 | 0 | 0 | 0 | 0 | 0 | 0 | 0 | 0 | 0 |
| Visiting Faculty | 0 | 0 | 0 | 0 | 0 | 0 | 0 | 0 | 0 | 0 | 0 | 0 | 0 | 0 | 0 | 0 | 0 |
| Industry Researchers | 0 | 0 | 0 | 0 | 0 | 0 | 0 | 0 | 0 | 0 | 0 | 0 | 0 | 0 | 0 | 0 | 0 |
| Post Docs | 0 | 0 | 0 | 0 | 0 | 0 | 0 | 0 | 0 | 0 | 0 | 0 | 0 | 0 | 0 | 0 | 0 |
| Doctoral Students | 0 | 0 | 0 | 0 | 0 | 0 | 0 | 0 | 0 | 0 | 0 | 0 | 0 | 0 | 0 | 0 | 0 |
| Master's Students | 0 | 0 | 0 | 0 | 0 | 0 | 0 | 0 | 0 | 0 | 0 | 0 | 0 | 0 | 0 | 0 | 0 |
| Undergraduate Students | 0 | 0 | 0 | 0 | 0 | 0 | 0 | 0 | 0 | 0 | 0 | 0 | 0 | 0 | 0 | 0 | 0 |
| **Subtotal** | **0** | **0** | **0** | **0** | **0** | **0** | **0** | **0** | **0** | **0** | **0** | **0** | **0** | **0** | **0** | **0** | **0** |
| **ERC REU Student** | | | | | | | | | | | | | | | | | |
| NSF REU Site Award | 0 | 0 | 0 | 0 | 0 | 0 | 0 | 0 | 0 | 0 | 0 | 0 | 0 | 0 | 0 | 0 | 0 |
| ERC's Own REU | 0 | 0 | 0 | 0 | 0 | 0 | 0 | 0 | 0 | 0 | 0 | 0 | 0 | 0 | 0 | 0 | 0 |
| Other Visiting College Students | 2 | 0 | 2 | 0 | 0 | 0 | 0 | 0 | 0 | 0 | 0 | 0 | 2 | 0 | 0 | 0 | 0 |
| **Subtotal** | **2** | **0** | **2** | **0** | **0** | **0** | **0** | **0** | **0** | **0** | **0** | **0** | **2** | **0** | **0** | **0** | **0** |
| **Pre-College (K-12)** | | | | | | | | | | | | | | | | | |
| Teachers (RET) | 0 | 0 | 0 | 0 | 0 | 0 | 0 | 0 | 0 | 0 | 0 | 0 | 0 | 0 | 0 | 0 | 0 |
| Teachers (non-RET) | 0 | 0 | 0 | 0 | 0 | 0 | 0 | 0 | 0 | 0 | 0 | 0 | 0 | 0 | 0 | 0 | 0 |
| **Subtotal** | **0** | **0** | **0** | **0** | **0** | **0** | **0** | **0** | **0** | **0** | **0** | **0** | **0** | **0** | **0** | **0** | **0** |

Fig. A39: Demographic information for the no cost extension period was updated for personnel who have left the ERC. However, no information on new personnel were collected and no information was collected from partner institutions.

| Personnel Type | Total | Gender | | | Citizenship Status | | | | | | | | | Ethnicity: Hispanic | | | Disability |
|---|---|---|---|---|---|---|---|---|---|---|---|---|---|---|---|---|---|
| | | | | | Race: U.S. citizens and permanent residents only | | | | | | | | | | | | |
| | | Male | Female | Gender Not Reported | AI/AN | NH/PI | B/AA | W | A | More than one race reported, minority | More than one race reported, non-minority | Not Provided | Other Non-U.S. | US/Perm | Temp | Not Reported | |
| **Total - All Core Partner** | | | | | | | | | | | | | | | | | |
| **Total *** | **2** | **2** | **0** | **0** | **0** | **0** | **0** | **2** | **0** | **0** | **0** | **0** | **0** | **0** | **0** | **0** | **0** |
| **Leadership/Administration** | | | | | | | | | | | | | | | | | |
| Directors | 0 | 0 | 0 | 0 | 0 | 0 | 0 | 0 | 0 | 0 | 0 | 0 | 0 | 0 | 0 | 0 | 0 |
| Thrust Leaders | 0 | 0 | 0 | 0 | 0 | 0 | 0 | 0 | 0 | 0 | 0 | 0 | 0 | 0 | 0 | 0 | 0 |
| Research Thrust Management and Strategic Planning | 0 | 0 | 0 | 0 | 0 | 0 | 0 | 0 | 0 | 0 | 0 | 0 | 0 | 0 | 0 | 0 | 0 |
| Industrial Liaison Officer (ILO) | 0 | 0 | 0 | 0 | 0 | 0 | 0 | 0 | 0 | 0 | 0 | 0 | 0 | 0 | 0 | 0 | 0 |
| Education Program Leaders | 0 | 0 | 0 | 0 | 0 | 0 | 0 | 0 | 0 | 0 | 0 | 0 | 0 | 0 | 0 | 0 | 0 |
| Administrative Director | 0 | 0 | 0 | 0 | 0 | 0 | 0 | 0 | 0 | 0 | 0 | 0 | 0 | 0 | 0 | 0 | 0 |
| Staff | 0 | 0 | 0 | 0 | 0 | 0 | 0 | 0 | 0 | 0 | 0 | 0 | 0 | 0 | 0 | 0 | 0 |
| **Subtotal** | **0** | **0** | **0** | **0** | **0** | **0** | **0** | **0** | **0** | **0** | **0** | **0** | **0** | **0** | **0** | **0** | **0** |
| **Research under Strategic Research Plan** | | | | | | | | | | | | | | | | | |
| Senior Faculty | 1 | 1 | 0 | 0 | 0 | 0 | 0 | 1 | 0 | 0 | 0 | 0 | 0 | 0 | 0 | 0 | 0 |
| Junior Faculty | 0 | 0 | 0 | 0 | 0 | 0 | 0 | 0 | 0 | 0 | 0 | 0 | 0 | 0 | 0 | 0 | 0 |
| Research Staff | 1 | 1 | 0 | 0 | 0 | 0 | 0 | 1 | 0 | 0 | 0 | 0 | 0 | 0 | 0 | 0 | 0 |
| Visiting Faculty | 0 | 0 | 0 | 0 | 0 | 0 | 0 | 0 | 0 | 0 | 0 | 0 | 0 | 0 | 0 | 0 | 0 |
| Industry Researchers | 0 | 0 | 0 | 0 | 0 | 0 | 0 | 0 | 0 | 0 | 0 | 0 | 0 | 0 | 0 | 0 | 0 |
| Post Docs | 0 | 0 | 0 | 0 | 0 | 0 | 0 | 0 | 0 | 0 | 0 | 0 | 0 | 0 | 0 | 0 | 0 |
| Doctoral Students | 0 | 0 | 0 | 0 | 0 | 0 | 0 | 0 | 0 | 0 | 0 | 0 | 0 | 0 | 0 | 0 | 0 |
| Master's Students | 0 | 0 | 0 | 0 | 0 | 0 | 0 | 0 | 0 | 0 | 0 | 0 | 0 | 0 | 0 | 0 | 0 |
| Undergraduate Students | 0 | 0 | 0 | 0 | 0 | 0 | 0 | 0 | 0 | 0 | 0 | 0 | 0 | 0 | 0 | 0 | 0 |
| **Subtotal** | **2** | **2** | **0** | **0** | **0** | **0** | **0** | **2** | **0** | **0** | **0** | **0** | **0** | **0** | **0** | **0** | **0** |
| **Curriculum Development and Outreach** | | | | | | | | | | | | | | | | | |
| Senior Faculty | 0 | 0 | 0 | 0 | 0 | 0 | 0 | 0 | 0 | 0 | 0 | 0 | 0 | 0 | 0 | 0 | 0 |
| Junior Faculty | 0 | 0 | 0 | 0 | 0 | 0 | 0 | 0 | 0 | 0 | 0 | 0 | 0 | 0 | 0 | 0 | 0 |
| Research Staff | 0 | 0 | 0 | 0 | 0 | 0 | 0 | 0 | 0 | 0 | 0 | 0 | 0 | 0 | 0 | 0 | 0 |
| Visiting Faculty | 0 | 0 | 0 | 0 | 0 | 0 | 0 | 0 | 0 | 0 | 0 | 0 | 0 | 0 | 0 | 0 | 0 |
| Industry Researchers | 0 | 0 | 0 | 0 | 0 | 0 | 0 | 0 | 0 | 0 | 0 | 0 | 0 | 0 | 0 | 0 | 0 |
| Post Docs | 0 | 0 | 0 | 0 | 0 | 0 | 0 | 0 | 0 | 0 | 0 | 0 | 0 | 0 | 0 | 0 | 0 |
| Doctoral Students | 0 | 0 | 0 | 0 | 0 | 0 | 0 | 0 | 0 | 0 | 0 | 0 | 0 | 0 | 0 | 0 | 0 |
| Master's Students | 0 | 0 | 0 | 0 | 0 | 0 | 0 | 0 | 0 | 0 | 0 | 0 | 0 | 0 | 0 | 0 | 0 |
| Undergraduate Students | 0 | 0 | 0 | 0 | 0 | 0 | 0 | 0 | 0 | 0 | 0 | 0 | 0 | 0 | 0 | 0 | 0 |
| **Subtotal** | **0** | **0** | **0** | **0** | **0** | **0** | **0** | **0** | **0** | **0** | **0** | **0** | **0** | **0** | **0** | **0** | **0** |
| **ERC REU Student** | | | | | | | | | | | | | | | | | |
| NSF REU Site Award | 0 | 0 | 0 | 0 | 0 | 0 | 0 | 0 | 0 | 0 | 0 | 0 | 0 | 0 | 0 | 0 | 0 |
| ERC's Own REU | 0 | 0 | 0 | 0 | 0 | 0 | 0 | 0 | 0 | 0 | 0 | 0 | 0 | 0 | 0 | 0 | 0 |
| Other Visiting College Students | 0 | 0 | 0 | 0 | 0 | 0 | 0 | 0 | 0 | 0 | 0 | 0 | 0 | 0 | 0 | 0 | 0 |
| **Subtotal** | **0** | **0** | **0** | **0** | **0** | **0** | **0** | **0** | **0** | **0** | **0** | **0** | **0** | **0** | **0** | **0** | **0** |
| **Pre-College (K-12)** | | | | | | | | | | | | | | | | | |
| Teachers (RET) | 0 | 0 | 0 | 0 | 0 | 0 | 0 | 0 | 0 | 0 | 0 | 0 | 0 | 0 | 0 | 0 | 0 |
| Teachers (non-RET) | 0 | 0 | 0 | 0 | 0 | 0 | 0 | 0 | 0 | 0 | 0 | 0 | 0 | 0 | 0 | 0 | 0 |
| **Subtotal** | **0** | **0** | **0** | **0** | **0** | **0** | **0** | **0** | **0** | **0** | **0** | **0** | **0** | **0** | **0** | **0** | **0** |
| **Total *** | **2** | **2** | **0** | **0** | **0** | **0** | **0** | **2** | **0** | **0** | **0** | **0** | **0** | **0** | **0** | **0** | **0** |
| **Research under Strategic Research Plan** | | | | | | | | | | | | | | | | | |
| Senior Faculty | 1 | 1 | 0 | 0 | 0 | 0 | 0 | 1 | 0 | 0 | 0 | 0 | 0 | 0 | 0 | 0 | 0 |
| Junior Faculty | 0 | 0 | 0 | 0 | 0 | 0 | 0 | 0 | 0 | 0 | 0 | 0 | 0 | 0 | 0 | 0 | 0 |
| Research Staff | 1 | 1 | 0 | 0 | 0 | 0 | 0 | 1 | 0 | 0 | 0 | 0 | 0 | 0 | 0 | 0 | 0 |
| Visiting Faculty | 0 | 0 | 0 | 0 | 0 | 0 | 0 | 0 | 0 | 0 | 0 | 0 | 0 | 0 | 0 | 0 | 0 |
| Industry Researchers | 0 | 0 | 0 | 0 | 0 | 0 | 0 | 0 | 0 | 0 | 0 | 0 | 0 | 0 | 0 | 0 | 0 |
| Post Docs | 0 | 0 | 0 | 0 | 0 | 0 | 0 | 0 | 0 | 0 | 0 | 0 | 0 | 0 | 0 | 0 | 0 |
| Doctoral Students | 0 | 0 | 0 | 0 | 0 | 0 | 0 | 0 | 0 | 0 | 0 | 0 | 0 | 0 | 0 | 0 | 0 |
| Master's Students | 0 | 0 | 0 | 0 | 0 | 0 | 0 | 0 | 0 | 0 | 0 | 0 | 0 | 0 | 0 | 0 | 0 |
| Undergraduate Students | 0 | 0 | 0 | 0 | 0 | 0 | 0 | 0 | 0 | 0 | 0 | 0 | 0 | 0 | 0 | 0 | 0 |
| **Subtotal** | **2** | **2** | **0** | **0** | **0** | **0** | **0** | **2** | **0** | **0** | **0** | **0** | **0** | **0** | **0** | **0** | **0** |
| **Massachusetts Institute of Technology - Core Partner** | | | | | | | | | | | | | | | | | |
| **Total *** | **0** | **0** | **0** | **0** | **0** | **0** | **0** | **0** | **0** | **0** | **0** | **0** | **0** | **0** | **0** | **0** | **0** |
| **Brigham and Women's Hospital - Core Partner** | | | | | | | | | | | | | | | | | |
| **Total *** | **0** | **0** | **0** | **0** | **0** | **0** | **0** | **0** | **0** | **0** | **0** | **0** | **0** | **0** | **0** | **0** | **0** |
| **Total - All Collaborating Institutions** | | | | | | | | | | | | | | | | | |
| **Total *** | **3** | **3** | **0** | **0** | **0** | **0** | **2** | **1** | **0** | **0** | **0** | **0** | **0** | **0** | **0** | **0** | **0** |
| **Leadership/Administration** | | | | | | | | | | | | | | | | | |
| Directors | 0 | 0 | 0 | 0 | 0 | 0 | 0 | 0 | 0 | 0 | 0 | 0 | 0 | 0 | 0 | 0 | 0 |
| Thrust Leaders | 0 | 0 | 0 | 0 | 0 | 0 | 0 | 0 | 0 | 0 | 0 | 0 | 0 | 0 | 0 | 0 | 0 |
| Research Thrust Management and Strategic Planning | 0 | 0 | 0 | 0 | 0 | 0 | 0 | 0 | 0 | 0 | 0 | 0 | 0 | 0 | 0 | 0 | 0 |
| Industrial Liaison Officer (ILO) | 0 | 0 | 0 | 0 | 0 | 0 | 0 | 0 | 0 | 0 | 0 | 0 | 0 | 0 | 0 | 0 | 0 |
| Education Program Leaders | 0 | 0 | 0 | 0 | 0 | 0 | 0 | 0 | 0 | 0 | 0 | 0 | 0 | 0 | 0 | 0 | 0 |
| Administrative Director | 0 | 0 | 0 | 0 | 0 | 0 | 0 | 0 | 0 | 0 | 0 | 0 | 0 | 0 | 0 | 0 | 0 |
| Staff | 0 | 0 | 0 | 0 | 0 | 0 | 0 | 0 | 0 | 0 | 0 | 0 | 0 | 0 | 0 | 0 | 0 |
| **Subtotal** | **0** | **0** | **0** | **0** | **0** | **0** | **0** | **0** | **0** | **0** | **0** | **0** | **0** | **0** | **0** | **0** | **0** |
| **Research under Strategic Research Plan** | | | | | | | | | | | | | | | | | |
| Senior Faculty | 1 | 1 | 0 | 0 | 0 | 0 | 1 | 0 | 0 | 0 | 0 | 0 | 0 | 0 | 0 | 0 | 0 |
| Junior Faculty | 1 | 1 | 0 | 0 | 0 | 0 | 0 | 1 | 0 | 0 | 0 | 0 | 0 | 0 | 0 | 0 | 0 |
| Research Staff | 0 | 0 | 0 | 0 | 0 | 0 | 0 | 0 | 0 | 0 | 0 | 0 | 0 | 0 | 0 | 0 | 0 |
| Visiting Faculty | 0 | 0 | 0 | 0 | 0 | 0 | 0 | 0 | 0 | 0 | 0 | 0 | 0 | 0 | 0 | 0 | 0 |
| Industry Researchers | 0 | 0 | 0 | 0 | 0 | 0 | 0 | 0 | 0 | 0 | 0 | 0 | 0 | 0 | 0 | 0 | 0 |
| Post Docs | 0 | 0 | 0 | 0 | 0 | 0 | 0 | 0 | 0 | 0 | 0 | 0 | 0 | 0 | 0 | 0 | 0 |
| Doctoral Students | 1 | 1 | 0 | 0 | 0 | 0 | 1 | 0 | 0 | 0 | 0 | 0 | 0 | 0 | 0 | 0 | 0 |
| Master's Students | 0 | 0 | 0 | 0 | 0 | 0 | 0 | 0 | 0 | 0 | 0 | 0 | 0 | 0 | 0 | 0 | 0 |
| Undergraduate Students | 0 | 0 | 0 | 0 | 0 | 0 | 0 | 0 | 0 | 0 | 0 | 0 | 0 | 0 | 0 | 0 | 0 |
| **Subtotal** | **3** | **3** | **0** | **0** | **0** | **0** | **2** | **1** | **0** | **0** | **0** | **0** | **0** | **0** | **0** | **0** | **0** |
| **Curriculum Development and Outreach** | | | | | | | | | | | | | | | | | |
| Senior Faculty | 0 | 0 | 0 | 0 | 0 | 0 | 0 | 0 | 0 | 0 | 0 | 0 | 0 | 0 | 0 | 0 | 0 |
| Junior Faculty | 0 | 0 | 0 | 0 | 0 | 0 | 0 | 0 | 0 | 0 | 0 | 0 | 0 | 0 | 0 | 0 | 0 |
| Research Staff | 0 | 0 | 0 | 0 | 0 | 0 | 0 | 0 | 0 | 0 | 0 | 0 | 0 | 0 | 0 | 0 | 0 |
| Visiting Faculty | 0 | 0 | 0 | 0 | 0 | 0 | 0 | 0 | 0 | 0 | 0 | 0 | 0 | 0 | 0 | 0 | 0 |
| Industry Researchers | 0 | 0 | 0 | 0 | 0 | 0 | 0 | 0 | 0 | 0 | 0 | 0 | 0 | 0 | 0 | 0 | 0 |
| Post Docs | 0 | 0 | 0 | 0 | 0 | 0 | 0 | 0 | 0 | 0 | 0 | 0 | 0 | 0 | 0 | 0 | 0 |
| Doctoral Students | 0 | 0 | 0 | 0 | 0 | 0 | 0 | 0 | 0 | 0 | 0 | 0 | 0 | 0 | 0 | 0 | 0 |
| Master's Students | 0 | 0 | 0 | 0 | 0 | 0 | 0 | 0 | 0 | 0 | 0 | 0 | 0 | 0 | 0 | 0 | 0 |
| Undergraduate Students | 0 | 0 | 0 | 0 | 0 | 0 | 0 | 0 | 0 | 0 | 0 | 0 | 0 | 0 | 0 | 0 | 0 |
| **Subtotal** | **0** | **0** | **0** | **0** | **0** | **0** | **0** | **0** | **0** | **0** | **0** | **0** | **0** | **0** | **0** | **0** | **0** |
| **ERC REU Student** | | | | | | | | | | | | | | | | | |
| NSF REU Site Award | 0 | 0 | 0 | 0 | 0 | 0 | 0 | 0 | 0 | 0 | 0 | 0 | 0 | 0 | 0 | 0 | 0 |
| ERC's Own REU | 0 | 0 | 0 | 0 | 0 | 0 | 0 | 0 | 0 | 0 | 0 | 0 | 0 | 0 | 0 | 0 | 0 |
| Other Visiting College Students | 0 | 0 | 0 | 0 | 0 | 0 | 0 | 0 | 0 | 0 | 0 | 0 | 0 | 0 | 0 | 0 | 0 |
| **Subtotal** | **0** | **0** | **0** | **0** | **0** | **0** | **0** | **0** | **0** | **0** | **0** | **0** | **0** | **0** | **0** | **0** | **0** |
| **Pre-College (K-12)** | | | | | | | | | | | | | | | | | |
| Teachers (RET) | 0 | 0 | 0 | 0 | 0 | 0 | 0 | 0 | 0 | 0 | 0 | 0 | 0 | 0 | 0 | 0 | 0 |
| Teachers (non-RET) | 0 | 0 | 0 | 0 | 0 | 0 | 0 | 0 | 0 | 0 | 0 | 0 | 0 | 0 | 0 | 0 | 0 |
| **Subtotal** | **0** | **0** | **0** | **0** | **0** | **0** | **0** | **0** | **0** | **0** | **0** | **0** | **0** | **0** | **0** | **0** | **0** |

Fig. A39 (continued)

Web Tables

| Personnel Type | Total | Gender | | | Citizenship Status | | | | | | | | | Ethnicity: Hispanic | | | Disability |
|---|---|---|---|---|---|---|---|---|---|---|---|---|---|---|---|---|---|
| | | | | | Race: U.S. citizens and permanent residents only | | | | | | | | | | | | |
| | | Male | Female | Gender Not Reported | AI/AN | NH/PI | B/AA | W | A | More than one race reported, minority | More than one race reported, non-minority | Not Provided | Other Non-U.S. | US/Perm | Temp | Not Reported | |
| **Morgan State University - Collaborating Institutions** | | | | | | | | | | | | | | | | | |
| **Total *** | **2** | **2** | **0** | **0** | **0** | **0** | **2** | **0** | **0** | **0** | **0** | **0** | **0** | **0** | **0** | **0** | **0** |
| **Research under Strategic Research Plan** | | | | | | | | | | | | | | | | | |
| Senior Faculty | 1 | 1 | 0 | 0 | 0 | 0 | 1 | 0 | 0 | 0 | 0 | 0 | 0 | 0 | 0 | 0 | 0 |
| Junior Faculty | 0 | 0 | 0 | 0 | 0 | 0 | 0 | 0 | 0 | 0 | 0 | 0 | 0 | 0 | 0 | 0 | 0 |
| Research Staff | 0 | 0 | 0 | 0 | 0 | 0 | 0 | 0 | 0 | 0 | 0 | 0 | 0 | 0 | 0 | 0 | 0 |
| Visiting Faculty | 0 | 0 | 0 | 0 | 0 | 0 | 0 | 0 | 0 | 0 | 0 | 0 | 0 | 0 | 0 | 0 | 0 |
| Industry Researchers | 0 | 0 | 0 | 0 | 0 | 0 | 0 | 0 | 0 | 0 | 0 | 0 | 0 | 0 | 0 | 0 | 0 |
| Post Docs | 0 | 0 | 0 | 0 | 0 | 0 | 0 | 0 | 0 | 0 | 0 | 0 | 0 | 0 | 0 | 0 | 0 |
| Doctoral Students | 1 | 1 | 0 | 0 | 0 | 0 | 1 | 0 | 0 | 0 | 0 | 0 | 0 | 0 | 0 | 0 | 0 |
| Master's Students | 0 | 0 | 0 | 0 | 0 | 0 | 0 | 0 | 0 | 0 | 0 | 0 | 0 | 0 | 0 | 0 | 0 |
| Undergraduate Students | 0 | 0 | 0 | 0 | 0 | 0 | 0 | 0 | 0 | 0 | 0 | 0 | 0 | 0 | 0 | 0 | 0 |
| **Subtotal** | **2** | **2** | **0** | **0** | **0** | **0** | **2** | **0** | **0** | **0** | **0** | **0** | **0** | **0** | **0** | **0** | **0** |
| **University of California, Berkeley - Collaborating Institutions** | | | | | | | | | | | | | | | | | |
| **Total *** | **1** | **1** | **0** | **0** | **0** | **0** | **0** | **1** | **0** | **0** | **0** | **0** | **0** | **0** | **0** | **0** | **0** |
| **Research under Strategic Research Plan** | | | | | | | | | | | | | | | | | |
| Senior Faculty | 0 | 0 | 0 | 0 | 0 | 0 | 0 | 0 | 0 | 0 | 0 | 0 | 0 | 0 | 0 | 0 | 0 |
| Junior Faculty | 1 | 1 | 0 | 0 | 0 | 0 | 0 | 1 | 0 | 0 | 0 | 0 | 0 | 0 | 0 | 0 | 0 |
| Research Staff | 0 | 0 | 0 | 0 | 0 | 0 | 0 | 0 | 0 | 0 | 0 | 0 | 0 | 0 | 0 | 0 | 0 |
| Visiting Faculty | 0 | 0 | 0 | 0 | 0 | 0 | 0 | 0 | 0 | 0 | 0 | 0 | 0 | 0 | 0 | 0 | 0 |
| Industry Researchers | 0 | 0 | 0 | 0 | 0 | 0 | 0 | 0 | 0 | 0 | 0 | 0 | 0 | 0 | 0 | 0 | 0 |
| Post Docs | 0 | 0 | 0 | 0 | 0 | 0 | 0 | 0 | 0 | 0 | 0 | 0 | 0 | 0 | 0 | 0 | 0 |
| Doctoral Students | 0 | 0 | 0 | 0 | 0 | 0 | 0 | 0 | 0 | 0 | 0 | 0 | 0 | 0 | 0 | 0 | 0 |
| Master's Students | 0 | 0 | 0 | 0 | 0 | 0 | 0 | 0 | 0 | 0 | 0 | 0 | 0 | 0 | 0 | 0 | 0 |
| Undergraduate Students | 0 | 0 | 0 | 0 | 0 | 0 | 0 | 0 | 0 | 0 | 0 | 0 | 0 | 0 | 0 | 0 | 0 |
| **Subtotal** | **1** | **1** | **0** | **0** | **0** | **0** | **0** | **1** | **0** | **0** | **0** | **0** | **0** | **0** | **0** | **0** | **0** |

*- If ERC Personnel were entered at the individual level  the Total may not equal the sum of the line items. This is because an individual may be reported in more than one personnel category but is only counted once for the purposes of the Total.

**Legend:**

AI/AN: American Indian or Alaska Native

NH/PI: Native Hawaiian or Other Pacific Islander

B/AA: Black/African American

W: White

A: Asian, e.g., Asian Indian, Chinese, Filipino, Japanese, Korean, Vietnamese, Other Asian

More than one race reported, minority - Personnel reporting a) two or more race categories and b) one or more of the reported categories includes American Indian or Alaska Native, Black or African American, or Native Hawaiian or Other Pacific Islander

More than one race reported, non-minority - Personnel reporting a) both White and Asian and b) no other categories in addition to White and Asian

US/Perm: U.S. citizens and legal permanent residents

Non-US: Non-U.S. citizens/Non-legal permanent residents

Fig. A39 (continued)

**Table 7a: Diversity Statistics for ERC faculty and students**

| | Total ERC Personnel [1] | | | | | Non-U.S. Citizen / Permanent Resident [2] | | | | |
|---|---|---|---|---|---|---|---|---|---|---|
| | Leadership Team [6] | Faculty [7] | Doctoral Students | Masters Students | Undergraduate Students | Leadership Team [6] | Faculty [7] | Doctoral Students | Masters Students | Undergraduate Students |
| **Center Total (Men, Women, Gender Not Reported, Persons with Disabilities, Underrepresented Racial Minorities, Hispanic/Latinos)** | **7** | **33** | **31** | **4** | **5** | **0** | **1** | **14** | **1** | **0** |
| **Women** | | | | | | | | | | |
| Category Total | 3 | 4 | 4 | 3 | 2 | 0 | 1 | 2 | 1 | 0 |
| Center Percent [3] | 43% | 12% | 13% | 75% | 40% | 0% | 3% | 6% | 25% | 0% |
| National Percent [4] | N/A | 11.8% | 20.8% | 22.4% | 18.1% | N/A | N/A | N/A | N/A | N/A |
| **Underrepresented Racial Minorities** | | | | | | | | | | |
| Category Total | 1 | 4 | 1 | 0 | 1 | 0 | 0 | 1 | 0 | 0 |
| Center Percent [3] | 14% | 12% | 3% | 0% | 20% | 0% | 0% | 3% | 0% | 0% |
| National Percent [4] | N/A | 6.0% | 7.3% | 10.0% | 12.0% | N/A | N/A | N/A | N/A | N/A |
| **Hispanic/Latinos** | | | | | | | | | | |
| Category Total | 1 | 0 | 0 | 0 | 0 | 0 | 0 | 0 | 0 | 0 |
| Center Percent [3] | 14% | 0% | 0% | 0% | 0% | 0% | 0% | 0% | 0% | 0% |
| National Percent [4] | N/A | 3.4% | 3.5% | 5.2% | 7.0% | 0% | N/A | N/A | N/A | N/A |
| **Persons with Disabilities** | | | | | | | | | | |
| Category Total | 0 | 0 | 0 | 0 | 0 | 0 | 0 | 0 | 0 | 0 |
| Center Percent [3] | 0% | 0% | 0% | 0% | 0% | 0 | 0% | 0% | 0% | 0% |
| National Percent [4,5] | N/A | 6.0% | 0.5% | 3.2% | 2.6% | N/A | N/A | N/A | N/A | N/A |

Fig. A40

## Figure 7b: Women in the ERC

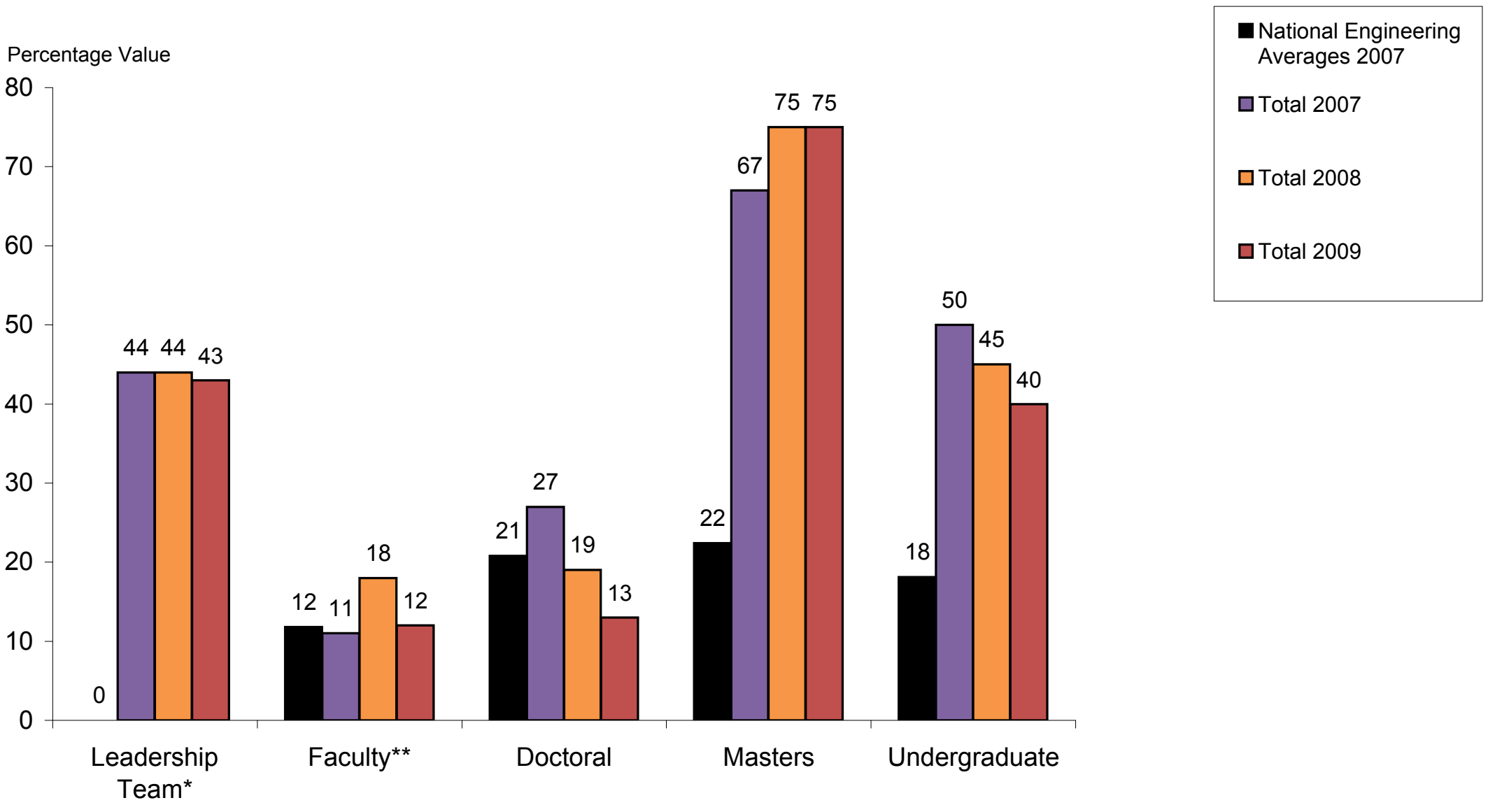

| Averages | Faculty | Doctoral | Masters | Undergraduate | Leadership Team |
|---|---|---|---|---|---|
| National Engineering Averages 2007 | 11.8% | 20.8% | 22.4% | 18.1% | N/A |
| All ERC's 2008 | 23.54% | 29.64% | 30.35% | 36.28% | 28.80% |

* - The Leadership Team Includes - Directors, Thrust Leaders, Education Program Leaders

** - Faculty Includes - Directors, Thrust Leaders, Education Program Leaders, Research - Senior Faculty, Research - Junior Faculty, Research - Visiting Faculty, Curriculum Development and Outreach - Senior Faculty, Curriculum Development and Outreach - Junior Faculty and, Curriculum Development and Outreach - Visiting Faculty

Fig. A41

## Figure 7c: Underrepresented Racial Minorities in the ERC

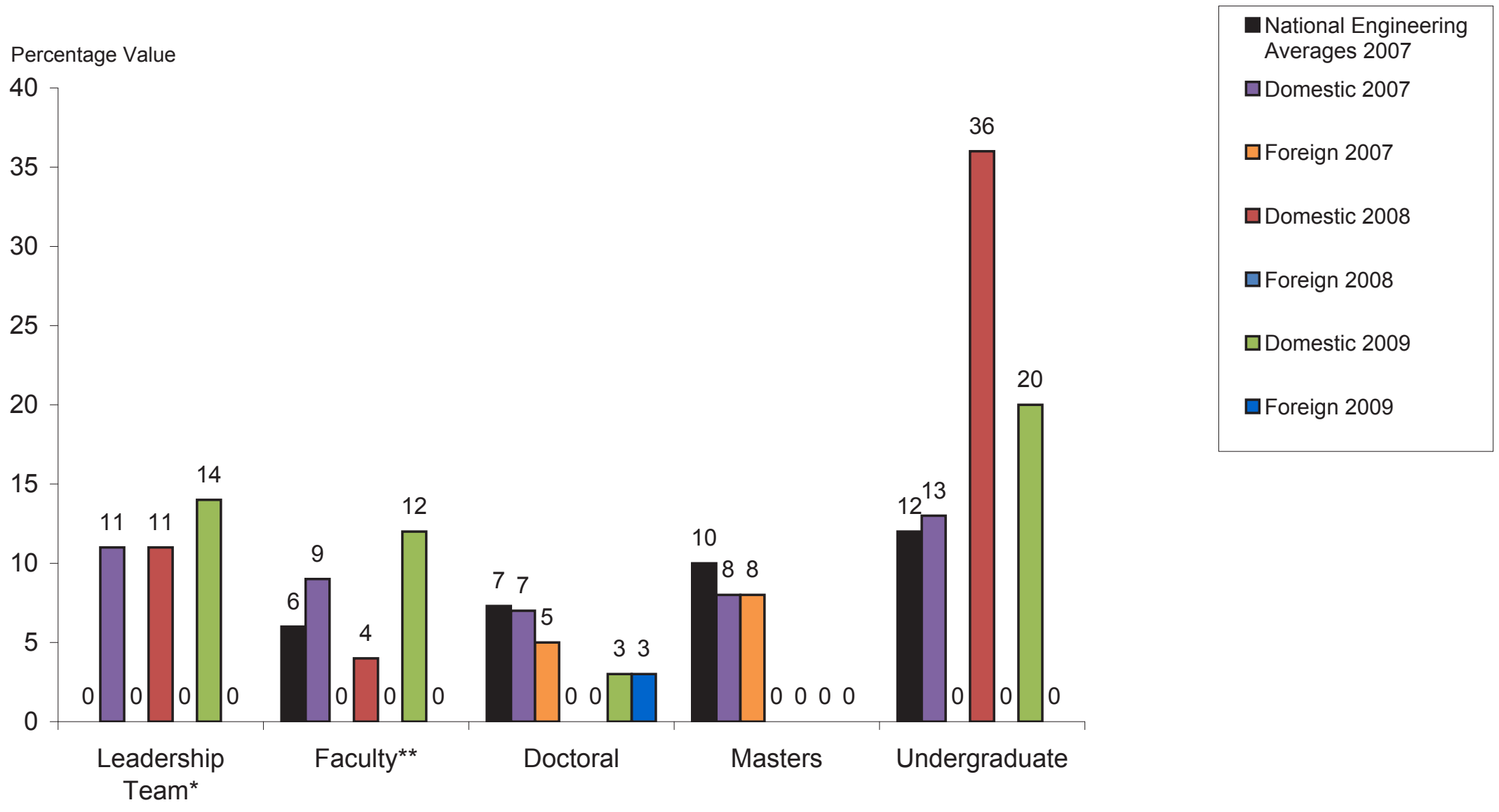

| Averages | Faculty | Doctoral | Masters | Undergraduate | Leadership Team |
|---|---|---|---|---|---|
| National Engineering Averages 2003 | 6.0% | 7.3% | 10.0% | 12.0% | N/A |
| All ERC's 2008 | 6.99% | 5.91% | 9.73% | 16.04% | 6.52% |

* - The Leadership Team Includes - Directors, Thrust Leaders, Education Program Leaders

** - Faculty Includes - Directors, Thrust Leaders, Education Program Leaders, Research - Senior Faculty, Research - Junior Faculty, Research - Visiting Faculty, Curriculum Development and Outreach - Senior Faculty, Curriculum Development and Outreach - Junior Faculty and, Curriculum Development and Outreach - Visiting Faculty

Fig. A42

## Figure 7d: Hispanics/Latinos in the ERC

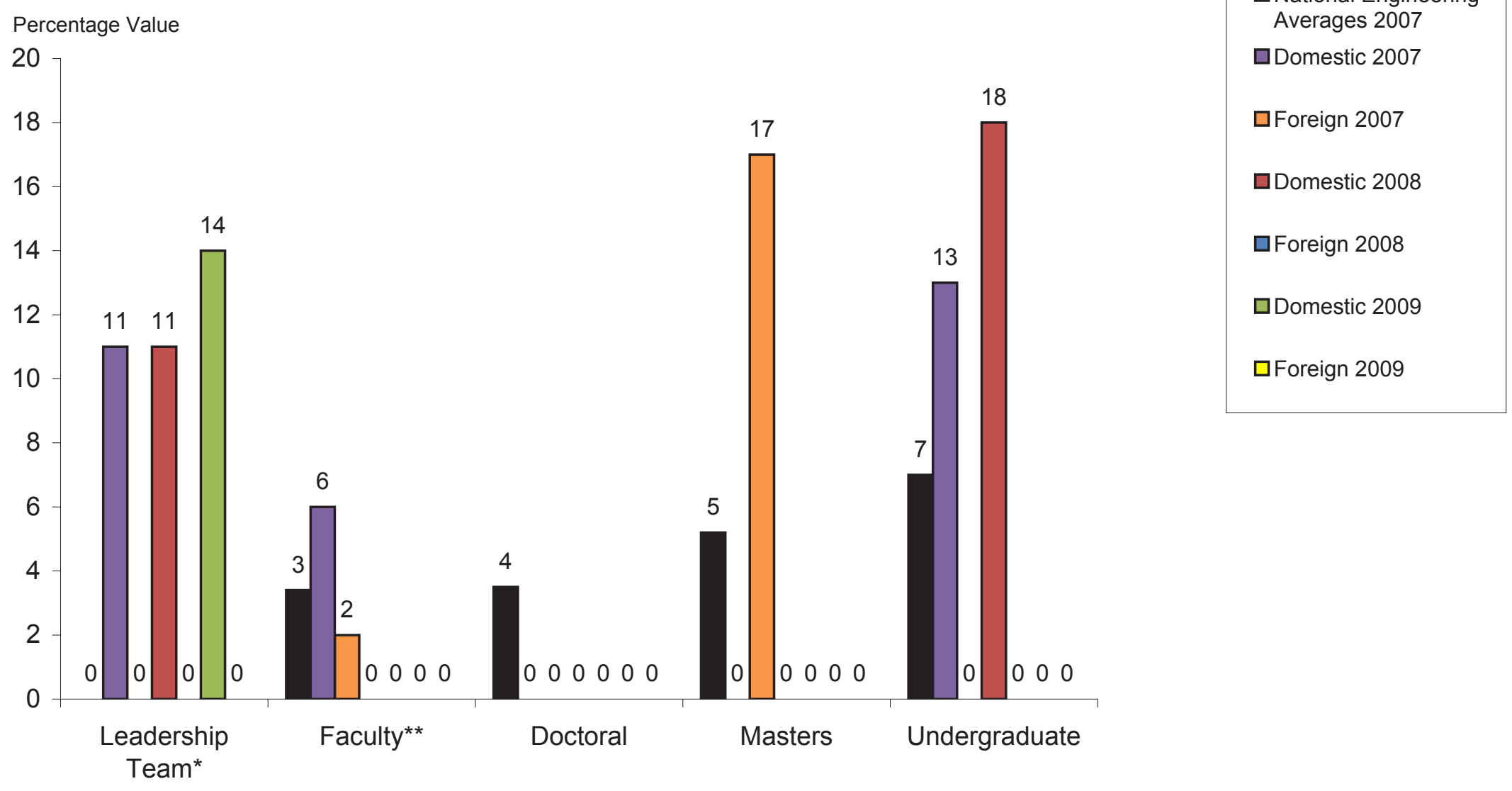

| Averages | Faculty | Doctoral | Masters | Undergraduate | Leadership Team |
|---|---|---|---|---|---|
| National Engineering Averages 2003 | 3.4% | 3.5% | 5.2% | 7.0% | N/A |
| All ERC's 2008 | 9.70% | 6.65% | 18.68% | 15.59% | 8.15% |

* - The Leadership Team Includes - Directors, Thrust Leaders, Education Program Leaders

** - Faculty Includes - Directors, Thrust Leaders, Education Program Leaders, Research - Senior Faculty, Research - Junior Faculty, Research - Visiting Faculty, Curriculum Development and Outreach - Senior Faculty, Curriculum Development and Outreach - Junior Faculty and, Curriculum Development and Outreach - Visiting Faculty

Fig. A43

**Figure 7e: Persons with Disabilities in the ERC**

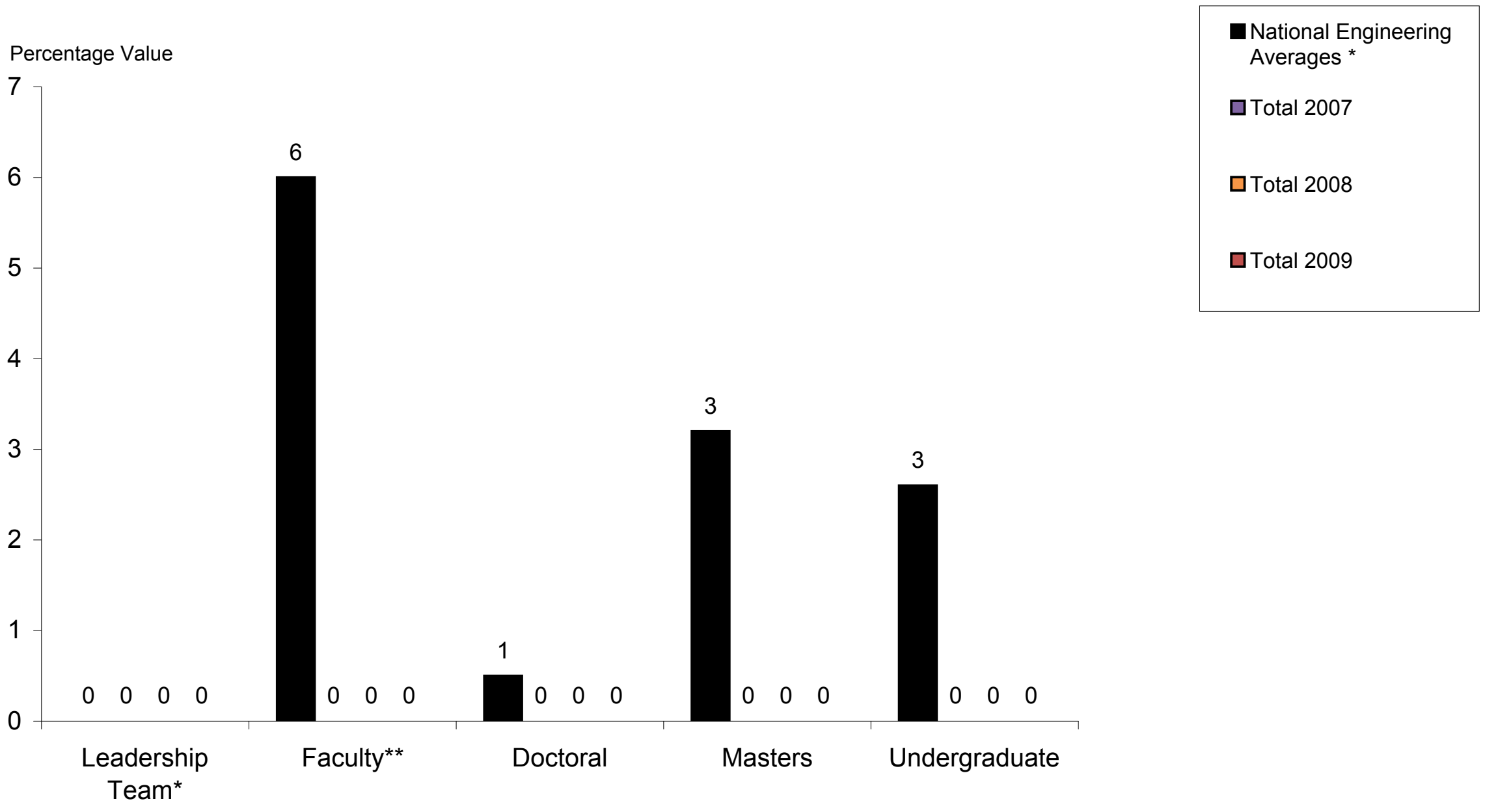

| Averages | Faculty | Doctoral | Masters | Undergraduate | Leadership Team |
|---|---|---|---|---|---|
| National Engineering Averages * | 6.0% | 0.5% | 3.2% | 2.6% | N/A |
| All ERC's 2008 | 2.28% | 0.74% | 1.17% | 1.05% | 4.35% |

* National Engineering Average data for disabled personnel reflects 2003 percentages (Faculty, Masters and Undergraduates), and 2005 percentages (Doctoral)

* - The Leadership Team Includes - Directors, Thrust Leaders, Education Program Leaders

** - Faculty Includes - Directors, Thrust Leaders, Education Program Leaders, Research - Senior Faculty, Research - Junior Faculty, Research - Visiting Faculty, Curriculum Development and Outreach - Senior Faculty, Curriculum Development and Outreach - Junior Faculty and, Curriculum Development and Outreach - Visiting Faculty

Fig. A44

**Table 7f: Center Diversity, by Institution**

| Institution | Females | | Underrepresented Racial Minorities [1] | | Hispanics [1] | |
|---|---|---|---|---|---|---|
| | # | % | # | % | # | % |
| **Lead Institution** | | | | | | |
| **Johns Hopkins University** | 22 | 24% | 4 | 6% | 2 | 2% |
| **Core Partner** | | | | | | |
| **Brigham and Women's Hospital** | 0 | 0% | 0 | 0% | 0 | 0% |
| **Carnegie Mellon University** | 0 | 0% | 0 | 0% | 0 | 0% |
| **Massachusetts Institute of Technology** | 0 | 0% | 0 | 0% | 0 | 0% |
| **Collaborating Institutions** | | | | | | |
| **Morgan State University** | 0 | 0% | 2 | 100% | 0 | 0% |
| **University of California, Berkeley** | 0 | 0% | 0 | 0% | 0 | 0% |

1 - This data only includes U.S. Citizens and Legal Permanent Residents.

Fig. A45

| Table 8: Functional Budget | | | | | | | | | | |
|---|---|---|---|---|---|---|---|---|---|---|
| Source of Support | | | | | | | | | | |
| Function | Direct Support | | | | | | | Direct Support Total | Associated Projects | Total |
| | ERC Program | Industry | State | University | Other NSF | Other Government | Other | | | |
| Engineering and Systems Infrastructure | $220,344 | $0 | $0 | $0 | $0 | $65,166 | $0 | $285,510 | $522,897 | $808,407 |
| Surgical Assistants | $156,325 | $49,655 | $0 | $0 | $51,742 | $4,732 | $0 | $262,454 | $842,698 | $1,105,152 |
| Surgical CAD/CAM | $80,768 | $119,073 | $0 | $77,792 | $0 | $118,271 | $112,201 | $508,105 | $464,246 | $972,351 |
| Research Total | $457,437 | $168,728 | $0 | $77,792 | $51,742 | $188,169 | $112,201 | $1,056,069 | $1,829,841 | $2,885,910 |
| General & Shared Equipment | $0 | $0 | $0 | $0 | $0 | $0 | $0 | $0 | $0 | $0 |
| New Facilities/ New Construction | $0 | $0 | $0 | $0 | $0 | $0 | $0 | $0 | $0 | $0 |
| Leadership/ Administration/ Management | $225,445 | $0 | $0 | $0 | $0 | $0 | $0 | $225,445 | $0 | $225,445 |
| Education Programs (excluding REU and RET Programs) | $100,669 | $0 | $0 | $0 | $54,174 | $0 | $0 | $154,843 | $71,197 | $226,040 |
| Research Experiences for Teachers Program | $0 | $0 | $0 | $0 | $0 | $0 | $0 | $0 | $0 | $0 |
| Research Experience for Undergraduates Program | $0 | $0 | $0 | $0 | $9,623 | $0 | $0 | $9,623 | $0 | $9,623 |
| Industrial Collaboration/Innovation Program | $7,767 | $0 | $0 | $0 | $0 | $0 | $0 | $7,767 | $0 | $7,767 |
| Center Related Travel | $0 | $0 | $0 | $0 | $0 | $0 | $0 | $0 | $0 | $0 |
| Residual Funds Remaining | $369,464 | $0 | $0 | $0 | $0 | $0 | $0 | $369,464 | N/A | $369,464 |
| Indirect Cost | $315,769 | $105,052 | $0 | $49,785 | $17,010 | $90,397 | $206 | $578,219 | N/A | $578,219 |
| Total | $1,476,551 | $273,780 | $0 | $127,577 | $132,549 | $278,566 | $112,407 | $2,401,430 | $1,901,038 | $4,302,468 |

Fig. A46

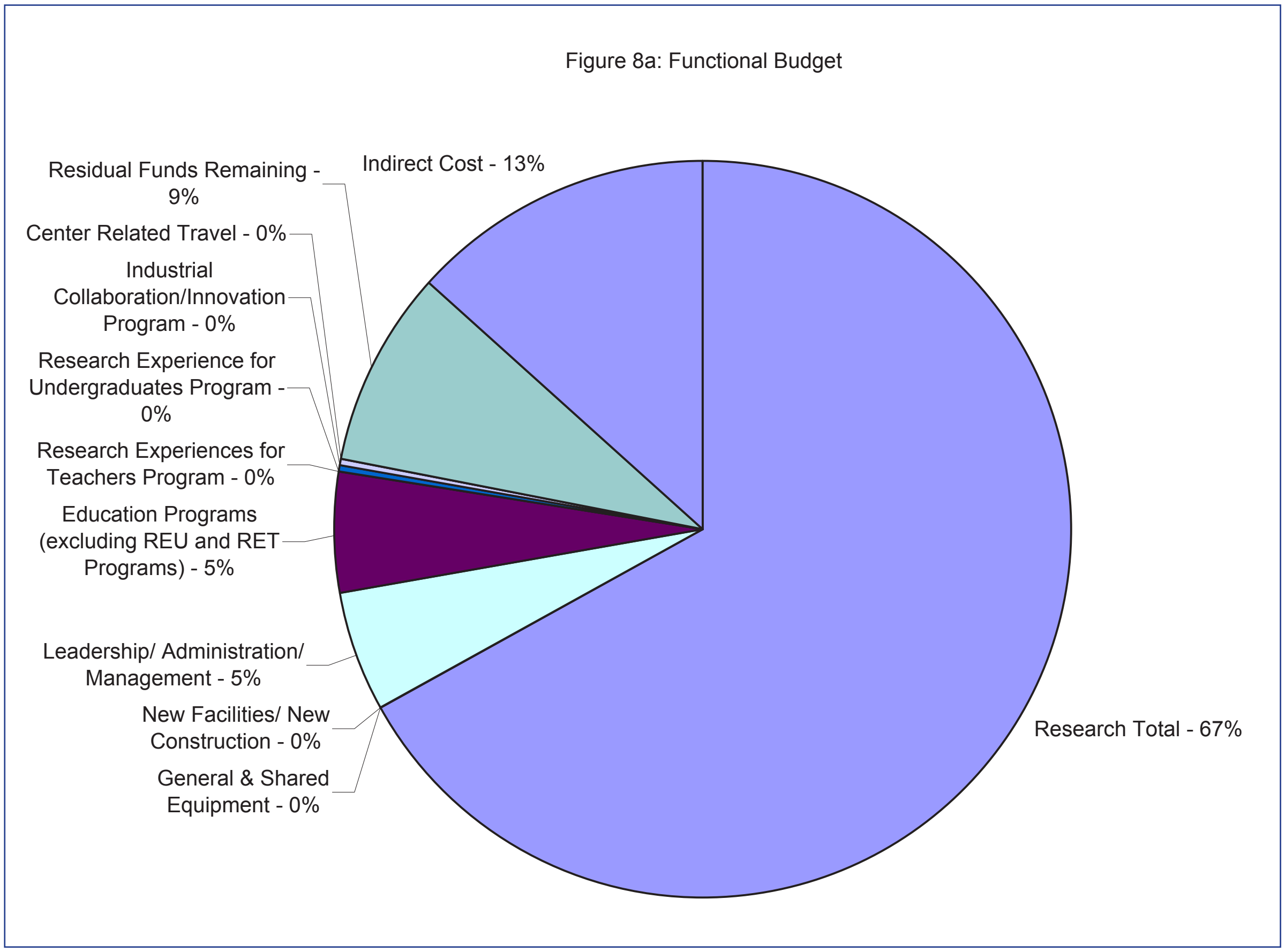

Fig. A47

| Table 9: Sources of Support | | | | | | | | | |
|---|---|---|---|---|---|---|---|---|---|
| Type of Support | Early Cumulative Total [1] | Sep 01, 2004 - Aug 31, 2005 | Sep 01, 2005 - Aug 31, 2006 | Sep 01, 2006 - Aug 31, 2007 | Sep 01, 2007 - Aug 31, 2008 | Sep 01, 2008 - Aug 31, 2009 | | | Cumul. Total [2] |
| | | | | | | Rec'd. | Prom. | Total | |
| NSF ERC Base Award | $18,066,289 | $1,380,989 | $3,006,277 | $3,166,945 | $1,733,330 | $0 | $0 | $0 | $27,353,830 |
| U.S. Industry | $1,127,371 | $10,000 | $10,000 | $0 | $0 | $0 | $0 | $0 | $1,147,371 |
| Foreign Industry | $0 | $0 | $0 | $0 | $0 | $0 | $0 | $0 | $0 |
| State | $0 | $0 | $0 | $0 | $0 | $0 | $0 | $0 | $0 |
| U.S. University | $7,475,562 | $0 | $0 | $0 | $1,763,465 | $0 | $0 | $0 | $9,239,027 |
| Foreign University | $0 | $0 | $0 | $0 | $0 | $0 | $0 | $0 | $0 |
| Other NSF (Not ERC Program) | $2,064,583 | $0 | $0 | $0 | $0 | $0 | $0 | $0 | $2,064,583 |
| Other U.S. Government (Not NSF) | $890,713 | $0 | $0 | $0 | $0 | $0 | $0 | $0 | $890,713 |
| Foreign Government | $0 | $0 | $0 | $0 | $0 | $0 | $0 | $0 | $0 |
| Other Source. | $0 | $0 | $0 | $0 | $0 | $0 | $0 | $0 | $0 |
| **TOTAL Unrestricted Cash** | **$29,624,518** | **$1,390,989** | **$3,016,277** | **$3,166,945** | **$3,496,795** | **$0** | **$0** | **$0** | **$40,695,524** |
| NSF ERC Program Special Purpose Awards and Supplements | $444,937 | $0 | $0 | $169,620 | $380,021 | $0 | $0 | $0 | $994,578 |
| U.S. Industry | $242,325 | $417,252 | $223,191 | $446,356 | $236,335 | $273,780 | $0 | $273,780 | $1,839,239 |
| Foreign Industry | $0 | $0 | $0 | $0 | $0 | $0 | $0 | $0 | $0 |
| State | $0 | $0 | $0 | $0 | $0 | $0 | $0 | $0 | $0 |
| U.S. University | $1,074,895 | $1,463,660 | $818,668 | $946,031 | $219,091 | $127,577 | $0 | $127,577 | $4,649,922 |
| Foreign University | $0 | $0 | $0 | $0 | $0 | $0 | $0 | $0 | $0 |
| Other NSF (Not ERC Program) | $369,028 | $338,125 | $310,879 | $435,656 | $318,044 | $132,549 | $0 | $132,549 | $1,904,281 |
| Other U.S. Government (Not NSF) | $298,255 | $754,598 | $942,860 | $1,234,420 | $970,482 | $269,720 | $8,846 | $278,566 | $4,479,181 |
| Foreign Government | $0 | $0 | $0 | $0 | $0 | $0 | $0 | $0 | $0 |
| Other Source. JHU APL<br>BCRF<br>Sloan Kettering | $151,915 | $94,048 | $127,877 | $67,663 | $45,285 | $112,407 | $0 | $112,407 | $599,195 |
| **TOTAL Restricted Cash** | **$2,581,355** | **$3,067,683** | **$2,423,475** | **$3,299,746** | **$2,169,258** | **$916,033** | **$8,846** | **$924,879** | **$14,466,396** |
| U.S. Industry | $0 | $0 | $0 | $60,660 | $100,299 | $26,028 | $0 | $26,028 | $186,987 |
| Foreign Industry | $0 | $0 | $0 | $0 | $26,417 | $694 | $0 | $694 | $27,111 |
| State | $0 | $0 | $0 | $0 | $0 | $0 | $0 | $0 | $0 |
| Other NSF (not ERC program) | $0 | $4,370 | $248,585 | $117,765 | $262,122 | $622,172 | $0 | $622,172 | $1,255,014 |
| Other US Government (not NSF) | $0 | $372,498 | $1,140,166 | $193,962 | $507,187 | $1,191,143 | $0 | $1,191,143 | $3,404,956 |
| Foreign Government | $0 | $0 | $0 | $0 | $0 | $0 | $0 | $0 | $0 |
| Other (specify source) | $0 | $0 | $0 | $79,930 | $73,840 | $61,001 | $0 | $61,001 | $214,771 |
| Foreign University | $0 | $0 | $0 | $0 | $0 | $0 | $0 | $0 | $0 |
| **TOTAL Associated Projects 4** | **$0** | **$376,868** | **$1,388,751** | **$452,317** | **$969,865** | **$1,901,038** | **$0** | **$1,901,038** | **$5,088,839** |
| **TOTAL Cash Support, All Sources 3** | **$33,133,150** | **$4,538,087** | **$5,439,752** | **$6,466,691** | **$7,964,571** | **$2,023,120** | **$378,310** | **$2,401,430** | **$55,161,920** |
| NSF/ERC Program 2 | $927,277 | $0 | $0 | $0 | $2,298,518 | $1,107,087 | $369,464 | $1,476,551 | N/A |
| U.S. Industry 2 | $0 | $79,415 | $0 | $0 | $0 | $0 | $0 | $0 | N/A |
| Foreign Industry 2 | $0 | $0 | $0 | $0 | $0 | $0 | $0 | $0 | N/A |
| State 2 | $0 | $0 | $0 | $0 | $0 | $0 | $0 | $0 | N/A |
| U.S. University 2 | $0 | $0 | $0 | $0 | $0 | $0 | $0 | $0 | N/A |
| Foreign University 2 | $0 | $0 | $0 | $0 | $0 | $0 | $0 | $0 | N/A |
| Other NSF (Not ERC Program) 2 | $0 | $0 | $0 | $0 | $0 | $0 | $0 | $0 | N/A |
| Other U.S. Government (Not NSF) 2 | $0 | $0 | $0 | $0 | $0 | $0 | $0 | $0 | N/A |
| Foreign Government 2 | $0 | $0 | $0 | $0 | $0 | $0 | $0 | $0 | N/A |
| Other Source. 2 | $0 | $0 | $0 | $0 | $0 | $0 | $0 | $0 | N/A |
| **TOTAL Residual Funds 2** | **$927,277** | **$79,415** | **$0** | **$0** | **$2,298,518** | **$1,107,087** | **$369,464** | **$1,476,551** | **N/A** |
| U.S. University | $0 | $0 | $0 | $12,210,000 | $0 | $0 | $0 | $0 | $12,210,000 |
| **TOTAL Value of New Construction** | **$0** | **$0** | **$0** | **$12,210,000** | **$0** | **$0** | **$0** | **$0** | **$12,210,000** |
| U.S. Industry | $1,200,000 | $295,000 | $0 | $0 | $0 | $0 | $0 | $0 | $1,495,000 |
| Foreign Industry | $88,185 | $0 | $0 | $0 | $0 | $0 | $0 | $0 | $88,185 |
| **TOTAL Value of In-Kind Equipment** | **$1,288,185** | **$295,000** | **$0** | **$0** | **$0** | **$0** | **$0** | **$0** | **$1,583,185** |
| U.S. Industry | $80,494 | $0 | $0 | $0 | $0 | $0 | $0 | $0 | $80,494 |
| **TOTAL Value of New Facilities in Existing Buildings** | **$80,494** | **$0** | **$0** | **$0** | **$0** | **$0** | **$0** | **$0** | **$80,494** |
| **TOTAL In-Kind Support, All Sources** | **$1,368,679** | **$295,000** | **$0** | **$12,210,000** | **$0** | **$0** | **$0** | **$0** | **$13,873,679** |
| **Percent Non-ERC Program Cash** | **42.52** | **69.03** | **44.74** | **48.40** | **62.70** | **100.00** | **100.00** | **100.00** | **48.61** |
| **Grand Total (Cash + In-Kind)** | **$34,501,829** | **$4,833,087** | **$5,439,752** | **$18,676,691** | **$7,964,571** | **$2,023,120** | **$378,310** | **$2,401,430** | **$73,817,360** |

1 - For Centers in operation for more than five years.

2 - No Residual amounts are included in the Cumulative Total column because the funds are by definition included in the year in which they were received.

3 - Cash Total = The sum of Unrestricted Cash, Restricted Cash, and Residual Funds for a particular NSF Award Year, but NOT Indirect Support for Associated Projects. This cash amount in Table 9 is also the total for the 'Expenditure' column pertaining to the same Award Year in Table 10: Annual Expenditures and Budgets.

4 - In 2003 -2004 Associated Projects Data was not collected.

**Explanation of Residual Funds entry in Direct Sources of Support - Cash**

Residual funds are reported for NSF cooperative agreement funding only. Of a total of $369,464 remaining, $149,460 are funds for the SAW project which is on no cost extension through Dec. 2009. $214,160 of the residual is from unused participant support costs and special purpose supplements over the life of the award. These funds require permission for rebudgeting.

Fig. A48

| Table 10: Annual Expenditures and Budgets | | | | | | | |
|---|---|---|---|---|---|---|---|
| Expenses Proposed and Residual Budget | Early Cumulative Total* | Sep 01, 2004 - Aug 31, 2005 Expend. | Sep 01, 2005 - Aug 31, 2006 Expend. | Sep 01, 2006 - Aug 31, 2007 Expend. | Sep 01, 2007 - Aug 31, 2008 Expend. | Sep 01, 2008 - Aug 31, 2009 Budget | Proposed Budget |
| **Salaries** | $10,386,132 | $2,515,149 | $2,317,778 | $2,467,688 | $3,749,424 | $784,108 | **$0** |
| **Faculty** | $3,900,915 | $486,708 | $478,141 | $1,010,614 | $701,557 | $202,995 | **$0** |
| **Postdocs** | $381,745 | $141,182 | $19,449 | $731,652 | $182,141 | $115,148 | **$0** |
| **Students** | $4,032,656 | $1,201,344 | $1,134,626 | $275,718 | $2,413,279 | $293,390 | **$0** |
| **Research Staff** | $742,358 | $381,008 | $409,408 | $154,389 | $168,845 | $40,481 | **$0** |
| **Administration/Management** | $1,316,115 | $304,907 | $276,154 | $295,315 | $283,602 | $132,094 | **$0** |
| **Other Salaries** | $12,343 | $0 | $0 | $0 | $0 | $0 | **$0** |
| **Fringe Benefits** | $1,606,551 | $385,284 | $332,043 | $382,913 | $396,974 | $124,899 | **$0** |
| **Salaries and Fringe Benefits Total** | $11,992,683 | $2,900,433 | $2,649,821 | $2,850,601 | $4,146,398 | $909,007 | **$0** |
| | | | | | | | |
| **Other Expenses** | $20,126,857 | $3,603,324 | $2,789,931 | $3,616,089 | $4,788,038 | $1,122,959 | **$0** |
| **General Operating Expenses** | $10,933,086 | $2,220,882 | $1,436,886 | $805,238 | $659,228 | $145,207 | **$0** |
| **Facilities** | $13,697 | $0 | $0 | $7,100 | $0 | $0 | **$0** |
| **Major Isolated Expenses** | $344,290 | $0 | $0 | $0 | $0 | $0 | **$0** |
| **Equipment** | $565,882 | $63,997 | $93,220 | $0 | $435,684 | $50,528 | **$0** |
| **Indirect Costs** | $8,269,902 | $1,318,445 | $1,259,825 | $1,373,290 | $1,866,641 | $580,084 | **$0** |
| **Other** | $0 | $0 | $0 | $1,430,461 | $1,826,485 | $347,140 | **$0** |
| | | | | | | | |
| **Residual Funds Remaining** | $4,043,748 | $0 | $0 | $0 | $0 | $369,464 | **$0** |
| | | | | | | | |
| **TOTAL Expenditures & Budgets** | $36,163,288 | $6,503,757 | $5,439,752 | $6,466,690 | $8,934,436 | $2,401,430 | **$0** |
| | | | | | | | |
| **Prior Award Year Residual Funds spent in Current Award Year** | $284,327 | $0 | $0 | $0 | $2,298,519 | $1,475,020 | **$0** |
| **ERC Program** | $284,327 | $0 | $0 | $0 | $2,298,519 | $1,475,020 | **$0** |
| **Other NSF** | $0 | $0 | $0 | $0 | $0 | $0 | **$0** |
| **Other Federal** | $0 | $0 | $0 | $0 | $0 | $0 | **$0** |
| **Industry** | $0 | $0 | $0 | $0 | $0 | $0 | **$0** |
| **Other** | $0 | $0 | $0 | $0 | $0 | $0 | **$0** |

* For Centers in operation for more than 5 years

**Explanation of Residual Funds entry in Annual Expenditures and Budget**

Residual funds are reported for NSF cooperative agreement funding only. Of a total of $369,464 remaining, $149,460 are funds for the SAW project which is on no cost extension through Dec. 2009. $214,160 of the residual is from unused participant support costs and special purpose supplements over the life of the award. These funds require permission for rebudgeting.

Fig. A49

**Table 11: Modes of Cash Support by Industry and Other Practitioner Organizations to the Center**

| Organization | Fees and Contributions | Sponsored Projects | Associated Projects [1] | Fees and Contributions | Sponsored Projects | Associated Projects [1] | Fees and Contributions | Sponsored Projects | Associated Projects [1] |
|---|---|---|---|---|---|---|---|---|---|
| Acoustic Medical Systems | $0 | $14,437 | | | | | | $132,332 | |
| American Shared Hospital Service | $10,000 | | | $0 | $0 | $0 | $0 | $0 | $0 |
| BCRF | | | | | | | | $24,284 | |
| Brigham & Women's Hospital | $0 | $15,666 | $0 | | | | | $127,577 | |
| Burdette Medical Systems | $0 | $6,532 | $0 | | $250 | | | $0 | |
| Department of Defense | $0 | $61,982 | $0 | | $51,574 | | | $18,529 | $14,504 |
| EADS Deutschland | | | | | | | | $694 | $694 |
| Foster Miller | $5,000 | $0 | $0 | | | | $0 | $0 | $0 |
| Hologic | $0 | $9,385 | $0 | | $47,217 | | | $60,013 | |
| IEEE | | | | | | | | $24,332 | |
| Ikona Medical | | | | | | | | $23,145 | $23,145 |
| Intuitive Surgical | $0 | $9,323 | $0 | | $85,339 | | | $91,703 | $10,267 |
| Invenios | | $2,245 | | | | | | | |
| Johns Hopkins University | $0 | $218,531 | $0 | | $867,585 | $34,765 | | $91,159 | $3,124 |
| Memorial Sloan Kettering Foundat | $0 | $5,685 | $0 | | $471 | | | $88 | |
| Morgan State University | | $11,662 | | | | | | | |
| NIH | | $333,175 | $602,598 | | $659,494 | $685,476 | | $1,917,086 | $1,651,297 |
| NSF | $1,325,985 | $116,110 | $41,175 | $3,214,159 | $184,724 | $220,784 | | $906,310 | $773,760 |
| Philips U.S.A. | $0 | $0 | $0 | $10,000 | | | | $5,990 | $5,990 |
| Siemens | $0 | $10,736 | $0 | | $4,122 | | $0 | $0 | $0 |
| **Total** | **$1,340,985** | **$815,469** | **$643,773** | **$3,224,159** | **$1,900,776** | **$941,025** | **$0** | **$3,423,242** | **$2,482,781** |

1 - In 2003 -2004 Associated Projects Data was not collected.

Fig. A50

JOHNS HOPKINS
UNIVERSITY